\documentclass[10pt,twocolumn,letterpaper]{article}
\usepackage{wacv}   %
\def\arxivbuild{1}    %
\def\plannedbuild{1}  %
\input{preamble}
\graphicspath{{figs/}{generated/}{planned/figs/}}
\newcommand{\PlannedBlocksRows}{N7 & Main grid and DINOv3 matched-rate arm with the SegFormer head & \RunsNSeven \\
N8 & Second learning-rate multiplier of \lvd, \dofa, \tm and \rn & \RunsNEight \\
N9 & The N7 cells with the Mask2Former head & \RunsNNine \\
N10 & The N7 cells with the SAM head & \RunsNTen \\
N11 & The N7 cells with the Mask2Former head and its matching objective & \RunsNEleven \\
N12 & Four types at the pair level \Lpair, five models and the DINOv3 matched-rate arm & \RunsNTwelve \\
N13 & Seeds 6 to 8 of \dofa, \tm and \rn on \burn, clean and \Ltwo & \RunsNThirteen \\
N14 & Clean and \Ltwo cells under generalized cross-entropy and soft bootstrapping & \RunsNFourteen \\
N15 & The N7 cells with the pretrained SAM head & \RunsNFifteen \\
D1 & Chesapeake, main grid, NLCD real arm, matched-rate arm and decoder set & \RunsDOne \\
D2 & CloudSEN12, main grid, s2cloudless real arm, matched-rate arm and decoder set & \RunsDTwo \\
D3 & DFC2020, main grid, MODIS real arm, matched-rate arm and decoder set & \RunsDThree \\
D4 & EnviroAtlas, main grid, NLCD real arm, matched-rate arm and decoder set & \RunsDFour \\
D5 & SpaceNet 2, main grid, matched-rate arm and decoder set & \RunsDFive \\
D6 & Kuro Siwo, main grid, CEMS real arm, matched-rate arm and decoder set & \RunsDSix \\
D7 & OpenEarthMap, main grid, matched-rate arm and decoder set & \RunsDSeven \\}
\newcommand{\PlannedBlocksCaption}{ Every planned run of N7 to N15 and D1 to D7 finished. N8 reuses \RunsNEightAliased{} cells of A2 and N2. A D block runs on its own dataset, so its runs count that dataset only.}
\providecommand{\ProvNSeven}{\PlannedProvenance}
\providecommand{\ProvNEight}{\PlannedProvenance}
\providecommand{\ProvNNine}{\PlannedProvenance}
\providecommand{\ProvNTen}{\PlannedProvenance}
\providecommand{\ProvNEleven}{\PlannedProvenance}
\providecommand{\ProvNTwelve}{\PlannedProvenance}

\providecommand{\ProvNFourteen}{\PlannedProvenance}
\providecommand{\ProvNFifteen}{\PlannedProvenance}
\providecommand{\ProvExtension}{\PlannedProvenance}
\providecommand{\AbstractPostReview}{Under other decoder heads the verdicts mostly hold on \flair and are decoder dependent on \burn. On four of five real label pairs the real label costs more than its matched simulated type.}

\definecolor{wacvblue}{rgb}{0.21,0.49,0.74}
\usepackage[pagebackref,breaklinks,colorlinks,allcolors=wacvblue]{hyperref}
\hypersetup{pdfauthor={},pdftitle={}}

\def\wacvPaperID{21}
\def\confName{WACV}
\def\confYear{2027}

\title{Benchmarking Geospatial Foundation Models under Calibrated Label Errors}

\author{Mollah Md Saif\textsuperscript{*} \quad Showrin Rahman\textsuperscript{*} \quad Anindya Majumder\\
Sowad Rahman \quad Md. Golam Rabiul Alam\\
Department of Computer Science and Engineering, BRAC University, Dhaka, Bangladesh\\
{\tt\small mollah.mdsaif@bracu.ac.bd \quad showrin.rahman@g.bracu.ac.bd \quad anindya.majumder@g.bracu.ac.bd}\\
{\tt\small sowad.rahman@g.bracu.ac.bd \quad rabiul.alam@bracu.ac.bd}\\
{\small \textsuperscript{*}Equal contribution}}

\begin{document}
\maketitle
\begin{abstract}
Earth-observation labels from maps, older products or hurried annotation arrive misaligned, coarse, loosely outlined or incomplete.
Geospatial foundation models (\gfms) are nevertheless benchmarked with labels taken as correct.
A practitioner therefore cannot tell which encoder to choose when the labels are imperfect.
We present a benchmark protocol for fine-tuning under such labels.
Four spatially structured error types are calibrated to equal label \miou and applied identically to every model and seed.
Every contrast is paired and read against an equivalence margin of $\pm$\TostMargin{} points and a minimum detectable effect.
We apply it to nine encoders, six decoder heads and ten datasets.
In an exploratory reading at a matched learning rate the satellite-pretrained and web-pretrained DINOv3 encoders are equivalent within the margin on \hlDsNSixEquiv{} of \hlDsNSixRead{} datasets.
Multispectral \gfms and an ImageNet ResNet-50 are equivalent within the margin on about half of them.
Labels with random offsets cost the least at equal label quality.
\AbstractPostReview{}
No choice of pretraining data makes an encoder consistently more robust, so the encoder is better chosen by its accuracy under imperfect labels.
\end{abstract}

\par\smallskip\begingroup\footnotesize\noindent\textit{This is the extended version. It adds the supplementary material of the venue version as appendices after the references.}\par\endgroup\medskip
\section{Introduction}
\label{sec:intro}

\begin{figure*}[t]
\centering
\IfFileExists{planned/figs/overview_planned.pdf}{\includegraphics[width=\textwidth]{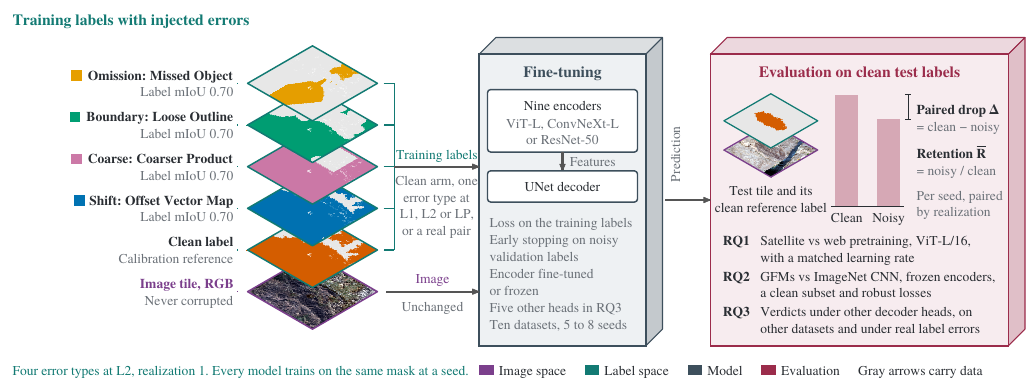}}{\phfig{5.6cm}{planned/figs/overview\_planned.pdf from figs/make\_overview\_fig.py --planned}}
\caption{\textbf{Study overview:} On the left, one \burn training tile with its clean label and the four corrupted labels of realization \CalSeed{} at \Ltwo. In the middle, every model trains on clean or corrupted training and validation labels, with a UNet decoder in the main grid. On the right, predictions are scored on the clean reference test labels. The scores give the paired drop \drop{} of \cref{eq:drop} and the retention \ret{} of \cref{eq:ret}. Each layer prints the label \miou{} of its whole mask set.}
\label{fig:overview}
\end{figure*}

Segmentation labels for Earth observation are rarely drawn for the image they are paired with.
Building footprints come from vector maps offset from the raster \cite{vargasmunoz2019correcting,kaiser2017learning}, and land cover from products coarser than the imagery.
Burned areas come from inventories with loose outlines \cite{kolden2012mapped,knezevic2026annual}, and rushed annotation misses small objects.
Such errors are spatially structured \cite{maiti2022effect,yao2023spatial,vargasmunoz2021osm}.
Imperfect labels are therefore the normal case for an agency that updates a land-cover map from an older product.
The practitioner has to choose an encoder, decide whether to freeze it, and decide whether to relabel a subset.

Geospatial foundation models (\gfms) are encoders pretrained with self-supervised or generative objectives on large Earth-observation archives \cite{bommasani2021foundation,mai2023opportunities}.
Their benchmarks, including \gbtwo \cite{simumba2025geobench2} and PANGAEA \cite{marsocci2024pangaea}, train on clean labels.
The gap is that no segmentation benchmark compares pretrained encoders in controlled pairs at equal label quality across error types.
Pant tests frozen against fine-tuned \gfms for equivalence \cite{pant2026auditing}, but no label-error benchmark pairs that test with a minimum detectable effect \cite{card2020little}.
This paper contributes a benchmark protocol for that gap, with calibrated error types, paired fine-tuning and an equivalence test for every contrast.

We apply it to nine encoders, six decoder heads and ten datasets (\cref{fig:overview}).
Four error types (\etype{shift}, \etype{coarse}, \etype{boundary} and \etype{omission}) are injected into the training and validation labels.
Each type is calibrated to equal label \miou against the clean masks at two levels (\Lone and \Ltwo).
A third level \Lpair{} is set to the label quality measured on real label pairs.
Six datasets carry such a pair, an operational or algorithmic label and a reference label on the same pixels.
At every seed all models train on the same corrupted mask and are scored on the clean reference test labels, so model contrasts are paired.
Each contrast is read against a pre-specified equivalence margin and a minimum detectable effect.
A null is therefore reported as equivalence, as undetermined with its minimum detectable effect, or as a detected difference.

We ask three research questions.
RQ1 asks whether pretraining on sub-meter satellite rather than web images makes fine-tuning more robust to label errors.
The DINOv3 ViT-L/16 encoders distilled from SAT-493M and from LVD-1689M teachers \cite{simeoni2025dinov3} test it at one learning rate.
RQ2 asks whether multispectral \gfms degrade less than an ImageNet CNN under label errors and whether a frozen encoder protects against them.
It also asks which error type costs most and whether a clean subset or a noise-robust loss recovers the drop.
RQ3 asks whether these verdicts hold under other decoder heads, on other datasets and under real label errors.

An analysis plan states nine hypotheses, four fixed before the noisy-label runs and five before the runs of their blocks (\Cref{sec:stats}\suppnote{, \stab{tab:hypotheses}}).

Our contributions are the following:
\begin{itemize}[leftmargin=*,itemsep=1pt,topsep=2pt]
\item \textbf{Calibrated Label Error Protocol:} Four spatially structured error types calibrated to equal label \miou at three levels and applied identically to every model at every seed.
\item \textbf{Equivalence Reading of Every Contrast:} Each contrast carries a $\pm$\TostMargin-point equivalence test and a minimum detectable effect, so a null is never read as no difference.
\item \textbf{Matched Conditions:} Every core model runs at two learning-rate multipliers, so each model contrast is read at one rate.
\item \textbf{Encoders, Decoders, Datasets and Real Labels:} Nine encoders in controlled pairs, six decoder heads, eight further datasets and six real label pairs, five of them trained on next to a matched simulated type.
\item \textbf{Practical Tests for Imperfect Labels:} Paired tests of frozen against fine-tuned encoders, of a \SubsetPct\,\% clean subset against all tiles with errors, and of two noise-robust losses.
\end{itemize}
We will release the code, plan, manifest, masks, tile lists and per-seed results.

\paragraph{Key Takeaways:}
Three readings answer RQ1 and RQ2.
In an exploratory reading at one learning rate the satellite-pretrained and web-pretrained encoders are equivalent within the margin on \hlDsNSixEquiv{} of \hlDsNSixRead{} datasets.
The multispectral \gfms and the ImageNet CNN are equivalent within the margin on \hlDsFTwoEquiv{} of \hlDsFTwoRead.
Labels with random offsets cost the least on \hlDsShiftLeast{} of \hlDsTypeRead{} datasets.
For region-type targets, label effort is therefore better spent on outlines, resolution and completeness.
The clean-label ranking of the encoders also predicts their ranking under label errors.

\section{Related Work}
\label{sec:related}

\paragraph{Benchmarks of Geospatial Foundation Models:}
GEO-Bench \cite{lacoste2023geobench}, \gbtwo \cite{simumba2025geobench2} and PANGAEA \cite{marsocci2024pangaea} rank \gfms on clean training labels.
Robustness studies of \gfms perturb the imagery rather than the labels \cite{li2025reobench,lehmann2026beyond,faruk2026rspdbench}.
Training on existing products such as OpenStreetMap \cite{vargasmunoz2019correcting,kaiser2017learning,maggiori2017convolutional} has had no controlled counterpart with pretrained \gfm encoders at equal label quality.
An audit of \gfm papers reports cross-paper gaps above seed variance and names label quality as an open axis \cite{corley2026noone}.

\paragraph{Label Noise in Segmentation and Remote Sensing:}
Label noise in natural and medical image segmentation is addressed by label correction \cite{liu2022adele,yao2023spatial} and noise taxonomies \cite{kimhi2024noisy}.
Mnih and Hinton modeled registration and omission errors in aerial labels \cite{mnih2012learning}.
Coarse labels have been studied as a resolution problem \cite{robinson2019large,rolf2022resolving}.
Maiti \etal injected simulated error types at fixed knobs into aerial labels for CNNs trained from scratch \cite{maiti2022effect}.
They rank \etype{shift} costliest, and ours ranks it cheapest on \hlDsShiftLeast{} of \hlDsTypeRead{} datasets at equal label \miou with pretrained encoders.
Burgert \etal compared noise types in multi-label classification \cite{burgert2022effects}.

Recent methods train CNNs on the noisy labels themselves \cite{liu2024taskspecific,liu2024aio2,liu2024cromss}.
Nogueira \etal benchmark label-noise estimators with UNet and SegFormer on seven simulated building-mask errors, graded by mask IoU but not equalized across types \cite{nogueira2026datacentric}.
Noise-robust losses \cite{zhang2018gce} and label bootstrapping \cite{reed2015bootstrapping} were designed for image-level labels, and we test both as drop-in baselines.
We measure the paired drop at calibrated label quality, whereas label-efficiency studies vary the quantity of labels \cite{marsocci2024pangaea,kaushik2026cryobench,kaiser2017learning} or their cost \cite{zlateski2018importance,das2023lowcost,mackowiak2018cereals}.

\paragraph{Pretraining, Freezing and Robustness:}
Pretraining improves robustness to label noise in classification \cite{hendrycks2019using,ghosh2021contrastive,zheltonozhskii2022contrast}, and in-domain self-supervision beats ImageNet and DINOv2 initialization at high noise \cite{szczecina2025pretrain}.
Concurrent work pairs satellite and web DINOv3 encoders on clean labels and finds the latter stronger on its semantic tasks \cite{sharma2026vhrbench,agrawal2026satellite}.
Full fine-tuning can distort pretrained features \cite{kumar2022lpft}.
Freezing has been proposed as a defense against noisy labels \cite{ahn2024turn,yu2024cufit}.
We measure robustness as a paired drop and plot retention against clean accuracy\suppnote{ (\sfig{fig:supp-er})}, the view of effective robustness for distribution shift \cite{taori2020measuring,shi2023effective}.

\section{Benchmark Protocol}
\label{sec:method}

This section defines the protocol, because every verdict of \Cref{sec:results} rests on how labels are corrupted, how models train and how contrasts are read.

\paragraph{Protocol in Five Steps:}
Corrupt the training and validation labels with one error type at a time.
Calibrate each type to a target label \miou, so types compare at equal label quality.
Fine-tune every encoder on the same corrupted masks at every seed with one recipe.
Score on the clean reference test labels and form paired drops.
Read each pre-specified contrast against an equivalence margin and a minimum detectable effect (\Cref{sec:stats}).

\subsection{Datasets}
\label{sec:datasets}

Two core datasets carry the full design, seven extension datasets test whether its verdicts transfer, and one dataset carries an algorithmic label.

\paragraph{Core Datasets:}
The \gbtwo version of \burn \cite{jakubik2023prithvi,simumba2025geobench2} pairs \ResBurnM\,m harmonized Landsat and Sentinel-2 tiles with binary burn masks.
It has \NBurnBands{} spectral bands and \NBurnTrain{} training, \NBurnVal{} validation and \NBurnTest{} test tiles of \TilePx\,px.
The \gbtwo version of \flair \cite{garioud2023flair2,simumba2025geobench2} pairs \ResFlairM\,m aerial tiles of France with RGB, near infrared and elevation channels.
All models train on one fixed subset of \NFlairTrain{} training and \NFlairVal{} validation tiles and are tested on the full \NFlairTest-tile test split.

\paragraph{Extension Datasets:}
Seven further datasets run a reduced grid, one block each (D1 to D7, \Cref{tab:ext-datasets}).
Five of them carry a second label on the same pixels from an operational product or an algorithm.
A real label error can therefore be measured and trained on next to the four simulated types.
Chesapeake \cite{robinson2019large} and EnviroAtlas \cite{rolf2022resolving} pair a 1\,m land-cover label with the 30\,m NLCD product over aerial tiles.
CloudSEN12 \cite{aybar2022cloudsen12} pairs expert cloud masks with the s2cloudless mask of the same Sentinel-2 patches.
DFC2020 \cite{schmitt2019sen12ms,schmitt2020weakly} pairs 10\,m semi-manual labels with the MODIS land-cover label of SEN12MS.
Kuro Siwo \cite{bountos2024kurosiwo} pairs an expert flood annotation with the Copernicus EMS (CEMS) polygons on Sentinel-1 SAR.
SpaceNet 2 \cite{vanetten2018spacenet} adds building footprints at 0.3\,m.
OpenEarthMap \cite{xia2023openearthmap} adds eight land-cover classes at 0.25 to 0.5\,m.
Every extension dataset trains on at most 1{,}000 tiles and 250 validation tiles drawn once with seed \CalSeed{}\suppnote{ (\ssec{sec:supp-ext})}.
\sfloods \cite{bonafilia2020sen1floods11} adds a tenth dataset whose tiles carry hand-labeled water masks and Otsu masks \cite{otsu1979threshold} from the coincident Sentinel-1 scene.
Against the hand masks the Otsu masks score a label \miou of \hlOtsuLabelMiouFracAll{}, near the \Ltwo target\suppnote{ (\ssec{sec:supp-stretch})}.

\begin{table*}[t]
\centering\footnotesize\setlength{\tabcolsep}{3pt}
\IfFileExists{planned/generated/table_ext_datasets.tex}{\begin{adjustbox}{max width=\textwidth}%
\providecommand{\gcvdry}[1]{\textcolor{red}{#1}}

\begin{tabular}{@{}llccclccc@{}}
\toprule
Block & Dataset & Sensor & GSD (m) & Classes & Tiles (train / val / test) & Real pair & Label \miou{} / flip (\%) & Runs \\
\midrule
D1 & Chesapeake~\cite{robinson2019large} & NAIP aerial RGBN & 1 & 4 & 1{,}000 / 250 / 1{,}000 & NLCD 30\,m vs 1\,m Conservancy label & \RealLabelMiouChes{} / \RealFlipRateChes & \RunsDOne \\
D2 & CloudSEN12~\cite{aybar2022cloudsen12} & S2 L1C, 12 bands & 10 & 4 & 1{,}000 / 250 / 975 & s2cloudless vs expert cloud mask & \RealLabelMiouCsTwelve{} / \RealFlipRateCsTwelve & \RunsDTwo \\
D3 & DFC2020~\cite{schmitt2019sen12ms,schmitt2020weakly} & S2 L1C 13 + S1 VV/VH & 10 & 8 & 886 / 100 / 5{,}128 & MODIS IGBP 500\,m vs 10\,m DFC2020 label & \RealLabelMiouDfc{} / \RealFlipRateDfc & \RunsDThree \\
D4 & EnviroAtlas~\cite{rolf2022resolving} & NAIP aerial RGBN & 1 & 4 & 1{,}000 / 250 / 1{,}000 & NLCD 30\,m vs EPA 1\,m label & \RealLabelMiouEnva{} / \RealFlipRateEnva & \RunsDFour \\
D5 & SpaceNet 2~\cite{vanetten2018spacenet} & WorldView-3, 8 bands & 0.3 & 2 & 1{,}000 / 250 / 2{,}961 & -- & -- & \RunsDFive \\
D6 & Kuro Siwo~\cite{bountos2024kurosiwo} & S1 GRD VV/VH + DEM & 10 & 3 & 1{,}000 / 250 / 2{,}000 & CEMS polygons vs expert re-annotation & \RealLabelMiouKuro{} / \RealFlipRateKuro & \RunsDSix \\
D7 & OpenEarthMap~\cite{xia2023openearthmap} & Aerial and satellite RGB & 0.25--0.5 & 8 & 1{,}000 / 250 / 2{,}000 & -- & -- & \RunsDSeven \\
\bottomrule
\end{tabular}
\end{adjustbox}}{\tabletbd[2.5cm]{planned/generated/table\_ext\_datasets.tex: run analysis/planned\_placeholders.py}}
\caption{\textbf{Extension datasets of blocks D1 to D7:} Sensor, ground sampling distance (GSD), class count and tile counts of the fixed splits. The last columns give the real label pair where one exists. The label \miou and flip rate of a pair are measured on its training and validation tiles. Runs count every run of the block over the main grid, the real arm, the matched-rate arm and the decoder set. A dash marks a dataset without a second label.}
\label{tab:ext-datasets}
\end{table*}

\subsection{Models and Decoder Heads}
\label{sec:models}

Nine encoders run, five in the full design and four as controlled extensions\suppnote{ (\sfig{fig:supp-roster})}.
\sat and \lvd are the DINOv3 ViT-L/16 encoders distilled from teachers pretrained on SAT-493M satellite images and on LVD-1689M web images \cite{simeoni2025dinov3}.
\dofa \cite{xiong2024dofa} and \tm \cite{jakubik2025terramind} are multispectral \gfms that receive every optical band their pretraining supports\suppnote{ (\stab{tab:supp-models})}.
\rn \cite{he2016resnet} is initialized from ImageNet and receives all dataset bands through an adapted first convolution.
\prithvi \cite{szwarcman2024prithvi2} is pretrained on the HLS sensor family of \burn.
\rnssl \cite{wang2022ssl4eo} is the \rn architecture pretrained on Sentinel-2 instead of ImageNet.
\vitr is the DINOv3 architecture without pretraining.
\cnxl is distilled from LVD-1689M.
All four run the main grid on both core datasets with their own learning-rate probes.
Each is read against core models as controls\suppnote{ (\ssec{sec:supp-stretch})}.

The main grid uses the UNet decoder \cite{ronneberger2015unet} of the \gbtwo recipe.
Five further heads repeat the grid on the same masks and seeds, so the decoder is a tested factor.
The first two are the all-MLP head of SegFormer \cite{xie2021segformer} and a Mask2Former head \cite{cheng2022mask2former} trained by per-pixel cross-entropy.
A third is the same Mask2Former head with its matching objective.
The SAM mask decoder \cite{kirillov2023segment} runs without prompts, once from scratch and once from its released weights.
\suppnote{\ssec{sec:supp-heads} specifies each head and \ssec{sec:supp-n7} its parameter count.}

\subsection{Label Corruption}
\label{sec:noise}

The four types are compared at equal label quality rather than at equal knob values, which makes their ranking meaningful.

\paragraph{Error Types:}
Each type imitates a known source of error in Earth-observation labels.
The \etype{shift} type models misalignment between a vector source and the raster \cite{vargasmunoz2019correcting,vargasmunoz2021osm,maiti2022effect}, and the \etype{coarse} type models labels drawn at a lower resolution \cite{malkin2019label,rolf2022resolving}.
The \etype{boundary} type models loosely drawn outlines \cite{kolden2012mapped}, and the \etype{omission} type models missed objects \cite{vargasmunoz2021osm,gutter2022impact}.
\Cref{fig:overview} shows one corrupted tile per type\suppnote{, and \ssec{sec:supp-calibration} gives each operator and its knob}.
Random pixel flips are excluded, because segmentation networks tolerate them better than structured errors \cite{maiti2022effect,heller2018imperfect}.

\paragraph{Calibration:}
Severity is defined by label quality, in the class-averaged unit of the evaluation metric.
For each type and dataset, a bisection search sets the knob of realization \CalSeed.
The corrupted training and validation masks then score a label \miou of \CalTargetLOne{} (\Lone) or \CalTargetLTwo{} (\Ltwo) against the clean masks, within $\pm$\CalTol.
On \burn the \Ltwo offsets and blocks exceed operational errors, so \etype{shift} and \etype{coarse} are stress tests there\suppnote{ (\ssec{sec:supp-calibration})}.
A third level \Lpair{} therefore targets a measured label quality.
Its label \miou of \LrTargetMiou{} is the median over the \LrTargetNPairs{} real label pairs of this study.

For each type and level, \NRealisations{} realizations are generated with the knob calibrated on realization \CalSeed.
Realization $s$ is used at training seed $s$.
The \etype{omission} realizations vary in severity\suppnote{ (\stab{tab:supp-realizations})}.
On \flair they are milder than the target.
Each affected drop is also rescaled to it\suppnote{ (\stab{tab:supp-omadj})}.

\paragraph{Real Label Pairs:}
On a pair dataset the operational label is aligned to the tile grid once and scored against the reference label by the same label \miou.
It then enters training as a fifth error type at every seed.
Its matched arm is the \etype{coarse} type at the level, \Lone or \Ltwo, whose target lies nearest to the measured label \miou.
The \etype{coarse} type is fixed for every pair, because the four types share one target per level.
A real and a simulated error at the nearest level are thus compared on the same tiles.

\subsection{Training Protocol}
\label{sec:protocol}

One fixed recipe per dataset keeps model contrasts paired.
Every model is fine-tuned with the \gbtwo recipe for its dataset \cite{simumba2025geobench2}, without further hyperparameter search.
The recipe is AdamW \cite{loshchilov2019adamw} with weight decay \WeightDecay, a constant learning rate and the \gbtwo batch size.
\lvd has no \gbtwo entry.
Its rate was chosen among one third, one and three times the \sat rate on clean validation labels, which gave \LvdLrMult{} on both datasets.

A rate picked for one encoder only would confound every model contrast.
Every core model therefore also runs at the multiplier it lacks, either the \gbtwo rate or one third of it.
Each model contrast is read at a matched rate, and the contrast at the selected rates is its sensitivity analysis.

Training uses random crops of \CropPx\,px with flips for at most \MaxEpochs{} epochs and stops after \PatienceChecks{} validation checks without improvement.
Early stopping and checkpoint selection use the noisy validation labels, because a project with noisy training labels rarely has a clean validation set.
The clean subset is one fixed random draw of \SubsetPct\,\% of the training tiles shared by all seeds.
In every run the seed $s$ fixes data order, decoder initialization and noise.

\paragraph{Noise Robust Losses:}
Two losses from the noisy-label literature replace the cross-entropy in one further block.
Generalized cross-entropy \cite{zhang2018gce} interpolates between the cross-entropy and the mean absolute error, and soft bootstrapping \cite{reed2015bootstrapping} mixes each label with the model prediction.
Both are applied per pixel at their published settings without tuning\suppnote{ (\ssec{sec:supp-heads})}.

\subsection{Runs and Hardware}
\label{sec:setup}

Of the \hlNRunsTotal{} finished runs, \hlNRunsExcluded{} probe, timing and replicate runs are excluded and \hlNRunsAnalysed{} enter the analysis\suppnote{ (\ssec{sec:supp-hardware})}.
Most cells hold \NSeeds{} seeds. Clean and \Ltwo cells hold eight for every core model on \burn and the DINOv3 pair on \flair.
The runs were spread over several GPU classes\suppnote{ (\stab{tab:supp-hw-blocks})}.
The scheduler aims to keep the clean run and the noisy runs of a seed on one class.
A contrast between two arms may still mix classes, and $\ddagger$ marks such rows in \Cref{tab:verdicts}.

\subsection{Metrics}
\label{sec:metrics}

The primary estimand is a paired drop, so every model contrast cancels the seed and the realization.
All metrics are computed on the clean reference test labels, whose own errors are a limitation (\Cref{sec:limits}).
For model $m$, dataset $d$, condition $c$ and seed $s$ the paired drop in points is
\begin{equation}
\drop_{m,d,c}(s) = \mathrm{mIoU}_{m,d,\mathrm{clean}}(s) - \mathrm{mIoU}_{m,d,c}(s).
\label{eq:drop}
\end{equation}
Retention is the mean of the per-seed ratio over $n$ seeds,
\begin{equation}
\ret_{m,d,c} = \frac{1}{n}\sum_{s=1}^{n} \frac{\mathrm{mIoU}_{m,d,c}(s)}{\mathrm{mIoU}_{m,d,\mathrm{clean}}(s)},
\label{eq:ret}
\end{equation}
which offsets compressed drops near a floor or ceiling.
Boundary \miou \cite{cheng2021boundary} and a memorization rate \cite{zhang2017understanding,arpit2017closer,liu2020elr} are logged as diagnostics\suppnote{ (\ssec{sec:supp-boundary} and \ssec{sec:supp-heads})}.

\subsection{Statistical Analysis}
\label{sec:stats}
\ifdefined\cvprbuild\else%
\providecommand{\hypref}[1]{\ifdefined\cvprbuild\msec{#1}\else\Cref{#1}\fi}
\begin{table*}[t]
\centering\footnotesize\setlength{\tabcolsep}{4pt}
\begin{tabular}{@{}llp{0.47\textwidth}ll@{}}
\toprule
ID & Question & Statement tested & Test & Answered in \\
\midrule
H1 & RQ2 & Misalignment is the most harmful error type & Type ranking by mean drop & \hypref{sec:types} \\
H2 & RQ2 & Omission costs more than boundary errors & Type ranking by mean drop & \hypref{sec:types} \\
H3 & RQ2 & Frozen encoders lose less than fine-tuned encoders & F3, paired by seed & \hypref{sec:budget} \\
H4 & RQ1 & Satellite pretraining loses less than web pretraining & F1 and N6 at one rate & \hypref{sec:rq1} \\
\midrule
H5 & RQ3 & Another decoder head changes the mean drop by less than the margin & Head minus UNet & \hypref{sec:decoders} \\
H6 & RQ2 & F2 at a common learning-rate multiplier is equivalent & F2 at one rate & \hypref{sec:rq2} \\
H7 & RQ3 & A real label costs as much as its matched simulated arm & Real minus simulated & \hypref{sec:ext} \\
H8 & RQ3 & The two Mask2Former objectives lose alike & Matching minus cross-entropy & \ifdefined\cvprbuild\Cref{sec:supp-n7}\else\hypref{sec:decoders}\fi{} \\
H9 & RQ3 & The type ranking at \Lpair{} equals the ranking at \Ltwo & Rank comparison & \hypref{sec:types} \\
\bottomrule
\end{tabular}
\caption{\textbf{Hypotheses and their tests:} Each row gives the research question a hypothesis belongs to, the statement tested, the contrast that tests it and the section with its result. H1 to H4 were fixed before the noisy-label runs and H5 to H9 before the runs of their blocks. F1 to F4 are the confirmatory families, and N6 is an exploratory contrast.}
\label{tab:hypotheses}
\end{table*}
\fi

Every contrast is read with a pre-specified equivalence margin, so a null is never left as no difference.
The analysis plan fixed H1 to H4 before the noisy-label runs.
It will be released with the code.
Readings outside it, among them N6 and the eight-seed F2 row, are exploratory.

Four confirmatory families each test one contrast on both core datasets with a paired $t$ test over seeds.
Holm's correction \cite{holm1979simple} is applied within each family of two tests at $\alpha = \AlphaFam$, and each dataset gets its own verdict.
F1 (RQ1) contrasts the drop of \sat with that of \lvd averaged over the four \Ltwo types at the selected rates.
N6 repeats it at one learning rate. It was added after F1 detected a difference and is exploratory.
F2 (RQ2) contrasts the mean drop of \dofa and \tm with that of \rn.
Its eight-seed row on \burn extends an undetermined five-seed result and is also exploratory.
F3 contrasts frozen with fine-tuned encoders at the selected \Ltwo type over \sat, \dofa and \tm.
F4 contrasts \miou after training on \SubsetPct\,\% of the tiles with clean labels against all tiles with the selected \Ltwo errors.

Every contrast carries two one-sided tests of equivalence (TOST) \cite{schuirmann1987tost,lakens2017equivalence} at $\pm$\TostMargin{} points with a 90\,\% interval.
The margin is the upper end of the pair difference the plan expected.
Equivalence is stated only where both tests pass.
A null elsewhere is reported as undetermined with its minimum detectable effect (MDE) \cite{card2020little} at \MdePower\,\% power and two-sided $\alpha = \MdeAlpha$.

H1 states that misalignment is the most harmful type and H2 that \etype{omission} costs more than \etype{boundary}.
Both rank the types by mean drop over models and seeds.
H3 states that frozen encoders lose less and H4 that satellite pretraining loses less than web pretraining, read from F3 and F1.
Five further hypotheses were fixed before the runs of their blocks\ifdefined\cvprbuild\suppnote{\suppnote{ (\stab{tab:hypotheses})}}\else\ (\Cref{tab:hypotheses})\fi.
H5 states that another decoder head changes the mean \Ltwo drop by less than \TostMargin{} points.
Under H6, F2 at a common learning-rate multiplier is equivalent within the margin.
H7 states that the drop under a real label and the drop under its matched simulated arm agree within the margin.
H8 states the same for the two Mask2Former objectives, and H9 that the type ranking at \Lpair{} equals the ranking at \Ltwo.
N6, F1 and F2 are repeated under every head and on every extension dataset outside Holm.
A verdict that flips between heads or datasets is reported as decoder dependent or dataset dependent.
Counts over datasets include every dataset that runs the contrast, and an undetermined verdict stays in the denominator.
Ablations use one selected \Ltwo type per dataset, \WorstBurn{} on \burn and \WorstFlair{} on \flair. Each is the worst type on seed \CalSeed{} with clean validation labels.

\section{Results}
\label{sec:results}

\begin{figure*}[t]
\centering
\IfFileExists{planned/figs/fig_datasets_heatmap.pdf}{\includegraphics[width=\textwidth]{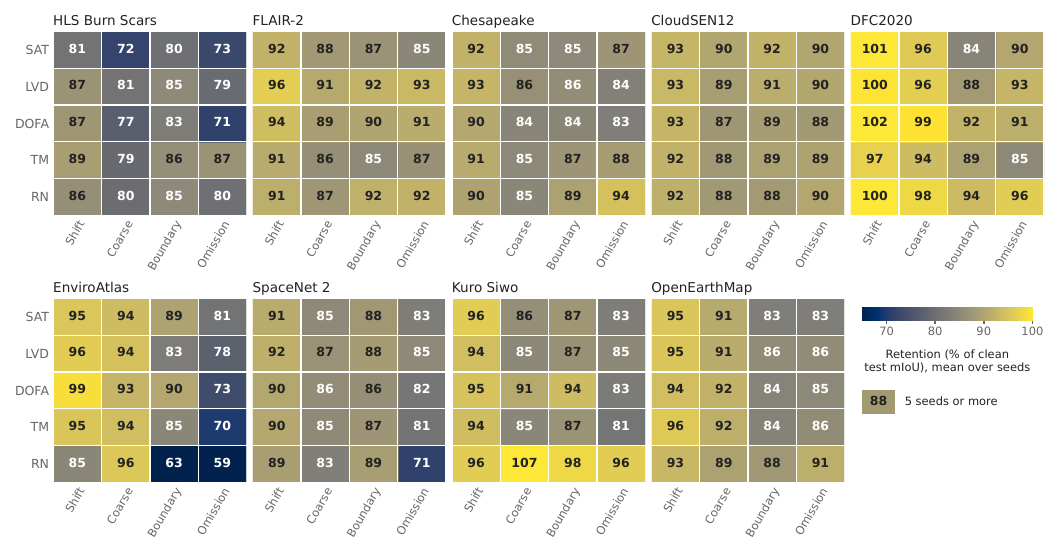}}{\phfig{5cm}{planned/figs/fig\_datasets\_heatmap.pdf from analysis/figures.py}}
\caption{\textbf{Retention at \Ltwo on nine datasets:} Each cell is the mean retention \ret{} of \cref{eq:ret} over the seeds of the cell in percent. Each panel is one dataset, with the five core models as rows and the four error types as columns. Models are abbreviated as SAT (\sat), LVD (\lvd), DOFA (\dofa), TM (\tm) and RN (\rn).}
\label{fig:datasets}
\end{figure*}

\paragraph{Key:}
\ifdefined\cvprbuild\else
N6 (RQ1) tests satellite against web pretraining at a matched learning rate.
F1 tests it at the selected rates.
F2 tests \gfms against \rn, F3 frozen against fine-tuned encoders and F4 the clean subset against all tiles.
H5 to H9 test the decoder head, the common learning rate, the real labels, the Mask2Former objective and the level \Lpair.
\fi
Each contrast carries a TOST equivalence test at $\pm$\TostMargin{} points.
Equivalence is stated only where it passes.
Every number is a mean over the seeds of its cell with \CiLevel\,\% $t$ intervals on the clean reference test labels.
The analysis keeps every run, including the \hlNCollapsedburn{} core-design runs on \burn whose burned-class test IoU fell below \CollapseIoU{}\suppnote{ (\stab{tab:supp-collapsed}, \stab{tab:supp-collapsed-sens})}.
The seven extension datasets run the five core models under the UNet decoder at \Ltwo (\Cref{tab:datasets}, Fig.~\ref{fig:datasets}).

\begin{table*}[t]
\centering\footnotesize\setlength{\tabcolsep}{4pt}
\IfFileExists{planned/generated/table_datasets_main.tex}{%
\providecommand{\gcvdry}[1]{\textcolor{red}{#1}}
\providecommand{\ProvTableDatasetsMain}{Five core models, main arm, means over seeds. Clean is the clean test mIoU averaged over the models, Drop the mean paired drop over the models and the four \Ltwo types in points and Worst type the type with the largest mean \Ltwo drop. N6 is the verdict of the matched-rate contrast of \sat (SAT) against \lvd (LVD) and F2 the verdict of \dofa and \tm (GFMs) against \rn (RN), both on the drop at $\pm$\TostMargin{} points. Real is the drop under the real label of the dataset and a dash a dataset without one. Red is a value of a block that is still running or has not run. Analysis of the finished runs.}
\begin{tabular}{@{}l@{\hspace{14pt}}c@{\hspace{14pt}}c@{\hspace{14pt}}l@{\hspace{14pt}}l@{\hspace{14pt}}l@{\hspace{14pt}}c@{}}
\toprule
Dataset & Clean & Drop & Worst type & N6 & F2 & Real \\
\midrule
\burn~\cite{jakubik2023prithvi,simumba2025geobench2} & 83.8 & 15.4 & \etype{Coarse} & Equivalent & Equivalent & -- \\
\flair~\cite{garioud2023flair2,simumba2025geobench2} & 47.1 & 4.7 & \etype{Coarse} & Equivalent & Equivalent & -- \\
\midrule
Chesapeake~\cite{robinson2019large} & 81.6 & 10.2 & \etype{Coarse} & Equivalent & RN loses less & 21.4 \\
CloudSEN12~\cite{aybar2022cloudsen12} & 67.1 & 6.7 & \etype{Coarse} & Equivalent & Equivalent & 32.4 \\
DFC2020~\cite{schmitt2019sen12ms,schmitt2020weakly} & 39.6 & 2.3 & \etype{Boundary} & Equivalent & Equivalent & 4.8 \\
EnviroAtlas~\cite{rolf2022resolving} & 54.2 & 7.7 & \etype{Omission} & Undetermined & GFMs lose less & 24.6 \\
SpaceNet 2~\cite{vanetten2018spacenet} & 84.2 & 11.8 & \etype{Omission} & Equivalent & GFMs lose less & -- \\
Kuro Siwo~\cite{bountos2024kurosiwo} & 51.4 & 5.5 & \etype{Omission} & Equivalent & RN loses less & 0.8 \\
OpenEarthMap~\cite{xia2023openearthmap} & 59.7 & 6.5 & \etype{Boundary} & Equivalent & Equivalent & -- \\
\bottomrule
\end{tabular}
}{\tabletbd[3cm]{planned/generated/table\_datasets\_main.tex: run the analysis}}
\caption{\textbf{Results per dataset:} Clean is the clean test \miou in points, averaged over the five core models, \sat and \lvd~\cite{simeoni2025dinov3}, \dofa~\cite{xiong2024dofa}, \tm~\cite{jakubik2025terramind} and \rn~\cite{he2016resnet}. Drop is the mean paired drop over the models and the four \Ltwo types. Worst type is the type with the largest mean \Ltwo drop. N6 and F2 are the verdicts of the matched-rate DINOv3 contrast and of multispectral \gfms against \rn (RN) at $\pm$\TostMargin{} points. The \burn F2 verdict is the exploratory eight-seed reading. The Kuro Siwo F2 verdict is a floor effect of \rn\suppnote{ (\ssec{sec:supp-ext})}. Real is the drop under the real label of the dataset, and a dash marks a dataset without one.}
\label{tab:datasets}
\end{table*}

\begin{table*}[t]
\centering\fontsize{8}{9}\selectfont\setlength{\tabcolsep}{3pt}%
\IfFileExists{generated/table_verdicts.tex}{\begin{adjustbox}{max width=\textwidth}%
\providecommand{\gcvdry}[1]{\textcolor{red}{#1}}
\begin{tabular}{@{}>{\raggedright\arraybackslash}p{0.125\textwidth}>{\raggedright\arraybackslash}p{0.365\textwidth}llll@{}}
\toprule
Test & Contrast & Dataset & Estimate [95\,\% CI] & Holm $p$ & Verdict \\
\midrule
RQ1, N6 (matched rate) & Drop \sat at the \lvd rate $-$ \lvd, four \Ltwo types$^{\ddagger}$ & \begin{tabular}[t]{@{}l@{}}\burn \\ {}\flair \\ {}\end{tabular} & \begin{tabular}[t]{@{}l@{}}$-$0.2 [$-$3.7, 3.3] \\ {}1.5 [0.6, 2.5] \\ {}\end{tabular} & \begin{tabular}[t]{@{}l@{}}-- \\ {}-- \\ {}\end{tabular} & \begin{tabular}[t]{@{}l@{}}Equivalent within $\pm$\TostMargin \\ {}Equivalent within $\pm$\TostMargin, \\ {}\lvd loses less\end{tabular} \\
\addlinespace[1.5pt]
RQ1, N6 TOST & Equivalence of N6 within $\pm$\TostMargin{} points, 90\,\% interval & \begin{tabular}[t]{@{}l@{}}\burn \\ {}\flair\end{tabular} & \begin{tabular}[t]{@{}l@{}}[$-$2.9, 2.5] \\ {}[0.8, 2.2]\end{tabular} & \begin{tabular}[t]{@{}l@{}}-- \\ {}--\end{tabular} & \begin{tabular}[t]{@{}l@{}}Equivalent \\ {}Equivalent\end{tabular} \\
\addlinespace[1.5pt]
RQ1, F1 (selected rates, H4) & Drop \sat $-$ \lvd, four \Ltwo types & \begin{tabular}[t]{@{}l@{}}\burn \\ {}\flair\end{tabular} & \begin{tabular}[t]{@{}l@{}}4.3 [1.3, 7.3] \\ {}2.8 [1.8, 3.8]\end{tabular} & \begin{tabular}[t]{@{}l@{}}0.011 \\ {}$<$0.001\end{tabular} & \begin{tabular}[t]{@{}l@{}}\lvd loses less \\ {}\lvd loses less\end{tabular} \\
\addlinespace[1.5pt]
RQ1, F1 TOST & Equivalence of F1 within $\pm$\TostMargin{} points, 90\,\% interval & \begin{tabular}[t]{@{}l@{}}\burn \\ {}\flair\end{tabular} & \begin{tabular}[t]{@{}l@{}}[1.9, 6.7] \\ {}[2.0, 3.6]\end{tabular} & \begin{tabular}[t]{@{}l@{}}-- \\ {}--\end{tabular} & \begin{tabular}[t]{@{}l@{}}Not shown \\ {}Not shown\end{tabular} \\
\addlinespace[1.5pt]
H4 & Retention \sat $-$ \lvd, four \Ltwo types, points & \begin{tabular}[t]{@{}l@{}}\burn \\ {}\flair\end{tabular} & \begin{tabular}[t]{@{}l@{}}$-$6.6 [$-$9.8, $-$3.4] \\ {}$-$5.2 [$-$7.1, $-$3.3]\end{tabular} & \begin{tabular}[t]{@{}l@{}}-- \\ {}--\end{tabular} & \begin{tabular}[t]{@{}l@{}}Not supported \\ {}Not supported\end{tabular} \\
\addlinespace[1.5pt]
RQ2, F2 & Drop mean of \dofa and \tm $-$ \rn, four \Ltwo types$^{\ddagger}$ & \begin{tabular}[t]{@{}l@{}}\burn \\ {}\flair\end{tabular} & \begin{tabular}[t]{@{}l@{}}0.0 [$-$2.8, 2.7]$^{\S}$ \\ {}0.6 [$-$2.0, 3.1]\end{tabular} & \begin{tabular}[t]{@{}l@{}}1.000 \\ {}1.000\end{tabular} & \begin{tabular}[t]{@{}l@{}}Equivalent within $\pm$\TostMargin \\ {}Equivalent within $\pm$\TostMargin\end{tabular} \\
\addlinespace[1.5pt]
RQ2, F2 TOST & Equivalence of F2 within $\pm$\TostMargin{} points, 90\,\% interval & \begin{tabular}[t]{@{}l@{}}\burn \\ {}\flair\end{tabular} & \begin{tabular}[t]{@{}l@{}}[$-$2.2, 2.1]$^{\S}$ \\ {}[$-$1.4, 2.5]\end{tabular} & \begin{tabular}[t]{@{}l@{}}-- \\ {}--\end{tabular} & \begin{tabular}[t]{@{}l@{}}Equivalent \\ {}Equivalent\end{tabular} \\
\addlinespace[1.5pt]
H1 & Drop of the type nearest \etype{shift} $-$ \etype{shift} at \Ltwo (\etype{boundary} on \burn, \etype{omission} on \flair) & \begin{tabular}[t]{@{}l@{}}\burn \\ {}\flair\end{tabular} & \begin{tabular}[t]{@{}l@{}}1.7 [0.6, 2.8] \\ {}1.9 [1.2, 2.5]\end{tabular} & \begin{tabular}[t]{@{}l@{}}-- \\ {}--\end{tabular} & \begin{tabular}[t]{@{}l@{}}Not supported \\ {}Not supported\end{tabular} \\
\addlinespace[1.5pt]
H2 & Drop \etype{omission} $-$ \etype{boundary} at \Ltwo & \begin{tabular}[t]{@{}l@{}}\burn \\ {}\flair \\ {}\end{tabular} & \begin{tabular}[t]{@{}l@{}}4.8 [2.6, 7.1] \\ {}0.0 [$-$0.7, 0.6] / \\ {}0.7 [0.1, 1.3]\end{tabular} & \begin{tabular}[t]{@{}l@{}}-- \\ {}-- \\ {}\end{tabular} & \begin{tabular}[t]{@{}l@{}}Supported (model, seed) \\ {}Undetermined$^{\dagger}$ \\ {}\end{tabular} \\
\addlinespace[1.5pt]
H3, F3 & Drop frozen $-$ fine-tuned encoder, selected \Ltwo type & \begin{tabular}[t]{@{}l@{}}\burn \\ {}\flair \\ {}\end{tabular} & \begin{tabular}[t]{@{}l@{}}$-$5.4 [$-$8.3, $-$2.4] \\ {}$-$2.8 [$-$3.8, $-$1.9] / \\ {}$-$2.8 [$-$3.8, $-$1.9]\end{tabular} & \begin{tabular}[t]{@{}l@{}}0.007 \\ {}0.002 / \\ {}0.002\end{tabular} & \begin{tabular}[t]{@{}l@{}}Supported \\ {}Supported$^{\dagger}$ \\ {}\end{tabular} \\
\addlinespace[1.5pt]
RQ2, F4 & \miou \SubsetPct\,\% clean tiles $-$ all tiles, selected \Ltwo type & \begin{tabular}[t]{@{}l@{}}\burn \\ {}\flair \\ {}\end{tabular} & \begin{tabular}[t]{@{}l@{}}14.4 [12.0, 16.8] \\ {}$-$1.2 [$-$2.3, $-$0.1] / \\ {}$-$0.6 [$-$2.0, 0.9]\end{tabular} & \begin{tabular}[t]{@{}l@{}}$<$0.001 \\ {}0.036 / \\ {}0.353\end{tabular} & \begin{tabular}[t]{@{}l@{}}Clean subset wins \\ {}Undetermined$^{\dagger}$ \\ {}\end{tabular} \\
\bottomrule
\end{tabular}
\end{adjustbox}}{\tabletbd[4.5cm]{generated/table\_verdicts.tex: run figs/make\_verdict\_table.py}}
\caption{\ifdefined\cvprbuild \textbf{Verdicts of the pre-specified tests and hypotheses:} Estimate in \miou points with its \CiLevel\,\% paired $t$ interval (CI). A positive drop contrast means the first arm loses more. TOST rows give the 90\,\% equivalence interval at $\pm$\TostMargin{} points, and MDE is the minimum detectable effect. N6 and the $\S$ rows are exploratory. $\dagger$ marks \flair omission rows printed raw / rescaled\suppnote{ (\stab{tab:supp-omadj})}, read as undetermined where the two disagree, a rule added after the runs. $\ddagger$ marks arms that ran on different GPU classes. A dash marks a missing quantity. \Cref{sec:stats} defines the families and the verdict words. \else \textbf{Verdicts of the pre-specified tests and hypotheses:} Estimate in \miou points with its unadjusted \CiLevel\,\% paired $t$ interval (CI) over its seeds. N6 is the RQ1 contrast at a matched learning rate, an exploratory reading added after the registered F1 at the selected rates. F1 to F4 are the confirmatory families with Holm-adjusted $p$ within each family of two tests. A TOST row gives the 90\,\% interval of the equivalence test at $\pm$\TostMargin{} points, and Equivalent means that it lies within the margin. A detected difference inside the margin is printed with both readings. Undetermined means that the interval includes zero and exceeds the margin. MDE is the minimum detectable effect in points at \MdePower\,\% power and $\alpha = \MdeAlpha$. H1 and H2 are read from the type rankings, H3 from F3 and H4 from the F1 retention. A positive drop contrast means the first arm loses more. $\dagger$ marks \flair omission rows printed as raw / rescaled to the target\suppnote{ (\stab{tab:supp-omadj})}. They read undetermined where the two disagree, a rule added after the runs. $\S$ marks the exploratory eight-seed F2 rows. $\ddagger$ marks a contrast whose two arms ran on different GPU classes for some or all seeds. \stab{tab:hypotheses} names the sections of H5 to H9. A dash marks a quantity the row does not have.\fi}
\label{tab:verdicts}
\end{table*}
\ifdefined\cvprbuild\else%
\begin{figure*}[t]
\centering
\IfFileExists{planned/figs/fig_an_verdict_forest.pdf}{\includegraphics[width=\textwidth]{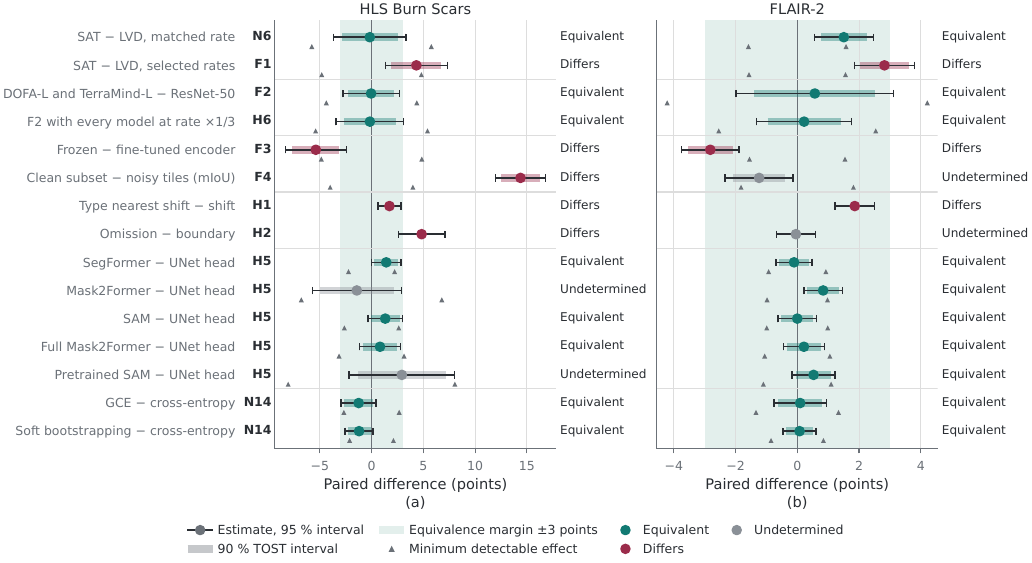}}{\phfig{4cm}{planned/figs/fig\_an\_verdict\_forest.pdf from figs/make\_analysis\_diagrams.py}}
\caption{\textbf{Pre-specified contrasts against the equivalence margin:} Panel (a) shows \burn and panel (b) \flair. The H5, H6 and N14 rows are the decoder, common-rate and loss contrasts of \Cref{tab:supp-verdicts-post}. The other rows are contrasts of \mtab{tab:verdicts}. SAT is \sat{}, LVD is \lvd{} and GCE is generalized cross-entropy. The dot is the estimate, the thin bar its \CiLevel\,\% interval and the thick bar its TOST interval. The shaded band is the margin of $\pm$\TostMargin{} points, and the small triangles mark the minimum detectable effect on both sides of zero. The word at the right is the reading. A contrast is equivalent when its TOST interval lies inside the band, and otherwise it differs when its \CiLevel\,\% interval excludes zero. On \flair the H2, F3 and F4 rows rest on \etype{omission} realizations, and they read undetermined where their raw and rescaled readings disagree. The rows H1 and H2 carry no TOST interval. F4 is a difference in test \miou, and every other row is a difference in drop.}
\label{fig:ax-an-verdict-forest}
\end{figure*}
\fi

\subsection{Satellite and Web Pretraining}
\label{sec:rq1}

Satellite pretraining does not make fine-tuning more robust than web pretraining.
At the matched learning rate N6 reads equivalent within $\pm$\TostMargin{} points on \hlDsNSixEquiv{} of the \hlDsNSixRead{} datasets that carry it (\Cref{tab:datasets}, \Cref{tab:verdicts}).
That margin is \hlMsShareflair\,\% of the mean \Ltwo drop on \flair, so we also read every verdict at \hlMsFractionPct\,\% of the drop.
At that scaled margin N6 stays equivalent on \hlMsNSixEquivScaled{} of \hlMsNSixRead{} datasets\suppnote{ (\stab{tab:supp-margin})}.
On the sub-meter optical datasets, where the satellite encoder is in domain, that count is \hlMsNSixEquivScaledSubm{} of \hlMsNSixReadSubm.
\ifdefined\cvprbuild\else\Cref{fig:ax-an-verdict-forest} draws the contrasts of \Cref{tab:verdicts} against the margin.
\fi
On \burn the contrast centers on zero with a 90\,\% interval of [\hlMlrDropTostCILoburn, \hlMlrDropTostCIHiburn].
On \flair \sat still loses \hlMlrDropDiffflair{} [\hlMlrDropDiffCILoflair, \hlMlrDropDiffCIHiflair] points more, a detected difference inside the margin.
At the selected rates \sat loses more on both datasets (F1), the opposite of H4.
Equivalence is not shown there.
\ifdefined\cvprbuild\else
The F1 gap on \flair halves at the matched rate, so part of it is a learning-rate effect.
\fi
At the \gbtwo rate of \sat the pair reads \vd{\hlMraVerdictFOneburn} on \burn and \vd{\hlMraVerdictFOneflair} on \flair. RQ1 is thus read at both multipliers\suppnote{ (\ssec{sec:supp-n8})}.

Under the SegFormer, Mask2Former and SAM heads N6 reads \vd{\hlDecVerdictNSixburn}, \vd{\hlDecVerdictNSixMtwoFburn} and \vd{\hlDecVerdictNSixSamburn} on \burn.
On \flair it reads \vd{\hlDecVerdictNSixflair}, \vd{\hlDecVerdictNSixMtwoFflair} and \vd{\hlDecVerdictNSixSamflair}\suppnote{ (\ssec{sec:supp-n7})}.
On SpaceNet 2 N6 is \hlExtNSixSnTwoEst{} [\hlExtNSixSnTwoCILo, \hlExtNSixSnTwoCIHi] points and reads \vd{\hlExtVerdictNSixSnTwo}.

\subsection{Foundation Models and a CNN Baseline}
\label{sec:rq2}

The multispectral \gfms do not consistently lose less than the ImageNet CNN.
The F2 contrast reads equivalent within $\pm$\TostMargin{} points on \hlDsFTwoEquiv{} of the \hlDsFTwoRead{} datasets (\Cref{tab:datasets}).
At the scaled margin it stays equivalent on \hlMsFTwoEquivScaled{} of \hlMsFTwoRead{}\suppnote{ (\stab{tab:supp-margin})}.
On \flair the multispectral \gfms and \rn are equivalent within the margin (\Cref{tab:verdicts}).
On \burn F2 at the five registered seeds reads \vd{\hlFTwoExtFiveVerdictburn}, with a 90\,\% interval of [\hlFTwoExtFiveburnTostCILo, \hlFTwoExtFiveburnTostCIHi].
Seeds 6 to 8 were registered next and added before the contrast was read again.
With eight seeds per model the \burn contrast is \hlFTwoExtburnEst{} [\hlFTwoExtburnCILo, \hlFTwoExtburnCIHi] points and reads \vd{\hlFTwoExtVerdictburn}.
\ifdefined\cvprbuild\else
\Cref{tab:verdicts} prints the eight-seed row for \burn.
\fi
At the common $\times$1/3 multiplier F2 reads \vd{\hlMraVerdictburn} on \burn and \vd{\hlMraVerdictflair} on \flair (H6).
\ifdefined\cvprbuild\else
The contrast there is \hlMraFTwoThirdburnEst{} [\hlMraFTwoThirdburnCILo, \hlMraFTwoThirdburnCIHi] points on \burn and \hlMraFTwoThirdflairEst{} [\hlMraFTwoThirdflairCILo, \hlMraFTwoThirdflairCIHi] on \flair.
\fi
Under the SegFormer, Mask2Former and SAM heads F2 reads \vd{\hlDecVerdictFTwoburn}, \vd{\hlDecVerdictFTwoMtwoFburn} and \vd{\hlDecVerdictFTwoSamburn} on \burn.
On \flair it reads \vd{\hlDecVerdictFTwoflair}, \vd{\hlDecVerdictFTwoMtwoFflair} and \vd{\hlDecVerdictFTwoSamflair}.

\ifdefined\cvprbuild\else
\begin{figure*}[t]
\centering
\IfFileExists{planned/figs/fig_an_model_agreement.pdf}{\includegraphics[width=\textwidth]{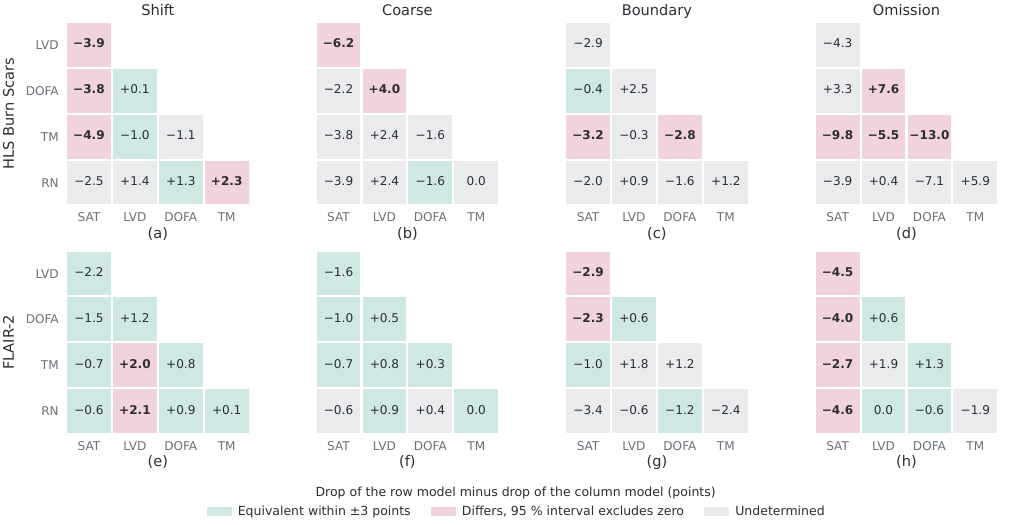}}{\phfig{4cm}{planned/figs/fig\_an\_model\_agreement.pdf from figs/make\_analysis\_diagrams.py}}
\caption{\textbf{Pairwise drop differences between the core models at \Ltwo:} Each cell is the paired drop of the row model minus that of the column model, in points, over the seeds both models have. The top row is \burn and the bottom row \flair, with one panel per error type. A green cell is equivalent, since its TOST interval lies within $\pm$\TostMargin{} points. A red cell with a bold value differs, since its \CiLevel\,\% interval excludes zero and its TOST interval leaves the margin. A gray cell is undetermined. SAT is \sat, LVD is \lvd, DOFA is \dofa, TM is \tm and RN is \rn.}
\label{fig:ax-an-model-agreement}
\end{figure*}

\Cref{fig:ax-an-model-agreement} reads every pair of core models per error type.
Most pairs on \flair are equivalent within the margin, and most pairs on \burn stay undetermined.
\fi

\subsection{Error Types and Their Severity}
\label{sec:types}

\ifdefined\cvprbuild\else
\begin{figure}[t]
\centering
\includegraphics[width=\linewidth]{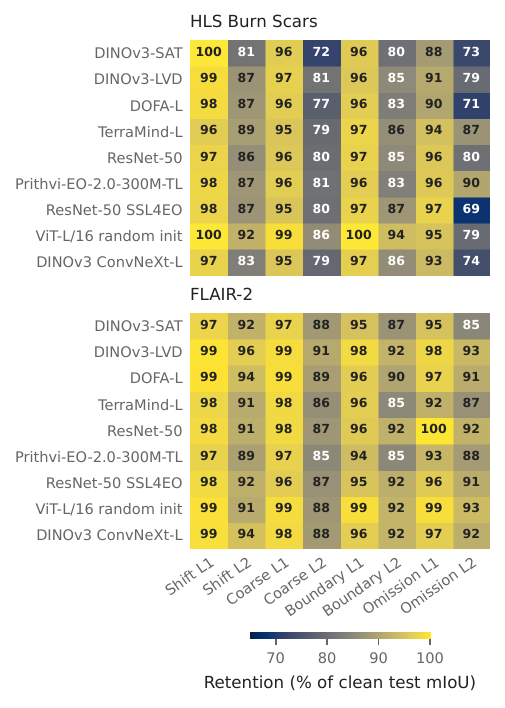}
\caption{\textbf{Retention by encoder, error type and level on the core datasets:} Each cell is the mean retention \ret{} of \cref{eq:ret} over the seeds of the cell in percent. The top panel shows \burn and the bottom panel \flair. Rows are the nine encoders and columns the four error types at \Lone and \Ltwo under the UNet decoder.}
\label{fig:heatmap}
\end{figure}
\fi

Labels with random offsets are the cheapest error type at equal label quality.
\Cref{fig:datasets} shows the \Ltwo retention of the five core models on the nine datasets that run all four types.
At equal label \miou \etype{shift} has the smallest mean drop on \hlDsShiftLeast{} of these \hlDsTypeRead{} datasets.
On \hlDsOmissionMost{} of them \etype{omission} costs the most and on \hlDsCoarseMost{} \etype{coarse} does (\Cref{tab:datasets}).
\ifdefined\cvprbuild\else%
\begin{figure*}[t]
\centering
\IfFileExists{planned/figs/fig_an_seed_spread.pdf}{\includegraphics[width=\textwidth]{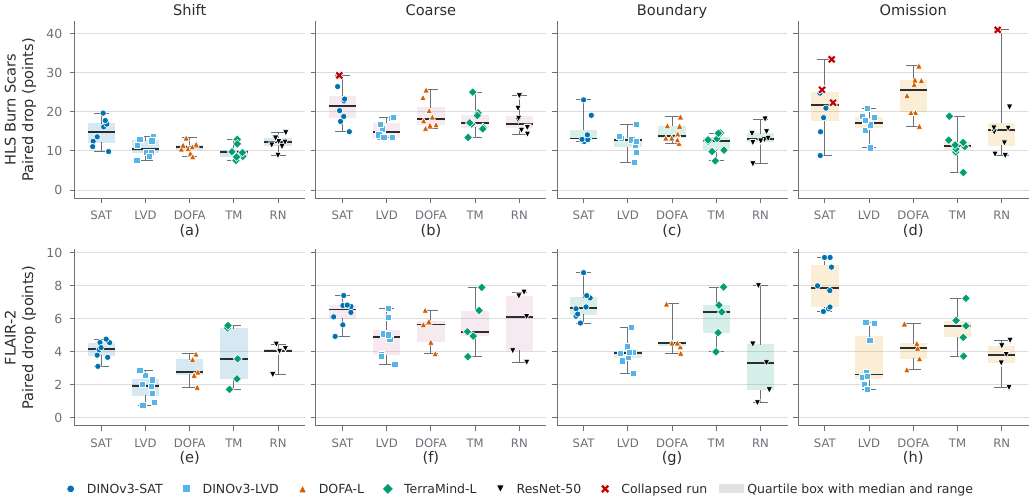}}{\phfig{4cm}{planned/figs/fig\_an\_seed\_spread.pdf from figs/make\_analysis\_diagrams.py}}
\caption{\textbf{Paired drops of single runs at \Ltwo:} Each marker is the paired drop of one seed, per core model and error type. Panels (a) to (d) show \burn and panels (e) to (h) show \flair. The box spans the quartiles, the line inside it is the median and the whiskers reach the extreme runs. A red cross marks a collapsed run on \burn. SAT is \sat, LVD is \lvd, DOFA is \dofa, TM is \tm and RN is \rn.}
\label{fig:ax-an-seed-spread}
\end{figure*}

\Cref{fig:heatmap} adds \Lone and all nine encoders on the core datasets.
\Cref{fig:ax-an-seed-spread} shows the paired drop of every seed, with the collapsed \burn runs marked.
\fi
On \burn \etype{\hlTypeRankburnOne} and \etype{\hlTypeRankburnTwo} cost the most and are not separated.
On \flair \etype{omission} ranks third raw and first after rescaling to the target\suppnote{ (\stab{tab:supp-omadj})}.
The \etype{shift} type ranks last at \Lone and \Ltwo on both core datasets, so H1 (misalignment most harmful) is not supported\suppnote{ (\ssec{sec:supp-grid})}.
H2 (omission worse than boundary) holds on \burn at (model, seed) units and is undetermined on \flair (\Cref{tab:verdicts}).
\ifdefined\cvprbuild\else
At a common flip rate \etype{omission} ranks first and \etype{shift} last on both datasets\suppnote{ (\ssec{sec:supp-fliprate})}.
\fi

\ifdefined\cvprbuild
At the pair level \Lpair{} H9 reads \vd{\hlLrRankVerdictburn} on \burn and \vd{\hlLrRankVerdictflair} on \flair\suppnote{ (\ssec{sec:supp-lr})}.
\else
At the pair level \Lpair{} the mean drop over models and types is \hlLrMeanDropburn{} points on \burn and \hlLrMeanDropflair{} on \flair.
The ranking at \Lpair{} puts \etype{\hlLrTypeRankburnOne} first on \burn and \etype{\hlLrTypeRankflairOne} first on \flair. H9 reads \vd{\hlLrRankVerdictburn} and \vd{\hlLrRankVerdictflair}\suppnote{ (\ssec{sec:supp-lr})}.
\fi

\subsection{Decoder Heads}
\label{sec:decoders}

\ifdefined\cvprbuild\else%
\begin{table*}[t]
\centering\footnotesize\setlength{\tabcolsep}{4pt}
\IfFileExists{planned/generated/table_decoders_main.tex}{\begin{adjustbox}{max width=\textwidth}%
\providecommand{\gcvdry}[1]{\textcolor{red}{#1}}
\providecommand{\ProvTableDecodersMain}{Units per contrast 25 to 25 (model, seed) pairs. Clean test mIoU and the mean paired drop over the four \Ltwo types in points, averaged over the five core models on seeds 1 to 5. The contrast is the head's drop minus the UNet drop paired by model and seed with the 95\,\% $t$ interval, and the verdict is its equivalence test within $\pm$\TostMargin{} points. Red cells are not run yet or rest on a block that is still running. Analysis of the finished runs.}
\begin{tabular}{@{}l@{\hspace{9pt}}c@{\hspace{5pt}}c@{\hspace{6pt}}c@{\hspace{6pt}}l@{\hspace{9pt}}c@{\hspace{5pt}}c@{\hspace{6pt}}c@{\hspace{6pt}}l@{}}
\toprule
& \multicolumn{4}{c}{\burn} & \multicolumn{4}{c}{\flair} \\
\cmidrule(lr){2-5}\cmidrule(l){6-9}
Head & Clean & Drop & Head $-$ UNet & Verdict & Clean & Drop & Head $-$ UNet & Verdict \\
\midrule
UNet & 83.5 & 15.7 & -- & -- & 47.1 & 4.8 & -- & -- \\
SegFormer (N7) & 85.2 & 17.1 & 1.4 [0.0, 2.8] & Equivalent within $\pm$\TostMargin & 47.2 & 4.7 & $-$0.1 [$-$0.7, 0.5] & Equivalent within $\pm$\TostMargin \\
Mask2Former (N9) & 76.3 & 14.3 & $-$1.4 [$-$5.7, 2.9] & Undetermined, MDE 6.8 & 46.9 & 5.6 & 0.8 [0.2, 1.5] & Equivalent within $\pm$\TostMargin \\
Full Mask2Former (N11) & 82.3 & 16.5 & 0.8 [$-$1.2, 2.8] & Equivalent within $\pm$\TostMargin & 46.5 & 5.0 & 0.2 [$-$0.5, 0.9] & Equivalent within $\pm$\TostMargin \\
SAM (N10) & 83.9 & 17.0 & 1.3 [$-$0.3, 3.0] & Equivalent within $\pm$\TostMargin & 44.8 & 4.8 & 0.0 [$-$0.6, 0.6] & Equivalent within $\pm$\TostMargin \\
Pretrained SAM (N15) & 78.1 & 18.6 & 2.9 [$-$2.2, 8.0] & Undetermined, MDE 8.0 & 42.9 & 5.3 & 0.5 [$-$0.2, 1.2] & Equivalent within $\pm$\TostMargin \\
\bottomrule
\end{tabular}
\end{adjustbox}}{\tabletbd[2.5cm]{planned/generated/table\_decoders\_main.tex: run the analysis}}
\caption{\textbf{Main grid under the decoder heads:} Clean test \miou and mean paired drop over the four \Ltwo types in points, averaged over the five core models on seeds 1 to 5. The contrast is the drop under the head minus the drop under the UNet decoder over (model, seed) units. It is printed with its \CiLevel\,\% interval and the H5 verdict of the $\pm$\TostMargin-point equivalence test without the direction, which the interval shows. MDE is the minimum detectable effect, and a dash marks the UNet row, which has no contrast. \ifdefined\cvprbuild\Cref{tab:n7}\else\stab{tab:n7}\fi{} gives the values per model.}
\label{tab:decoders}
\end{table*}
\fi

The decoder contrast is the mean \Ltwo drop under a head minus the drop under the UNet decoder (H5\ifdefined\cvprbuild\suppnote{, \stab{tab:supp-verdicts-post}}\else, \Cref{tab:decoders}\fi).
\ifdefined\cvprbuild
Under the SegFormer head it reads \vd{\hlDecVerdictburn} on \burn and \vd{\hlDecVerdictflair} on \flair.
Under the Mask2Former head it reads \vd{\hlDecVerdictMtwoFburn} and \vd{\hlDecVerdictMtwoFflair}.
Under the SAM head it reads \vd{\hlDecVerdictSamburn} and \vd{\hlDecVerdictSamflair}.
\else
Under the SegFormer head it is \hlDecInvburnEst{} [\hlDecInvburnCILo, \hlDecInvburnCIHi] points on \burn and \hlDecInvflairEst{} [\hlDecInvflairCILo, \hlDecInvflairCIHi] on \flair.
It reads \vd{\hlDecVerdictburn} and \vd{\hlDecVerdictflair}.
Under the Mask2Former head it is \hlDecInvMtwoFburnEst{} [\hlDecInvMtwoFburnCILo, \hlDecInvMtwoFburnCIHi] and \hlDecInvMtwoFflairEst{} [\hlDecInvMtwoFflairCILo, \hlDecInvMtwoFflairCIHi] points.
It reads \vd{\hlDecVerdictMtwoFburn} and \vd{\hlDecVerdictMtwoFflair}.
Under the SAM head trained from scratch it is \hlDecInvSamburnEst{} [\hlDecInvSamburnCILo, \hlDecInvSamburnCIHi] and \hlDecInvSamflairEst{} [\hlDecInvSamflairCILo, \hlDecInvSamflairCIHi] points.
It reads \vd{\hlDecVerdictSamburn} and \vd{\hlDecVerdictSamflair}.
\fi

\ifdefined\cvprbuild
\else
The matching objective changes the drop of the Mask2Former head by \hlDecFullVsLiteburnEst{} [\hlDecFullVsLiteburnCILo, \hlDecFullVsLiteburnCIHi] points on \burn.
On \flair the change is \hlDecFullVsLiteflairEst{} [\hlDecFullVsLiteflairCILo, \hlDecFullVsLiteflairCIHi] points, and H8 reads \vd{\hlDecVerdictFullVsLiteburn} and \vd{\hlDecVerdictFullVsLiteflair}.
The released weights change the drop of the SAM head by \hlDecPretrainedVsScratchburnEst{} [\hlDecPretrainedVsScratchburnCILo, \hlDecPretrainedVsScratchburnCIHi] points on \burn.
On \flair the change is \hlDecPretrainedVsScratchflairEst{} [\hlDecPretrainedVsScratchflairCILo, \hlDecPretrainedVsScratchflairCIHi] points\suppnote{ (\ssec{sec:supp-n7})}.
\fi

\subsection{Real Labels}
\label{sec:ext}

H7 compares each real label with its matched simulated arm\suppnote{ (\sfig{fig:ext-real})}.
The real label costs more than the matched arm on \hlDsRealMore{} of the \hlDsRealRead{} pairs with a matched arm.
The matched arm sits at the nearest calibrated level, so the two labels differ in label \miou.
Read at the measured label \miou, the Chesapeake difference is \hlRealInterpChesEst{} [\hlRealInterpChesCILo, \hlRealInterpChesCIHi] points and reads \vd{\hlRealInterpChesVerdict}\suppnote{ (\stab{tab:supp-real-interp})}.
On Chesapeake the NLCD label agrees with the 1\,m label at a label \miou of \RealLabelMiouChes.
Training on it costs \hlRealDropChes{} points against \hlRealMatchedDropChes{} under the matched \etype{\hlRealMatchedTypeChes} type, and H7 reads \vd{\hlRealVerdictChes}.
The same pairing on EnviroAtlas sits at \RealLabelMiouEnva{} and costs \hlRealDropEnva{} points, and H7 reads \vd{\hlRealVerdictEnva}.
On CloudSEN12 the s2cloudless mask sits at \RealLabelMiouCsTwelve{} and costs \hlRealDropCsTwelve{} points, and H7 reads \vd{\hlRealVerdictCsTwelve}.
On DFC2020 the MODIS label sits at \RealLabelMiouDfc{} and costs \hlRealDropDfc{} points, and H7 reads \vd{\hlRealVerdictDfc}.
The CEMS polygons of Kuro Siwo sit at \RealLabelMiouKuro{} and cost \hlRealDropKuro{} points, and H7 reads \vd{\hlRealVerdictKuro}.
\suppnote{The values per model and the intervals behind each verdict follow in \stab{tab:ext-results} and \ssec{sec:supp-ext}.}

\subsection{All Nine Encoders}
\label{sec:encoders}

\begin{table}[t]
\centering\footnotesize\setlength{\tabcolsep}{3pt}
\IfFileExists{planned/generated/table_all_models.tex}{%
\providecommand{\gcvdry}[1]{\textcolor{red}{#1}}
\providecommand{\ProvTableAllModels}{Seeds per cell 5 to 8. Clean is the clean test mIoU, Drop the mean paired drop over the four \Ltwo types in points (every noisy run paired with the clean run of its seed) and Subset the clean test mIoU after training on the 25\,\% of the training tiles with clean labels, all means over seeds. The core models are the main grid (A1, subset arm A4), the extension encoders the blocks S3 to S6, \vitr with random initialisation. Bold marks the largest clean and subset mIoU and the smallest drop per column. A dash is a cell without a run, red a value of a block that is still running. Hardware (GPU classes of the analysed runs): A1, A4 and S3 ran on GB10, RTX 4090 and RTX PRO 6000, S4 on RTX 4090 and RTX PRO 6000, S5 and S6 on RTX PRO 6000. Analysis of the finished runs.}
\begin{tabular}{@{}l@{\hspace{6pt}}c@{\hspace{4pt}}c@{\hspace{4pt}}c@{\hspace{6pt}}c@{\hspace{4pt}}c@{\hspace{4pt}}c@{}}
\toprule
& \multicolumn{3}{c}{\burn} & \multicolumn{3}{c}{\flair} \\
\cmidrule(lr){2-4} \cmidrule(lr){5-7}
Model & Clean & Drop & Subset & Clean & Drop & Subset \\
\midrule
\sat & 77.0 & 18.1 & 68.3 & \textbf{52.4} & 6.3 & 44.6 \\
\lvd & 81.8 & 13.8 & 77.5 & 50.7 & 3.5 & \textbf{46.5} \\
\dofa & 85.5 & 17.3 & 79.9 & 47.5 & 4.3 & 43.1 \\
\tm & 86.3 & 12.7 & 82.0 & 40.5 & 5.2 & 31.3 \\
\rn & 88.3 & 15.1 & 86.4 & 44.5 & 4.2 & 38.9 \\
\midrule
\prithvi & \textbf{88.6} & 13.0 & 86.0 & 39.4 & 5.2 & 32.7 \\
\rnssl & 88.2 & 16.9 & \textbf{87.3} & 42.5 & 4.2 & 36.8 \\
\vitr & 69.0 & \textbf{8.5} & 61.7 & 27.5 & \textbf{2.5} & 21.1 \\
\cnxl & 81.6 & 16.0 & 74.6 & 47.6 & 4.0 & 43.5 \\
\bottomrule
\end{tabular}
}{\tabletbd[4cm]{planned/generated/table\_all\_models.tex: run the analysis}}
\caption{\textbf{Every encoder on the core datasets:} Clean test \miou and mean paired drop over the four \Ltwo types in points per encoder and dataset. Subset is the clean test \miou after training on the \SubsetPct\,\% clean subset. The first block holds the five core models and the second the four extension encoders. Bold marks the largest \miou and the smallest drop of each column. Core-model cells average all seeds of a cell, while the contrasts of \Cref{sec:encoders} pair the first \NSeeds{} seeds.}
\label{tab:all-models}
\end{table}

\Cref{tab:all-models} lists the clean \miou, the mean \Ltwo drop and the clean-subset \miou of every encoder.
The clean-label ranking of the nine encoders predicts their \Ltwo test \miou ranking.
Its Spearman correlation is \hlRankRhoburn{} on \burn and \hlRankRhoflair{} on \flair\suppnote{ (\sfig{fig:supp-rank})}.
\ifdefined\cvprbuild
On point estimates only \vitr and \prithvi lower the mean \Ltwo drop by more than \TostMargin{} points against a control\suppnote{ (\ssec{sec:supp-stretch})}.
The latter does so only on \burn.
\else
Each extension contrast is the \Ltwo drop averaged over the four types and paired by seed.
\prithvi minus \dofa is \hlSThreeDropDOFAburnEst{} [\hlSThreeDropDOFAburnCILo, \hlSThreeDropDOFAburnCIHi] points on \burn and \hlSThreeDropDOFAflairEst{} [\hlSThreeDropDOFAflairCILo, \hlSThreeDropDOFAflairCIHi] on \flair, with mixed GPU classes on \burn.
\rnssl minus \rn is \hlSFourDropRNburnEst{} [\hlSFourDropRNburnCILo, \hlSFourDropRNburnCIHi] and \hlSFourDropRNflairEst{} [\hlSFourDropRNflairCILo, \hlSFourDropRNflairCIHi] points, so a Sentinel-2 pretraining effect is not detected.
\vitr minus \sat is \hlSFiveDropSATburnEst{} [\hlSFiveDropSATburnCILo, \hlSFiveDropSATburnCIHi] and \hlSFiveDropSATflairEst{} [\hlSFiveDropSATflairCILo, \hlSFiveDropSATflairCIHi] points.
The untrained encoder loses fewer points from a far lower clean \miou\suppnote{, as the retention view shows\suppnote{ (\sfig{fig:supp-er})}}.
\cnxl minus \sat is \hlSSixDropSATburnEst{} [\hlSSixDropSATburnCILo, \hlSSixDropSATburnCIHi] and \hlSSixDropSATflairEst{} [\hlSSixDropSATflairCILo, \hlSSixDropSATflairCIHi] points.
\fi
\ifdefined\cvprbuild\else
On \sfloods the Otsu labels cost \hlSOneDropMeansenEst{} [\hlSOneDropMeansenCILo, \hlSOneDropMeansenCIHi] points over the five core models, below the \burn \Ltwo type means\suppnote{ (\stab{tab:supp-sen1floods})}.
\fi

\subsection{Freezing, Labeling Budget and Robust Losses}
\label{sec:budget}

Frozen encoders start lower and lose less, as H3 predicts (\Cref{tab:verdicts}\suppnote{, \stab{tab:supp-frozen}}).
On \flair frozen minus fine-tuned noisy-label \miou runs from \hlVeFrozenSATflairEst{} to \hlVeFrozenTMflairEst{} points and is negative for every model.
The clean subset beats all tiles with errors at the \WorstBurn{} \Ltwo stress test on \burn (F4\suppnote{, \stab{tab:supp-budget}}).
On \flair the contrast rescaled to the target is \hlFFourAdjflairEst{} [\hlFFourAdjflairCILo, \hlFFourAdjflairCIHi] with Holm $p$ \hlFFourAdjflairPHolm, so the \flair half is undetermined.

\ifdefined\cvprbuild\else%
\begin{table}[t]
\centering\footnotesize\setlength{\tabcolsep}{3pt}
\IfFileExists{planned/generated/table_n14.tex}{\begin{adjustbox}{max width=\columnwidth}%
\providecommand{\gcvdry}[1]{\textcolor{red}{#1}}
\providecommand{\ProvTableNFourteen}{Seeds per cell 5 (loss cells). Clean test mIoU and the mean paired drop over the four \Ltwo types in points. CE is cross-entropy, GCE is generalized cross-entropy, Boot is soft bootstrapping. Every noisy run is paired with the clean run of its own loss and seed. The CE columns are the main-grid cells on the seeds a model's loss arms have run. The contrast is the drop under the loss minus the drop under CE, mean $\pm$ 95\,\% $t$ half-width, paired by seed per model and by model and seed in the mean row, so a negative value means the loss reduces the drop. The mean row averages the five models. Models are abbreviated as SAT (\sat), LVD (\lvd), DOFA (\dofa), TM (\tm) and RN (\rn). Red cells are not run yet or rest on a block that is still running. Analysis of the finished runs.}
\begin{tabular}{@{}l@{\hspace{5pt}}c@{\hspace{4pt}}c@{\hspace{4pt}}c@{\hspace{6pt}}c@{\hspace{4pt}}c@{\hspace{4pt}}c@{\hspace{6pt}}c@{\hspace{5pt}}c@{}}
\toprule
& \multicolumn{3}{c}{Clean mIoU} & \multicolumn{3}{c}{Drop at \Ltwo} & \multicolumn{2}{c}{Loss $-$ CE drop} \\
\cmidrule(lr){2-4}\cmidrule(lr){5-7}\cmidrule(l){8-9}
Model & CE & GCE & Boot & CE & GCE & Boot & GCE & Boot \\
\midrule
\multicolumn{9}{@{}l}{\textit{\burn}} \\
SAT & 75.8 & 70.6 & 72.2 & 17.8 & 11.1 & 13.3 & $-$6.6 $\pm$ 6.4 & $-$4.5 $\pm$ 4.5 \\
LVD & 82.0 & 81.3 & 82.0 & 14.1 & 14.1 & 14.1 & 0.0 $\pm$ 1.7 & 0.0 $\pm$ 3.6 \\
DOFA & 85.5 & 86.8 & 86.4 & 17.9 & 17.1 & 17.8 & $-$0.7 $\pm$ 3.5 & $-$0.1 $\pm$ 1.6 \\
TM & 86.2 & 87.2 & 86.3 & 12.9 & 14.0 & 12.9 & 1.0 $\pm$ 2.6 & $-$0.1 $\pm$ 1.5 \\
RN & 88.1 & 88.7 & 88.3 & 15.9 & 16.0 & 14.5 & 0.1 $\pm$ 4.4 & $-$1.4 $\pm$ 5.6 \\
\addlinespace[1.5pt]
Mean & 83.5 & 82.9 & 83.0 & 15.7 & 14.5 & 14.5 & $-$1.2 $\pm$ 1.7 & $-$1.2 $\pm$ 1.3 \\
\midrule
\multicolumn{9}{@{}l}{\textit{\flair}} \\
SAT & 52.2 & 49.9 & 51.7 & 6.5 & 5.5 & 5.9 & $-$1.0 $\pm$ 3.0 & $-$0.6 $\pm$ 0.9 \\
LVD & 50.8 & 50.4 & 51.3 & 3.6 & 4.3 & 4.2 & 0.7 $\pm$ 0.6 & 0.6 $\pm$ 1.6 \\
DOFA & 47.5 & 43.8 & 47.2 & 4.3 & 5.8 & 4.6 & 1.5 $\pm$ 3.3 & 0.2 $\pm$ 1.2 \\
TM & 40.5 & 37.1 & 40.8 & 5.2 & 5.1 & 5.6 & $-$0.1 $\pm$ 2.5 & 0.4 $\pm$ 1.6 \\
RN & 44.5 & 40.0 & 44.3 & 4.2 & 3.6 & 3.9 & $-$0.6 $\pm$ 2.1 & $-$0.3 $\pm$ 2.6 \\
\addlinespace[1.5pt]
Mean & 47.1 & 44.2 & 47.1 & 4.8 & 4.9 & 4.8 & 0.1 $\pm$ 0.8 & 0.1 $\pm$ 0.5 \\
\bottomrule
\end{tabular}
\end{adjustbox}}{\tabletbd[3cm]{planned/generated/table\_n14.tex: run the analysis}}
\caption{\textbf{Noise-robust losses against the cross-entropy:} Clean test \miou and mean paired drop over the four \Ltwo types in points per core model on seeds 1 to 5. The losses are the cross-entropy (CE), generalized cross-entropy (GCE) and soft bootstrapping (Boot). The contrast is the drop under the loss minus the CE drop, paired by seed, with the half-width of its \CiLevel\,\% interval. The top block shows \burn and the bottom block \flair. Models are abbreviated as SAT (\sat), LVD (\lvd), DOFA (\dofa), TM (\tm) and RN (\rn).}
\label{tab:n14}
\end{table}
\fi

\ifdefined\cvprbuild
Against the cross-entropy, generalized cross-entropy reads \vd{\hlLossVerdictGceburn} on \burn and \vd{\hlLossVerdictGceflair} on \flair.
Soft bootstrapping reads \vd{\hlLossVerdictBootburn} and \vd{\hlLossVerdictBootflair}\suppnote{ (\stab{tab:n14})}.
\else
Generalized cross-entropy changes the mean \Ltwo drop by \hlLossInvGceburnEst{} [\hlLossInvGceburnCILo, \hlLossInvGceburnCIHi] points on \burn against the cross-entropy.
On \flair the change is \hlLossInvGceflairEst{} [\hlLossInvGceflairCILo, \hlLossInvGceflairCIHi] points, and the loss reads \vd{\hlLossVerdictGceburn} and \vd{\hlLossVerdictGceflair}.
Soft bootstrapping changes the drop by \hlLossInvBootburnEst{} [\hlLossInvBootburnCILo, \hlLossInvBootburnCIHi] points on \burn and \hlLossInvBootflairEst{} [\hlLossInvBootflairCILo, \hlLossInvBootflairCIHi] on \flair.
It reads \vd{\hlLossVerdictBootburn} and \vd{\hlLossVerdictBootflair} (\Cref{tab:n14}).
\fi

\section{Discussion}
\label{sec:discussion}

\paragraph{Pretraining Data:}
At one learning rate the two DINOv3 encoders degrade alike within the margin on \hlDsNSixEquiv{} of \hlDsNSixRead{} datasets (\Cref{sec:rq1}, \Cref{tab:datasets}).
The web-pretrained one keeps a small lead on \flair.
The pair differs in more than pretraining data (\Cref{sec:limits}), so the choice can rest on the noisy-label \miou each encoder reaches.
\lvd is \hlVeNoisyLVDSATNSixflairEst{} points higher on \flair\suppnote{ (\ssec{sec:supp-stats})}.
\ifdefined\cvprbuild\else
The four extension encoders put that reading in context (\Cref{sec:encoders}).
On point estimates only \vitr{} and \prithvi on \burn lower the drop by more than the margin against a control.
\fi

\paragraph{Error Types:}
Labels with random offsets cost the least on \hlDsShiftLeast{} of \hlDsTypeRead{} datasets.
Incomplete labels cost more on both core datasets (\Cref{sec:types}).
For region-type targets such as burned areas or farmland, effort is therefore better spent on resolution, outlines and completeness than on registration.
That advice has six conditions, among them the random offset per tile, the metric and the target class\suppnote{ (\ssec{sec:supp-advice})}.

\paragraph{Decoders and Datasets:}
Where H5 holds for a head, the mean drop moves by less than the margin, yet a model contrast can still flip (\Cref{sec:rq2}).
Where it fails, the per-type contrasts say which type that head absorbs\suppnote{ (\ssec{sec:supp-n7})}.
The same reading applies across datasets, and a verdict that flips in \Cref{tab:datasets} is reported as dataset dependent.

\ifdefined\cvprbuild\else
\paragraph{Freezing, Budget and Losses:}
Frozen encoders lose less but end lower on \flair (\Cref{sec:budget}), so fine-tuning remains the default there at \Ltwo.
The clean subset wins on \burn at \Ltwo and trails at \Lone\suppnote{ (\ssec{sec:supp-budget})}.
The two losses are baselines without tuning, so their contrast states what a drop-in loss recovers at \Ltwo.
\fi

\section{Limitations}
\label{sec:limits}

The four error types are simulated, and the real pairs cover five operational labels and one algorithmic label.
The \etype{shift} and \etype{boundary} operators are zero-mean and the other two one-sided, which may favor the first two in the ranking.
One-sided variants of both operators keep \etype{shift} the cheapest type\suppnote{ (\ssec{sec:supp-robust})}.
Commission, thematic and temporal-change errors are not simulated.
Types are applied one at a time.
The \etype{boundary} operator is symmetric, whereas inventory outlines are mostly over-inclusive \cite{kolden2012mapped}.
Realizations of the same \etype{omission} knob vary in severity, which flags the \flair H2, F3 and F4 rows of \Cref{tab:verdicts}.
The reference labels are themselves imperfect \cite{cao2026contrastive,kolden2012mapped}, and type rankings are read against their convention.

On \flair and the extension datasets all models train on a fixed subset of tiles, which lowers clean \miou by a model-dependent amount.
\sat and \lvd differ mainly but not only in pretraining data, because normalization, training length and distillation also differ.
Learning rates are the clean-label \gbtwo values and one third of them, and a rate search per condition is untested.
The Mask2Former and SAM heads keep the encoder learning rate, and the SAM head runs without prompts.
The extension datasets run a reduced grid, with the decoder set on the DINOv3 pair only.

The runs span several GPU classes\suppnote{ (\stab{tab:supp-hw-blocks})}.
A paired drop whose two runs used different classes carries a hardware term, and so does a contrast between arms on different classes.
On the seeds of one class N6 reads \vd{\hlHwCtlNSixburnVerdict} on \burn and \vd{\hlHwCtlNSixflairVerdict} on \flair, so the term is not excluded\suppnote{ (\ssec{sec:supp-robust})}.
In the main grid the mean drops per class are up to \hlHwDropDiffMaxburn{} and \hlHwDropDiffMaxflair{} points apart\suppnote{ (\stab{tab:supp-hwmix})}.

The paired $t$ tests assume normality, and an exact sign-flip test cannot go below $p = \hlVeFOneSignFlipburnFloorFive$ at five seeds\suppnote{ (\ssec{sec:supp-stats})}.
The \SubsetPct\,\% comparison assumes a cost ratio of \CostRatio{} between clean and noisy training labels.
Pretraining data may overlap the evaluation regions, which the paired drops partly control.
Intervals are over seeds on fixed test splits, and each training subset is a single draw.

\section{Conclusion}
\label{sec:conclusion}

We presented a benchmark protocol for fine-tuning under imperfect labels and applied it to nine encoders, six decoder heads and ten datasets.
\ifdefined\cvprbuild\else
Its verdicts rest on paired contrasts with a pre-specified equivalence margin and a stated minimum detectable effect.
\fi
In the exploratory reading at the $\times$1/3 rate the satellite-pretrained and web-pretrained DINOv3 encoders are equivalent within $\pm$\TostMargin{} points on \hlDsNSixEquiv{} of \hlDsNSixRead{} datasets.
\lvd keeps a small detected lead on \flair, and H4 is not supported.
The multispectral \gfms and \rn are equivalent within the margin on \hlDsFTwoEquiv{} of \hlDsFTwoRead{} datasets.
Labels with random offsets cost the least at \Ltwo on \hlDsShiftLeast{} of \hlDsTypeRead{} datasets.
At a margin scaled to the drop both equivalence readings hold on fewer datasets.

\ifdefined\cvprbuild\else
Under the SegFormer head the decoder contrast reads \vd{\hlDecVerdictburn} on \burn and \vd{\hlDecVerdictflair} on \flair (\Cref{sec:decoders}).
\fi
Frozen encoders lose less but end lower on \flair, and the clean subset beats all tiles at \Ltwo on \burn only.
No choice of pretraining data makes an encoder consistently more robust, so the encoder is better chosen by its accuracy under imperfect labels.
\ifdefined\cvprbuild\else
New encoders can join on the released masks if each one gets both learning-rate multipliers.
\fi

\section*{Ethics and Data Statement}
\label{sec:ethics}

\gbtwo keeps the licenses of \burn (CC BY 4.0) and \flair (Etalab Open Licence 2.0). The \sfloods mirror on Hugging Face states Apache 2.0.
Chesapeake is released under CDLA-Permissive 1.0, CloudSEN12 under CC0 1.0, and DFC2020, EnviroAtlas and Kuro Siwo under CC BY 4.0.
SpaceNet 2 is released under CC BY-SA 4.0, and OpenEarthMap under a per-region license of CC BY 4.0 or CC BY-NC-SA 4.0.
People are not identifiable in the imagery.
Models partly reproduce their training labels\suppnote{ (\ssec{sec:supp-stats})}, so a product should not be assessed against them or other labels from the same source.
The compute footprint is \hlGpuHours{} GPU-hours.

\ifdefined\cvprbuild
\section*{LLM Usage Statement}
\label{sec:llm}

An LLM assistant (Claude, Anthropic) was used throughout under the direction of the authors.
It wrote and edited code, ran and monitored training, and drafted and edited text, tables and figure scripts.
It helped to plan the study and analysis and gave reviewer-style feedback on drafts.
The authors take full responsibility for all content.
\fi

{
    \small
    \bibliographystyle{ieeenat_fullname}
    \bibliography{refs}
}
\clearpage
\appendix
\makeatletter
\newcommand{\gcvAlph}[1]{\ifnum#1>26 A\@Alph{\numexpr#1-26\relax}\else\@Alph{#1}\fi}
\makeatother
\renewcommand{\thesection}{\gcvAlph{\value{section}}}\renewcommand{\theHsection}{\gcvAlph{\value{section}}}
\section*{Appendix}\suppressfloats[t]
\addcontentsline{toc}{section}{Appendix}
\noindent The appendix sections that follow are the supplementary material of the venue version. They hold the calibration
record and every table that the page budget of the venue version moved out of the main text.
They also hold the extension blocks, the figure galleries and the run manifest. Numbers come from the same analysis run as the main text.

\ifdefined\cvprbuild
\input{planned/sec/supp_cvpr_guide}
\section{Robustness of the Main Readings}
\label{sec:supp-robust}
\ifdefined\cvprbuild%
\providecommand{\hypref}[1]{\ifdefined\cvprbuild\msec{#1}\else\Cref{#1}\fi}
\begin{table*}[t]
\centering\footnotesize\setlength{\tabcolsep}{4pt}
\begin{tabular}{@{}llp{0.47\textwidth}ll@{}}
\toprule
ID & Question & Statement tested & Test & Answered in \\
\midrule
H1 & RQ2 & Misalignment is the most harmful error type & Type ranking by mean drop & \hypref{sec:types} \\
H2 & RQ2 & Omission costs more than boundary errors & Type ranking by mean drop & \hypref{sec:types} \\
H3 & RQ2 & Frozen encoders lose less than fine-tuned encoders & F3, paired by seed & \hypref{sec:budget} \\
H4 & RQ1 & Satellite pretraining loses less than web pretraining & F1 and N6 at one rate & \hypref{sec:rq1} \\
\midrule
H5 & RQ3 & Another decoder head changes the mean drop by less than the margin & Head minus UNet & \hypref{sec:decoders} \\
H6 & RQ2 & F2 at a common learning-rate multiplier is equivalent & F2 at one rate & \hypref{sec:rq2} \\
H7 & RQ3 & A real label costs as much as its matched simulated arm & Real minus simulated & \hypref{sec:ext} \\
H8 & RQ3 & The two Mask2Former objectives lose alike & Matching minus cross-entropy & \ifdefined\cvprbuild\Cref{sec:supp-n7}\else\hypref{sec:decoders}\fi{} \\
H9 & RQ3 & The type ranking at \Lpair{} equals the ranking at \Ltwo & Rank comparison & \hypref{sec:types} \\
\bottomrule
\end{tabular}
\caption{\textbf{Hypotheses and their tests:} Each row gives the research question a hypothesis belongs to, the statement tested, the contrast that tests it and the section with its result. H1 to H4 were fixed before the noisy-label runs and H5 to H9 before the runs of their blocks. F1 to F4 are the confirmatory families, and N6 is an exploratory contrast.}
\label{tab:hypotheses}
\end{table*}

\Cref{tab:hypotheses} lists the nine hypotheses with their tests.
\fi

This section tests the main readings against five objections.
One absolute margin may be loose where the drop is small, and runs on different GPU classes may carry a hardware term.
A real label and its simulated arm may differ in label quality.
The type ranking may reflect the operators rather than the error kind.
Equivalence findings give no positive statement about the encoders.

\begin{table*}[t]
\centering\scriptsize\setlength{\tabcolsep}{4pt}
\IfFileExists{planned/generated/table_margin_sensitivity.tex}{\begin{adjustbox}{max width=\textwidth}%
\providecommand{\gcvdry}[1]{\textcolor{red}{#1}}
\providecommand{\ProvTableMarginSensitivity}{Paired units per contrast 5 to 25. Estimate and 90\,\% interval of the contrast in points of drop. The first verdict uses the registered margin of $\pm$\TostMargin{} points. The scaled margin is \hlMsFractionPct\,\% of the mean \Ltwo drop of the two arms on that dataset and never above the registered margin, and it is treated as a fixed number. Equivalent means the 90\,\% interval lies inside the margin, Differs means the 95\,\% interval excludes zero and Undetermined is the rest. Red value: its block is still running or has not run. Analysis of the finished runs.}
\begin{tabular}{@{}llcclcl@{}}
\toprule
Contrast & Dataset & Estimate & 90\,\% interval & Verdict at $\pm$\TostMargin & Scaled margin & Verdict at scaled margin \\
\midrule
N6 & \burn & $-$0.2 & [$-$2.9, 2.5] & Equivalent & 2.77 & Undetermined \\
F1 & \burn & 4.3 & [1.9, 6.7] & Differs & 3.00 & Differs \\
F2 & \burn & 0.0 & [$-$2.2, 2.1] & Equivalent & 3.00 & Equivalent \\
H5, SegFormer head & \burn & 1.4 & [0.3, 2.6] & Equivalent & 3.00 & Equivalent \\
H5, Mask2Former head & \burn & $-$1.4 & [$-$5.0, 2.1] & Undetermined & 3.00 & Undetermined \\
H5, SAM head & \burn & 1.3 & [$-$0.1, 2.7] & Equivalent & 3.00 & Equivalent \\
H5, Full Mask2Former head & \burn & 0.8 & [$-$0.8, 2.5] & Equivalent & 3.00 & Equivalent \\
H5, Pretrained SAM head & \burn & 2.9 & [$-$1.3, 7.2] & Undetermined & 3.00 & Undetermined \\
RQ1 at $\times$1 & \burn & $-$1.7 & [$-$4.4, 0.9] & Undetermined & 3.00 & Undetermined \\
RQ1 at $\times$1/3 & \burn & $-$0.2 & [$-$2.9, 2.5] & Equivalent & 2.80 & Undetermined \\
H6, F2 at $\times$1 & \burn & $-$0.5 & [$-$4.3, 3.3] & Undetermined & 3.00 & Undetermined \\
H6, F2 at $\times$1/3 & \burn & $-$0.2 & [$-$2.7, 2.4] & Equivalent & 2.81 & Equivalent \\
N14, GCE & \burn & $-$1.2 & [$-$2.6, 0.2] & Equivalent & 3.00 & Equivalent \\
N14, Soft bootstrapping & \burn & $-$1.2 & [$-$2.3, $-$0.1] & Equivalent & 3.00 & Equivalent \\
\midrule
N6 & \flair & 1.5 & [0.8, 2.2] & Equivalent & 0.86 & Differs \\
F1 & \flair & 2.8 & [2.0, 3.6] & Differs & 0.98 & Differs \\
F2 & \flair & 0.6 & [$-$1.4, 2.5] & Equivalent & 0.90 & Undetermined \\
H5, SegFormer head & \flair & $-$0.1 & [$-$0.6, 0.4] & Equivalent & 0.94 & Equivalent \\
H5, Mask2Former head & \flair & 0.8 & [0.3, 1.4] & Equivalent & 1.04 & Differs \\
H5, SAM head & \flair & 0.0 & [$-$0.5, 0.5] & Equivalent & 0.95 & Equivalent \\
H5, Full Mask2Former head & \flair & 0.2 & [$-$0.3, 0.8] & Equivalent & 0.97 & Equivalent \\
H5, Pretrained SAM head & \flair & 0.5 & [0.0, 1.1] & Equivalent & 1.01 & Undetermined \\
RQ1 at $\times$1 & \flair & 2.5 & [1.4, 3.6] & Differs & 1.05 & Differs \\
RQ1 at $\times$1/3 & \flair & 1.5 & [0.8, 2.2] & Equivalent & 0.87 & Differs \\
H6, F2 at $\times$1 & \flair & 0.6 & [$-$1.4, 2.5] & Equivalent & 0.90 & Undetermined \\
H6, F2 at $\times$1/3 & \flair & 0.2 & [$-$1.0, 1.4] & Equivalent & 0.79 & Undetermined \\
N14, GCE & \flair & 0.1 & [$-$0.6, 0.8] & Equivalent & 0.96 & Equivalent \\
N14, Soft bootstrapping & \flair & 0.1 & [$-$0.4, 0.5] & Equivalent & 0.96 & Equivalent \\
\midrule
N6 & Chesapeake & $-$0.7 & [$-$1.6, 0.2] & Equivalent & 1.97 & Equivalent \\
F1 & Chesapeake & $-$0.1 & [$-$1.6, 1.5] & Equivalent & 2.04 & Equivalent \\
F2 & Chesapeake & 2.3 & [1.4, 3.1] & Differs & 1.96 & Differs \\
H5, three heads & Chesapeake & 0.8 & [0.1, 1.4] & Equivalent & 2.11 & Equivalent \\
\midrule
N6 & CloudSEN12 & 0.3 & [$-$0.6, 1.3] & Equivalent & 1.31 & Equivalent \\
F1 & CloudSEN12 & $-$0.5 & [$-$2.1, 1.1] & Equivalent & 1.23 & Undetermined \\
F2 & CloudSEN12 & 0.4 & [$-$0.6, 1.5] & Equivalent & 1.40 & Undetermined \\
H5, three heads & CloudSEN12 & 0.3 & [$-$0.2, 0.8] & Equivalent & 1.25 & Equivalent \\
\midrule
N6 & DFC2020 & 0.5 & [$-$1.2, 2.2] & Equivalent & 0.56 & Undetermined \\
F1 & DFC2020 & 0.4 & [$-$0.5, 1.3] & Equivalent & 0.55 & Undetermined \\
F2 & DFC2020 & 1.2 & [$-$0.1, 2.6] & Equivalent & 0.36 & Undetermined \\
H5, three heads & DFC2020 & $-$0.1 & [$-$0.8, 0.6] & Equivalent & 0.54 & Undetermined \\
\midrule
N6 & EnviroAtlas & $-$1.5 & [$-$4.0, 1.0] & Undetermined & 1.40 & Undetermined \\
F1 & EnviroAtlas & $-$1.5 & [$-$3.8, 0.9] & Undetermined & 1.40 & Undetermined \\
F2 & EnviroAtlas & $-$5.9 & [$-$7.9, $-$3.9] & Differs & 1.82 & Differs \\
H5, three heads & EnviroAtlas & $-$3.3 & [$-$4.7, $-$2.0] & Differs & 1.07 & Differs \\
\midrule
N6 & SpaceNet 2 & 0.1 & [$-$1.1, 1.3] & Equivalent & 2.08 & Equivalent \\
F1 & SpaceNet 2 & 1.1 & [0.7, 1.5] & Equivalent & 2.18 & Equivalent \\
F2 & SpaceNet 2 & $-$2.1 & [$-$3.6, $-$0.6] & Differs & 2.56 & Differs \\
H5, three heads & SpaceNet 2 & 2.9 & [2.3, 3.4] & Differs & 2.47 & Differs \\
\midrule
N6 & Kuro Siwo & $-$1.3 & [$-$3.0, 0.3] & Equivalent & 1.35 & Undetermined \\
F1 & Kuro Siwo & $-$0.9 & [$-$3.5, 1.6] & Undetermined & 1.39 & Undetermined \\
F2 & Kuro Siwo & 5.9 & [2.2, 9.6] & Differs & 0.70 & Differs \\
H5, three heads & Kuro Siwo & 0.9 & [$-$1.4, 3.1] & Undetermined & 1.48 & Undetermined \\
\midrule
N6 & OpenEarthMap & $-$0.1 & [$-$0.6, 0.3] & Equivalent & 1.34 & Equivalent \\
F1 & OpenEarthMap & 0.9 & [0.3, 1.4] & Equivalent & 1.44 & Differs \\
F2 & OpenEarthMap & 1.0 & [0.0, 2.0] & Equivalent & 1.17 & Undetermined \\
H5, three heads & OpenEarthMap & $-$0.5 & [$-$1.0, 0.0] & Equivalent & 1.39 & Equivalent \\
\bottomrule
\end{tabular}
\end{adjustbox}}{\tabletbd[4cm]{planned/generated/table\_margin\_sensitivity.tex}}
\caption{\textbf{Equivalence verdicts at two margins:} Estimate and 90\,\% interval of each contrast in points of drop. The first verdict uses the registered margin of $\pm$\TostMargin{} points. The scaled margin is \hlMsFractionPct\,\% of the mean \Ltwo drop of the two arms on that dataset and never above \TostMargin{} points. Equivalent means the 90\,\% interval lies inside the margin. Otherwise Differs means the \CiLevel\,\% interval excludes zero, and Undetermined is the rest. On an extension dataset the H5 row pools three heads on the DINOv3 pair. The learning-rate multipliers are $\times$1 and $\times$1/3, and GCE is generalized cross-entropy.}
\label{tab:supp-margin}
\end{table*}

\paragraph{Scaled Margin:}
The registered margin of $\pm$\TostMargin{} points was sized on \burn, where it is \hlMsShareburn\,\% of the mean \Ltwo drop.
On \flair it is \hlMsShareflair\,\% of the drop, so equivalence there is a weak statement.
\Cref{tab:supp-margin} therefore repeats every verdict at a margin of \hlMsFractionPct\,\% of the drop.
N6 stays equivalent on \hlMsNSixEquivScaled{} of \hlMsNSixRead{} datasets and F2 on \hlMsFTwoEquivScaled{} of \hlMsFTwoRead.
On \burn N6 reads \vd{\hlMsVerdictNSixburn} at the scaled margin and on \flair it reads \vd{\hlMsVerdictNSixflair}.
The main paper therefore states the registered verdict with the size of the drop.

\begin{table*}[t]
\centering\scriptsize\setlength{\tabcolsep}{4pt}
\IfFileExists{planned/generated/table_hardware_control.tex}{\begin{adjustbox}{max width=\textwidth}%
\providecommand{\gcvdry}[1]{\textcolor{red}{#1}}
\providecommand{\ProvTableHardwareControl}{Upper block: the contrast in points of drop on the seeds whose runs sit on one GPU class, with the number of such seeds of all seeds, the 90\,\% interval and the verdict at $\pm$\TostMargin{} points. All runs on one class removes the hardware term. One class per model removes mixed clean and noisy pairs and keeps the term between the arms. Lower block: every configuration and seed found on two GPU classes or more among all finished runs, with the excluded probe and timing runs. These are \hlHwCtlDupEpochs{}-epoch clean-label timing runs. The columns are the pairs of runs on different classes, the mean absolute difference in test mIoU across classes, the mean signed difference against \hlHwCtlDupRefClass{} and the mean absolute difference between runs on one class, in points. No noisy run was repeated on a second class. Analysis of the finished runs.}
\begin{tabular}{@{}lllcccl@{}}
\toprule
Contrast and seed subset & Dataset & GPU classes & Seeds & Estimate & 90\,\% interval & Verdict \\
\midrule
N6, all runs on one class & \burn & RTX PRO 6000 & 3 of 5 & $-$1.2 & [$-$5.0, 2.6] & Undetermined \\
N6, one class per model & \burn & RTX PRO 6000 & 3 of 5 & $-$1.2 & [$-$5.0, 2.6] & Undetermined \\
N6, all runs on one class & \flair & RTX PRO 6000 & 2 of 5 & 2.0 & [$-$2.0, 5.9] & Undetermined \\
N6, one class per model & \flair & RTX 4090 / RTX PRO 6000 & 3 of 5 & 2.0 & [0.9, 3.0] & Differs \\
F2, all runs on one class & \burn & -- & 0 of 8 & -- & -- & No such subset \\
F2, one class per model & \burn & 5 classes & 8 of 8 & 0.0 & [$-$2.2, 2.1] & Equivalent \\
F2, all runs on one class & \flair & -- & 0 of 5 & -- & -- & No such subset \\
F2, one class per model & \flair & GB10 / RTX 4090 & 5 of 5 & 0.6 & [$-$1.4, 2.5] & Equivalent \\
\midrule
Same seed on two classes & Runs & GPU classes & Pairs & Across classes & Signed & Within class \\
\midrule
\sat, \burn & 19 runs & 7 classes & 144 & 11.05 & 3.04 & 13.14 \\
\sat, \flair & 9 runs & 5 classes & 31 & 0.10 & $-$0.04 & 0.09 \\
\dofa, Chesapeake & 2 runs & RTX 3080 / RTX PRO 6000 & 1 & 0.52 & 0.52 & -- \\
\dofa, \flair & 9 runs & 7 classes & 34 & 0.27 & 0.27 & 0.07 \\
\rn, \burn & 10 runs & 7 classes & 42 & 2.05 & $-$0.89 & 0.77 \\
\rn, Chesapeake & 2 runs & RTX 3080 / RTX PRO 6000 & 1 & 0.22 & $-$0.22 & -- \\
\tm, \burn & 8 runs & 5 classes & 25 & 0.16 & $-$0.04 & 0.08 \\
Mean over 7 configurations & 59 runs & 7 classes & 278 & 2.05 & 0.38 & 2.83 \\
\bottomrule
\end{tabular}
\end{adjustbox}}{\tabletbd[3cm]{planned/generated/table\_hardware\_control.tex}}
\caption{\textbf{Primary contrasts on seeds of one GPU class:} N6 and F2 recomputed on the seeds whose runs share one GPU class, with the estimate and the 90\,\% interval in points of drop. The first reading asks for one class over all runs of a seed and the second for one class per model. A dash marks a contrast without such a subset or a configuration without two runs on one class. The lower block holds same-seed clean timing runs of \hlHwCtlDupEpochs{} epochs on two or more GPU classes. Pairs counts the run pairs on different classes. Across classes and Within class give the mean absolute test \miou difference in points between classes and within one class. Signed is the mean signed difference against the \hlHwCtlDupRefClass{}.}
\label{tab:supp-hwctl}
\end{table*}

\paragraph{Hardware Control:}
On \burn the seeds \hlHwCtlNSixburnSeeds{} ran N6 on one GPU class.
With them N6 is \hlHwCtlNSixburnEst{} [\hlHwCtlNSixburnCILo, \hlHwCtlNSixburnCIHi] points and reads \vd{\hlHwCtlNSixburnVerdict} (\Cref{tab:supp-hwctl}).
On \flair the seeds \hlHwCtlNSixflairSeeds{} give \hlHwCtlNSixflairEst{} [\hlHwCtlNSixflairCILo, \hlHwCtlNSixflairCIHi] points, which reads \vd{\hlHwCtlNSixflairVerdict}.
The subsets are too small to confirm equivalence. F2 has no subset with one class over all runs.
Same-seed repeats of clean timing runs differ by \hlHwCtlDupAbs{} points across classes against \hlHwCtlDupWithin{} points within a class.
The hardware term of a paired drop is therefore not identified from these timing runs.
A replicate block repeats the N6 cells on a second GPU class to measure it.

\begin{table*}[t]
\centering\scriptsize\setlength{\tabcolsep}{4pt}
\IfFileExists{planned/generated/table_real_matched_interp.tex}{\begin{adjustbox}{max width=\textwidth}%
\providecommand{\gcvdry}[1]{\textcolor{red}{#1}}
\providecommand{\ProvTableRealMatchedInterp}{Drop under the real label and under each simulated type at the measured label mIoU of the real pair, in points, over the (model, seed) units that have both. The simulated drop of a unit is a piecewise linear function of label mIoU through zero at clean labels and its drops at the levels the dataset has run. A label mIoU below the lowest level continues the last segment, and the row says by how much label mIoU it extrapolates. Difference with its 95\,\% interval, and the verdict at $\pm$\TostMargin{} points. Red value: its block is still running or has not run. Analysis of the finished runs.}
\begin{tabular}{@{}lcllccccl@{}}
\toprule
Dataset & Label mIoU & Type & Reading & Real & Simulated & Difference & Units & Verdict \\
\midrule
Chesapeake & 0.571 & \etype{Shift} & Extrapolated by 0.13 & 21.4 & 12.6 & 8.8 [7.8, 9.8] & 25 & Real costs more \\
 &  & \etype{Coarse} &  & 21.4 & 21.2 & 0.2 [$-$0.8, 1.1] & 25 & Equivalent \\
 &  & \etype{Boundary} &  & 21.4 & 18.1 & 3.3 [1.6, 4.9] & 25 & Real costs more \\
 &  & \etype{Omission} &  & 21.4 & 16.3 & 5.0 [1.8, 8.3] & 25 & Real costs more \\
 &  & Mean of four &  & 21.4 & 17.1 & 4.3 [3.1, 5.5] & 25 & Real costs more \\
\midrule
CloudSEN12 & 0.331 & \etype{Shift} & Extrapolated by 0.37 & 32.4 & 15.7 & 16.7 [15.1, 18.3] & 25 & Real costs more \\
 &  & \etype{Coarse} &  & 32.4 & 22.9 & 9.5 [7.6, 11.3] & 25 & Real costs more \\
 &  & \etype{Boundary} &  & 32.4 & 18.3 & 14.0 [11.4, 16.7] & 25 & Real costs more \\
 &  & \etype{Omission} &  & 32.4 & 16.2 & 16.2 [14.5, 17.8] & 25 & Real costs more \\
 &  & Mean of four &  & 32.4 & 18.3 & 14.1 [12.8, 15.4] & 25 & Real costs more \\
\midrule
DFC2020 & 0.675 & \etype{Shift} & Extrapolated by 0.03 & 4.8 & 0.1 & 4.7 [3.5, 5.9] & 25 & Real costs more \\
 &  & \etype{Coarse} &  & 4.8 & 1.5 & 3.2 [2.0, 4.5] & 25 & Real costs more \\
 &  & \etype{Boundary} &  & 4.8 & 4.8 & 0.0 [$-$1.5, 1.5] & 25 & Equivalent \\
 &  & \etype{Omission} &  & 4.8 & 3.8 & 0.9 [$-$0.5, 2.4] & 25 & Equivalent \\
 &  & Mean of four &  & 4.8 & 2.6 & 2.2 [1.1, 3.3] & 25 & Real costs more \\
\midrule
EnviroAtlas & 0.483 & \etype{Shift} & Extrapolated by 0.22 & 24.6 & 8.7 & 15.9 [11.5, 20.4] & 25 & Real costs more \\
 &  & \etype{Coarse} &  & 24.6 & 9.2 & 15.4 [11.6, 19.2] & 25 & Real costs more \\
 &  & \etype{Boundary} &  & 24.6 & 19.8 & 4.8 [$-$1.7, 11.3] & 25 & Undetermined \\
 &  & \etype{Omission} &  & 24.6 & 26.0 & $-$1.4 [$-$8.6, 5.8] & 25 & Undetermined \\
 &  & Mean of four &  & 24.6 & 15.9 & 8.7 [5.6, 11.8] & 25 & Real costs more \\
\midrule
Kuro Siwo & 0.893 & \etype{Shift} & Interpolated & 0.8 & 1.4 & $-$0.6 [$-$2.3, 1.1] & 25 & Equivalent \\
 &  & \etype{Coarse} &  & 0.8 & 0.0 & 0.7 [$-$1.0, 2.5] & 25 & Equivalent \\
 &  & \etype{Boundary} &  & 0.8 & 0.9 & $-$0.2 [$-$1.8, 1.5] & 25 & Equivalent \\
 &  & \etype{Omission} &  & 0.8 & 2.3 & $-$1.6 [$-$3.3, 0.1] & 25 & Equivalent \\
 &  & Mean of four &  & 0.8 & 1.2 & $-$0.4 [$-$2.0, 1.2] & 25 & Equivalent \\
\bottomrule
\end{tabular}
\end{adjustbox}}{\tabletbd[3cm]{planned/generated/table\_real\_matched\_interp.tex}}
\caption{\textbf{Real labels against simulated types at the measured label quality:} For each pair dataset the simulated drop of every type is read at the label \miou of the real label. The reading is linear from clean labels through the calibrated levels and is marked as extrapolated below \Ltwo. The difference is the real drop minus the simulated drop in points with its \CiLevel\,\% interval over model and seed units. Equivalent means the 90\,\% interval lies inside $\pm$\TostMargin{} points. Otherwise the verdict is Real costs more where the \CiLevel\,\% interval lies above zero and Undetermined where it includes zero.}
\label{tab:supp-real-interp}
\end{table*}

\paragraph{Real Labels at Matched Label Quality:}
The H7 test of the main paper pairs each real label with the \etype{coarse} type at the nearest calibrated level.
The two labels then differ in label \miou, by \hlRealInterpChesGap{} on Chesapeake.
\Cref{tab:supp-real-interp} reads the simulated drop at the measured label \miou of each pair instead.
On Chesapeake the real label then costs \hlRealInterpChesEst{} [\hlRealInterpChesCILo, \hlRealInterpChesCIHi] points more than the \etype{coarse} type and reads \vd{\hlRealInterpChesVerdict}.
That reading is \hlRealInterpChesFlag{} from the calibrated levels.

\begin{table*}[t]
\centering\scriptsize\setlength{\tabcolsep}{4pt}
\IfFileExists{planned/generated/table_real_matched_lm.tex}{\begin{adjustbox}{max width=\textwidth}%
\providecommand{\gcvdry}[1]{\textcolor{red}{#1}}
\providecommand{\ProvTableRealMatchedLm}{Drop under the real label and under the simulated arm calibrated to the measured label mIoU of the real pair, in points, over the (model, seed) units that have both. Label mIoU of the real pair and of the simulated arm (mean of its five realisations). Difference with its 95\,\% interval, the 90\,\% interval of the equivalence test at $\pm$\TostMargin{} points, the minimum detectable effect and the verdict. Red value: its block is still running or has not run. Analysis of the finished runs.}
\begin{tabular}{@{}llccccccccl@{}}
\toprule
Dataset & Type & Real label mIoU & Arm label mIoU & Real & Simulated & Difference & 90\,\% interval & MDE & Units & Verdict \\
\midrule
Chesapeake & Coarse & 0.571 & 0.574 & 21.4 & 22.7 & $-$1.4 [$-$2.3, $-$0.4] & [$-$2.2, $-$0.6] & 1.5 & 25 & Equivalent \\
\midrule
CloudSEN12 & Omission & 0.331 & 0.331 & 32.4 & 32.7 & $-$0.3 [$-$2.4, 1.7] & [$-$2.0, 1.4] & 3.2 & 25 & Equivalent \\
\midrule
DFC2020 & Coarse & 0.675 & 0.676 & 4.8 & 1.6 & 3.2 [1.9, 4.4] & [2.1, 4.2] & 2.0 & 25 & Real costs more \\
\midrule
EnviroAtlas & Coarse & 0.483 & 0.485 & 24.6 & 15.9 & 8.8 [5.7, 11.8] & [6.2, 11.3] & 4.8 & 25 & Real costs more \\
\midrule
Kuro Siwo & Coarse & 0.893 & 0.898 & 0.8 & 1.2 & $-$0.5 [$-$2.0, 1.1] & [$-$1.8, 0.8] & 2.5 & 25 & Equivalent \\
 & Boundary &  & 0.895 & 0.8 & 0.4 & 0.3 [$-$1.1, 1.7] & [$-$0.8, 1.5] & 2.2 & 25 & Equivalent \\
\bottomrule
\end{tabular}
\end{adjustbox}}{\tabletbd[3cm]{planned/generated/table\_real\_matched\_lm.tex}}
\caption{\textbf{Real labels against simulated arms trained at the measured label quality:} Each simulated arm is calibrated to the label \miou of the real label of its dataset. The difference is the real drop minus the simulated drop in points, paired by model and seed, with its \CiLevel\,\% interval and the 90\,\% equivalence interval. MDE is the minimum detectable effect in points.}
\label{tab:supp-real-lm}
\end{table*}

\paragraph{Simulated Arms Trained at the Measured Label Quality:}
Block N16 replaces the linear reading by training.
Each pair dataset gets a simulated arm calibrated to the label \miou of its real label (\Cref{tab:supp-real-lm}).
On Chesapeake the real label then costs \hlRealLmChesEst{} [\hlRealLmChesCILo, \hlRealLmChesCIHi] points against the \etype{\hlRealLmChesType} arm and reads \vd{\hlRealLmChesVerdict}.
The contrast covers \hlRealLmRead{} of \hlRealLmTotal{} pair datasets.

\begin{table}[t]
\centering\scriptsize\setlength{\tabcolsep}{3pt}
\IfFileExists{planned/generated/table_hardware_replicates.tex}{\begin{adjustbox}{max width=\columnwidth}%
\providecommand{\gcvdry}[1]{\textcolor{red}{#1}}
\providecommand{\ProvTableHardwareReplicates}{Replicate minus original of the same model, dataset, condition and seed, in points. The replicates repeat the cells of the matched-rate DINOv3 contrast on a second GPU class. Clean and noisy test mIoU per run, and the paired drop with each drop taken inside one class. Mean with its 95\,\% interval, mean absolute difference with its 95\,\% interval, largest absolute difference and number of pairs. The last three rows of a dataset give the contrast on each class alone over the same seeds and its difference per seed. The replicates enter no other statistic. Red value: the replicates are still running or have not run. Analysis of the finished runs.}
\begin{tabular}{@{}llcccc@{}}
\toprule
Dataset & Quantity & Mean & Mean absolute & Largest absolute & Pairs \\
\midrule
\burn & Clean test mIoU & 0.80 [$-$0.70, 2.30] & 0.98 [$-$0.38, 2.33] & 2.70 & 6 \\
 & Noisy test mIoU & $-$0.37 [$-$1.51, 0.77] & 2.23 [1.59, 2.87] & 5.65 & 24 \\
 & Paired drop & 1.17 [$-$0.04, 2.38] & 2.26 [1.38, 3.13] & 6.69 & 24 \\
 & N6, replicates & 0.15 [$-$0.95, 1.25] & -- & -- & 3 \\
 & N6, originals & $-$1.22 [$-$6.83, 4.38] & -- & -- & 3 \\
 & N6, difference & 1.37 [$-$3.71, 6.46] & 1.76 [$-$2.07, 5.59] & 3.50 & 3 \\
\midrule
\flair & Clean test mIoU & 0.14 [$-$1.09, 1.38] & 0.53 [$-$0.26, 1.33] & 1.14 & 4 \\
 & Noisy test mIoU & 0.27 [$-$0.06, 0.61] & 0.47 [0.21, 0.73] & 1.78 & 16 \\
 & Paired drop & $-$0.13 [$-$0.64, 0.38] & 0.74 [0.42, 1.05] & 2.50 & 16 \\
 & N6, replicates & 0.73 [0.00, 1.46] & -- & -- & 2 \\
 & N6, originals & 1.97 [$-$6.01, 9.96] & -- & -- & 2 \\
 & N6, difference & $-$1.24 [$-$9.95, 7.47] & 1.24 [$-$7.47, 9.95] & 1.93 & 2 \\
\midrule
Pooled & Clean test mIoU & 0.54 [$-$0.33, 1.40] & 0.80 [0.06, 1.54] & 2.70 & 10 \\
 & Noisy test mIoU & $-$0.11 [$-$0.80, 0.57] & 1.53 [1.05, 2.00] & 5.65 & 40 \\
 & Paired drop & 0.65 [$-$0.11, 1.40] & 1.65 [1.07, 2.22] & 6.69 & 40 \\
\bottomrule
\end{tabular}
\end{adjustbox}}{\tabletbd[3cm]{planned/generated/table\_hardware\_replicates.tex}}
\caption{\textbf{Same-seed replicates on a second GPU class:} Difference between a replicate on the \hlHwRepClass{} and its original on the \hlHwRepOrigClass{} for the same model, dataset, condition and seed, in points. The last rows give N6 on the replicates and on the originals of the same seeds. Brackets give the \CiLevel\,\% interval, and Pooled combines both datasets. Mean absolute and Largest absolute summarize the absolute pair differences. A dash marks a quantity that a contrast does not have.}
\label{tab:supp-hwrep}
\end{table}

\paragraph{Replicates on a Second GPU Class:}
A replicate block repeats \hlHwRepPlannedPooled{} runs of the N6 cells on a second GPU class with the same seeds (\Cref{tab:supp-hwrep}).
A replicate differs from its original only in hardware, so the difference measures the hardware term of a paired drop.
The replicates enter no other statistic of the paper.

\begin{table*}[t]
\centering\scriptsize\setlength{\tabcolsep}{4pt}
\IfFileExists{planned/generated/table_operator_variants.tex}{\begin{adjustbox}{max width=\textwidth}%
\providecommand{\gcvdry}[1]{\textcolor{red}{#1}}
\providecommand{\ProvTableOperatorVariants}{Mean \Ltwo{} drop under a one-sided operator variant and under its zero-mean parent, in points, over the (model, seed) units of the five core models that have both, at equal label mIoU. Difference with its 95\,\% interval, the 90\,\% interval of the equivalence test at $\pm$\TostMargin{} points, the minimum detectable effect and the verdict. The two order rows list the four types from the largest to the smallest mean drop on the same units, with the variants in place of their parents and with the parents. Red value: its block is still running or has not run. Analysis of the finished runs.}
\begin{tabular}{@{}llccccccl@{}}
\toprule
Dataset & Contrast & Variant & Parent & Difference & 90\,\% interval & MDE & Units & Verdict \\
\midrule
\burn & Fixed shift minus shift & 13.0 & 11.4 & 1.6 [0.4, 2.8] & [0.6, 2.6] & 1.9 & 25 & Equivalent \\
 & Boundary dilation minus boundary & 18.8 & 13.4 & 5.4 [3.8, 7.1] & [4.1, 6.8] & 2.6 & 25 & Variant costs more \\
 & Order with the variants & \multicolumn{7}{l}{Coarse 19.1, omission 19.0, boundary dilation 18.8, fixed shift 13.0} \\
 & Order of the parents & \multicolumn{7}{l}{Coarse 19.1, omission 19.0, boundary 13.4, shift 11.4} \\
\midrule
\flair & Fixed shift minus shift & 4.2 & 3.3 & 0.9 [0.4, 1.4] & [0.5, 1.3] & 0.8 & 25 & Equivalent \\
 & Boundary dilation minus boundary & 5.1 & 5.2 & $-$0.1 [$-$0.7, 0.6] & [$-$0.6, 0.5] & 1.0 & 25 & Equivalent \\
 & Order with the variants & \multicolumn{7}{l}{Coarse 5.5, boundary dilation 5.1, omission 5.0, fixed shift 4.2} \\
 & Order of the parents & \multicolumn{7}{l}{Coarse 5.5, boundary 5.2, omission 5.0, shift 3.3} \\
\bottomrule
\end{tabular}
\end{adjustbox}}{\tabletbd[3cm]{planned/generated/table\_operator\_variants.tex}}
\caption{\textbf{One-sided variants of the two zero-mean operators:} A fixed shift moves every tile of a realization in one direction, and a boundary dilation only widens outlines. Both are calibrated to the \Ltwo label \miou. The difference is the drop under the variant minus the drop under its parent type in points, paired by model and seed. Variant and Parent give the mean drops, and brackets the \CiLevel\,\% interval. MDE is the minimum detectable effect and Units the number of model and seed pairs. A verdict is Equivalent when the 90\,\% interval lies inside $\pm$\TostMargin{} points. Otherwise Variant costs more or less names the side of zero where the \CiLevel\,\% interval lies, and Undetermined marks an interval that covers zero. The order rows rank the four types by mean drop, largest first.}
\label{tab:supp-variants}
\end{table*}

\paragraph{One Sided Operator Variants:}
The \etype{shift} and \etype{boundary} operators have no preferred direction, whereas \etype{coarse} and \etype{omission} act in one direction.
The type ranking could therefore reflect the operators and not the error kind.
\Cref{tab:supp-variants} tests this with a fixed shift direction and with outlines that only widen, both at the \Ltwo label \miou.
The ranking is repeated with the variants in place of their parent types.

\begin{figure}[t]
\centering
\IfFileExists{planned/figs/fig_an_rank_agreement.pdf}{\includegraphics[width=\linewidth]{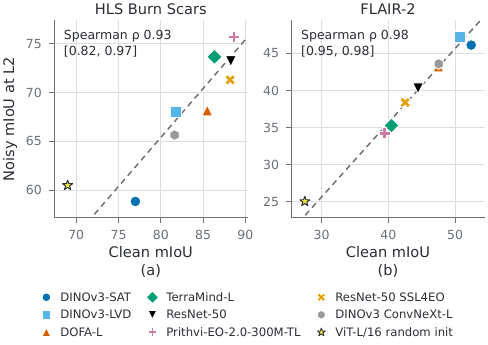}}{\phfig{4cm}{planned/figs/fig\_an\_rank\_agreement.pdf}}
\caption{\textbf{Clean and noisy accuracy of the nine encoders:} Each marker is one encoder, with its clean test \miou on the horizontal axis and its mean \Ltwo test \miou on the vertical axis. The dashed line is the identity lowered by the mean \Ltwo drop, so an encoder above it loses less than average. Brackets give the \CiLevel\,\% interval of the Spearman correlation $\rho$ from a bootstrap over seeds. Panel (a) shows \burn and panel (b) \flair.}
\label{fig:supp-rank}
\end{figure}

\paragraph{Clean and Noisy Rankings:}
The ranking of the nine encoders on clean labels predicts their ranking under \Ltwo label errors (\Cref{fig:supp-rank}).
The Spearman correlation is \hlRankRhoburn{} [\hlRankRhoCILoburn, \hlRankRhoCIHiburn] on \burn.
On \flair it is \hlRankRhoflair{} [\hlRankRhoCILoflair, \hlRankRhoCIHiflair].
Because the intervals resample seeds within each encoder, they bound seed noise and not the choice of encoders.
On Chesapeake, with five encoders, the correlation is \hlRankRhoChes{} and is not separated from zero.

\begin{table*}[t]
\centering\scriptsize\setlength{\tabcolsep}{4pt}
\IfFileExists{planned/generated/table_n19_encoders.tex}{\begin{adjustbox}{max width=\textwidth}%
\providecommand{\gcvdry}[1]{\textcolor{red}{#1}}
\providecommand{\ProvTableNNineteenEncoders}{Seeds per cell 5. Clean is the clean test mIoU and Drop the mean paired drop over the four \Ltwo types in points, both means over the seeds with all four types. The five rows above the rule are the core models and the four rows below it the extension encoders, \vitr with random initialization. A superscript o marks a pair that is out of domain, where no input channel or no band beyond the visible ones has a pretrained counterpart. A superscript p marks a pair with partial bands, where part of the pretraining band set is present or input channels start from fresh first-layer weights. Red is a value of a block that is still running and a red dash a cell that has not run. Analysis of the finished runs}
\begin{tabular}{@{}l@{\hspace{7pt}}c@{\hspace{4pt}}c@{\hspace{7pt}}c@{\hspace{4pt}}c@{\hspace{7pt}}c@{\hspace{4pt}}c@{\hspace{7pt}}c@{\hspace{4pt}}c@{\hspace{7pt}}c@{\hspace{4pt}}c@{\hspace{7pt}}c@{\hspace{4pt}}c@{\hspace{7pt}}c@{\hspace{4pt}}c@{}}
\toprule
& \multicolumn{2}{c}{Chesapeake} & \multicolumn{2}{c}{CloudSEN12} & \multicolumn{2}{c}{DFC2020} & \multicolumn{2}{c}{EnviroAtlas} & \multicolumn{2}{c}{SpaceNet 2} & \multicolumn{2}{c}{Kuro Siwo} & \multicolumn{2}{c}{OpenEarthMap} \\
\cmidrule(lr){2-3} \cmidrule(lr){4-5} \cmidrule(lr){6-7} \cmidrule(lr){8-9} \cmidrule(lr){10-11} \cmidrule(lr){12-13} \cmidrule(lr){14-15}
Model & Clean & Drop & Clean & Drop & Clean & Drop & Clean & Drop & Clean & Drop & Clean & Drop & Clean & Drop \\
\midrule
\sat & 81.3 & 10.2 & 66.6 & 5.9 & 40.5 & 2.9 & 61.8 & 6.3 & 85.8 & 11.4 & 53.0 & 6.5 & 63.5 & 7.6 \\
\lvd & 79.3 & 10.2 & 69.6 & 6.4 & 43.2 & 2.5 & 63.0 & 7.7 & 86.0 & 10.4 & 59.8 & 7.4 & 63.4 & 6.7 \\
\dofa & 83.0 & 12.0 & 65.9 & 7.2 & 36.5 & 1.4 & 48.1 & 5.4 & 84.2 & 11.8 & 53.5 & 5.1 & 61.0 & 6.8 \\
\tm & 82.0 & 9.9 & 69.5 & 7.2 & 38.9 & 3.4 & 49.0 & 6.9 & 83.4 & 11.7 & 58.0 & 7.7 & 55.2 & 6.0 \\
\rn & 82.5 & 8.7 & 64.0 & 6.8 & 38.8 & 1.2 & 49.1 & 12.1 & 81.7 & 13.8 & 32.5 & 0.5 & 55.6 & 5.4 \\
\midrule
\prithvi & 83.9$^{\mathrm{p}}$ & 10.6 & 67.2 & 6.0 & 35.7 & 3.2 & 49.9$^{\mathrm{p}}$ & 9.3 & 83.0$^{\mathrm{p}}$ & 14.8 & 49.7$^{\mathrm{o}}$ & 3.9 & 54.0$^{\mathrm{o}}$ & 6.5 \\
\rnssl & 82.6$^{\mathrm{p}}$ & 9.5 & 62.3 & 7.0 & 39.3$^{\mathrm{p}}$ & 1.6 & 49.2$^{\mathrm{p}}$ & 14.8 & 81.0$^{\mathrm{p}}$ & 12.5 & 33.1$^{\mathrm{o}}$ & $-$1.1 & 49.8$^{\mathrm{o}}$ & 4.6 \\
\vitr & 66.8 & 7.5 & 57.0 & 5.8 & 30.9 & 1.0 & 33.3 & 8.2 & 71.6 & 11.9 & 38.7 & 0.4 & 37.4 & 3.6 \\
\cnxl & 80.7 & 10.6 & 68.2 & 6.8 & 42.1 & 3.1 & 49.3 & $-$2.7 & 86.3 & 11.9 & 58.8$^{\mathrm{o}}$ & 9.3 & 61.5 & 6.8 \\
\bottomrule
\end{tabular}
\end{adjustbox}}{\tabletbd[3cm]{planned/generated/table\_n19\_encoders.tex}\providecommand{\ProvTableNNineteenEncoders}{\gcvdry{PLANNED, no runs yet}}}
\caption{\textbf{Nine encoders on the seven extension datasets:} Clean test \miou and mean \Ltwo drop in points for each encoder and dataset. The four rows below the rule are the extension encoders. A superscript o marks an out-of-domain pair and a superscript p a pair with partial bands.}
\label{tab:supp-n19}
\end{table*}

\ifdefined\cvprbuild\else
\begin{figure*}[t]
\centering
\IfFileExists{planned/figs/fig_an_rank_agreement_ext.pdf}{\includegraphics[width=\textwidth]{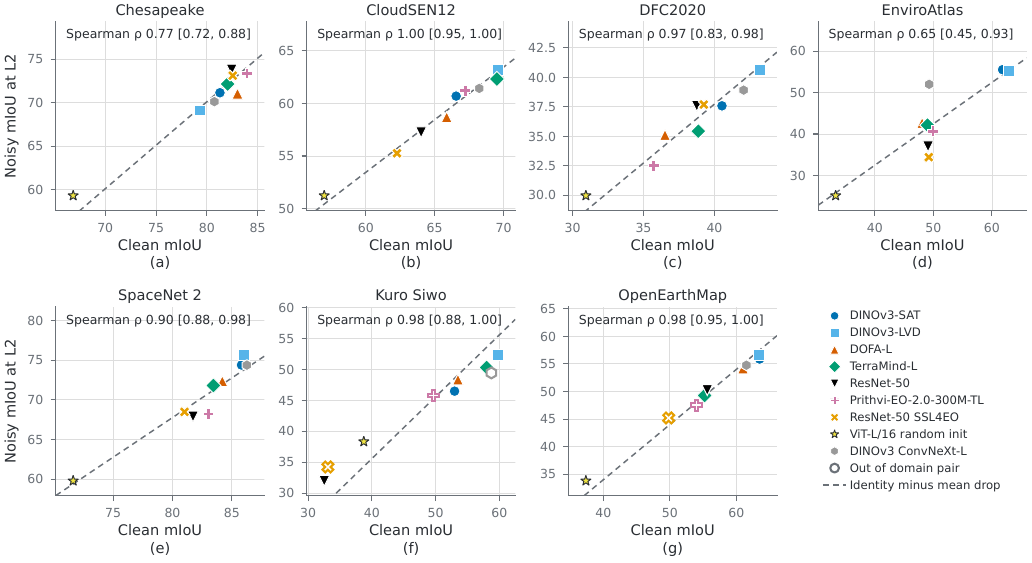}}{\phfig{4cm}{planned/figs/fig\_an\_rank\_agreement\_ext.pdf}}
\caption{\textbf{Clean and noisy accuracy on the extension datasets:} Each marker is one encoder, with its clean test \miou on the horizontal axis and its mean \Ltwo test \miou on the vertical axis. Panels (a) to (g) show the seven extension datasets. A hollow marker is an out-of-domain pair. The dashed line is parallel to the identity.}
\label{fig:supp-rank-ext}
\end{figure*}

\fi
\paragraph{Extension Encoders on the Extension Datasets:}
A further block trains the four extension encoders on the seven extension datasets with the design of the core models (\Cref{tab:supp-n19}).
Every extension dataset then has nine encoders, as the two core datasets do\ifdefined\cvprbuild\else{} (\Cref{fig:supp-rank-ext})\fi{}.
On Chesapeake the Spearman correlation over nine encoders is \hlRankNineChesRho{} [\hlRankNineChesRhoCILo, \hlRankNineChesRhoCIHi].
Its mean over the datasets with nine encoders is \hlRankNinePooledRho{} [\hlRankNinePooledRhoCILo, \hlRankNinePooledRhoCIHi].
A permutation test separates the correlation from zero on \hlRankNineSepCount{} of the \hlRankNineRead{} datasets.

Some pairs are out of domain, because the inputs of the dataset do not match the bands of the pretraining.
Kuro Siwo is one case, where \hlRankNineInDomKuroExcluded{} extension encoders read SAR channels without pretrained weights.
Without those pairs the correlation on Kuro Siwo is \hlRankNineInDomKuroRho{}, against \hlRankNineKuroRho{} with them.
We therefore count the contrasts of the extension encoders against their reference models in two groups.
They cover \hlNNineteenInDomRead{} of \hlNNineteenInDomPairs{} in-domain pairs and \hlNNineteenOutDomRead{} of \hlNNineteenOutDomPairs{} out-of-domain pairs.
On Chesapeake the contrast of \prithvi against \dofa reads \vd{\hlSThreeDropDOFAChesVerdict}.

\section{Where Label Errors Act in the Encoder}
\label{sec:supp-layers}

\ifdefined\cvprbuild
A layer ablation asks where label errors act in the encoder.
We train \sat and \lvd with the last \hlLayArmSix, the last \hlLayArmTwelve{} or all \hlLayDepth{} blocks trainable (\Cref{tab:supp-layers}).
A frozen encoder is the reference, and it uses its own learning rate.
On \flair the drop of \sat under \etype{\hlLayWorstflair} changes by \hlLayStepSATflairTwentyFourTwelveEst{} [\hlLayStepSATflairTwentyFourTwelveCILo, \hlLayStepSATflairTwentyFourTwelveCIHi] points from \hlLayArmTwelve{} to \hlLayDepth{} blocks.
For \lvd the change is \hlLayStepLVDflairTwentyFourTwelveEst{} [\hlLayStepLVDflairTwentyFourTwelveCILo, \hlLayStepLVDflairTwentyFourTwelveCIHi] points.
The ablation holds \hlLayRunsPlanned{} runs over two encoders and two datasets, and all \hlLayRunsDone{} enter the analysis.
\else
This section asks where label errors act in the encoder.
Block N20 trains \sat and \lvd on both core datasets with the last \hlLayArmSix, the last \hlLayArmTwelve{} or all \hlLayDepth{} blocks trainable.
These three arms update \hlLayParamsSix, \hlLayParamsTwelve{} and \hlLayParamsAll{} encoder parameters.
A frozen encoder is the reference without any adaptation.
The labels are clean, corrupted by \etype{shift} at \Ltwo, or corrupted by the selected \Ltwo type of the dataset.
On \burn that type is \etype{\hlLayWorstburn}, and on \flair it is \etype{\hlLayWorstflair}.
The partial arms run \hlLaySeeds{} seeds, and the arm that repeats all trainable blocks with diagnostics runs \hlLaySeedsAll.
The block holds \hlLayRunsPlanned{} runs, and all \hlLayRunsDone{} enter the analysis.
\fi

\begin{table*}[t]
\centering\scriptsize\setlength{\tabcolsep}{4pt}
\IfFileExists{planned/generated/table_n20_depth.tex}{\begin{adjustbox}{max width=\textwidth}%
\providecommand{\gcvdry}[1]{\textcolor{red}{#1}}
\providecommand{\ProvTableNTwentyDepth}{Seeds per cell 5. Clean is the clean test mIoU and Drop the paired drop in points, means over seeds. The last three columns give the drop at the deeper setting minus the drop at the shallower one, paired by seed, with the half-width of the 95\,\% interval. The 24 block columns are the existing runs with all blocks trainable and the 0 block columns follow the frozen recipe. Red is a value of an arm that is still running and a red dash a cell that has not run. Analysis of the finished runs}
\begin{tabular}{@{}lll@{\hspace{7pt}}c@{\hspace{4pt}}c@{\hspace{7pt}}c@{\hspace{4pt}}c@{\hspace{7pt}}c@{\hspace{4pt}}c@{\hspace{7pt}}c@{\hspace{4pt}}c@{\hspace{8pt}}ccc@{}}
\toprule
& & & \multicolumn{2}{c}{0 blocks} & \multicolumn{2}{c}{6 blocks} & \multicolumn{2}{c}{12 blocks} & \multicolumn{2}{c}{24 blocks} & \multicolumn{3}{c}{Change in drop} \\
\cmidrule(lr){4-5} \cmidrule(lr){6-7} \cmidrule(lr){8-9} \cmidrule(lr){10-11} \cmidrule(l){12-14}
Dataset & Type & Model & Clean & Drop & Clean & Drop & Clean & Drop & Clean & Drop & 24 $-$ 12 & 12 $-$ 6 & 6 $-$ 0 \\
\midrule
\burn & \etype{Shift} & \sat & 74.5 & 9.1 & 80.6 & 10.0 & 83.7 & 9.7 & 81.3 & 11.9 & 2.2 $\pm$ 4.9 & $-$0.3 $\pm$ 2.6 & 0.9 $\pm$ 1.2 \\
\burn & \etype{Shift} & \lvd & 74.1 & 9.8 & 78.8 & 11.2 & 81.9 & 11.5 & 82.0 & 10.6 & $-$0.9 $\pm$ 4.2 & 0.3 $\pm$ 2.0 & 1.4 $\pm$ 3.3 \\
\burn & \etype{Coarse} & \sat & 74.5 & 15.0 & 80.6 & 15.3 & 83.7 & 15.5 & 81.3 & 15.2 & $-$0.4 $\pm$ 4.6 & 0.3 $\pm$ 2.2 & 0.3 $\pm$ 2.7 \\
\burn & \etype{Coarse} & \lvd & 74.1 & 16.7 & 78.8 & 15.3 & 81.9 & 15.9 & 82.0 & 15.9 & $-$0.1 $\pm$ 4.0 & 0.7 $\pm$ 5.1 & $-$1.4 $\pm$ 6.6 \\
\midrule
\flair & \etype{Shift} & \sat & 45.5 & 3.1 & 48.5 & 2.1 & 50.3 & 2.6 & 50.5 & 3.0 & 0.4 $\pm$ 0.7 & 0.4 $\pm$ 0.2 & $-$1.0 $\pm$ 1.2 \\
\flair & \etype{Shift} & \lvd & 49.8 & 3.6 & 50.8 & 2.5 & 51.0 & 2.2 & 50.8 & 1.7 & $-$0.5 $\pm$ 1.6 & $-$0.3 $\pm$ 0.2 & $-$1.1 $\pm$ 1.3 \\
\flair & \etype{Omission} & \sat & 45.5 & 2.8 & 48.5 & 4.8 & 50.3 & 5.9 & 50.5 & 6.1 & 0.2 $\pm$ 1.0 & 1.1 $\pm$ 1.2 & 2.0 $\pm$ 2.2 \\
\flair & \etype{Omission} & \lvd & 49.8 & 3.1 & 50.8 & 3.5 & 51.0 & 3.3 & 50.8 & 3.6 & 0.3 $\pm$ 1.9 & $-$0.2 $\pm$ 0.6 & 0.4 $\pm$ 1.1 \\
\bottomrule
\end{tabular}
\end{adjustbox}}{\tabletbd[3cm]{planned/generated/table\_n20\_depth.tex}}
\caption{\textbf{Drop at \Ltwo against the number of trainable encoder blocks:} Clean test \miou and paired drop in points for \sat and \lvd with a frozen encoder and with the last \hlLayArmSix, the last \hlLayArmTwelve{} or all \hlLayDepth{} blocks trainable. The last three columns give the change in drop between neighboring depths, paired by seed, with the half-width of the \CiLevel\,\% interval. The columns for \hlLayDepth{} blocks are the existing runs with all blocks trainable at the $\times$1/3 learning-rate multiplier on seeds 1 to 5. Their \sat cells are therefore those of the N6 arm and differ from the main-grid cells at the \gbtwo rate. The columns for 0 blocks are new frozen runs with their own learning rate and batch size, separate from those of \Cref{tab:supp-frozen}.}
\label{tab:supp-layers}
\end{table*}

\ifdefined\cvprbuild\else
\begin{figure*}[t]
\centering
\IfFileExists{planned/figs/fig_an_layer_ablation.pdf}{\includegraphics[width=\textwidth]{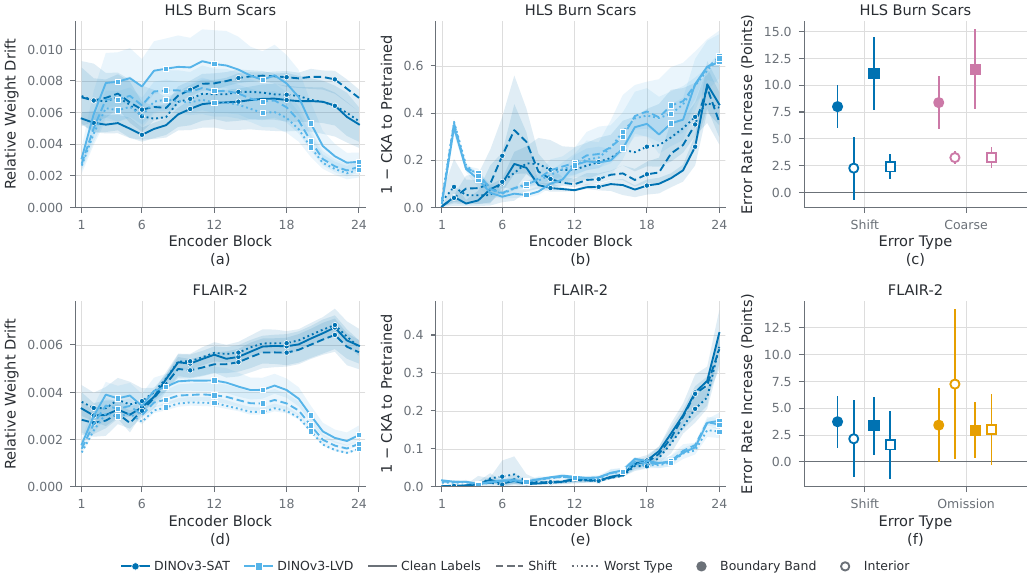}}{\phfig{4cm}{planned/figs/fig\_an\_layer\_ablation.pdf}}
\caption{\textbf{Weight drift and representation change per encoder block:} Panels (a) and (d) show the relative weight drift of each block when all blocks are trainable. Panels (b) and (e) show one minus the linear CKA between the trained and the pretrained encoder. Solid lines are clean labels, dashed lines \etype{shift} and dotted lines the selected type of the dataset. Shaded bands span the seeds. Panels (c) and (f) show the increase of the error rate from clean to noisy labels with \CiLevel\,\% intervals. A filled marker is the boundary band and a hollow marker the interior. The top row is \burn and the bottom row \flair.}
\label{fig:supp-layers}
\end{figure*}

\paragraph{Diagnostics:}
Each run stores three diagnostics of the selected checkpoint.
The drift of a block is the norm of its weight change divided by the norm of its pretrained weights.
Linear centered kernel alignment (CKA) compares the output of each block before and after training on the same test crops.
The third diagnostic splits the test error into a boundary band and the interior.
The band holds every valid pixel within \hlLayBandPx{} pixels of another class in the clean test label.
It covers \hlLayBandFracburn\,\% of the valid test pixels on \burn and \hlLayBandFracflair\,\% on \flair.

\paragraph{Drop Against Trainable Depth:}
\Cref{tab:supp-layers} gives the paired drop at each depth.
The arms with \hlLayArmSix, \hlLayArmTwelve{} and \hlLayDepth{} trainable blocks share one recipe, so we read their trend first.
On \burn the drop of \sat under \etype{\hlLayWorstburn} changes by \hlLayStepSATburnTwentyFourTwelveEst{} [\hlLayStepSATburnTwentyFourTwelveCILo, \hlLayStepSATburnTwentyFourTwelveCIHi] points from \hlLayArmTwelve{} to \hlLayDepth{} blocks.
From \hlLayArmSix{} to \hlLayArmTwelve{} blocks it changes by \hlLayStepSATburnTwelveSixEst{} [\hlLayStepSATburnTwelveSixCILo, \hlLayStepSATburnTwelveSixCIHi] points.
For \lvd the two steps are \hlLayStepLVDburnTwentyFourTwelveEst{} [\hlLayStepLVDburnTwentyFourTwelveCILo, \hlLayStepLVDburnTwentyFourTwelveCIHi] and \hlLayStepLVDburnTwelveSixEst{} [\hlLayStepLVDburnTwelveSixCILo, \hlLayStepLVDburnTwelveSixCIHi] points.
On \flair under \etype{\hlLayWorstflair} the two steps of \sat are \hlLayStepSATflairTwentyFourTwelveEst{} [\hlLayStepSATflairTwentyFourTwelveCILo, \hlLayStepSATflairTwentyFourTwelveCIHi] and \hlLayStepSATflairTwelveSixEst{} [\hlLayStepSATflairTwelveSixCILo, \hlLayStepSATflairTwelveSixCIHi] points.
Those of \lvd are \hlLayStepLVDflairTwentyFourTwelveEst{} [\hlLayStepLVDflairTwentyFourTwelveCILo, \hlLayStepLVDflairTwentyFourTwelveCIHi] and \hlLayStepLVDflairTwelveSixEst{} [\hlLayStepLVDflairTwelveSixCILo, \hlLayStepLVDflairTwelveSixCIHi] points.

The frozen encoder trains with another learning rate and batch size.
We therefore treat it as a reference and not as a point of the trend.
From the frozen encoder to \hlLayArmSix{} trainable blocks, the drop of \sat changes by \hlLayStepSATburnSixZeroEst{} [\hlLayStepSATburnSixZeroCILo, \hlLayStepSATburnSixZeroCIHi] points on \burn.
On \flair it changes by \hlLayStepSATflairSixZeroEst{} [\hlLayStepSATflairSixZeroCILo, \hlLayStepSATflairSixZeroCIHi] points.
For \lvd the changes are \hlLayStepLVDburnSixZeroEst{} [\hlLayStepLVDburnSixZeroCILo, \hlLayStepLVDburnSixZeroCIHi] and \hlLayStepLVDflairSixZeroEst{} [\hlLayStepLVDflairSixZeroCILo, \hlLayStepLVDflairSixZeroCIHi] points.

The same table reads the contrast of the two encoders against depth.
With a frozen encoder, the drop of \sat minus the drop of \lvd is \hlLayPairburnZeroEst{} [\hlLayPairburnZeroCILo, \hlLayPairburnZeroCIHi] points on \burn.
On \flair it is \hlLayPairflairZeroEst{} [\hlLayPairflairZeroCILo, \hlLayPairflairZeroCIHi] points.
With all blocks trainable the differences are \hlLayPairburnTwentyFourEst{} [\hlLayPairburnTwentyFourCILo, \hlLayPairburnTwentyFourCIHi] and \hlLayPairflairTwentyFourEst{} [\hlLayPairflairTwentyFourCILo, \hlLayPairflairTwentyFourCIHi] points.

\paragraph{Drift and CKA per Block:}
\Cref{fig:supp-layers} shows both diagnostics for the arm with all blocks trainable.
We compare each run under the selected type with the clean run of the same seed.
On \burn the drift of \sat is \hlLayDriftSATburnEst{} [\hlLayDriftSATburnCILo, \hlLayDriftSATburnCIHi] times its clean value over all blocks.
The ratio departs most from one in the \vd{\hlLayDriftSATburnThird} third of the blocks, where it is \hlLayDriftSATburnThirdEst.
For \lvd the ratio is \hlLayDriftLVDburnEst{} [\hlLayDriftLVDburnCILo, \hlLayDriftLVDburnCIHi], and \hlLayDriftLVDburnThirdEst{} in the \vd{\hlLayDriftLVDburnThird} third.
On \flair the ratio of \sat is \hlLayDriftSATflairEst{} [\hlLayDriftSATflairCILo, \hlLayDriftSATflairCIHi], and \hlLayDriftSATflairThirdEst{} in the \vd{\hlLayDriftSATflairThird} third.
That of \lvd is \hlLayDriftLVDflairEst{} [\hlLayDriftLVDflairCILo, \hlLayDriftLVDflairCIHi], and \hlLayDriftLVDflairThirdEst{} in the \vd{\hlLayDriftLVDflairThird} third.

One minus CKA measures how far the output of a block moved from the pretrained encoder.
On \burn it changes by \hlLayCkaSATburnEst{} [\hlLayCkaSATburnCILo, \hlLayCkaSATburnCIHi] for \sat under the selected type, most in the \vd{\hlLayCkaSATburnThird} third.
For \lvd it changes by \hlLayCkaLVDburnEst{} [\hlLayCkaLVDburnCILo, \hlLayCkaLVDburnCIHi], most in the \vd{\hlLayCkaLVDburnThird} third.
On \flair the changes are \hlLayCkaSATflairEst{} [\hlLayCkaSATflairCILo, \hlLayCkaSATflairCIHi] for \sat and \hlLayCkaLVDflairEst{} [\hlLayCkaLVDflairCILo, \hlLayCkaLVDflairCIHi] for \lvd.
They are largest in the \vd{\hlLayCkaSATflairThird} third for \sat and the \vd{\hlLayCkaLVDflairThird} third for \lvd.

Drift grows with the number of optimizer steps, and early stopping selects another epoch under label errors.
On \burn the selected epoch of \sat is \hlLayEpochSATburnClean{} with clean labels and \hlLayEpochSATburnNoisy{} under the selected type.
For \lvd the two epochs are \hlLayEpochLVDburnClean{} and \hlLayEpochLVDburnNoisy.
On \flair they are \hlLayEpochSATflairClean{} and \hlLayEpochSATflairNoisy{} for \sat, and \hlLayEpochLVDflairClean{} and \hlLayEpochLVDflairNoisy{} for \lvd.
Per step, the drift ratio is \hlLayDriftStepSATburn{} for \sat and \hlLayDriftStepLVDburn{} for \lvd on \burn.
On \flair it is \hlLayDriftStepSATflair{} and \hlLayDriftStepLVDflair.

\paragraph{Boundary Band Against Interior:}
We ask which share of the added wrong pixels lies in the band, with the band fraction as the chance level.
On \burn the share under \etype{shift} is \hlLayShareSATburnShift\,\% for \sat and \hlLayShareLVDburnShift\,\% for \lvd.
Under \etype{\hlLayWorstburn} it is \hlLayShareSATburnWorst\,\% and \hlLayShareLVDburnWorst\,\%.
On \flair the share under \etype{shift} is \hlLayShareSATflairShift\,\% for \sat and \hlLayShareLVDflairShift\,\% for \lvd.
Under \etype{\hlLayWorstflair} it is \hlLayShareSATflairWorst\,\% and \hlLayShareLVDflairWorst\,\%.

The paired test compares the rise of the error rate in the band with the rise in the interior.
On \burn under \etype{shift}, the band rate of \sat rises by \hlLayBandIntSATburnShiftEst{} [\hlLayBandIntSATburnShiftCILo, \hlLayBandIntSATburnShiftCIHi] points more than the interior rate.
Under \etype{\hlLayWorstburn} the difference is \hlLayBandIntSATburnWorstEst{} [\hlLayBandIntSATburnWorstCILo, \hlLayBandIntSATburnWorstCIHi] points.
For \lvd the two differences are \hlLayBandIntLVDburnShiftEst{} [\hlLayBandIntLVDburnShiftCILo, \hlLayBandIntLVDburnShiftCIHi] and \hlLayBandIntLVDburnWorstEst{} [\hlLayBandIntLVDburnWorstCILo, \hlLayBandIntLVDburnWorstCIHi] points.
On \flair those of \sat are \hlLayBandIntSATflairShiftEst{} [\hlLayBandIntSATflairShiftCILo, \hlLayBandIntSATflairShiftCIHi] and \hlLayBandIntSATflairWorstEst{} [\hlLayBandIntSATflairWorstCILo, \hlLayBandIntSATflairWorstCIHi] points.
Those of \lvd are \hlLayBandIntLVDflairShiftEst{} [\hlLayBandIntLVDflairShiftCILo, \hlLayBandIntLVDflairShiftCIHi] and \hlLayBandIntLVDflairWorstEst{} [\hlLayBandIntLVDflairWorstCILo, \hlLayBandIntLVDflairWorstCIHi] points.

Freezing more blocks could protect the interior.
With \hlLayArmSix{} trainable blocks, the interior error rate of \sat rises by \hlLayIntSATburnSix{} points under the selected type on \burn.
With all blocks trainable it rises by \hlLayIntSATburnTwentyFour{} points.
For \lvd the two rises are \hlLayIntLVDburnSix{} and \hlLayIntLVDburnTwentyFour{} points.
On \flair they are \hlLayIntSATflairSix{} and \hlLayIntSATflairTwentyFour{} points for \sat, and \hlLayIntLVDflairSix{} and \hlLayIntLVDflairTwentyFour{} for \lvd.

\paragraph{Reproducibility and Caveats:}
The arm with all blocks trainable repeats existing cells with the diagnostics switched on.
On \burn its test \miou differs from the existing run by \hlLayReproburnAbs{} points on average over \hlLayReproburnN{} pairs, and by at most \hlLayReproburnMax.
On \flair the mean difference is \hlLayReproflairAbs{} points over \hlLayReproflairN{} pairs, and the largest is \hlLayReproflairMax.

Five limits apply to this section.
The frozen encoder has its own learning rate and batch size, so the step from zero to \hlLayArmSix{} blocks changes both.
Drift is confounded with the number of optimizer steps, which the ratio per step only partly removes.
Below the frozen part, a drift of zero and a CKA of one hold by construction.
Over \hlLayFrozenRuns{} runs the largest drift of a frozen group is \hlLayFrozenMaxDrift{} and the smallest such CKA is \hlLayFrozenMinCka.
All diagnostics use the clean test labels, and the block covers two encoders and two datasets only.
\fi

\section{Further Analysis Figures}
\label{sec:ax-analysis}

This section adds diagrams that read the existing result tables from angles the earlier figures do not cover.
Each figure answers one question, and its plotted values will be released with the code.

\paragraph{Verdict Forest:}
\Cref{fig:ax-an-verdict-forest} shows contrasts of \mtab{tab:verdicts} and \Cref{tab:supp-verdicts-post} against the equivalence margin.
The matched-rate contrast of the DINOv3 pair lies inside the margin on both datasets, while the contrast at the selected rates leaves it.
For the \gfms against \rn the contrast is equivalent on both datasets.
On \flair the decoder and loss contrasts are equivalent, and two head contrasts stay undetermined on \burn.

\ifdefined\cvprbuild%
\begin{figure*}[t]
\centering
\IfFileExists{planned/figs/fig_an_verdict_forest.pdf}{\includegraphics[width=\textwidth]{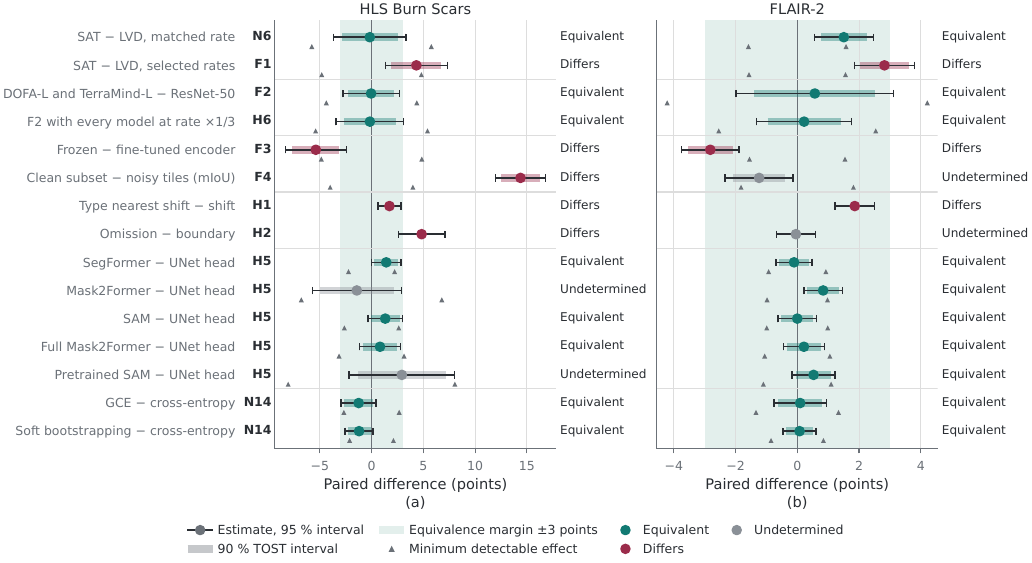}}{\phfig{4cm}{planned/figs/fig\_an\_verdict\_forest.pdf from figs/make\_analysis\_diagrams.py}}
\caption{\textbf{Pre-specified contrasts against the equivalence margin:} Panel (a) shows \burn and panel (b) \flair. The H5, H6 and N14 rows are the decoder, common-rate and loss contrasts of \Cref{tab:supp-verdicts-post}. The other rows are contrasts of \mtab{tab:verdicts}. SAT is \sat{}, LVD is \lvd{} and GCE is generalized cross-entropy. The dot is the estimate, the thin bar its \CiLevel\,\% interval and the thick bar its TOST interval. The shaded band is the margin of $\pm$\TostMargin{} points, and the small triangles mark the minimum detectable effect on both sides of zero. The word at the right is the reading. A contrast is equivalent when its TOST interval lies inside the band, and otherwise it differs when its \CiLevel\,\% interval excludes zero. On \flair the H2, F3 and F4 rows rest on \etype{omission} realizations, and they read undetermined where their raw and rescaled readings disagree. The rows H1 and H2 carry no TOST interval. F4 is a difference in test \miou, and every other row is a difference in drop.}
\label{fig:ax-an-verdict-forest}
\end{figure*}
\fi

\paragraph{Spread over Seeds:}
Mean bars hide how far single runs lie apart.
\Cref{fig:ax-an-seed-spread} therefore shows every paired drop at \Ltwo.
The spread is widest under \etype{omission} and \etype{coarse} on \burn, where the collapsed runs also sit.
On \flair the drops of one model stay within a few points of each other.
The order of the models is therefore read with more confidence on \flair than on \burn.

\ifdefined\cvprbuild%
\begin{figure*}[t]
\centering
\IfFileExists{planned/figs/fig_an_seed_spread.pdf}{\includegraphics[width=\textwidth]{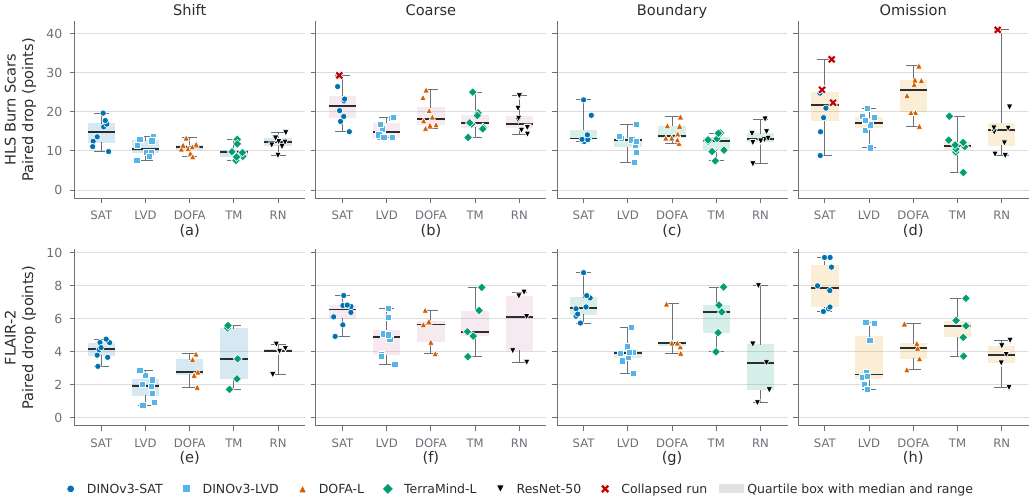}}{\phfig{4cm}{planned/figs/fig\_an\_seed\_spread.pdf from figs/make\_analysis\_diagrams.py}}
\caption{\textbf{Paired drops of single runs at \Ltwo:} Each marker is the paired drop of one seed, per core model and error type. Panels (a) to (d) show \burn and panels (e) to (h) show \flair. The box spans the quartiles, the line inside it is the median and the whiskers reach the extreme runs. A red cross marks a collapsed run on \burn. SAT is \sat, LVD is \lvd, DOFA is \dofa, TM is \tm and RN is \rn.}
\label{fig:ax-an-seed-spread}
\end{figure*}
\fi

\ifdefined\cvprbuild\else
\paragraph{Rank of the Error Types:}
The rank of the four error types is stable at the bottom and unstable at the top (\Cref{fig:ax-an-type-rank}).
The type \etype{shift} costs least for most models at both levels. At \Ltwo it ranks last in every seed on both datasets.
The costliest type changes between models and between \Lone and \Ltwo.
A single worst type per dataset is thus a weaker statement than a single mildest type.

\begin{figure*}[t]
\centering
\IfFileExists{planned/figs/fig_an_type_rank.pdf}{\includegraphics[width=\textwidth]{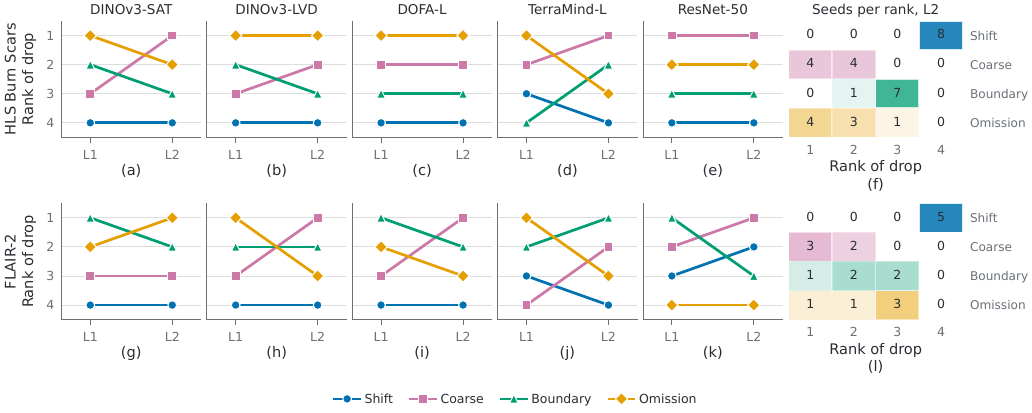}}{\phfig{4cm}{planned/figs/fig\_an\_type\_rank.pdf from figs/make\_analysis\_diagrams.py}}
\caption{\textbf{Rank of the four error types per model and level:} Panels (a) to (e) and (g) to (k) show the rank of the mean paired drop of each type at \Lone and \Ltwo. Rank one is the largest drop. The top row is \burn and the bottom row \flair. Panels (f) and (l) count the seeds in which a type takes a rank at \Ltwo. That rank is taken on the mean drop over the five core models, over the seeds that every model has.}
\label{fig:ax-an-type-rank}
\end{figure*}

\paragraph{Classes of \flair:}
The loss on \flair is concentrated in two built classes and not in the rare classes (\Cref{fig:ax-an-perclass}).
\emph{Building} and \emph{Impervious surface} lose most.
Their largest loss comes under \etype{coarse}.
The pixel share of a class does not order the drops, because rare and frequent vegetation classes lose similar amounts.
The class \emph{Other} has a clean IoU near zero and so has nothing to lose.

\begin{figure*}[t]
\centering
\IfFileExists{planned/figs/fig_an_perclass_flair.pdf}{\includegraphics[width=\textwidth]{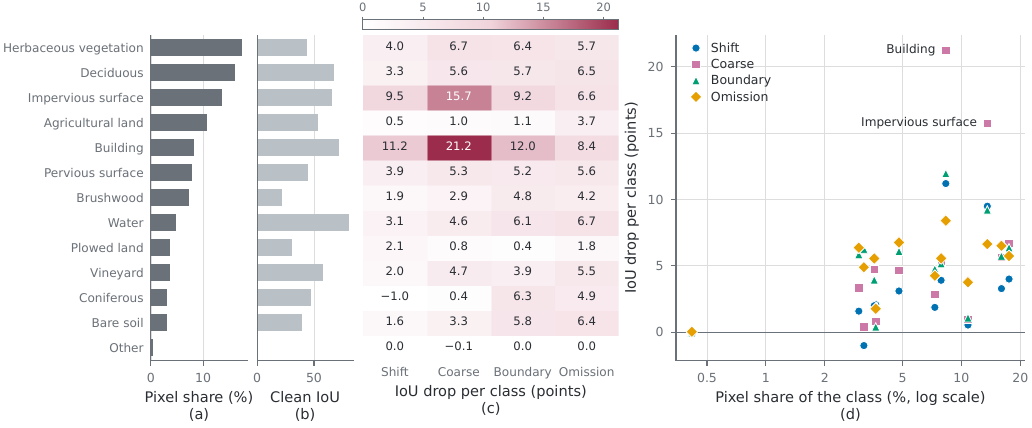}}{\phfig{4cm}{planned/figs/fig\_an\_perclass\_flair.pdf from figs/make\_analysis\_diagrams.py}}
\caption{\textbf{Per-class IoU drop on \flair at \Ltwo:} Classes are ordered by their pixel share, which panel (a) shows. Panel (b) gives the clean IoU of each class and panel (c) its IoU drop under each error type. Both are means over the five core models and their seeds. Panel (d) plots the same drops against the pixel share on a logarithmic axis, with one marker per class and type.}
\label{fig:ax-an-perclass}
\end{figure*}

\fi
\paragraph{Agreement Between Models:}
Most pairs of models are equivalent on \flair and undetermined on \burn (\Cref{fig:ax-an-model-agreement}).
On \flair the pairs that differ under \etype{boundary} and \etype{omission} all involve \sat, which loses more than its partner.
On \burn few pairs reach equivalence, because the seed spread is wide against the margin.
The figure therefore supports a ranking of models only where a cell reads as a difference.

\ifdefined\cvprbuild%
\begin{figure*}[t]
\centering
\IfFileExists{planned/figs/fig_an_model_agreement.pdf}{\includegraphics[width=\textwidth]{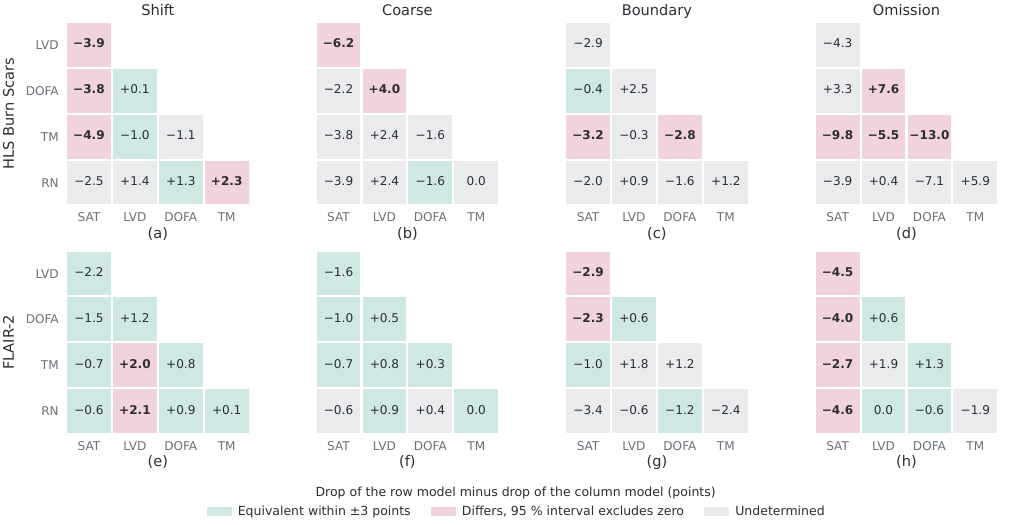}}{\phfig{4cm}{planned/figs/fig\_an\_model\_agreement.pdf from figs/make\_analysis\_diagrams.py}}
\caption{\textbf{Pairwise drop differences between the core models at \Ltwo:} Each cell is the paired drop of the row model minus that of the column model, in points, over the seeds both models have. The top row is \burn and the bottom row \flair, with one panel per error type. A green cell is equivalent, since its TOST interval lies within $\pm$\TostMargin{} points. A red cell with a bold value differs, since its \CiLevel\,\% interval excludes zero and its TOST interval leaves the margin. A gray cell is undetermined. SAT is \sat, LVD is \lvd, DOFA is \dofa, TM is \tm and RN is \rn.}
\label{fig:ax-an-model-agreement}
\end{figure*}
\fi

\ifdefined\cvprbuild\else
\paragraph{Resolution of the Contrasts:}
Whether a contrast can be called equivalent depends on its minimum detectable effect (\Cref{fig:ax-an-power}).
On \flair nearly every contrast resolves the margin.
On \burn the only contrasts that do so pool the five models.
An undetermined row on \burn is thus a statement about resolution and not about the models.

\begin{figure}[t]
\centering
\IfFileExists{planned/figs/fig_an_power.pdf}{\includegraphics[width=\linewidth]{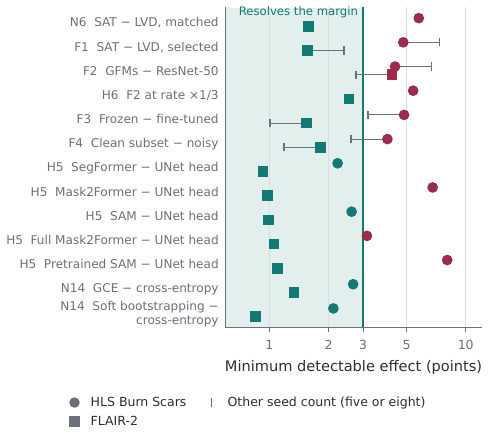}}{\phfig{4cm}{planned/figs/fig\_an\_power.pdf from figs/make\_analysis\_diagrams.py}}
\caption{\textbf{Minimum detectable effect of each contrast against the margin:} Circles are \burn and squares \flair, on a logarithmic axis. The vertical line is the margin of $\pm$\TostMargin{} points, and a green marker in the shaded region resolves it. For F1 to F4 a tick marks the minimum detectable effect at the other seed count of the design.}
\label{fig:ax-an-power}
\end{figure}

\paragraph{Memorization Across Models:}
Memorization depends more on the error type than on the model (\Cref{fig:ax-an-memorisation}).
Every core model memorizes \etype{coarse} labels most on both datasets, and the models differ most under \etype{omission}.
Runs that memorize more also lose more.
This relation is closer on \burn than on \flair.
The figure extends the memorization table of the DINOv3 pair to all five core models and all four types.

\begin{figure*}[t]
\centering
\IfFileExists{planned/figs/fig_an_memorisation.pdf}{\includegraphics[width=\textwidth]{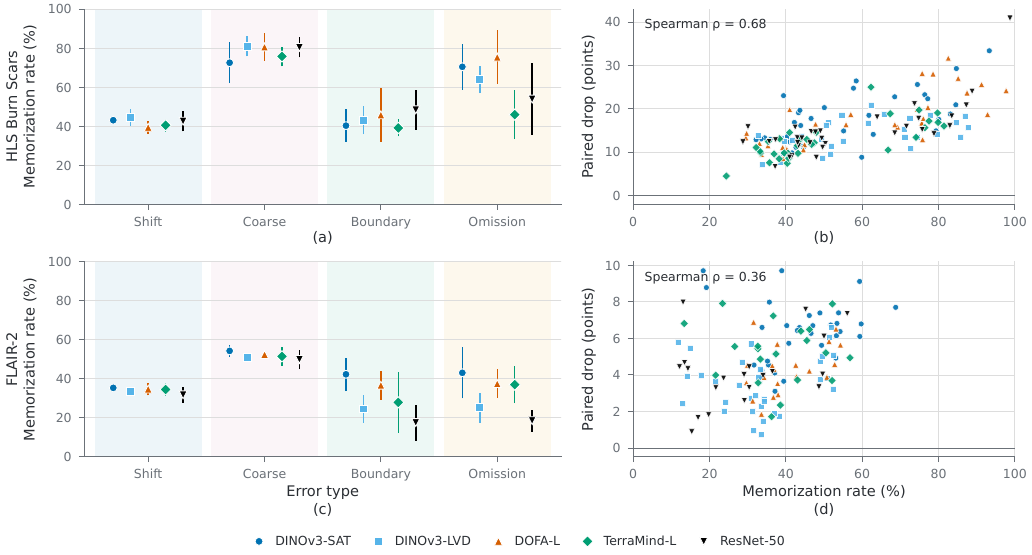}}{\phfig{4cm}{planned/figs/fig\_an\_memorisation.pdf from figs/make\_analysis\_diagrams.py}}
\caption{\textbf{Memorization rate of the five core models at \Ltwo:} Panels (a) and (c) show the memorization rate \memrate{} at the selected checkpoint per error type, as the mean over seeds with its \CiLevel\,\% interval. Panels (b) and (d) plot the paired drop of every run against its memorization rate. The printed value is the Spearman rank correlation over all runs of the dataset. The top row is \burn and the bottom row \flair.}
\label{fig:ax-an-memorisation}
\end{figure*}

\paragraph{Checkpoint Selection under Label Errors:}
Label errors move the selected checkpoint and bias the validation score that selects it (\Cref{fig:ax-an-training-dynamics}).
Most models stop earlier at \Ltwo than on clean labels.
Under \etype{coarse} the validation \miou on corrupted labels is higher than on clean labels for every model.
Under the other three types it is lower for most models, so the corrupted validation split understates the model.

\begin{figure*}[t]
\centering
\IfFileExists{planned/figs/fig_an_training_dynamics.pdf}{\includegraphics[width=\textwidth]{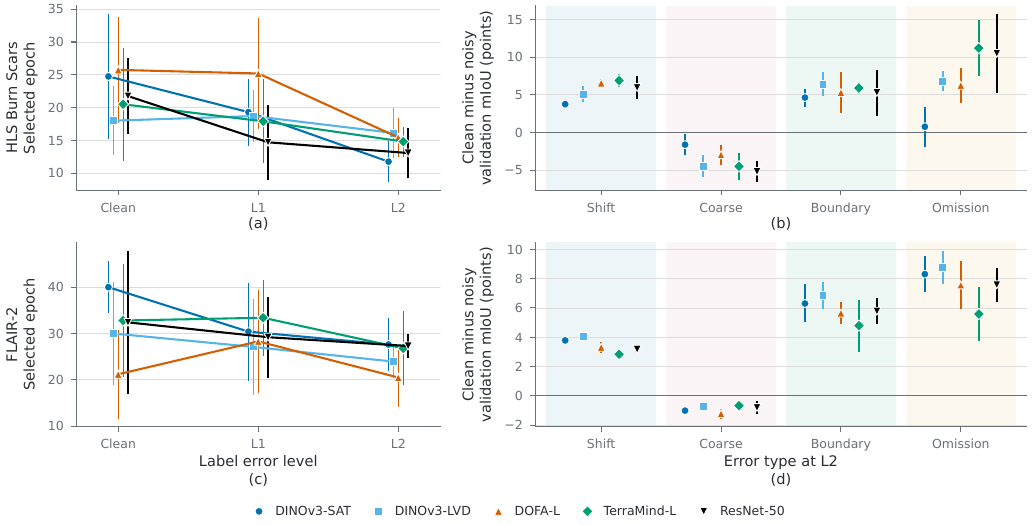}}{\phfig{4cm}{planned/figs/fig\_an\_training\_dynamics.pdf from figs/make\_analysis\_diagrams.py}}
\caption{\textbf{Selected epoch and validation gap of the five core models:} Panels (a) and (c) show the epoch of the selected checkpoint on clean labels, at \Lone and at \Ltwo, averaged over the four error types. Panels (b) and (d) show the validation \miou on clean labels minus that on corrupted labels at the selected checkpoint, per type at \Ltwo. Bars are \CiLevel\,\% intervals over seeds. The top row is \burn and the bottom row \flair.}
\label{fig:ax-an-training-dynamics}
\end{figure*}

\paragraph{Spread of the Noise Realizations:}
The calibrated knobs hit their label \miou target for three types and miss it for \etype{omission} (\Cref{fig:ax-an-realisations}).
Several \etype{omission} realizations leave the tolerance band on both datasets.
The paired drop falls as the realized label \miou rises, which is the slope used for the adjusted contrasts.
The \etype{omission} rows of \mtab{tab:verdicts} therefore carry this caveat.

\begin{figure*}[t]
\centering
\IfFileExists{planned/figs/fig_an_realisations.pdf}{\includegraphics[width=\textwidth]{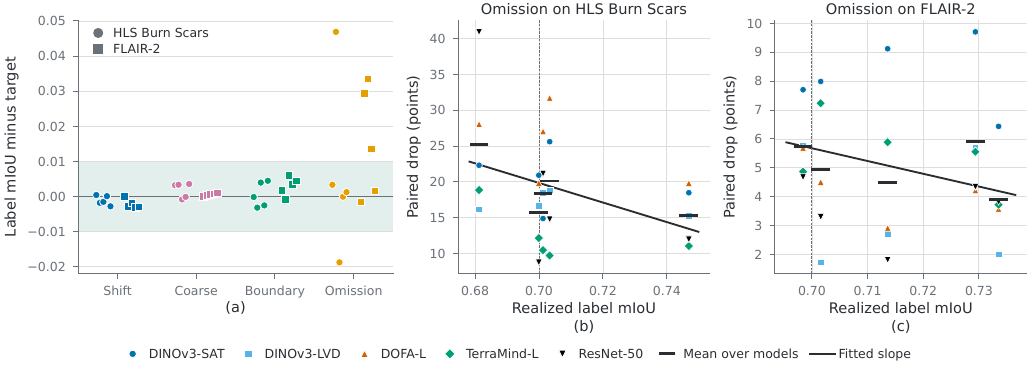}}{\phfig{4cm}{planned/figs/fig\_an\_realisations.pdf from figs/make\_analysis\_diagrams.py}}
\caption{\textbf{Realized label \miou per noise realization and its effect on the drop:} Panel (a) shows the label \miou of each \Ltwo realization minus its target, with circles for \burn and squares for \flair. The shaded band is the calibration tolerance. Panels (b) and (c) plot the paired \etype{omission} drop of each core model against the realized label \miou of its noise realization. The short bars are means over models, the line is the fitted slope and the dashed line is the target.}
\label{fig:ax-an-realisations}
\end{figure*}

\paragraph{Decoder Heads per Model:}
The effect of the decoder head depends on the model on \burn and is absent on \flair (\Cref{fig:ax-an-decoder-interaction}).
On \flair nearly every head and model is equivalent to the UNet decoder.
On \burn the Mask2Former head lowers the drop of \sat and \rn.
It raises the drop of \lvd.
The pooled contrast of that head therefore hides opposite changes of single models.

\begin{figure*}[t]
\centering
\IfFileExists{planned/figs/fig_an_decoder_interaction.pdf}{\includegraphics[width=\textwidth]{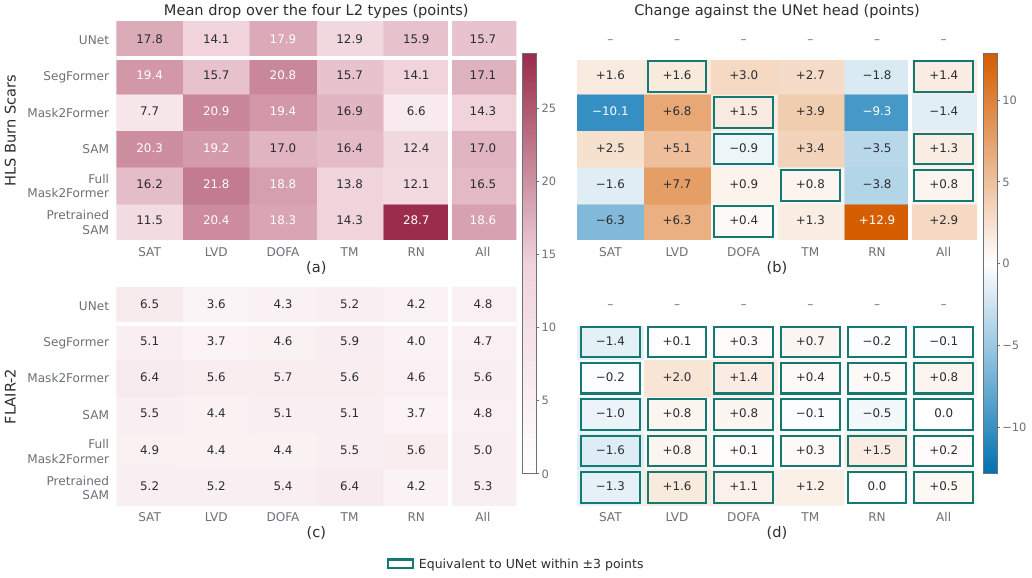}}{\phfig{4cm}{planned/figs/fig\_an\_decoder\_interaction.pdf from figs/make\_analysis\_diagrams.py}}
\caption{\textbf{Drop per decoder head and core model:} Panels (a) and (c) show the mean paired drop over the four \Ltwo types per head and model. Panels (b) and (d) show the change against the UNet head on the same seeds. The top row is \burn and the bottom row \flair. The column All pools the five models. A green frame marks a change that is equivalent to zero within $\pm$\TostMargin{} points. A dash marks the UNet reference row. SAT is \sat, LVD is \lvd, DOFA is \dofa, TM is \tm and RN is \rn.}
\label{fig:ax-an-decoder-interaction}
\end{figure*}

\paragraph{Hardware Classes:}
Cells that ran on two GPU classes show no systematic offset between the classes (\Cref{fig:ax-an-hardware-mix}).
The mean difference over cells stays inside the margin for every model and class pair.
Single cells leave the band, most often for \sat on \burn.
The seed spread is also widest there.
The two classes of a cell hold different seeds, so each difference includes seed variation.

\begin{figure*}[t]
\centering
\IfFileExists{planned/figs/fig_an_hardware_mix.pdf}{\includegraphics[width=\textwidth]{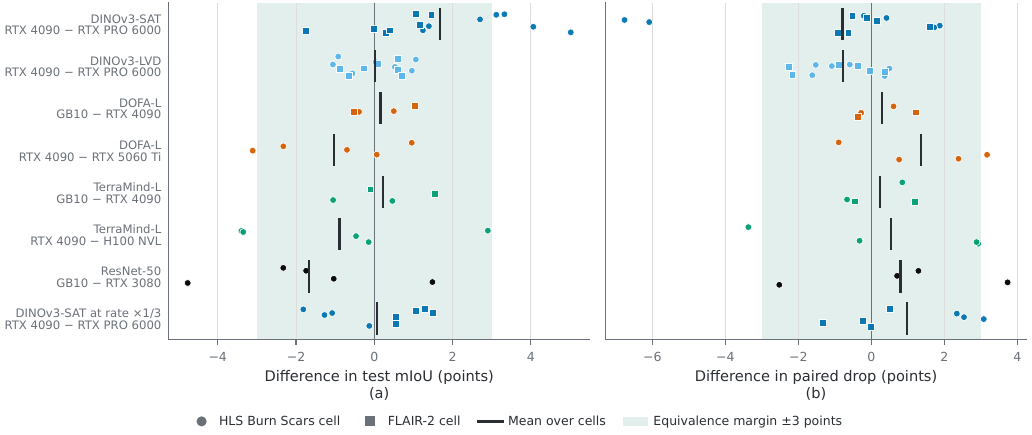}}{\phfig{4cm}{planned/figs/fig\_an\_hardware\_mix.pdf from figs/make\_analysis\_diagrams.py}}
\caption{\textbf{Differences between GPU classes within one cell:} Each marker is one cell of the design that ran on two GPU classes. Its value is the mean over the seeds on the first class minus the mean over the seeds on the second class. Panel (a) shows test \miou and panel (b) the paired drop. Circles are \burn cells and squares \flair cells, the black bar is the mean over cells and the shaded band the margin of $\pm$\TostMargin{} points.}
\label{fig:ax-an-hardware-mix}
\end{figure*}

\paragraph{Model Profiles:}
No core model is best on every axis of the study (\Cref{fig:ax-an-model-profile}).
On \flair \sat has the highest clean \miou and the largest drops, while \lvd leads on the drop axes.
On \burn \rn leads on clean \miou and on the clean subset, and \tm has the smallest drop at \Ltwo.
A choice between the models thus depends on which axis the application weighs.

\begin{figure*}[t]
\centering
\IfFileExists{planned/figs/fig_an_model_profile.pdf}{\includegraphics[width=\textwidth]{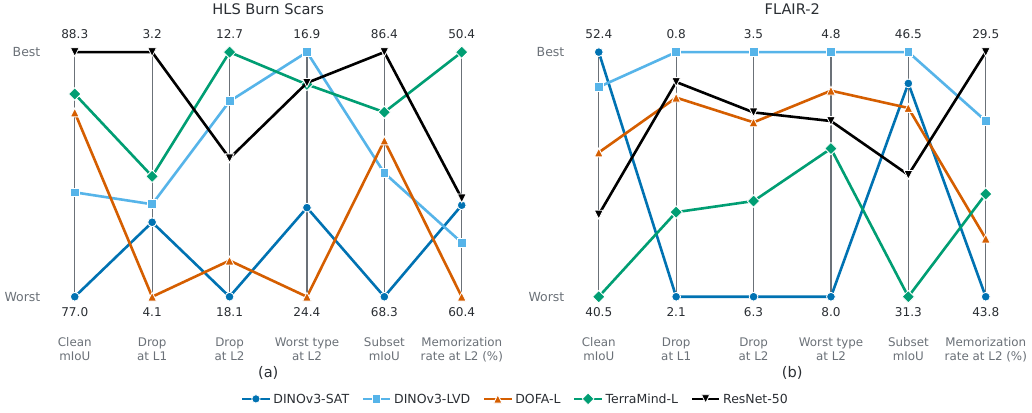}}{\phfig{4cm}{planned/figs/fig\_an\_model\_profile.pdf from figs/make\_analysis\_diagrams.py}}
\caption{\textbf{Profiles of the five core models:} Each vertical axis is one quantity, scaled between the worst and the best model of the dataset with the best at the top. The values printed above and below an axis are those of the best and the worst model. The first three axes are the clean test \miou and the mean drop at \Lone and at \Ltwo. The last three are the drop under the worst \Ltwo type, the clean subset test \miou and the \Ltwo memorization rate. Panel (a) is \burn and panel (b) \flair.}
\label{fig:ax-an-model-profile}
\end{figure*}

\paragraph{Estimates Against the Number of Seeds:}
The two family contrasts settle at different seed counts (\Cref{fig:ax-an-seed-convergence}).
For F1 the interval excludes zero after three or four seeds, and later seeds only narrow it.
On \burn the F2 interval enters the margin only with the last added seeds.
The equivalence reading of F2 on \burn therefore rests on the extended seeds.

\begin{figure*}[t]
\centering
\IfFileExists{planned/figs/fig_an_seed_convergence.pdf}{\includegraphics[width=\textwidth]{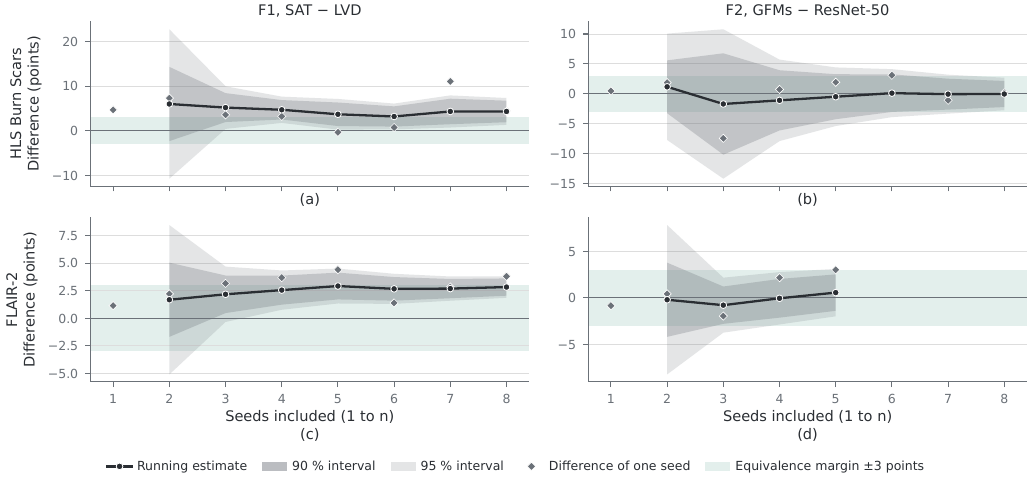}}{\phfig{4cm}{planned/figs/fig\_an\_seed\_convergence.pdf from figs/make\_analysis\_diagrams.py}}
\caption{\textbf{Running estimates of F1 and F2 as seeds are added:} The line is the mean paired difference over the first $n$ seeds, the dark band its TOST interval and the light band its \CiLevel\,\% interval. Diamonds are the differences of single seeds, and the shaded horizontal band is the margin of $\pm$\TostMargin{} points. The top row is \burn and the bottom row \flair. F1 is \sat minus \lvd and F2 is the mean of \dofa and \tm minus \rn.}
\label{fig:ax-an-seed-convergence}
\end{figure*}
\fi

\else
\ifdefined\arxivbuild The appendix adds views of the main-text results in the order of its sections, from calibration to statistics.
\else
This supplement adds views of the main-paper results in the order of its sections, from calibration to statistics.
\fi
It then details the extension blocks with S1 on \sfloods{} (\Cref{sec:supp-stretch}) and closes with the block table, input bands and hardware.
\fi

\section{Label Error Calibration}
\label{sec:supp-calibration}

\begin{table}[t]
\centering\footnotesize\setlength{\tabcolsep}{3pt}
\IfFileExists{generated/table_calibration.tex}{%
\begin{tabular}{@{}llrrrrrr@{}}
\toprule
 & & \multicolumn{3}{c}{L1} & \multicolumn{3}{c}{L2} \\
\cmidrule(lr){3-5}\cmidrule(l){6-8}
Type & Knob & Value & mIoU & Flip (\%) & Value & mIoU & Flip (\%) \\
\midrule
\multicolumn{8}{@{}l}{\textit{HLS Burn Scars}} \\
Shift & Offset (px) & 20 & 0.848 & 3.4 & 60 & 0.700 & 7.8 \\
Coarse & Block (px) & 57 & 0.850 & 3.3 & 192 & 0.703 & 6.7 \\
Boundary & Width (px) & 11 & 0.850 & 3.4 & 32 & 0.700 & 8.2 \\
Omission & Drop prob. & 0.28 & 0.849 & 3.0 & 0.53 & 0.703 & 5.9 \\
\midrule
\multicolumn{8}{@{}l}{\textit{FLAIR-2}} \\
Shift & Offset (px) & 10 & 0.848 & 9.5 & 29 & 0.700 & 21.1 \\
Coarse & Block (px) & 29 & 0.849 & 9.6 & 105 & 0.700 & 21.2 \\
Boundary & Width (px) & 16 & 0.846 & 8.4 & 52 & 0.702 & 18.2 \\
Omission & Drop prob. & 0.07 & 0.849 & 7.4 & 0.18 & 0.698 & 17.4 \\
\bottomrule
\end{tabular}
}{\tabletbd[4cm]{generated/table\_calibration.tex: run scripts/make\_config\_macros.py}}
\caption{\textbf{Calibrated knob values:} For each dataset, error type and level, the knob value in pixels (px) or as a drop probability. The mIoU column is the label \miou of the corrupted training and validation masks against the clean masks for realization \CalSeed. The flip rate is the percentage of labeled pixels whose label changed. Targets are \CalTargetLOne{} at \Lone and \CalTargetLTwo{} at \Ltwo with tolerance $\pm$\CalTol. A drop probability $p$ keeps a share $1-p$ of the objects. In ground units at \Lone and \Ltwo the offset is \KnobMBurnShiftLOne\,\KnobMUnitBurnShiftLOne{} and \KnobMBurnShiftLTwo\,\KnobMUnitBurnShiftLTwo{} on \burn and \KnobMFlairShiftLOne\,\KnobMUnitFlairShiftLOne{} and \KnobMFlairShiftLTwo\,\KnobMUnitFlairShiftLTwo{} on \flair. The block is \KnobMBurnCoarseLOne\,\KnobMUnitBurnCoarseLOne{} and \KnobMBurnCoarseLTwo\,\KnobMUnitBurnCoarseLTwo{} on \burn and \KnobMFlairCoarseLOne\,\KnobMUnitFlairCoarseLOne{} and \KnobMFlairCoarseLTwo\,\KnobMUnitFlairCoarseLTwo{} on \flair. The width is \KnobMBurnBoundaryLOne\,\KnobMUnitBurnBoundaryLOne{} and \KnobMBurnBoundaryLTwo\,\KnobMUnitBurnBoundaryLTwo{} on \burn and \KnobMFlairBoundaryLOne\,\KnobMUnitFlairBoundaryLOne{} and \KnobMFlairBoundaryLTwo\,\KnobMUnitFlairBoundaryLTwo{} on \flair. Over all \NRealisations{} realizations, label \miou lies between \RealMiouMinLOne{} and \RealMiouMaxLOne{} at \Lone and between \RealMiouMinLTwo{} and \RealMiouMaxLTwo{} at \Ltwo.}
\label{tab:supp-calibration}
\end{table}

\ifdefined\cvprbuild\else
\begin{figure}[t]
\centering
\IfFileExists{figs/noise_mechanics.pdf}{\includegraphics[width=\linewidth]{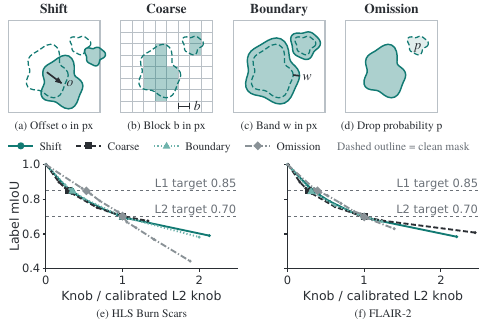}}{\phfig{6.2cm}{figs/noise\_mechanics.pdf from figs/make\_noise\_mechanics\_fig.py}}
\caption{\textbf{Label corruption mechanics and calibration:} (a) to (d) The knob of each error type on a schematic mask. The dashed outline is the clean mask and the filled shape the corrupted one. (a) \etype{Shift} translates every label by an offset $o$ in a random direction. (b) \etype{Coarse} replaces each block of $b\times b$ pixels by its majority label. (c) \etype{Boundary} dilates or erodes one class by a band of width $w$. (d) \etype{Omission} removes each connected object with probability $p$. (e) and (f) The bisection history of the calibration on \burn and \flair. The horizontal axis is the knob relative to its calibrated \Ltwo value and the vertical axis the label \miou of realization \CalSeed. Small dots are the probes of the search, large markers the accepted \Lone and \Ltwo knobs and the dashed lines the targets \CalTargetLOne{} and \CalTargetLTwo.}
\label{fig:supp-mechanics}
\end{figure}
\fi

\begin{table}[t]
\centering\footnotesize\setlength{\tabcolsep}{4pt}
\begin{tabular}{@{}llccccc@{}}
\toprule
Dataset & Type & R1 & R2 & R3 & R4 & R5 \\
\midrule
\burn & \etype{Shift} & \RealMiouBurnShiftLTwoROne & \RealMiouBurnShiftLTwoRTwo & \RealMiouBurnShiftLTwoRThree & \RealMiouBurnShiftLTwoRFour & \RealMiouBurnShiftLTwoRFive \\
 & \etype{Coarse} & \RealMiouBurnCoarseLTwoROne & \RealMiouBurnCoarseLTwoRTwo & \RealMiouBurnCoarseLTwoRThree & \RealMiouBurnCoarseLTwoRFour & \RealMiouBurnCoarseLTwoRFive \\
 & \etype{Boundary} & \RealMiouBurnBoundaryLTwoROne & \RealMiouBurnBoundaryLTwoRTwo & \RealMiouBurnBoundaryLTwoRThree & \RealMiouBurnBoundaryLTwoRFour & \RealMiouBurnBoundaryLTwoRFive \\
 & \etype{Omission} & \RealMiouBurnOmissionLTwoROne & \RealMiouBurnOmissionLTwoRTwo & \RealMiouBurnOmissionLTwoRThree & \RealMiouBurnOmissionLTwoRFour & \RealMiouBurnOmissionLTwoRFive \\
\addlinespace[2pt]
\flair & \etype{Shift} & \RealMiouFlairShiftLTwoROne & \RealMiouFlairShiftLTwoRTwo & \RealMiouFlairShiftLTwoRThree & \RealMiouFlairShiftLTwoRFour & \RealMiouFlairShiftLTwoRFive \\
 & \etype{Coarse} & \RealMiouFlairCoarseLTwoROne & \RealMiouFlairCoarseLTwoRTwo & \RealMiouFlairCoarseLTwoRThree & \RealMiouFlairCoarseLTwoRFour & \RealMiouFlairCoarseLTwoRFive \\
 & \etype{Boundary} & \RealMiouFlairBoundaryLTwoROne & \RealMiouFlairBoundaryLTwoRTwo & \RealMiouFlairBoundaryLTwoRThree & \RealMiouFlairBoundaryLTwoRFour & \RealMiouFlairBoundaryLTwoRFive \\
 & \etype{Omission} & \RealMiouFlairOmissionLTwoROne & \RealMiouFlairOmissionLTwoRTwo & \RealMiouFlairOmissionLTwoRThree & \RealMiouFlairOmissionLTwoRFour & \RealMiouFlairOmissionLTwoRFive \\
\bottomrule
\end{tabular}
\caption{\textbf{Label \miou of \Ltwo realizations R1 to R5:} Label \miou in points of the corrupted training and validation masks against the clean masks. These realizations train seeds 1 to 5. The knob of each type was calibrated on realization \CalSeed{} to the target \CalTargetLTwo{} with tolerance $\pm$\CalTol, and the other realizations reuse it. Target and tolerance are fractions, and the table prints points, a factor of 100 apart. The \etype{omission} rows span \RealRangeBurnOmissionLTwo{} points on \burn and \RealRangeFlairOmissionLTwo{} on \flair. Every \flair omission realization outside the tolerance is milder than the target. The drops of those realizations are rescaled to the target in \Cref{tab:supp-omadj} and their verdicts marked in the verdict table of the main paper. Values are read from the calibration files.}
\label{tab:supp-realizations}
\end{table}

\ifdefined\cvprbuild
\Cref{tab:supp-calibration} gives the calibrated knobs and \Cref{tab:supp-realizations} the \Ltwo label \miou of realizations 1 to 5.
\else
\Cref{fig:supp-mechanics} shows the knob of each error type and the calibration histories, \Cref{tab:supp-calibration} the calibrated knobs and \Cref{tab:supp-realizations} the label \miou of every realization.
\fi
The main paper shows one \burn training tile under the four types.
The \etype{shift} operator translates the label map of each tile by an offset in a random direction.
It fills the uncovered border with the nearest valid label.
The \etype{coarse} operator replaces each block of pixels by its majority label.
Its block grid has a random offset per tile, and no-data pixels are excluded from the vote.
The \etype{boundary} operator picks one class present in the tile, the burned area on \burn.
It dilates or erodes that class by a fixed width with equal probability.
The \etype{omission} operator removes each connected component of a class with a fixed probability and fills it with the surrounding label.
The class is the burned area on \burn, and every class is eligible on \flair.
At \Ltwo the selected type, \etype{coarse}, leaves the burned class of \burn at a label IoU of \CalIouBurnedBurnCoarseLTwo\,\% in realization \CalSeed. The \flair classes keep \CalIouClassMinFlairOmissionLTwo{} to \CalIouClassMaxFlairOmissionLTwo\,\% under \etype{omission}.
No-data pixels never move, are never copied into labeled regions and are never removed from them.

Realizations of the same knob vary little in severity for \etype{shift}, \etype{coarse} and \etype{boundary} and much more for \etype{omission}.
The range of label \miou is taken over realizations 1 to 5. For the first three types, it is at most \RealRangeBurnBoundaryLTwo{} points on \burn and \RealRangeFlairBoundaryLTwo{} on \flair.
The \etype{omission} type spans \RealRangeBurnOmissionLTwo{} points on \burn.
There realizations \RealOffTolBurnOmissionLTwo{} fall outside the tolerance with \RealMiouBurnOmissionLTwoRTwo{} and \RealMiouBurnOmissionLTwoRThree{} points against the \CalTargetLTwo{} target.
On \flair the \etype{omission} type spans \RealRangeFlairOmissionLTwo{} points. Realizations \RealOffTolFlairOmissionLTwo{} fall outside the tolerance, all milder than the target.
Realizations 6 to 8 serve the eight-seed cells. Among them \RealOffTolExtBurnOmissionLTwo{} of \etype{omission} on \burn fall outside the tolerance.
On \flair so do \RealOffTolExtFlairBoundaryLTwo{} of \etype{boundary} and \RealOffTolExtFlairOmissionLTwo{} of \etype{omission}.
The \etype{omission} knob is a drop probability applied to a few hundred connected objects of unequal size.
Which objects a realization removes therefore moves the label \miou by several points.
The other three knobs act on every pixel or every boundary of the tile and average out.
Realization \RealHarshestBurnOmissionLTwo{}, the harshest of R1 to R5 on \burn, is the outlier seed of \rn and \tm in their \etype{omission} \Ltwo cells.
That \rn cell has the widest interval of the main grid ($\pm$\dropCIRNburnOmissionLTwo{} points, \Cref{tab:supp-main}).
Part of its seed variance is therefore realization severity rather than training noise, which a label \miou covariate would absorb.

That covariate defines the realization adjustment.
The per-model-centered drop changes by \hlRealSlopeburnOmission{} points per label \miou point on \burn and by \hlRealSlopeflairOmission{} on \flair.
\Cref{tab:supp-omadj} rescales every \etype{omission} drop to the target with that slope.
The slope is common to all models, so the adjustment is the same for every model at a seed. It leaves the between-model contrasts F1 and F2 unchanged.
The adjusted intervals treat the slope as known and are therefore conditional on it.
Over (model, seed) units the omission mean is \hlOmRawMeanburn{} points raw and \hlOmAdjMeanburn{} adjusted on \burn, and \hlOmRawMeanflair{} and \hlOmAdjMeanflair{} on \flair.
Adjusted, the H2 contrast is \hlHTwoAdjburn{} [\hlHTwoAdjCILoburn, \hlHTwoAdjCIHiburn] on \burn and \hlHTwoAdjflair{} [\hlHTwoAdjCILoflair, \hlHTwoAdjCIHiflair] on \flair.
The \burn verdict therefore stands.
The \flair row is undetermined because its raw and adjusted readings disagree. \mtab{tab:verdicts} applies this rule, which was added after the runs. %
With the seed as the unit (\Cref{sec:supp-grid}) the adjusted \flair contrast is \hlHTwoAdjSeedflairEst{} [\hlHTwoAdjSeedflairCILo, \hlHTwoAdjSeedflairCIHi] and includes zero.
On \flair the selected type is \etype{omission}.
The F3 and F4 arms there carry the same realizations and are rescaled per (model, seed) in the same way.
For the frozen arm of F3 the rescaling uses its own slope of \hlFThreeAdjflairSlopeArm{} points per point from the A3 runs.
The clean-subset arm of F4 needs no rescaling.
F3 is \hlFThreeRawflairEst{} [\hlFThreeRawflairCILo, \hlFThreeRawflairCIHi] raw and \hlFThreeAdjflairEst{} [\hlFThreeAdjflairCILo, \hlFThreeAdjflairCIHi] adjusted, and keeps its verdict.
F4 moves from \hlFFourRawflairEst{} [\hlFFourRawflairCILo, \hlFFourRawflairCIHi] to \hlFFourAdjflairEst{} [\hlFFourAdjflairCILo, \hlFFourAdjflairCIHi] with Holm $p$ \hlFFourAdjflairPHolm, so its verdict in the main paper is undetermined.
After adjustment \etype{omission} moves from rank \hlOmRawRankflair{} to rank \hlOmAdjRankflair{} of the four \Ltwo types on \flair.

Every between-model contrast pairs models on the same realization.
The collapsed \burn runs are described in \Cref{sec:supp-collapsed}.
The reference labels are themselves imperfect, because \flair acknowledges photo-interpretation errors \cite{garioud2023flair2} and burn masks derived from regional inventories include unburned land \cite{kolden2012mapped}.

\begin{table}[t]
\centering\footnotesize\setlength{\tabcolsep}{2pt}
\IfFileExists{generated/table_omission_adjusted.tex}{\begin{adjustbox}{max width=\columnwidth}%
\providecommand{\gcvdry}[1]{\textcolor{red}{#1}}
\providecommand{\ProvOmissionAdjusted}{Paired drop in \miou points at \etype{omission} \Ltwo. Adjusted = raw + slope $\times$ (target $-$ label \miou of the realization), with the per-dataset slope of the per-model-centered drop on the realization label \miou (-1.36 on \burn, -0.44 on \flair points per point). The H2 row gives the contrast over (model, seed) units with its 95\,\% $t$ interval, the same units as the verdict table. The F3 and F4 rows give the family contrast over seeds on the dataset whose selected type is \etype{omission}, with the Holm $p$ within the family of two datasets. A dash marks a dataset whose selected type carries no adjustment. Bold marks the smallest drop of a column among the five models. The frozen arm of F3 is rescaled with its own slope (-0.42 on \flair). Analysis of the finished runs.}
\begin{tabular}{@{}lcccc@{}}
\toprule
 & \multicolumn{2}{c}{\burn} & \multicolumn{2}{c}{\flair} \\
\cmidrule(lr){2-3}\cmidrule(l){4-5}
Model & Raw & Adjusted & Raw & Adjusted \\
\midrule
\sat & 21.2 & 21.1 & 8.0 & 8.8 \\
\lvd & 16.9 & 16.8 & \textbf{3.4} & \textbf{4.2} \\
\dofa & 24.4 & 24.4 & 4.2 & 4.9 \\
\tm & \textbf{11.4} & \textbf{11.4} & 5.5 & 6.1 \\
\rn & 17.3 & 17.3 & 3.6 & 4.3 \\
Mean & 18.2 & 18.2 & 5.1 & 5.8 \\
\addlinespace[2pt]
H2, \etype{omission} $-$ \etype{boundary} & 4.8 & 4.8 & 0.0 & 0.7 \\
95\,\% interval & [2.6, 7.1] & [2.7, 6.9] & [$-$0.7, 0.6] & [0.1, 1.3] \\
\addlinespace[2pt]
F3, drop frozen $-$ fine-tuned & -- & -- & $-$2.8 & $-$2.8 \\
95\,\% interval & -- & -- & [$-$3.8, $-$1.9] & [$-$3.8, $-$1.9] \\
Holm $p$ & -- & -- & 0.002 & 0.002 \\
\addlinespace[2pt]
F4, \miou clean subset $-$ noisy & -- & -- & $-$1.2 & $-$0.6 \\
95\,\% interval & -- & -- & [$-$2.3, $-$0.1] & [$-$2.0, 0.9] \\
Holm $p$ & -- & -- & 0.036 & 0.353 \\
\bottomrule
\end{tabular}
\end{adjustbox}}{\tabletbd[2.6cm]{generated/table\_omission\_adjusted.tex: run the analysis harness}}
\caption{\textbf{Raw and realization-adjusted \etype{omission} drops at \Ltwo:} Paired drop in \miou points per core model and the mean over (model, seed) units. On \flair that mean weights the eight-seed DINOv3 cells more than the model-equal mean of \Cref{tab:supp-main}. Adjusted drops are rescaled to the calibration target with the per-dataset slope of the per-model-centered drop on the realization label \miou. Bold marks the smallest drop of a column among the five models. The H2 row is \etype{omission} minus \etype{boundary} over (model, seed) units of \mtab{tab:verdicts}, with its \CiLevel\,\% $t$ interval. The F3 and F4 rows give the family contrasts over seeds on \flair, whose selected type is \etype{omission}, with the Holm $p$ of the family. A dash marks the F3 and F4 rows on \burn, where the selected type is \etype{coarse}. \ProvSlot{\ResultsProvenance}} %
\label{tab:supp-omadj}
\end{table}

\ifdefined\cvprbuild\else
\section{Clean Accuracy Against GEO-Bench-2}
\label{sec:supp-acceptance}

\begin{table}[t]
\centering\footnotesize\setlength{\tabcolsep}{3pt}
\IfFileExists{generated/table_acceptance.tex}{\begin{adjustbox}{max width=\columnwidth}%
\providecommand{\gcvdry}[1]{\textcolor{red}{#1}}
\providecommand{\ProvAcceptance}{Seeds per cell 5 to 8 for our clean runs, 5 for N4. Ours is mean $\pm$ 95\,\% $t$ half-width. Analysis of the finished runs.}
\begin{tabular}{@{}lcccc@{}}
\toprule
 & \multicolumn{2}{c}{\burn} & \multicolumn{2}{c}{\flair} \\
\cmidrule(lr){2-3}\cmidrule(l){4-5}
Model & Ours & \gbtwo & Ours & \gbtwo \\
\midrule
\sat & 77.0 $\pm$ 2.6 & 69.4 & \textbf{52.4} $\pm$ 0.3 & \textbf{54.7} \\
\lvd & 81.8 $\pm$ 1.3 & -- & 50.7 $\pm$ 0.7 & -- \\
\dofa & 85.5 $\pm$ 0.7 & \textbf{89.2} & 47.5 $\pm$ 1.0 & 52.7 \\
\tm & 86.3 $\pm$ 1.0 & 88.4 & 40.5 $\pm$ 1.2 & 46.8 \\
\rn & \textbf{88.3} $\pm$ 0.8 & 74.4 & 44.5 $\pm$ 1.7 & 46.7 \\
\midrule
\rn RGB only (N4) & 78.9 $\pm$ 1.7 & 74.4 & -- & -- \\
\gbtwo multispectral \rn (range) & -- & 87.9--89.3 & -- & -- \\
\bottomrule
\end{tabular}
\end{adjustbox}}{\tabletbd[2.6cm]{generated/table\_acceptance.tex: run the analysis harness}}
\caption{\textbf{Clean test \miou against \gbtwo:} Rows are the core models, columns give our clean \miou and the \gbtwo mean over its \NSeeds{} seeds for \burn and \flair. On \flair our models train on the \NFlairTrain-tile subset. The \rn row is compared with both the RGB-only \rn of \gbtwo and its multispectral \rn entries. Our values are means with their \CiLevel\,\% $t$ half-width. The N4 row is our RGB-only \rn on \burn. Bold marks the best value of a column among the five core models. A dash marks a cell without an entry. \ProvSlot{\ResultsProvenance}}
\label{tab:supp-acceptance}
\end{table}

\Cref{tab:supp-acceptance} compares the clean test \miou of every core model with its \gbtwo entry.
Four deviations from the \gbtwo entries stand out. They are \sat on \burn, the \rn band set on \burn, the \flair training subset and the \dofa and \tm recipe on \burn.

\fi
\section{Main Grid Results}
\label{sec:supp-grid}

\begin{table*}[t]
\centering\footnotesize
\IfFileExists{generated/table_main.tex}{%
\providecommand{\gcvdry}[1]{\textcolor{red}{#1}}
\providecommand{\ProvMain}{Seeds per cell 5 to 8. Values are mean $\pm$ 95\,\% $t$ half-width, drops in points. Bold marks the largest clean mIoU and the smallest drop per column within a dataset. Analysis of the finished runs.}
\begin{tabular}{@{}lccccc@{}}
\toprule
Model & Clean mIoU & Shift & Coarse & Boundary & Omission \\
\midrule
\multicolumn{6}{@{}l}{\textit{\burn}} \\
\sat & 77.0 $\pm$ 2.6 & 14.7 $\pm$ 2.9 & 21.7 $\pm$ 4.0 & 15.1 $\pm$ 3.2 & 21.2 $\pm$ 6.2 \\
\lvd & 81.8 $\pm$ 1.3 & 10.8 $\pm$ 1.9 & \textbf{15.5} $\pm$ 1.8 & 12.2 $\pm$ 2.4 & 16.9 $\pm$ 2.5 \\
\dofa & 85.5 $\pm$ 0.7 & 10.9 $\pm$ 1.2 & 19.4 $\pm$ 3.0 & 14.7 $\pm$ 1.9 & 24.4 $\pm$ 4.4 \\
\tm & 86.3 $\pm$ 1.0 & \textbf{9.8} $\pm$ 1.5 & 17.9 $\pm$ 2.9 & \textbf{11.9} $\pm$ 2.1 & \textbf{11.4} $\pm$ 3.3 \\
\rn & \textbf{88.3} $\pm$ 0.8 & 12.1 $\pm$ 1.5 & 17.8 $\pm$ 2.7 & 13.1 $\pm$ 2.7 & 17.3 $\pm$ 8.7 \\
Mean over models & 83.8 & 11.7 & 18.4 & 13.4 & 18.2 \\
\midrule
\multicolumn{6}{@{}l}{\textit{\flair}} \\
\sat & \textbf{52.4} $\pm$ 0.3 & 4.1 $\pm$ 0.5 & 6.3 $\pm$ 0.7 & 6.9 $\pm$ 0.8 & 8.0 $\pm$ 1.2 \\
\lvd & 50.7 $\pm$ 0.7 & \textbf{1.8} $\pm$ 0.6 & \textbf{4.8} $\pm$ 1.0 & 3.9 $\pm$ 0.7 & \textbf{3.4} $\pm$ 1.4 \\
\dofa & 47.5 $\pm$ 1.0 & 2.9 $\pm$ 1.0 & 5.3 $\pm$ 1.3 & 4.8 $\pm$ 1.5 & 4.2 $\pm$ 1.3 \\
\tm & 40.5 $\pm$ 1.2 & 3.7 $\pm$ 2.2 & 5.6 $\pm$ 2.0 & 6.1 $\pm$ 1.9 & 5.5 $\pm$ 1.6 \\
\rn & 44.5 $\pm$ 1.7 & 3.8 $\pm$ 0.9 & 5.7 $\pm$ 2.4 & \textbf{3.7} $\pm$ 3.5 & 3.6 $\pm$ 1.4 \\
Mean over models & 47.1 & 3.3 & 5.6 & 5.1 & 4.9 \\
\bottomrule
\end{tabular}
}{\tabletbd[3.8cm]{generated/table\_main.tex: run the analysis harness}}
\caption{\textbf{Paired drop in \miou points at \Ltwo:} For each core model and dataset, the clean \miou and the paired drop of each error type. Half-widths are the \CiLevel\,\% $t$ interval over the $n$ seeds of the cell\ifdefined\cvprbuild . The top block\else{} (\Cref{tab:supp-grid-burn,tab:supp-grid-flair}). The top block\fi{} shows \burn and the bottom block \flair. The last row of each block averages over models. Bold marks the largest clean \miou and the smallest drop per column within a dataset. \ProvSlot{\ResultsProvenance}}
\label{tab:supp-main}
\end{table*}

\ifdefined\cvprbuild\else
\begin{figure*}[t]
\centering
\IfFileExists{figs/fig_dose_response.pdf}{\includegraphics[width=\textwidth]{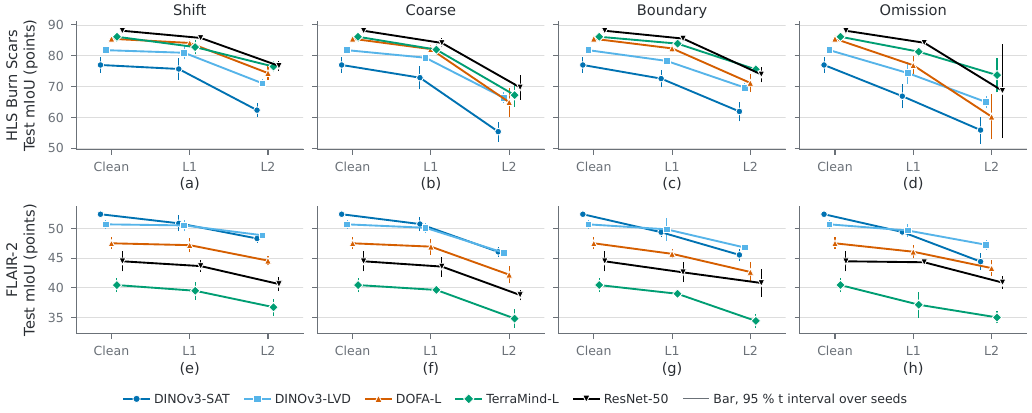}}{\phfig{5cm}{figs/fig\_dose\_response.pdf from figs/make\_dose\_response\_fig.py}}
\caption{\textbf{Test \miou against label error level:} Mean test \miou on the clean reference labels at the clean condition and at levels \Lone and \Ltwo for the five core models. Bars are the \CiLevel\,\% $t$ interval over the seeds of the cell. (a) to (d) \burn and (e) to (h) \flair, with one panel per error type in the order \etype{shift}, \etype{coarse}, \etype{boundary} and \etype{omission}. Each model keeps its color and marker in every figure of the paper and \ifdefined\arxivbuild these appendices\else the supplement\fi.}
\label{fig:supp-dose}
\end{figure*}

\begin{figure}[t]
\centering
\IfFileExists{figs/fig_type_drop.pdf}{\includegraphics[width=\linewidth]{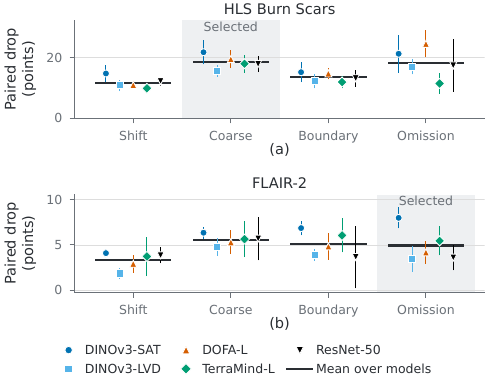}}{\phfig{6.5cm}{figs/fig\_type\_drop.pdf from figs/make\_type\_drop\_fig.py}}
\caption{\textbf{Paired drop by error type at \Ltwo:} One marker per core model with its \CiLevel\,\% $t$ interval over the seeds of the cell and a black bar for the mean over models. (a) \burn and (b) \flair. The shaded column is the selected type of each dataset.}
\label{fig:supp-typedrop}
\end{figure}
\fi
\ifdefined\cvprbuild\else

\Cref{tab:supp-main} lists the \Ltwo paired drops, \cref{fig:supp-dose} shows the test \miou of every cell against the error level and \cref{fig:supp-typedrop} the drops per model with intervals.
The retention heatmap of the main paper shows the retention of every main-grid cell.
\Cref{tab:supp-grid-burn,tab:supp-grid-flair} give the test \miou, drop and retention of every cell at both levels.
\fi
\ifdefined\cvprbuild
\Cref{tab:supp-main} lists the \Ltwo paired drops of the five core models.
\fi
On \burn \etype{\hlTypeRankburnOne} and \etype{\hlTypeRankburnTwo} are not separated, \hlTypeRankDiffburnOne{} [\hlTypeRankDiffCILoburnOne, \hlTypeRankDiffCIHiburnOne] points apart. Both exceed the other two types.
Every error type costs more on \burn than on \flair.
\burn has \NBurnClasses{} classes, so every corrupted burned pixel moves one of two IoU terms, whereas \flair spreads the budget over \NFlairClasses{} classes.

The H1 and H2 intervals of \mtab{tab:verdicts} use (model, seed) units. %
These units share the realization of a seed across models, and on \flair they weight the eight-seed DINOv3 cells more.
With the seed as the unit, the drop of each type is averaged over the five core models within each common seed. There are \hlHTwoSeedburnN{} common seeds on \burn and \hlHTwoSeedflairN{} on \flair.
The contrast is then paired across seeds.
H1 is read as the drop of the type ranked just above \etype{shift} minus the drop of \etype{shift}, so a positive value contradicts H1.
It is then \hlHOneSeedburnEst{} [\hlHOneSeedburnCILo, \hlHOneSeedburnCIHi] on \burn and \hlHOneSeedflairEst{} [\hlHOneSeedflairCILo, \hlHOneSeedflairCIHi] on \flair.
H2 is \hlHTwoSeedburnEst{} [\hlHTwoSeedburnCILo, \hlHTwoSeedburnCIHi] on \burn and \hlHTwoSeedflairEst{} [\hlHTwoSeedflairCILo, \hlHTwoSeedflairCIHi] on \flair.
After the realization adjustment of \Cref{sec:supp-calibration} it is \hlHTwoAdjSeedburnEst{} [\hlHTwoAdjSeedburnCILo, \hlHTwoAdjSeedburnCIHi] on \burn and \hlHTwoAdjSeedflairEst{} [\hlHTwoAdjSeedflairCILo, \hlHTwoAdjSeedflairCIHi] on \flair.
At the seed unit both \burn H2 intervals exclude zero, whereas both \flair intervals include zero.
The H1 interval excludes zero on both datasets at the seed unit.

On \burn the \Lone ranking by mean drop is \etype{\hlTypeRankLOneburnOne}, \etype{\hlTypeRankLOneburnTwo}, \etype{\hlTypeRankLOneburnThree} and \etype{\hlTypeRankLOneburnFour}. On \flair it is \etype{\hlTypeRankLOneflairOne}, \etype{\hlTypeRankLOneflairTwo}, \etype{\hlTypeRankLOneflairThree} and \etype{\hlTypeRankLOneflairFour}.
It rests on \Lone cells of which \hlHwMixedAOneLOneNburn{} per dataset hold runs from two GPU classes (\Cref{sec:supp-hardware}).
The ranking at a common flip rate is given in \Cref{sec:supp-fliprate}.

\ifdefined\cvprbuild\else
Seed counts differ between cells.
Seeds \hlSeedExtraSeeds{} are held in \hlSeedExtraCellsN{} cells. These are the clean and \Ltwo cells of all five core models on \burn and of the DINOv3 pair on \flair.
The N5 block is complete.
\Cref{tab:supp-grid-burn,tab:supp-grid-flair} print $n$ per cell.
Every paired contrast uses the seeds common to its arms and the clean run of the same seed as reference.

\begin{table}[t]
\centering\footnotesize\setlength{\tabcolsep}{3pt}
\IfFileExists{generated/table_full_grid_burn.tex}{%
\providecommand{\gcvdry}[1]{\textcolor{red}{#1}}
\providecommand{\ProvFullGrid}{Seeds per cell 5 to 8. Mean $\pm$ standard deviation over seeds. Bold marks the largest mIoU, the smallest drop and the largest retention of each type and level across the five models. Analysis of the finished runs.}
\begin{tabular}{@{}lllcccc@{}}
\toprule
Model & Type & Level & $n$ & mIoU & Drop & Retention (\%) \\
\midrule
SAT & Clean & -- & 8 & 77.0 $\pm$ 3.1 & -- & -- \\
 & Shift & L1 & 5 & 75.7 $\pm$ 2.8 & \textbf{0.1} $\pm$ 3.9 & \textbf{99.9} $\pm$ 5.1 \\
 & Shift & L2 & 8 & 62.3 $\pm$ 2.6 & 14.7 $\pm$ 3.5 & 81.0 $\pm$ 4.0 \\
 & Coarse & L1 & 5 & 72.9 $\pm$ 3.0 & 3.0 $\pm$ 3.5 & 96.1 $\pm$ 4.6 \\
 & Coarse & L2 & 8 & 55.3 $\pm$ 3.9 & 21.7 $\pm$ 4.7 & 72.0 $\pm$ 5.6 \\
 & Boundary & L1 & 5 & 72.6 $\pm$ 2.3 & 3.3 $\pm$ 3.5 & 95.8 $\pm$ 4.6 \\
 & Boundary & L2 & 8 & 61.9 $\pm$ 3.8 & 15.1 $\pm$ 3.9 & 80.4 $\pm$ 4.7 \\
 & Omission & L1 & 5 & 66.9 $\pm$ 3.2 & 9.0 $\pm$ 2.7 & 88.2 $\pm$ 3.7 \\
 & Omission & L2 & 8 & 55.8 $\pm$ 5.2 & 21.2 $\pm$ 7.4 & 72.8 $\pm$ 8.7 \\
LVD & Clean & -- & 8 & 81.8 $\pm$ 1.6 & -- & -- \\
 & Shift & L1 & 5 & 81.0 $\pm$ 1.7 & 1.1 $\pm$ 2.4 & 98.7 $\pm$ 2.9 \\
 & Shift & L2 & 8 & 71.0 $\pm$ 1.6 & 10.8 $\pm$ 2.2 & 86.9 $\pm$ 2.5 \\
 & Coarse & L1 & 5 & 79.4 $\pm$ 0.8 & \textbf{2.7} $\pm$ 1.2 & \textbf{96.8} $\pm$ 1.4 \\
 & Coarse & L2 & 8 & 66.4 $\pm$ 2.1 & \textbf{15.5} $\pm$ 2.2 & \textbf{81.1} $\pm$ 2.6 \\
 & Boundary & L1 & 5 & 78.3 $\pm$ 0.6 & 3.7 $\pm$ 1.0 & 95.5 $\pm$ 1.2 \\
 & Boundary & L2 & 8 & 69.6 $\pm$ 1.6 & 12.2 $\pm$ 2.9 & 85.2 $\pm$ 3.3 \\
 & Omission & L1 & 5 & 74.4 $\pm$ 2.8 & 7.6 $\pm$ 2.8 & 90.7 $\pm$ 3.4 \\
 & Omission & L2 & 8 & 64.9 $\pm$ 2.4 & 16.9 $\pm$ 3.0 & 79.4 $\pm$ 3.5 \\
DOFA & Clean & -- & 8 & 85.5 $\pm$ 0.8 & -- & -- \\
 & Shift & L1 & 5 & 84.2 $\pm$ 1.1 & 1.4 $\pm$ 0.5 & 98.4 $\pm$ 0.6 \\
 & Shift & L2 & 8 & 74.6 $\pm$ 1.5 & 10.9 $\pm$ 1.5 & 87.3 $\pm$ 1.7 \\
 & Coarse & L1 & 5 & 82.2 $\pm$ 1.0 & 3.4 $\pm$ 1.7 & 96.1 $\pm$ 2.0 \\
 & Coarse & L2 & 8 & 66.1 $\pm$ 3.5 & 19.4 $\pm$ 3.6 & 77.3 $\pm$ 4.1 \\
 & Boundary & L1 & 5 & 82.4 $\pm$ 0.8 & 3.1 $\pm$ 1.3 & 96.4 $\pm$ 1.5 \\
 & Boundary & L2 & 8 & 70.8 $\pm$ 2.0 & 14.7 $\pm$ 2.3 & 82.8 $\pm$ 2.6 \\
 & Omission & L1 & 5 & 76.9 $\pm$ 2.3 & 8.6 $\pm$ 3.0 & 90.0 $\pm$ 3.4 \\
 & Omission & L2 & 8 & 61.1 $\pm$ 5.4 & 24.4 $\pm$ 5.3 & 71.4 $\pm$ 6.2 \\
TM & Clean & -- & 8 & 86.3 $\pm$ 1.1 & -- & -- \\
 & Shift & L1 & 5 & 82.8 $\pm$ 1.9 & 3.4 $\pm$ 2.4 & 96.1 $\pm$ 2.7 \\
 & Shift & L2 & 8 & \textbf{76.5} $\pm$ 1.2 & \textbf{9.8} $\pm$ 1.8 & \textbf{88.7} $\pm$ 2.0 \\
 & Coarse & L1 & 5 & 82.0 $\pm$ 1.5 & 4.2 $\pm$ 2.2 & 95.2 $\pm$ 2.5 \\
 & Coarse & L2 & 8 & 68.5 $\pm$ 3.1 & 17.9 $\pm$ 3.5 & 79.3 $\pm$ 3.9 \\
 & Boundary & L1 & 5 & 83.9 $\pm$ 1.0 & \textbf{2.2} $\pm$ 1.3 & \textbf{97.4} $\pm$ 1.5 \\
 & Boundary & L2 & 8 & 74.5 $\pm$ 1.8 & \textbf{11.9} $\pm$ 2.5 & \textbf{86.3} $\pm$ 2.8 \\
 & Omission & L1 & 5 & 81.3 $\pm$ 1.3 & 4.9 $\pm$ 1.8 & 94.4 $\pm$ 2.0 \\
 & Omission & L2 & 8 & \textbf{75.0} $\pm$ 4.4 & \textbf{11.4} $\pm$ 4.0 & \textbf{86.8} $\pm$ 4.7 \\
RN & Clean & -- & 8 & \textbf{88.3} $\pm$ 0.9 & -- & -- \\
 & Shift & L1 & 5 & \textbf{85.8} $\pm$ 0.5 & 2.3 $\pm$ 0.6 & 97.4 $\pm$ 0.6 \\
 & Shift & L2 & 8 & 76.1 $\pm$ 1.4 & 12.1 $\pm$ 1.7 & 86.3 $\pm$ 1.9 \\
 & Coarse & L1 & 5 & \textbf{84.2} $\pm$ 1.2 & 4.0 $\pm$ 1.4 & 95.5 $\pm$ 1.6 \\
 & Coarse & L2 & 8 & \textbf{70.5} $\pm$ 2.7 & 17.8 $\pm$ 3.3 & 79.8 $\pm$ 3.5 \\
 & Boundary & L1 & 5 & \textbf{85.6} $\pm$ 0.6 & 2.5 $\pm$ 1.3 & 97.1 $\pm$ 1.5 \\
 & Boundary & L2 & 8 & \textbf{75.2} $\pm$ 3.0 & 13.1 $\pm$ 3.2 & 85.2 $\pm$ 3.6 \\
 & Omission & L1 & 5 & \textbf{84.2} $\pm$ 1.0 & \textbf{3.9} $\pm$ 1.0 & \textbf{95.6} $\pm$ 1.1 \\
 & Omission & L2 & 8 & 71.0 $\pm$ 10.0 & 17.3 $\pm$ 10.4 & 80.5 $\pm$ 11.6 \\
\bottomrule
\end{tabular}
}{\tabletbd[6cm]{generated/table\_full\_grid\_burn.tex: run scripts/reflow\_tables.py}}
\caption{\textbf{Every main-grid cell on \burn:} Test \miou, paired drop and retention for every core model, error type and level. Cells give the mean $\pm$ standard deviation over the $n$ seeds. Bold marks the largest \miou, the smallest drop and the largest retention of each type and level across the five models. A dash marks a quantity that does not exist for the clean row. Models are abbreviated as SAT (\sat), LVD (\lvd), DOFA (\dofa), TM (\tm) and RN (\rn). \ProvSlot{\ResultsProvenance}}
\label{tab:supp-grid-burn}
\end{table}

\begin{table}[t]
\centering\footnotesize\setlength{\tabcolsep}{3pt}
\IfFileExists{generated/table_full_grid_flair.tex}{%
\providecommand{\gcvdry}[1]{\textcolor{red}{#1}}
\providecommand{\ProvFullGrid}{Seeds per cell 5 to 8. Mean $\pm$ standard deviation over seeds. Bold marks the largest mIoU, the smallest drop and the largest retention of each type and level across the five models. Analysis of the finished runs.}
\begin{tabular}{@{}lllcccc@{}}
\toprule
Model & Type & Level & $n$ & mIoU & Drop & Retention (\%) \\
\midrule
SAT & Clean & -- & 8 & \textbf{52.4} $\pm$ 0.4 & -- & -- \\
 & Shift & L1 & 5 & \textbf{50.9} $\pm$ 1.1 & 1.3 $\pm$ 1.0 & 97.5 $\pm$ 1.9 \\
 & Shift & L2 & 8 & 48.3 $\pm$ 0.9 & 4.1 $\pm$ 0.6 & 92.2 $\pm$ 1.1 \\
 & Coarse & L1 & 5 & \textbf{50.8} $\pm$ 1.0 & 1.5 $\pm$ 0.8 & 97.2 $\pm$ 1.5 \\
 & Coarse & L2 & 8 & \textbf{46.1} $\pm$ 0.9 & 6.3 $\pm$ 0.8 & 87.9 $\pm$ 1.5 \\
 & Boundary & L1 & 5 & 49.4 $\pm$ 0.6 & 2.8 $\pm$ 0.6 & 94.6 $\pm$ 1.2 \\
 & Boundary & L2 & 8 & 45.6 $\pm$ 1.1 & 6.9 $\pm$ 1.0 & 86.9 $\pm$ 1.9 \\
 & Omission & L1 & 5 & 49.4 $\pm$ 0.5 & 2.8 $\pm$ 0.3 & 94.7 $\pm$ 0.6 \\
 & Omission & L2 & 8 & 44.4 $\pm$ 1.6 & 8.0 $\pm$ 1.4 & 84.7 $\pm$ 2.7 \\
LVD & Clean & -- & 8 & 50.7 $\pm$ 0.9 & -- & -- \\
 & Shift & L1 & 5 & 50.5 $\pm$ 0.8 & \textbf{0.3} $\pm$ 1.4 & \textbf{99.5} $\pm$ 2.6 \\
 & Shift & L2 & 8 & \textbf{48.9} $\pm$ 0.8 & \textbf{1.8} $\pm$ 0.8 & \textbf{96.4} $\pm$ 1.4 \\
 & Coarse & L1 & 5 & 50.1 $\pm$ 0.7 & 0.7 $\pm$ 1.5 & 98.7 $\pm$ 2.9 \\
 & Coarse & L2 & 8 & 45.9 $\pm$ 0.8 & \textbf{4.8} $\pm$ 1.2 & \textbf{90.6} $\pm$ 2.2 \\
 & Boundary & L1 & 5 & \textbf{49.9} $\pm$ 1.6 & \textbf{1.0} $\pm$ 1.9 & \textbf{98.2} $\pm$ 3.6 \\
 & Boundary & L2 & 8 & \textbf{46.8} $\pm$ 0.6 & 3.9 $\pm$ 0.8 & \textbf{92.3} $\pm$ 1.5 \\
 & Omission & L1 & 5 & \textbf{49.7} $\pm$ 0.9 & 1.2 $\pm$ 1.4 & 97.8 $\pm$ 2.8 \\
 & Omission & L2 & 8 & \textbf{47.3} $\pm$ 1.0 & \textbf{3.4} $\pm$ 1.7 & \textbf{93.2} $\pm$ 3.2 \\
DOFA & Clean & -- & 5 & 47.5 $\pm$ 0.8 & -- & -- \\
 & Shift & L1 & 5 & 47.2 $\pm$ 1.0 & 0.3 $\pm$ 1.3 & 99.4 $\pm$ 2.8 \\
 & Shift & L2 & 5 & 44.6 $\pm$ 0.7 & 2.9 $\pm$ 0.8 & 93.9 $\pm$ 1.6 \\
 & Coarse & L1 & 5 & 47.0 $\pm$ 1.1 & \textbf{0.6} $\pm$ 1.4 & \textbf{98.8} $\pm$ 2.9 \\
 & Coarse & L2 & 5 & 42.2 $\pm$ 1.2 & 5.3 $\pm$ 1.1 & 88.9 $\pm$ 2.2 \\
 & Boundary & L1 & 5 & 45.7 $\pm$ 0.6 & 1.8 $\pm$ 1.2 & 96.3 $\pm$ 2.5 \\
 & Boundary & L2 & 5 & 42.7 $\pm$ 1.4 & 4.8 $\pm$ 1.2 & 89.8 $\pm$ 2.5 \\
 & Omission & L1 & 5 & 46.1 $\pm$ 0.9 & 1.4 $\pm$ 1.1 & 97.0 $\pm$ 2.3 \\
 & Omission & L2 & 5 & 43.3 $\pm$ 1.1 & 4.2 $\pm$ 1.0 & 91.2 $\pm$ 2.2 \\
TM & Clean & -- & 5 & 40.5 $\pm$ 1.0 & -- & -- \\
 & Shift & L1 & 5 & 39.5 $\pm$ 1.3 & 1.0 $\pm$ 0.6 & 97.6 $\pm$ 1.4 \\
 & Shift & L2 & 5 & 36.7 $\pm$ 1.2 & 3.7 $\pm$ 1.8 & 90.8 $\pm$ 4.2 \\
 & Coarse & L1 & 5 & 39.6 $\pm$ 0.5 & 0.8 $\pm$ 1.2 & 98.0 $\pm$ 3.0 \\
 & Coarse & L2 & 5 & 34.8 $\pm$ 1.3 & 5.6 $\pm$ 1.6 & 86.1 $\pm$ 3.8 \\
 & Boundary & L1 & 5 & 39.0 $\pm$ 0.5 & 1.5 $\pm$ 1.0 & 96.4 $\pm$ 2.4 \\
 & Boundary & L2 & 5 & 34.4 $\pm$ 1.0 & 6.1 $\pm$ 1.5 & 85.1 $\pm$ 3.5 \\
 & Omission & L1 & 5 & 37.1 $\pm$ 1.8 & 3.3 $\pm$ 1.5 & 91.8 $\pm$ 3.8 \\
 & Omission & L2 & 5 & 35.0 $\pm$ 0.8 & 5.5 $\pm$ 1.3 & 86.6 $\pm$ 3.0 \\
RN & Clean & -- & 5 & 44.5 $\pm$ 1.4 & -- & -- \\
 & Shift & L1 & 5 & 43.7 $\pm$ 0.8 & 0.8 $\pm$ 0.9 & 98.2 $\pm$ 2.1 \\
 & Shift & L2 & 5 & 40.6 $\pm$ 1.0 & 3.8 $\pm$ 0.7 & 91.4 $\pm$ 1.5 \\
 & Coarse & L1 & 5 & 43.6 $\pm$ 1.4 & 0.9 $\pm$ 1.9 & 98.1 $\pm$ 4.2 \\
 & Coarse & L2 & 5 & 38.8 $\pm$ 0.7 & 5.7 $\pm$ 1.9 & 87.3 $\pm$ 4.0 \\
 & Boundary & L1 & 5 & 42.6 $\pm$ 1.3 & 1.9 $\pm$ 1.2 & 95.8 $\pm$ 2.7 \\
 & Boundary & L2 & 5 & 40.8 $\pm$ 1.9 & \textbf{3.7} $\pm$ 2.8 & 91.9 $\pm$ 6.0 \\
 & Omission & L1 & 5 & 44.3 $\pm$ 0.4 & \textbf{0.2} $\pm$ 1.5 & \textbf{99.7} $\pm$ 3.5 \\
 & Omission & L2 & 5 & 40.9 $\pm$ 0.9 & 3.6 $\pm$ 1.1 & 92.0 $\pm$ 2.4 \\
\bottomrule
\end{tabular}
}{\tabletbd[6cm]{generated/table\_full\_grid\_flair.tex: run scripts/reflow\_tables.py}}
\caption{\textbf{Every main-grid cell on \flair:} Columns as in \Cref{tab:supp-grid-burn}. Bold marks the best value of each type and level as in \Cref{tab:supp-grid-burn}. \ProvSlot{\ResultsProvenance}}
\label{tab:supp-grid-flair}
\end{table}

\fi
\section{Collapsed Runs}
\label{sec:supp-collapsed}

\ifdefined\cvprbuild\else
\begin{table}[t]
\centering\footnotesize\setlength{\tabcolsep}{3pt}
\IfFileExists{generated/table_collapsed.tex}{\begin{adjustbox}{max width=\columnwidth}%
\providecommand{\gcvdry}[1]{\textcolor{red}{#1}}
\providecommand{\ProvCollapsed}{9 runs. Burned IoU and test \miou in points on the clean reference test labels at the selected checkpoint, Epoch its training epoch. Analysis of the finished runs.}
\begin{tabular}{@{}lllcccc@{}}
\toprule
Model & Arm & Type & Seed & Burned IoU & Test \miou & Epoch \\
\midrule
\sat & N2, lr $\times$3 & \etype{Coarse} L2 & 2 & 0.0 & 44.8 & 2 \\
\sat & A5, clean val. & \etype{Coarse} L2 & 4 & 0.0 & 44.8 & 2 \\
\sat & Main grid & \etype{Coarse} L2 & 2 & 5.2 & 47.6 & 10 \\
\sat & Main grid & \etype{Omission} L2 & 1 & 12.7 & 51.6 & 6 \\
\sat & Main grid & \etype{Omission} L2 & 3 & 14.5 & 52.5 & 16 \\
\sat & Main grid & \etype{Omission} L2 & 7 & 6.1 & 48.0 & 8 \\
\rn & A6, random init & \etype{Omission} L2 & 3 & 14.8 & 52.9 & 2 \\
\rn & A6, random init & \etype{Omission} L2 & 5 & 7.0 & 48.6 & 6 \\
\rn & Main grid & \etype{Omission} L2 & 3 & 6.3 & 48.2 & 6 \\
\bottomrule
\end{tabular}
\end{adjustbox}}{\tabletbd[2.6cm]{generated/table\_collapsed.tex: run the analysis harness}}
\caption{\textbf{Collapsed runs on \burn:} Every analyzed noisy-label run of a core model whose burned-class test IoU fell below \CollapseIoU{} at the selected checkpoint. Columns give the arm, error type and seed, the burned-class IoU and test \miou in points, and the epoch of the selected checkpoint. \ProvSlot{\ResultsProvenance}}
\label{tab:supp-collapsed}
\end{table}

\fi
\begin{table}[t]
\centering\footnotesize\setlength{\tabcolsep}{3pt}
\IfFileExists{generated/table_collapsed_sens.tex}{\begin{adjustbox}{max width=\columnwidth}%
\providecommand{\gcvdry}[1]{\textcolor{red}{#1}}
\providecommand{\ProvCollapsedSens}{Estimates in \miou points with 95\,\% paired $t$ intervals on \burn at \Ltwo, $n$ = seeds for F1 to F4 and (model, seed) units for H2. The type means are model-equal means of the cell means and carry no $n$. Without the collapsed runs, F1 and F2 average the types present in each seed, F3 and F4 lose the seed of a collapsed arm. Analysis of the finished runs.}
\begin{tabular}{@{}lcccc@{}}
\toprule
Contrast & All runs & $n$ & Without collapsed & $n$ \\
\midrule
F1, drop \sat $-$ \lvd & 4.3 [1.3, 7.3] & 8 & 3.1 [1.0, 5.2] & 8 \\
F2, drop \gfms $-$ \rn & 0.0 [$-$2.8, 2.7] & 8 & 0.7 [$-$0.7, 2.0] & 8 \\
F3, drop frozen $-$ fine-tuned & $-$5.4 [$-$8.3, $-$2.4] & 5 & $-$4.5 [$-$7.0, $-$2.0] & 4 \\
F4, \miou clean subset $-$ noisy & 14.4 [12.0, 16.8] & 5 & 14.7 [11.3, 18.1] & 4 \\
H2, drop \etype{omission} $-$ \etype{boundary} & 4.8 [2.6, 7.1] & 40 & 3.6 [1.6, 5.6] & 36 \\
Model-equal mean drop, \etype{shift} & 11.7 & -- & 11.7 & -- \\
Model-equal mean drop, \etype{coarse} & 18.4 & -- & 18.2 & -- \\
Model-equal mean drop, \etype{boundary} & 13.4 & -- & 13.4 & -- \\
Model-equal mean drop, \etype{omission} & 18.2 & -- & 16.8 & -- \\
\bottomrule
\end{tabular}
\end{adjustbox}}{\tabletbd[2.6cm]{generated/table\_collapsed\_sens.tex: run the analysis harness}}
\caption{\textbf{\burn contrasts with and without the collapsed runs:} Estimates in \miou points with \CiLevel\,\% paired $t$ intervals. F1 and F2 keep their seeds and average the types left in each seed. F3 and F4 lose the seed of a collapsed arm, and H2 and the type means lose the collapsed (model, seed) units. The inclusive result is primary. The $n$ columns give the number of paired units, and a dash marks a row that reports none. \ProvSlot{\ResultsProvenance}}
\label{tab:supp-collapsed-sens}
\end{table}

\ifdefined\cvprbuild\else
\Cref{tab:supp-collapsed} lists the \hlNCollapsedburn{} collapsed runs.
\fi
\ifdefined\cvprbuild
The core design on \burn holds \hlNCollapsedburn{} collapsed runs.
\fi
A collapsed run predicts background almost everywhere.
Its burned-class IoU is therefore at most \hlCollapsedIoUMax\,\% and its test \miou is close to half the background IoU.
The selected checkpoints lie between epochs \hlCollapsedEpochMin{} and \hlCollapsedEpochMax, so early stopping left them collapsed.
Of these runs, \hlCollapsedMainGridN{} lie in the main grid, all of them \sat and \rn under \etype{coarse} or \etype{omission} at \Ltwo.
The rest lie in the N2, A5 and A6 ablations of \Cref{tab:supp-blocks}.
The same screen over the extension encoders finds \hlCollapsedExtN{} collapsed runs. They are \hlCollapsedExtList.
All of them are at \hlCollapsedExtTypes{} and none is a \prithvi run.
\Cref{tab:supp-collapsed-sens} recomputes the \burn contrasts without the core-design collapsed runs.
F1 moves from \hlCollFOneAllEst{} [\hlCollFOneAllCILo, \hlCollFOneAllCIHi] to \hlCollFOneWithoutEst{} [\hlCollFOneWithoutCILo, \hlCollFOneWithoutCIHi] and keeps its verdict.
F2 narrows from \hlCollFTwoAllEst{} [\hlCollFTwoAllCILo, \hlCollFTwoAllCIHi] to \hlCollFTwoWithoutEst{} [\hlCollFTwoWithoutCILo, \hlCollFTwoWithoutCIHi], so the \burn precision of the F2 null is set by one collapsed \rn run.
H2 moves from \hlCollHTwoAllEst{} to \hlCollHTwoWithoutEst{} points and the \etype{omission} mean from \hlCollTypeMeanOmissionAll{} to \hlCollTypeMeanOmissionWithout, so \etype{coarse} leads alone without the collapsed runs.
No verdict changes.

\ifdefined\cvprbuild\else
\section{Per Class Results on FLAIR-2}
\label{sec:supp-perclass}

\begin{table}[t]
\centering\footnotesize\setlength{\tabcolsep}{3pt}
\IfFileExists{generated/table_perclass_flair.tex}{\begin{adjustbox}{max width=\columnwidth}%
\providecommand{\gcvdry}[1]{\textcolor{red}{#1}}
\providecommand{\ProvPerclassFlair}{Seeds per cell 5 to 8. Retention in \% at the worst \Ltwo type (omission), mean over seeds of the per-seed ratio of noisy to clean class IoU. A dash marks a class whose clean IoU is below 0.05 in a seed. Bold marks the best model per class. Analysis of the finished runs.}
\begin{tabular}{@{}lcccccc@{}}
\toprule
Class & Pixels (\%) & \sat & \lvd & \dofa & \tm & \rn \\
\midrule
Herbaceous vegetation & 17.5 & 80.9 & \textbf{93.5} & 90.5 & 79.8 & 90.9 \\
Deciduous & 16.0 & 86.0 & \textbf{93.5} & 91.4 & 91.4 & 90.2 \\
Impervious surface & 13.6 & 88.7 & \textbf{93.1} & 90.4 & 85.0 & 90.4 \\
Agricultural land & 10.8 & 88.2 & 96.9 & \textbf{97.7} & 96.0 & 87.9 \\
Building & 8.3 & 88.7 & \textbf{94.5} & 89.6 & 73.7 & 89.0 \\
Pervious surface & 7.9 & 84.7 & 89.8 & 89.7 & 81.2 & \textbf{92.7} \\
Brushwood & 7.3 & 71.8 & \textbf{88.8} & 85.5 & 65.6 & 86.6 \\
Water & 4.8 & 86.9 & 92.9 & 93.9 & \textbf{97.0} & 90.1 \\
Plowed land & 3.6 & 82.4 & \textbf{114.4} & 83.7 & 94.6 & 104.8 \\
Vineyard & 3.6 & 88.4 & 91.4 & 93.2 & 85.8 & \textbf{94.1} \\
Coniferous & 3.2 & 76.9 & 93.2 & 95.3 & 89.0 & \textbf{110.8} \\
Bare soil & 3.0 & 80.5 & 86.2 & 84.3 & \textbf{94.9} & 88.0 \\
Other & 0.4 & -- & -- & -- & -- & -- \\
\bottomrule
\end{tabular}
\end{adjustbox}}{\tabletbd[5cm]{generated/table\_perclass\_flair.tex: run the analysis harness}}
\caption{\textbf{Per-class retention on \flair at the selected \Ltwo type:} Classes are ordered by their pixel frequency in the training subset. Retention is the per-seed ratio of noisy to clean class IoU in percent, averaged over the seeds of each cell. The cells have \NSeeds{} seeds and \statFOneburnN{} for the DINOv3 pair. Bold marks the best model per class, and a dash marks a class whose clean IoU is below 0.05 in a seed. \ProvSlot{\ResultsProvenance}}
\label{tab:supp-perclass}
\end{table}

Under \etype{omission} on \flair the drops of rare classes are not separated from those of frequent ones.
Retention in \Cref{tab:supp-perclass} does not fall with pixel share, and the lowest value belongs to brushwood for \sat and \tm.
Two rare classes exceed full retention for some models because their clean class IoU is small and the ratio is unstable.
The rarest class is shown as a dash for the same reason.
At the selected rates \lvd retains the most in half of the classes, in line with its smaller mean drop.
The per-class picture gives no reason to weight rare classes in the loss, which the S2 block tests directly (\Cref{sec:supp-stretch}).

\begin{table}[t]
\centering\footnotesize\setlength{\tabcolsep}{2.5pt}
\IfFileExists{generated/table_perclass_drop_flair.tex}{\begin{adjustbox}{max width=\columnwidth}%
\providecommand{\gcvdry}[1]{\textcolor{red}{#1}}
\providecommand{\ProvPerclassDropFlair}{Units per cell 31 to 31 (model, seed) pairs. Paired drop in class IoU points at \Ltwo, clean minus noisy class IoU of the same model and seed, mean $\pm$ 95\,\% $t$ half-width over the units of the five core models. Analysis of the finished runs.}
\begin{tabular}{@{}lccccc@{}}
\toprule
Class & Pixels (\%) & \etype{Shift} & \etype{Coarse} & \etype{Boundary} & \etype{Omission} \\
\midrule
Herbaceous vegetation & 17.5 & 4.0 $\pm$ 1.1 & 6.7 $\pm$ 1.0 & 6.4 $\pm$ 1.1 & 5.7 $\pm$ 1.5 \\
Deciduous & 16.0 & 3.3 $\pm$ 0.6 & 5.6 $\pm$ 0.7 & 5.7 $\pm$ 0.9 & 6.5 $\pm$ 1.5 \\
Impervious surface & 13.6 & 9.5 $\pm$ 0.7 & 15.7 $\pm$ 0.8 & 9.2 $\pm$ 1.3 & 6.6 $\pm$ 0.8 \\
Agricultural land & 10.8 & 0.5 $\pm$ 0.7 & 1.0 $\pm$ 0.8 & 1.1 $\pm$ 0.9 & 3.7 $\pm$ 1.6 \\
Building & 8.3 & 11.2 $\pm$ 1.1 & 21.2 $\pm$ 1.6 & 12.0 $\pm$ 2.0 & 8.4 $\pm$ 2.0 \\
Pervious surface & 7.9 & 3.9 $\pm$ 0.9 & 5.3 $\pm$ 1.0 & 5.2 $\pm$ 1.0 & 5.6 $\pm$ 1.2 \\
Brushwood & 7.3 & 1.9 $\pm$ 0.6 & 2.9 $\pm$ 0.5 & 4.8 $\pm$ 0.8 & 4.2 $\pm$ 0.8 \\
Water & 4.8 & 3.1 $\pm$ 1.1 & 4.6 $\pm$ 1.1 & 6.1 $\pm$ 1.3 & 6.7 $\pm$ 1.7 \\
Plowed land & 3.6 & 2.1 $\pm$ 2.2 & 0.8 $\pm$ 1.8 & 0.4 $\pm$ 1.3 & 1.8 $\pm$ 2.2 \\
Vineyard & 3.6 & 2.0 $\pm$ 1.1 & 4.7 $\pm$ 0.9 & 3.9 $\pm$ 1.4 & 5.5 $\pm$ 1.6 \\
Coniferous & 3.2 & $-$1.0 $\pm$ 1.8 & 0.4 $\pm$ 2.3 & 6.3 $\pm$ 3.1 & 4.9 $\pm$ 2.8 \\
Bare soil & 3.0 & 1.6 $\pm$ 1.6 & 3.3 $\pm$ 2.2 & 5.8 $\pm$ 2.1 & 6.4 $\pm$ 2.5 \\
Other & 0.4 & 0.0 $\pm$ 0.2 & $-$0.1 $\pm$ 0.2 & 0.0 $\pm$ 0.3 & 0.0 $\pm$ 0.1 \\
\bottomrule
\end{tabular}
\end{adjustbox}}{\tabletbd[5cm]{generated/table\_perclass\_drop\_flair.tex: run the analysis harness}}
\caption{\textbf{Per-class drop on \flair by error type at \Ltwo:} Paired drop in class IoU points, clean minus noisy class IoU of the same model and seed. Cells give the mean $\pm$ \CiLevel\,\% $t$ half-width over the (model, seed) units of the five core models. Classes are ordered by pixel share in the training subset. \ProvSlot{\ResultsProvenance}}
\label{tab:supp-perclass-drop}
\end{table}

\Cref{tab:supp-perclass-drop} gives the per-class drop under every type for the small-object question of the main paper.
Buildings lose \hlPcDropBuildingShift{} points under \etype{shift}, \hlPcDropBuildingCoarse{} under \etype{coarse}, \hlPcDropBuildingBoundary{} under \etype{boundary} and \hlPcDropBuildingOmission{} under \etype{omission}.
Impervious surfaces lose \hlPcDropImperviousShift, \hlPcDropImperviousCoarse, \hlPcDropImperviousBoundary{} and \hlPcDropImperviousOmission.
The two object classes therefore lose far more than the class-averaged \miou under every type.
The \etype{coarse} type costs them the most, and \etype{shift} costs them more than \etype{omission}.

\fi
\section{Boundary mIoU by Error Type}
\label{sec:supp-boundary}

\begin{table}[t]
\centering\footnotesize\setlength{\tabcolsep}{3pt}
\IfFileExists{generated/table_boundary_drop.tex}{%
\providecommand{\gcvdry}[1]{\textcolor{red}{#1}}
\providecommand{\ProvBoundaryDrop}{Seeds per cell 5 to 8. Paired drop in boundary mIoU points at \Ltwo, clean minus noisy boundary mIoU of the same seed, mean over seeds. The mean row averages over models. Bold marks the smallest drop of a column within each block among the five models. Analysis of the finished runs.}
\begin{tabular}{@{}lcccc@{}}
\toprule
Model & \etype{Shift} & \etype{Coarse} & \etype{Boundary} & \etype{Omission} \\
\midrule
\multicolumn{5}{@{}l}{\burn, \BoundaryDPxNarrow\,px band} \\
\sat & 7.8 & 6.5 & 11.6 & 8.2 \\
\lvd & \textbf{5.4} & \textbf{5.1} & \textbf{10.7} & \textbf{8.0} \\
\dofa & 8.7 & 10.2 & 13.2 & 12.2 \\
\tm & 8.8 & 8.7 & 13.3 & 10.4 \\
\rn & 13.7 & 13.0 & 17.3 & 17.1 \\
Mean & 8.9 & 8.7 & 13.2 & 11.2 \\
\addlinespace[2pt]
\multicolumn{5}{@{}l}{\burn, \BoundaryDPxBurn\,px band} \\
\sat & \textbf{17.8} & \textbf{18.2} & \textbf{20.2} & 17.4 \\
\lvd & 17.9 & 19.3 & 20.9 & 16.5 \\
\dofa & 22.5 & 27.9 & 26.2 & 26.1 \\
\tm & 21.6 & 24.3 & 25.2 & \textbf{16.1} \\
\rn & 25.6 & 27.4 & 29.4 & 23.1 \\
Mean & 21.1 & 23.4 & 24.4 & 19.9 \\
\addlinespace[2pt]
\multicolumn{5}{@{}l}{\flair, \BoundaryDPxNarrow\,px band} \\
\sat & 3.4 & 4.1 & 4.2 & 4.7 \\
\lvd & \textbf{1.7} & 3.1 & 2.8 & 3.0 \\
\dofa & 1.9 & \textbf{2.9} & \textbf{2.5} & \textbf{2.5} \\
\tm & 2.8 & 3.3 & 3.2 & 2.8 \\
\rn & 4.0 & 4.4 & 3.4 & 3.1 \\
Mean & 2.8 & 3.6 & 3.2 & 3.2 \\
\addlinespace[2pt]
\multicolumn{5}{@{}l}{\flair, \BoundaryDPxBurn\,px band} \\
\sat & 8.6 & 11.8 & 9.1 & 8.5 \\
\lvd & \textbf{5.2} & 9.0 & 5.8 & 5.0 \\
\dofa & 6.0 & 9.5 & 6.2 & 4.9 \\
\tm & 6.7 & \textbf{8.9} & 6.3 & 5.0 \\
\rn & 8.3 & 10.8 & \textbf{5.6} & \textbf{4.7} \\
Mean & 6.9 & 10.0 & 6.6 & 5.6 \\
\bottomrule
\end{tabular}
}{\tabletbd[6cm]{generated/table\_boundary\_drop.tex: run the analysis harness}}
\caption{\textbf{Paired drop in boundary \miou by error type at \Ltwo:} Clean minus noisy boundary \miou \cite{cheng2021boundary} of the same seed in points, mean over seeds. The bands are \BoundaryDPxNarrow\,px and 2\,\% of the tile diagonal (\BoundaryDPxBurn\,px). In ground units they are \BoundaryDMNarrowBurn\,m and \BoundaryDMBurn\,m on \burn and \BoundaryDMNarrowFlair\,m and \BoundaryDMFlair\,m on \flair. The mean row averages over models. Bold marks the smallest drop of a column within each block among the five models. \ProvSlot{\ResultsProvenance}}
\label{tab:supp-boundary}
\end{table}

\Cref{tab:supp-boundary} re-ranks the error types by the paired drop in boundary \miou \cite{cheng2021boundary}, which scores only pixels near a boundary.
At the \BoundaryDPxNarrow\,px band the ranking is \hlBdropRankNarrowburn{} on \burn and \hlBdropRankNarrowflair{} on \flair.
The \etype{shift} type stays last on \flair and ranks third on \burn.
Its gap to the next type is \hlBdropGapShiftNarrowburn{} [\hlBdropGapShiftCILoNarrowburn, \hlBdropGapShiftCIHiNarrowburn] points on \burn and \hlBdropGapShiftNarrowflair{} [\hlBdropGapShiftCILoNarrowflair, \hlBdropGapShiftCIHiNarrowflair] on \flair.
On \burn that interval excludes zero, so \etype{shift} is separated from the type ranked above it, and only \etype{coarse} ranks below it.
At the \BoundaryDPxBurn\,px band the ranking is \hlBdropRankWideburn{} on \burn and \hlBdropRankWideflair{} on \flair.
On both datasets \etype{omission} has the smallest mean drop.
The paired gap \etype{shift} minus \etype{omission} is \hlBdropGapOmissionWideburn{} [\hlBdropGapOmissionWideburnCILo, \hlBdropGapOmissionWideburnCIHi] points on \burn and \hlBdropGapOmissionWideflair{} [\hlBdropGapOmissionWideflairCILo, \hlBdropGapOmissionWideflairCIHi] on \flair.
For \etype{shift} minus \etype{boundary} the gap is \hlBdropGapBoundaryWideburn{} [\hlBdropGapBoundaryWideburnCILo, \hlBdropGapBoundaryWideburnCIHi] on \burn and \hlBdropGapBoundaryWideflair{} [\hlBdropGapBoundaryWideflairCILo, \hlBdropGapBoundaryWideflairCIHi] on \flair.
The \etype{shift} type is therefore separated from \etype{omission} on \flair only and from \etype{boundary} on \burn only.
The type nearest above it leads by \hlBdropGapShiftWideburn{} [\hlBdropGapShiftCILoWideburn, \hlBdropGapShiftCIHiWideburn] and \hlBdropGapShiftWideflair{} [\hlBdropGapShiftCILoWideflair, \hlBdropGapShiftCIHiWideflair] points.
We do not isolate why \etype{shift} costs more than \etype{omission} there.
The error type advice holds under \miou.
On the \flair evidence, for a product scored by boundary \miou at a wide band the completeness part fails and the resolution part holds.

\ifdefined\cvprbuild\section{Matched Learning Rates}\else\section{Matched Learning Rates and Ablations}\fi
\label{sec:supp-ablation}

\begin{table}[t]
\centering\footnotesize\setlength{\tabcolsep}{3pt}
\IfFileExists{generated/table_matched_lr_reflow.tex}{%
\providecommand{\gcvdry}[1]{\textcolor{red}{#1}}
\providecommand{\ProvMatchedLr}{Seeds per cell 5 to 8 (paired seeds of each contrast). Mean over the four \Ltwo types per seed. Drop in points, retention in \% of clean mIoU, contrasts with the 95\,\% paired $t$ interval. The N6 arm pairs each noisy run with the $\times$1/3 clean run of the same seed. Sensitivity analysis, not Holm corrected. Bold marks the smallest drop and the largest retention among the three arms. Analysis of the finished runs.}
\begin{tabular}{@{}lcc@{}}
\toprule
\multicolumn{3}{@{}l}{\textit{\burn}} \\
Arm & Drop & Retention (\%) \\
\midrule
SAT $\times$1 (F1 arm) & 18.1 & 76.5 \\
SAT $\times$1/3 (N6) & 13.9 & 82.9 \\
LVD $\times$1/3 (main arm) & \textbf{13.8} & \textbf{83.1} \\
\midrule
SAT $\times$1 $-$ LVD (F1 arms) & 4.3 [1.3, 7.3] & $-$6.6 [$-$9.8, $-$3.4] \\
SAT $\times$1/3 $-$ LVD (N6) & $-$0.2 [$-$3.7, 3.3] & 0.1 [$-$3.7, 3.9] \\
\bottomrule
\end{tabular}
\par\vspace{4pt}
\begin{tabular}{@{}lcc@{}}
\toprule
\multicolumn{3}{@{}l}{\textit{\flair}} \\
Arm & Drop & Retention (\%) \\
\midrule
SAT $\times$1 (F1 arm) & 6.3 & 87.9 \\
SAT $\times$1/3 (N6) & 5.1 & 89.9 \\
LVD $\times$1/3 (main arm) & \textbf{3.5} & \textbf{93.1} \\
\midrule
SAT $\times$1 $-$ LVD (F1 arms) & 2.8 [1.8, 3.8] & $-$5.2 [$-$7.1, $-$3.3] \\
SAT $\times$1/3 $-$ LVD (N6) & 1.5 [0.6, 2.5] & $-$3.0 [$-$4.9, $-$1.2] \\
\bottomrule
\end{tabular}
}{\tabletbd[2.4cm]{generated/table\_matched\_lr\_reflow.tex: run scripts/reflow\_tables.py}}
\caption{\textbf{RQ1 at matched learning rates:} Mean drop in \miou points and retention in percent of clean \miou over the four \Ltwo types. The top block shows \burn and the bottom block \flair. The N6 contrast pairs seeds 1 to 5, so it is not the difference of the printed arm means. Rows are \sat at its \gbtwo rate (the F1 arm), \sat at the \lvd rate (N6) and \lvd. The last rows are the two \sat minus \lvd contrasts with their \CiLevel\,\% paired intervals. The F1 arms and the F1 contrast hold \statFOneburnN{} seeds, and the N6 arm and contrast hold \NSeeds. N6 is an exploratory contrast outside the Holm family. Bold marks the smallest drop and the largest retention among the three arms. Models are abbreviated as SAT (\sat) and LVD (\lvd). Row labels give the learning rate (lr) as a multiple of the \gbtwo rate of \sat.}
\label{tab:supp-mlr}
\end{table}

\ifdefined\cvprbuild\else
\begin{table}[t]
\centering\footnotesize\setlength{\tabcolsep}{4pt}
\IfFileExists{generated/table_ablation_reflow.tex}{%
\providecommand{\gcvdry}[1]{\textcolor{red}{#1}}
\providecommand{\ProvAblation}{Seeds per cell 2 to 5 for the ablation cells, $n$ per cell. Drop in points at the worst \Ltwo type, mean $\pm$ 95\,\% $t$ half-width, main = main-grid arm of the same model. N2 multiplies the learning rate of the main arm (\lvd runs at 1/3 of its \gbtwo rate). The N6 and N8 rows measure the drop against the clean run of their own rate and seed ($\times$1/3 for \sat, $\times$1 for \lvd). Bold marks the smaller drop of each pair. Analysis of the finished runs.}
\begin{tabular}{@{}llccc@{}}
\toprule
\multicolumn{5}{@{}l}{\textit{\burn, coarse}} \\
Ablation & Model & Ablation & $n$ & Main \\
\midrule
A5 clean validation labels & SAT & \textbf{19.5} $\pm$ 10.7 & 5 & 21.7 \\
A6 random initialization & RN & \textbf{12.7} $\pm$ 2.3 & 5 & 17.8 \\
A7 RGB input & DOFA & \textbf{14.5} $\pm$ 2.3 & 5 & 19.4 \\
N4 RGB input & RN & \textbf{14.2} $\pm$ 3.3 & 5 & 17.8 \\
N6 lr $\times$1/3 (matched clean) & SAT & \textbf{15.2} $\pm$ 2.7 & 5 & 21.7 \\
N2 lr $\times$3 & SAT & 27.0 $\pm$ 64.1 & 2 & \textbf{21.7} \\
N2 lr $\times$1/3 & LVD & \textbf{13.6} $\pm$ 15.9 & 2 & 15.5 \\
N8 lr $\times$1 (matched clean) & LVD & 26.3 $\pm$ 7.9 & 5 & \textbf{15.5} \\
N3 no early stopping & SAT & \textbf{16.5} $\pm$ 33.8 & 2 & 21.7 \\
\bottomrule
\end{tabular}
\par\vspace{4pt}
\begin{tabular}{@{}llccc@{}}
\toprule
\multicolumn{5}{@{}l}{\textit{\flair, omission}} \\
Ablation & Model & Ablation & $n$ & Main \\
\midrule
A5 clean validation labels & SAT & \textbf{7.1} $\pm$ 0.8 & 5 & 8.0 \\
A6 random initialization & RN & \textbf{0.9} $\pm$ 3.7 & 5 & 3.6 \\
A7 RGB input & DOFA & -- & -- & 4.2 \\
N4 RGB input & RN & -- & -- & 3.6 \\
N6 lr $\times$1/3 (matched clean) & SAT & \textbf{6.1} $\pm$ 1.8 & 5 & 8.0 \\
N2 lr $\times$3 & SAT & \textbf{7.2} $\pm$ 0.1 & 2 & 8.0 \\
N2 lr $\times$1/3 & LVD & 5.9 $\pm$ 1.3 & 2 & \textbf{3.4} \\
N8 lr $\times$1 (matched clean) & LVD & 3.5 $\pm$ 1.7 & 5 & \textbf{3.4} \\
N3 no early stopping & SAT & \textbf{7.9} $\pm$ 9.0 & 2 & 8.0 \\
\bottomrule
\end{tabular}
}{\tabletbd[3.0cm]{generated/table\_ablation\_reflow.tex: run scripts/reflow\_tables.py}}
\caption{\textbf{Ablations at the selected \Ltwo type:} Paired drop in \miou points with its \CiLevel\,\% interval and the number of seeds $n$. Each ablation is set against the main-grid arm (Main) of the same model. The top block shows \burn with \etype{coarse} and the bottom block \flair with \etype{omission}. A5 stops on clean validation labels, A6 trains \rn from random initialization, and A7 and N4 give \dofa and \rn RGB input on \burn. N6 runs \sat at the \lvd learning rate (lr) against its own clean run at that rate. N2 scales the learning rate of the DINOv3 pair, and N3 trains for \MaxEpochs{} epochs without early stopping. Bold marks the smaller drop of each pair. A dash marks a cell that was not run. The Main column averages all seeds of the main-grid cell, \statFOneburnN{} on \burn and for the DINOv3 pair on \flair and \NSeeds{} otherwise. The N2 factors multiply the main-grid rate of the named model, and N8 is the second learning-rate multiplier of \lvd. The two-seed intervals of N2 and N3 have one degree of freedom and show the spread only. Models are abbreviated as SAT (\sat), LVD (\lvd), DOFA (\dofa) and RN (\rn).}
\label{tab:supp-ablation}
\end{table}

\fi
\Cref{tab:supp-mlr} gives the matched-rate contrasts read in the main paper.
The learning-rate probes at seed \CalSeed{} on clean validation labels chose \LvdLrMult{} for \lvd on both datasets and \AsixLrMult{} for the random-initialization \rn.
For \sat, which kept the \gbtwo rate, the \LvdLrMult{} probe scored \hlSATLrLrThirdburn{} against \hlSATLrMainburn{} on \burn and \hlSATLrLrThirdflair{} against \hlSATLrMainflair{} on \flair.
The same rule would therefore have changed the \sat rate on \burn only.
\ifdefined\cvprbuild\else
In \Cref{tab:supp-ablation}, each arm carries its own interval, and the paired ablation minus main contrasts over five seeds are the following.
A5 is \hlAblContrastAFiveburnEst{} [\hlAblContrastAFiveburnCILo, \hlAblContrastAFiveburnCIHi] on \burn and \hlAblContrastAFiveflairEst{} [\hlAblContrastAFiveflairCILo, \hlAblContrastAFiveflairCIHi] on \flair.
A6 is \hlAblContrastASixburnEst{} [\hlAblContrastASixburnCILo, \hlAblContrastASixburnCIHi] on \burn and \hlAblContrastASixflairEst{} [\hlAblContrastASixflairCILo, \hlAblContrastASixflairCIHi] on \flair.
\fi
The N6 arm contrast is the \sat drop at the \lvd rate minus its main-grid drop over the four \Ltwo types (\Cref{tab:n8}).
It is \hlAblContrastNSixburnEst{} [\hlAblContrastNSixburnCILo, \hlAblContrastNSixburnCIHi] on \burn and \hlAblContrastNSixflairEst{} [\hlAblContrastNSixflairCILo, \hlAblContrastNSixflairCIHi] on \flair.
\ifdefined\cvprbuild\else
The \flair N6 interval excludes zero, and the A5 and A6 effects are point readings at this precision.
On a point reading, random initialization (A6) gives a smaller \rn drop on \burn, from a clean \miou of \miouRNburnCleanVarInitRandom{} rather than \miouRNburnClean.
\fi
\ifdefined\cvprbuild
On \flair the interval of this arm contrast excludes zero.
\fi

\ifdefined\plannedbuild\section{Decoder Heads, Losses and Further Blocks}
\label{sec:supp-heads}

\ifdefined\cvprbuild\else
\begin{figure*}[t]
\centering
\IfFileExists{figs/noise_examples.pdf}{\includegraphics[width=\textwidth]{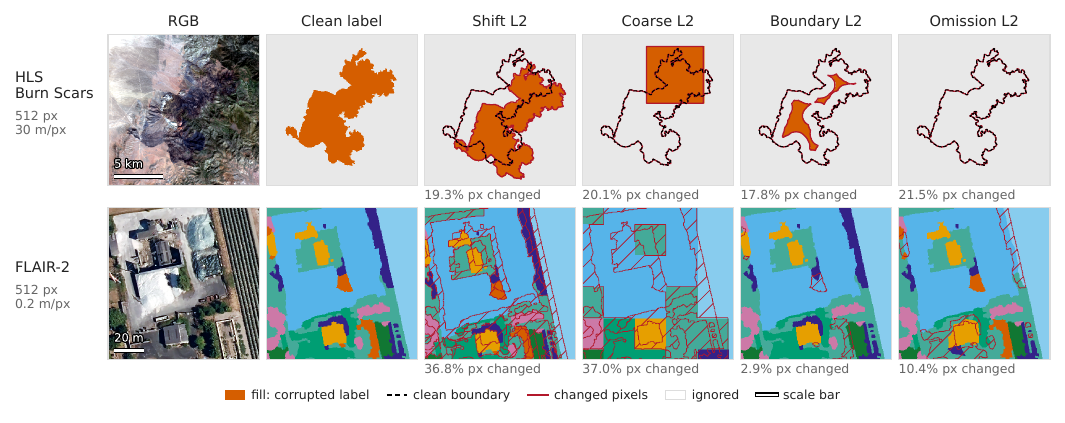}}{\phfig{3.6cm}{figs/noise\_examples.pdf from analysis/figures.py}}
\caption{\textbf{The four error types at \Ltwo:} One training tile per core dataset, \burn in the top row and \flair in the bottom row. Columns show the image with a scale bar, the clean label and the four corrupted labels of realization \CalSeed. The value under each corrupted panel is the percentage of labeled pixels of the tile whose label changed. The fill is the corrupted label and white marks ignored pixels. Black dashes trace the clean burned area on \burn. Dark red outlines the changed pixels, with a hatch on \flair.}
\label{fig:examples}
\end{figure*}

\fi
This section specifies the blocks that the main text names only, in the order of its protocol section.
Every block keeps the \gbtwo recipe, the masks of each realization and the seeds of the main grid unless stated.

\paragraph{SegFormer Head (N7):}
Block N7 repeats the main grid with the all-MLP decode head of SegFormer \cite{xie2021segformer} in place of the UNet decoder.
The head projects each of the four pyramid levels to one width with a 1$\times$1 convolution.
It then upsamples the four maps to the stride-4 level and fuses them with one further 1$\times$1 convolution.
It has no skip convolutions and no learned upsampling path, so it is the lightest common head for the five encoders.
Its parameter count per encoder is listed next to that of the UNet decoder in \Cref{sec:supp-n7}.
The recipe, the learning rate, the masks of each realization and the seeds are held fixed.
The \gbtwo recipe has one rate for encoder and decoder, and N7 keeps it so the only change is the head.
N7 holds the clean condition and the four types at both levels for the five core models.
It adds the matched-rate arm of the DINOv3 pair at the clean condition and \Ltwo.

\paragraph{Mask2Former Head (N9):}
N9 repeats the N7 cells with a query-based mask head in the form of Mask2Former \cite{cheng2022mask2former}, reduced to semantic segmentation.
A light feature-pyramid decoder brings the four levels to one width at stride 4.
Three transformer decoder layers run masked cross-attention from a fixed set of learned queries to the stride-32, 16 and 8 maps in turn.
Each layer uses the mask of the previous layer as its attention mask.
Each query predicts a class distribution and a mask embedding, and the mask logits are the dot product with the stride-4 features.
The per-pixel class scores are the query masks weighted by the query class probabilities, the semantic inference step of Mask2Former.
Their logarithm enters the same cross-entropy and \miou path as every other head, so no Hungarian matching loss is used.
The query count is 16 on the binary \burn task and 32 on the 13 classes of \flair.

\paragraph{Full Mask2Former Head (N11):}
Block N11 repeats the N7 cells with the complete Mask2Former training objective \cite{cheng2022mask2former}.
The head of N9 is kept, and the objective adds the set prediction loss of Mask2Former.
A Hungarian matching assigns the queries to the training label masks of the classes present.
The loss then sums a class term, a binary mask term and a dice term, with deep supervision over every decoder layer.
The per-pixel cross-entropy on the semantic inference output stays as an auxiliary term so the validation path and the memorization diagnostics are unchanged.
N11 therefore separates the head from its objective, which N9 could not.

\paragraph{SAM Head (N10):}
N10 repeats the N7 cells with the mask decoder of the Segment Anything Model \cite{kirillov2023segment}, run without prompts.
The stride-16 map of the encoder, projected to 256 channels, plays the role of the SAM image embedding.
One learned output token per class and one IoU token pass through two two-way transformer layers with the image embedding.
Two transposed convolutions bring the embedding to stride 4.
A hypernetwork MLP per class turns its token into the weights that read the mask logits from it.
The IoU prediction head of SAM is dropped, because no IoU target exists without a matching step.
The output is one logit map per class, so the loss and the metrics are unchanged.
Every head keeps the \gbtwo recipe, rates, masks and seeds of N7. The parameter counts of all heads are listed in \Cref{sec:supp-n7}.

\paragraph{Pretrained SAM Head (N15):}
Block N15 repeats the N7 cells with the SAM head of N10 started from the released SAM ViT-B mask decoder \cite{kirillov2023segment}.
The two-way transformer layers, the upscaling convolutions and the IoU token are loaded from the released weights.
The projection of the encoder map to 256 channels is new for each encoder.
Its normalization starts from the statistics of the SAM image embedding.
The per-class output tokens and the hypernetwork MLPs start from a random initialization, as in N10.
N15 therefore separates the SAM decoder architecture from its pretraining, which N10 could not.

\ifdefined\cvprbuild\else
\paragraph{Matched Rates for Every Model (N8):}
Block N8 extends the N6 pattern to the three encoders that have one learning rate only, and gives \lvd the \sat rate.
Each model trains at the multiplier it does not yet have on both datasets.
The cells are the clean condition and the four \Ltwo types at seeds 1 to 5.
\dofa, \tm and \rn run at one third of their \gbtwo rate and \lvd at the full \sat rate.
After N8 every core model has both multipliers at the clean condition and \Ltwo.
F2 can then be read at one common multiplier and RQ1 at either.

\paragraph{Seed Extension of F2 (N13):}
The F2 contrast on \burn was undetermined at five seeds.
Block N13 adds seeds 6 to 8 of \dofa, \tm and \rn on \burn at the clean condition and the four \Ltwo types.
With the N5 seeds of the DINOv3 pair, every core model then holds eight seeds on \burn at the clean condition and \Ltwo.
The added seeds were registered before they ran, and F2 on seeds 1 to 5 stays the confirmatory test.

\fi
\paragraph{Noise Robust Losses (N14):}
Block N14 repeats the clean condition and the four \Ltwo types of the main grid under two losses.
It covers the five core models over five seeds.
Generalized cross-entropy \cite{zhang2018gce} is $(1 - p_y^{\,q})/q$ for the probability $p_y$ of the labeled class, with $q = \GceQ$.
Soft bootstrapping \cite{reed2015bootstrapping} is the cross-entropy against the target $\beta\,y + (1 - \beta)\,p$ with $\beta = \BootBeta$, where $y$ is the one-hot label and $p$ the prediction.
Both are applied per pixel with the ignore value of the dataset, and neither setting was tuned.
Each loss has its own clean runs, so its drop is paired within the loss.

\ifdefined\cvprbuild\else
\paragraph{Pair Level (N12):}
The \Ltwo level exceeds operational label errors on \burn.
Block N12 adds a third level \Lpair{} on both core datasets whose target label \miou is measured rather than chosen.
The target is the median label \miou of the real pairs of D1, D2, D3, D4 and D6 and the Otsu pair of S1.
It holds the four types at \Lpair{} for the five core models over five seeds and the matched-rate arm of the DINOv3 pair.
It reuses the clean cells of the main grid.
The level is therefore defined only by measured pairs and is printed as \LrTargetMiou{} over \LrTargetNPairs{} pairs.

\fi
\paragraph{Memorization Rate:}
The memorization rate \cite{zhang2017understanding,arpit2017closer,liu2020elr} is measured at every validation check on a fixed set of \MemCrops{} training crops,
\begin{equation}
\memrate = \frac{\left|\{p \in \mathcal{C} : \hat{y}_p = \tilde{y}_p\}\right|}{|\mathcal{C}|}.
\label{eq:mem}
\end{equation}
Here $\mathcal{C}$ is the set of corrupted pixels, $\tilde{y}_p$ the corrupted label and $\hat{y}_p$ the prediction.
By the selected checkpoint the core models reproduce \hlMemRateRNflair\,\% to \hlMemRateLVDburn\,\% of the corrupted pixels, upper bounds that include chance agreement (\Cref{tab:mem}).
\section{Decoder Heads (N7, N9, N10, N11, N15)}
\label{sec:supp-n7}

\ifdefined\cvprbuild\fi
\ifdefined\cvprbuild\else
This section gives the decoder contrast of \ifdefined\cvprbuild\Cref{tab:decoders}\else\mtab{tab:decoders}\fi{} per model and per type, and every cell of the five decoder blocks. %
\fi
\ifdefined\cvprbuild
This section gives the decoder contrast of \Cref{tab:decoders} per model in \Cref{tab:n7} and the type ranking under every head.
\fi
The heads are specified in \Cref{sec:supp-heads}.

\begin{table*}[t]
\centering\footnotesize\setlength{\tabcolsep}{3pt}
\IfFileExists{planned/generated/table_n7.tex}{\begin{adjustbox}{max width=\textwidth}%
\providecommand{\gcvdry}[1]{\textcolor{red}{#1}}
\providecommand{\ProvTableNSeven}{Seeds per cell 5 (second-head cells). Clean mIoU and the mean paired drop over the four \Ltwo types in points, per decoder head (UNet = main grid, SegFormer = N7, Mask2Former = N9, SAM = N10, Full Mask2Former = N11, Pretrained SAM = N15). A model's UNet row takes the seeds its heads have run (seeds 1 to 5, the seeds of the decoder blocks, while none has), the pooled UNet row seeds 1 to 5. The contrast is the head's drop minus the UNet drop paired by seed (per model) or by model and seed (pooled rows) with the 95\,\% $t$ interval. Red cells are not run yet or rest on a block that is still running. Analysis of the finished runs.}
\begin{tabular}{@{}llccc@{\hspace{8pt}}ccc@{}}
\toprule
& & \multicolumn{3}{c}{\burn} & \multicolumn{3}{c}{\flair} \\
\cmidrule(lr){3-5}\cmidrule(l){6-8}
Model & Head & Clean mIoU & Drop at \Ltwo & Head $-$ UNet & Clean mIoU & Drop at \Ltwo & Head $-$ UNet \\
\midrule
\sat & UNet & 75.8 & 17.8 & -- & 52.2 & 6.5 & -- \\
 & SegFormer & 80.0 & 19.4 & 1.6 [$-$1.8, 5.0] & 50.5 & 5.1 & $-$1.4 [$-$2.6, $-$0.2] \\
 & Mask2Former & 60.0 & 7.7 & $-$10.1 [$-$29.6, 9.4] & 50.2 & 6.4 & $-$0.2 [$-$1.5, 1.2] \\
 & SAM & 73.7 & 20.3 & 2.5 [1.0, 4.0] & 46.3 & 5.5 & $-$1.0 [$-$2.4, 0.5] \\
 & Full Mask2Former & 70.5 & 16.2 & $-$1.6 [$-$4.4, 1.2] & 46.5 & 4.9 & $-$1.6 [$-$3.3, 0.0] \\
 & Pretrained SAM & 64.1 & 11.5 & $-$6.3 [$-$19.6, 7.0] & 44.8 & 5.2 & $-$1.3 [$-$2.4, $-$0.1] \\
\addlinespace[1.5pt]
\lvd & UNet & 82.0 & 14.1 & -- & 50.8 & 3.6 & -- \\
 & SegFormer & 83.4 & 15.7 & 1.6 [0.5, 2.8] & 51.3 & 3.7 & 0.1 [$-$1.3, 1.5] \\
 & Mask2Former & 83.4 & 20.9 & 6.8 [5.2, 8.5] & 50.4 & 5.6 & 2.0 [0.5, 3.6] \\
 & SAM & 82.5 & 19.2 & 5.1 [1.8, 8.3] & 46.2 & 4.4 & 0.8 [$-$1.2, 2.8] \\
 & Full Mask2Former & 79.0 & 21.8 & 7.7 [3.2, 12.3] & 49.2 & 4.4 & 0.8 [$-$0.7, 2.4] \\
 & Pretrained SAM & 73.9 & 20.4 & 6.3 [$-$9.0, 21.5] & 45.1 & 5.2 & 1.6 [0.5, 2.8] \\
\addlinespace[1.5pt]
\dofa & UNet & 85.5 & 17.9 & -- & 47.5 & 4.3 & -- \\
 & SegFormer & 86.8 & 20.8 & 3.0 [1.5, 4.5] & 47.6 & 4.6 & 0.3 [$-$1.3, 1.8] \\
 & Mask2Former & 86.4 & 19.4 & 1.5 [0.6, 2.5] & 48.2 & 5.7 & 1.4 [$-$0.5, 3.3] \\
 & SAM & 87.1 & 17.0 & $-$0.9 [$-$3.5, 1.8] & 47.5 & 5.1 & 0.8 [$-$0.1, 1.7] \\
 & Full Mask2Former & 86.8 & 18.8 & 0.9 [$-$3.2, 5.0] & 48.0 & 4.4 & 0.1 [$-$0.5, 0.8] \\
 & Pretrained SAM & 85.4 & 18.3 & 0.4 [$-$1.9, 2.8] & 45.1 & 5.4 & 1.1 [$-$0.4, 2.6] \\
\addlinespace[1.5pt]
\tm & UNet & 86.2 & 12.9 & -- & 40.5 & 5.2 & -- \\
 & SegFormer & 87.4 & 15.7 & 2.7 [$-$2.1, 7.6] & 41.9 & 5.9 & 0.7 [$-$0.8, 2.1] \\
 & Mask2Former & 88.2 & 16.9 & 3.9 [$-$2.2, 10.0] & 41.4 & 5.6 & 0.4 [$-$0.9, 1.7] \\
 & SAM & 88.7 & 16.4 & 3.4 [$-$1.1, 8.0] & 40.2 & 5.1 & $-$0.1 [$-$2.5, 2.3] \\
 & Full Mask2Former & 87.4 & 13.8 & 0.8 [$-$1.9, 3.5] & 42.8 & 5.5 & 0.3 [$-$1.9, 2.5] \\
 & Pretrained SAM & 87.8 & 14.3 & 1.3 [$-$1.9, 4.5] & 39.3 & 6.4 & 1.2 [$-$1.6, 3.9] \\
\addlinespace[1.5pt]
\rn & UNet & 88.1 & 15.9 & -- & 44.5 & 4.2 & -- \\
 & SegFormer & 88.2 & 14.1 & $-$1.8 [$-$8.2, 4.6] & 44.6 & 4.0 & $-$0.2 [$-$2.6, 2.3] \\
 & Mask2Former & 63.7 & 6.6 & $-$9.3 [$-$19.9, 1.3] & 44.0 & 4.6 & 0.5 [$-$1.9, 2.8] \\
 & SAM & 87.6 & 12.4 & $-$3.5 [$-$7.6, 0.6] & 43.8 & 3.7 & $-$0.5 [$-$2.5, 1.4] \\
 & Full Mask2Former & 88.0 & 12.1 & $-$3.8 [$-$8.0, 0.4] & 45.8 & 5.6 & 1.5 [$-$0.4, 3.4] \\
 & Pretrained SAM & 79.4 & 28.7 & 12.9 [$-$11.6, 37.3] & 40.5 & 4.2 & 0.0 [$-$1.9, 2.0] \\
\midrule
Mean over models & UNet & 83.5 & 15.7 & -- & 47.1 & 4.8 & -- \\
 & SegFormer & 85.2 & 17.1 & 1.4 [0.0, 2.8] & 47.2 & 4.7 & $-$0.1 [$-$0.7, 0.5] \\
 & Mask2Former & 76.3 & 14.3 & $-$1.4 [$-$5.7, 2.9] & 46.9 & 5.6 & 0.8 [0.2, 1.5] \\
 & SAM & 83.9 & 17.0 & 1.3 [$-$0.3, 3.0] & 44.8 & 4.8 & 0.0 [$-$0.6, 0.6] \\
 & Full Mask2Former & 82.3 & 16.5 & 0.8 [$-$1.2, 2.8] & 46.5 & 5.0 & 0.2 [$-$0.5, 0.9] \\
 & Pretrained SAM & 78.1 & 18.6 & 2.9 [$-$2.2, 8.0] & 42.9 & 5.3 & 0.5 [$-$0.2, 1.2] \\
\bottomrule
\end{tabular}
\end{adjustbox}}{\tabletbd[3cm]{planned/generated/table\_n7.tex}}
\caption{\textbf{Main grid under the decoder heads per model:} Clean test \miou and mean paired drop over the four \Ltwo types in points on seeds 1 to 5 for every core model and head. The contrast is the drop under the head minus the drop under the UNet decoder, paired by seed, with its \CiLevel\,\% interval. The last rows average the five models per head, and a dash marks the UNet rows, which have no contrast. The model blocks are \sat, \lvd, \dofa, \tm and \rn from top to bottom. \ProvSlot{\ProvNSeven}}
\label{tab:n7}
\end{table*}

\begin{table*}[t]
\centering\footnotesize
\IfFileExists{planned/generated/table_verdicts_planned.tex}{\begin{adjustbox}{max width=\textwidth}%
\providecommand{\gcvdry}[1]{\textcolor{red}{#1}}
\begin{tabular}{@{}lllllc@{}}
\toprule
Test & Contrast & Dataset & Estimate [95\,\% CI] & Holm $p$ & Verdict \\
\midrule
H5, N7 (SegFormer head) & Drop with the SegFormer head $-$ UNet decoder, four \Ltwo types, five models & \burn & 1.4 [0.0, 2.8] & 0.097 & Equivalent within $\pm$\TostMargin, SegFormer head loses more \\
 &  & \flair & $-$0.1 [$-$0.7, 0.5] & 0.721 & Equivalent within $\pm$\TostMargin \\
\addlinespace[1.5pt]
H5, N7 TOST & Equivalence of the decoder contrast within $\pm$\TostMargin{} points, 90\,\% interval & \burn & [0.3, 2.6] & -- & Equivalent \\
 &  & \flair & [$-$0.6, 0.4] & -- & Equivalent \\
\addlinespace[1.5pt]
RQ1, N7 (N6 repeated) & Drop \sat at \lvd rate $-$ \lvd, SegFormer head, four \Ltwo types & \burn & 1.4 [$-$2.1, 5.0] & -- & Undetermined, MDE 5.8 \\
 &  & \flair & 0.5 [$-$1.4, 2.3] & -- & Equivalent within $\pm$\TostMargin \\
\addlinespace[1.5pt]
RQ2, N7 (F2 repeated) & Drop mean of \dofa and \tm $-$ \rn, SegFormer head, four \Ltwo types & \burn & 4.2 [1.2, 7.1] & -- & \rn loses less \\
 &  & \flair & 1.2 [$-$0.2, 2.7] & -- & Equivalent within $\pm$\TostMargin \\
\addlinespace[1.5pt]
H5, N9 (Mask2Former head) & Drop with the Mask2Former head $-$ UNet decoder, four \Ltwo types, five models & \burn & $-$1.4 [$-$5.7, 2.9] & 0.502 & Undetermined, MDE 6.8 \\
 &  & \flair & 0.8 [0.2, 1.5] & 0.020 & Equivalent within $\pm$\TostMargin, Mask2Former head loses more \\
\addlinespace[1.5pt]
H5, N9 TOST & Equivalence of the decoder contrast within $\pm$\TostMargin{} points, 90\,\% interval & \burn & [$-$5.0, 2.1] & -- & Not shown \\
 &  & \flair & [0.3, 1.4] & -- & Equivalent \\
\addlinespace[1.5pt]
RQ1, N9 (N6 repeated) & Drop \sat at \lvd rate $-$ \lvd, Mask2Former head, four \Ltwo types & \burn & $-$2.3 [$-$4.6, 0.0] & -- & Undetermined, MDE 3.8 \\
 &  & \flair & $-$0.2 [$-$1.1, 0.6] & -- & Equivalent within $\pm$\TostMargin \\
\addlinespace[1.5pt]
RQ2, N9 (F2 repeated) & Drop mean of \dofa and \tm $-$ \rn, Mask2Former head, four \Ltwo types & \burn & 11.5 [0.9, 22.1] & -- & \rn loses less \\
 &  & \flair & 1.0 [$-$2.3, 4.4] & -- & Undetermined, MDE 5.6 \\
\addlinespace[1.5pt]
H5, N10 (SAM head) & Drop with the SAM head $-$ UNet decoder, four \Ltwo types, five models & \burn & 1.3 [$-$0.3, 3.0] & 0.226 & Equivalent within $\pm$\TostMargin \\
 &  & \flair & 0.0 [$-$0.6, 0.6] & 0.996 & Equivalent within $\pm$\TostMargin \\
\addlinespace[1.5pt]
H5, N10 TOST & Equivalence of the decoder contrast within $\pm$\TostMargin{} points, 90\,\% interval & \burn & [$-$0.1, 2.7] & -- & Equivalent \\
 &  & \flair & [$-$0.5, 0.5] & -- & Equivalent \\
\addlinespace[1.5pt]
RQ1, N10 (N6 repeated) & Drop \sat at \lvd rate $-$ \lvd, SAM head, four \Ltwo types & \burn & $-$6.5 [$-$14.4, 1.4] & -- & Undetermined, MDE 13.0 \\
 &  & \flair & 0.4 [$-$1.1, 1.9] & -- & Equivalent within $\pm$\TostMargin \\
\addlinespace[1.5pt]
RQ2, N10 (F2 repeated) & Drop mean of \dofa and \tm $-$ \rn, SAM head, four \Ltwo types & \burn & 4.3 [2.3, 6.3] & -- & \rn loses less \\
 &  & \flair & 1.4 [$-$0.5, 3.3] & -- & Equivalent within $\pm$\TostMargin \\
\addlinespace[1.5pt]
H5, N11 (full Mask2Former head) & Drop with the full Mask2Former head $-$ UNet decoder, four \Ltwo types, five models & \burn & 0.8 [$-$1.2, 2.8] & 0.809 & Equivalent within $\pm$\TostMargin \\
 &  & \flair & 0.2 [$-$0.5, 0.9] & 0.809 & Equivalent within $\pm$\TostMargin \\
\addlinespace[1.5pt]
H5, N11 TOST & Equivalence of the decoder contrast within $\pm$\TostMargin{} points, 90\,\% interval & \burn & [$-$0.8, 2.5] & -- & Equivalent \\
 &  & \flair & [$-$0.3, 0.8] & -- & Equivalent \\
\addlinespace[1.5pt]
RQ1, N11 (N6 repeated) & Drop \sat at \lvd rate $-$ \lvd, full Mask2Former head, four \Ltwo types & \burn & $-$5.4 [$-$14.3, 3.6] & -- & Undetermined, MDE 14.8 \\
 &  & \flair & 0.7 [$-$0.7, 2.0] & -- & Equivalent within $\pm$\TostMargin \\
\addlinespace[1.5pt]
RQ2, N11 (F2 repeated) & Drop mean of \dofa and \tm $-$ \rn, full Mask2Former head, four \Ltwo types & \burn & 4.2 [1.4, 6.9] & -- & \rn loses less \\
 &  & \flair & $-$0.7 [$-$2.0, 0.7] & -- & Equivalent within $\pm$\TostMargin \\
\addlinespace[1.5pt]
H5, N15 (pretrained SAM head) & Drop with the pretrained SAM head $-$ UNet decoder, four \Ltwo types, five models & \burn & 2.9 [$-$2.2, 8.0] & 0.260 & Undetermined, MDE 8.0 \\
 &  & \flair & 0.5 [$-$0.2, 1.2] & 0.260 & Equivalent within $\pm$\TostMargin \\
\addlinespace[1.5pt]
H5, N15 TOST & Equivalence of the decoder contrast within $\pm$\TostMargin{} points, 90\,\% interval & \burn & [$-$1.3, 7.2] & -- & Not shown \\
 &  & \flair & [0.0, 1.1] & -- & Equivalent \\
\addlinespace[1.5pt]
RQ1, N15 (N6 repeated) & Drop \sat at \lvd rate $-$ \lvd, pretrained SAM head, four \Ltwo types & \burn & $-$0.4 [$-$19.4, 18.7] & -- & Undetermined, MDE 31.5 \\
 &  & \flair & $-$0.5 [$-$1.6, 0.5] & -- & Equivalent within $\pm$\TostMargin \\
\addlinespace[1.5pt]
RQ2, N15 (F2 repeated) & Drop mean of \dofa and \tm $-$ \rn, pretrained SAM head, four \Ltwo types & \burn & $-$12.5 [$-$35.4, 10.5] & -- & Undetermined, MDE 37.9 \\
 &  & \flair & 1.7 [0.7, 2.6] & -- & Equivalent within $\pm$\TostMargin, \rn loses less \\
\addlinespace[1.5pt]
H6, N8 (common rate) & Drop mean of \dofa and \tm $-$ \rn, every model at $\times$1/3, four \Ltwo types & \burn & $-$0.2 [$-$3.4, 3.1] & 1.000 & Equivalent within $\pm$\TostMargin \\
 &  & \flair & 0.2 [$-$1.3, 1.8] & 1.000 & Equivalent within $\pm$\TostMargin \\
\addlinespace[1.5pt]
H6, N8 TOST & Equivalence of F2 at $\times$1/3 within $\pm$\TostMargin{} points, 90\,\% interval & \burn & [$-$2.7, 2.4] & -- & Equivalent \\
 &  & \flair & [$-$1.0, 1.4] & -- & Equivalent \\
\addlinespace[1.5pt]
RQ1, N8 (both at $\times$1) & Drop \sat $-$ \lvd at the \sat rate, four \Ltwo types & \burn & $-$1.7 [$-$5.2, 1.7] & -- & Undetermined, MDE 5.7 \\
 &  & \flair & 2.5 [1.1, 4.0] & -- & \lvd loses less \\
\addlinespace[1.5pt]
H8, N11 $-$ N9 & Drop with the full Mask2Former head $-$ Mask2Former head, four \Ltwo types, five models & \burn & 2.2 [$-$1.2, 5.7] & 0.197 & Undetermined, MDE 5.5 \\
 &  & \flair & $-$0.6 [$-$1.3, 0.1] & 0.148 & Equivalent within $\pm$\TostMargin \\
\addlinespace[1.5pt]
H8, N11 $-$ N9 TOST & Equivalence of the head pair within $\pm$\TostMargin{} points, 90\,\% interval & \burn & [$-$0.6, 5.1] & -- & Not shown \\
 &  & \flair & [$-$1.2, $-$0.1] & -- & Equivalent \\
\addlinespace[1.5pt]
N15 $-$ N10 & Drop with the pretrained SAM head $-$ SAM head, four \Ltwo types, five models & \burn & 1.6 [$-$3.9, 7.1] & 0.550 & Undetermined, MDE 8.6 \\
 &  & \flair & 0.5 [0.0, 1.1] & 0.103 & Equivalent within $\pm$\TostMargin \\
\addlinespace[1.5pt]
N15 $-$ N10 TOST & Equivalence of the head pair within $\pm$\TostMargin{} points, 90\,\% interval & \burn & [$-$2.9, 6.1] & -- & Not shown \\
 &  & \flair & [0.1, 1.0] & -- & Equivalent \\
\addlinespace[1.5pt]
N12 (level \Lpair) & Drop at the pair level \Lpair{} $-$ drop at \Ltwo, four types, five models & \burn & 6.6 [5.3, 7.9] & $<$0.001 & \Lpair{} loses more than \Ltwo \\
 &  & \flair & 3.1 [2.8, 3.4] & $<$0.001 & \Lpair{} loses more than \Ltwo \\
\addlinespace[1.5pt]
RQ2, N13 (F2, every seed) & Drop mean of \dofa and \tm $-$ \rn, four \Ltwo types, seeds 1 to 8 on \burn & \burn & 0.0 [$-$2.8, 2.7] & 1.000 & Equivalent within $\pm$\TostMargin \\
 &  & \flair & 0.6 [$-$2.0, 3.1] & 1.000 & Equivalent within $\pm$\TostMargin \\
\addlinespace[1.5pt]
N14 (GCE) & Drop with generalized cross-entropy $-$ drop with cross-entropy, four \Ltwo types, five models & \burn & $-$1.2 [$-$2.9, 0.4] & 0.285 & Equivalent within $\pm$\TostMargin \\
 &  & \flair & 0.1 [$-$0.8, 0.9] & 0.819 & Equivalent within $\pm$\TostMargin \\
\addlinespace[1.5pt]
N14 (Soft bootstrapping) & Drop with soft bootstrapping $-$ drop with cross-entropy, four \Ltwo types, five models & \burn & $-$1.2 [$-$2.5, 0.1] & 0.153 & Equivalent within $\pm$\TostMargin \\
 &  & \flair & 0.1 [$-$0.5, 0.6] & 0.774 & Equivalent within $\pm$\TostMargin \\
\addlinespace[1.5pt]
H7 (real against simulated) & Drop under the real label $-$ drop under its matched simulated type, five models & Chesapeake & 9.2 [8.5, 10.0] & $<$0.001 & Real costs more \\
 &  & CloudSEN12 & 24.6 [23.6, 25.5] & $<$0.001 & Real costs more \\
 &  & DFC2020 & 3.4 [2.2, 4.6] & $<$0.001 & Real costs more \\
 &  & EnviroAtlas & 21.4 [19.1, 23.7] & $<$0.001 & Real costs more \\
 &  & Kuro Siwo & 0.7 [$-$1.4, 2.8] & 1.000 & Equivalent within $\pm$\TostMargin \\
\bottomrule
\end{tabular}
\end{adjustbox}}{\tabletbd[3cm]{planned/generated/table\_verdicts\_planned.tex}}
\caption{\textbf{Verdicts of the later hypotheses and blocks:} Rows in the layout of \mtab{tab:verdicts} for H5 to H8 and every further contrast of the later blocks. A verdict names a direction where the unadjusted \CiLevel\,\% interval excludes zero, whatever its Holm $p$. A positive drop contrast means the first arm loses more. A TOST row gives the 90\,\% equivalence interval at $\pm$\TostMargin{} points, and Not shown means that the interval leaves the margin. MDE is the minimum detectable effect, and a dash marks a quantity the row does not have.} %
\label{tab:supp-verdicts-post}
\end{table*}

\paragraph{Parameter Counts:}
The SegFormer head \cite{xie2021segformer} has \ParamsDecSegfSAT{} million parameters on the DINOv3 pair, \ParamsDecSegfDOFA{} on \dofa, \ParamsDecSegfTM{} on \tm and \ParamsDecSegfRN{} on \rn.
The Mask2Former head \cite{cheng2022mask2former} has \ParamsDecMtwoFSAT, \ParamsDecMtwoFDOFA, \ParamsDecMtwoFTM{} and \ParamsDecMtwoFRN{} million on \flair, where it holds 32 queries over 13 classes.
On \burn it has \ParamsDecMtwoFBurnSAT, \ParamsDecMtwoFBurnDOFA, \ParamsDecMtwoFBurnTM{} and \ParamsDecMtwoFBurnRN{} million with 16 queries.
The matching objective of N11 adds no parameter to it.
The SAM head \cite{kirillov2023segment} has \ParamsDecSamSAT, \ParamsDecSamDOFA, \ParamsDecSamTM{} and \ParamsDecSamRN{} million on \flair, with one hypernetwork per class.
On \burn it has \ParamsDecSamBurnSAT, \ParamsDecSamBurnDOFA, \ParamsDecSamBurnTM{} and \ParamsDecSamBurnRN{} million.
The UNet decoder has \ParamsDecUNetSAT, \ParamsDecUNetDOFA, \ParamsDecUNetTM{} and \ParamsDecUNetRN{} million on the same encoders.
The analysis keeps the decoder cells apart from the main-grid cells.

\paragraph{Decoder Contrast with Equivalence Intervals:}
The 90\,\% equivalence intervals of the SegFormer contrast are [\hlDecInvburnTostCILo, \hlDecInvburnTostCIHi] on \burn and [\hlDecInvflairTostCILo, \hlDecInvflairTostCIHi] on \flair.
Both lie inside the margin, and the tests resolve \hlDecInvburnMde{} and \hlDecInvflairMde{} points.
For the Mask2Former head the intervals are [\hlDecInvMtwoFburnTostCILo, \hlDecInvMtwoFburnTostCIHi] and [\hlDecInvMtwoFflairTostCILo, \hlDecInvMtwoFflairTostCIHi].
For the SAM head they are [\hlDecInvSamburnTostCILo, \hlDecInvSamburnTostCIHi] and [\hlDecInvSamflairTostCILo, \hlDecInvSamflairTostCIHi].
With \CiLevel\,\% intervals, the full Mask2Former head changes the mean \Ltwo drop against the UNet decoder by \hlDecInvMtwoFFburnEst{} [\hlDecInvMtwoFFburnCILo, \hlDecInvMtwoFFburnCIHi] points on \burn.
On \flair the change is \hlDecInvMtwoFFflairEst{} [\hlDecInvMtwoFFflairCILo, \hlDecInvMtwoFFflairCIHi] points, so H5 reads \vd{\hlDecVerdictMtwoFFburn} and \vd{\hlDecVerdictMtwoFFflair} under it.
The pretrained SAM head changes the drop by \hlDecInvSamPreburnEst{} [\hlDecInvSamPreburnCILo, \hlDecInvSamPreburnCIHi] points on \burn and \hlDecInvSamPreflairEst{} [\hlDecInvSamPreflairCILo, \hlDecInvSamPreflairCIHi] on \flair.
It reads \vd{\hlDecVerdictSamPreburn} and \vd{\hlDecVerdictSamPreflair}.
The clean \miou averages \hlDecCleanSegfburn{} with the SegFormer head, \hlDecCleanMtwoFburn{} with the Mask2Former head and \hlDecCleanSamburn{} with the SAM head on \burn.
The UNet decoder reaches \hlDecCleanUNetburn{} there.
On \flair the four values are \hlDecCleanSegfflair, \hlDecCleanMtwoFflair, \hlDecCleanSamflair{} and \hlDecCleanUNetflair.

\ifdefined\cvprbuild\else
\begin{figure}[t]
\centering
\IfFileExists{planned/figs/fig_n7_decoder.pdf}{\includegraphics[width=\linewidth]{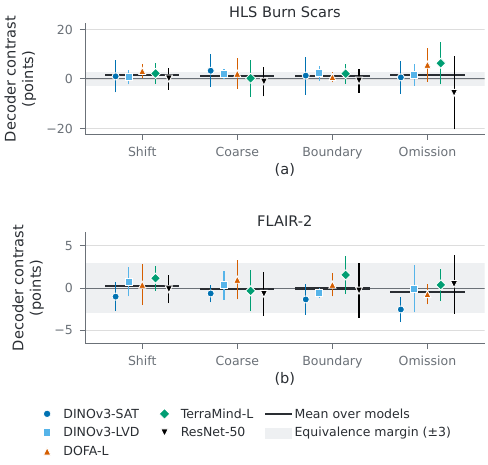}}{\phfig{5cm}{planned/figs/fig\_n7\_decoder.pdf}}
\caption{\textbf{SegFormer decoder contrast per error type:} Paired drop under the SegFormer head minus the drop under the UNet decoder in points at \Ltwo, for every core model and type. Whiskers are \CiLevel\,\% intervals and black bars the means over models. Panel (a) is \burn and (b) \flair. The gray band is the equivalence margin of $\pm$\TostMargin{} points. \ProvSlot{\ProvNSeven}}
\label{fig:n7}
\end{figure}

\begin{figure}[t]
\centering
\IfFileExists{planned/figs/fig_n9_decoder.pdf}{\includegraphics[width=\linewidth]{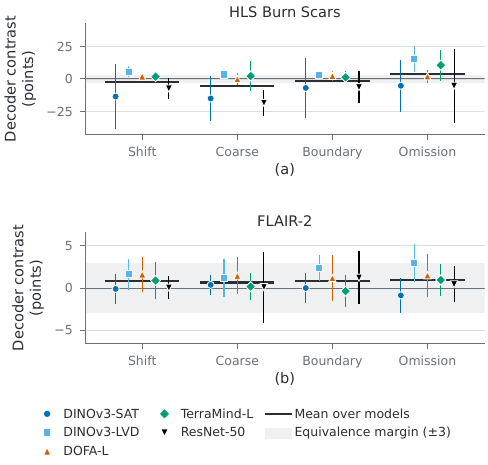}}{\phfig{5cm}{planned/figs/fig\_n9\_decoder.pdf}}
\caption{\textbf{Mask2Former decoder contrast per error type:} Paired drop under the Mask2Former head minus the drop under the UNet decoder in points at \Ltwo, for every core model and type. A short black bar marks the mean over models. The layout is that of \Cref{fig:n7}. \ProvSlot{\ProvNNine}}
\label{fig:n9}
\end{figure}

\begin{figure}[t]
\centering
\IfFileExists{planned/figs/fig_n11_decoder.pdf}{\includegraphics[width=\linewidth]{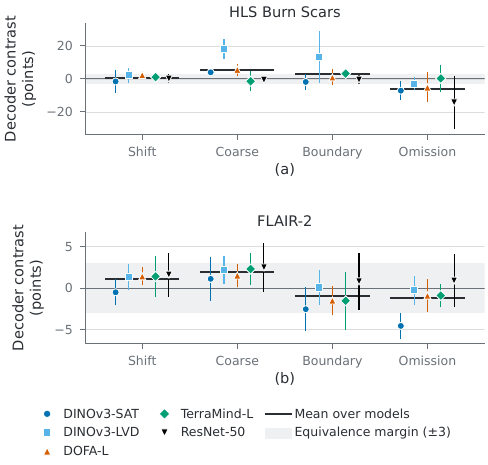}}{\phfig{5cm}{planned/figs/fig\_n11\_decoder.pdf}}
\caption{\textbf{Full Mask2Former decoder contrast per error type:} Paired drop under the full Mask2Former head minus the drop under the UNet decoder in points at \Ltwo, for every core model and type. A short black bar marks the mean over models. The layout is that of \Cref{fig:n7}. \ProvSlot{\ProvNEleven}}
\label{fig:n11}
\end{figure}

\begin{figure}[t]
\centering
\IfFileExists{planned/figs/fig_n10_decoder.pdf}{\includegraphics[width=\linewidth]{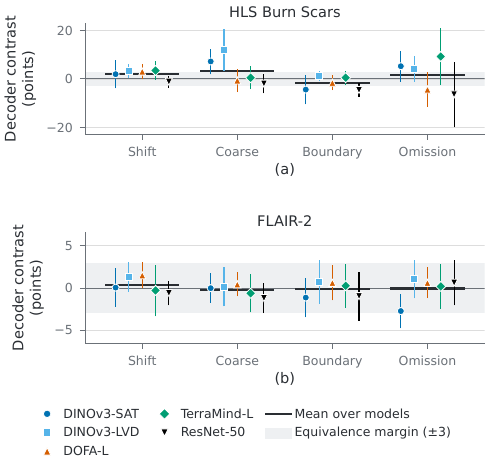}}{\phfig{5cm}{planned/figs/fig\_n10\_decoder.pdf}}
\caption{\textbf{SAM decoder contrast per error type:} Paired drop under the SAM head minus the drop under the UNet decoder in points at \Ltwo, for every core model and type. A short black bar marks the mean over models. The layout is that of \Cref{fig:n7}. \ProvSlot{\ProvNTen}}
\label{fig:n10}
\end{figure}

\begin{figure}[t]
\centering
\IfFileExists{planned/figs/fig_n15_decoder.pdf}{\includegraphics[width=\linewidth]{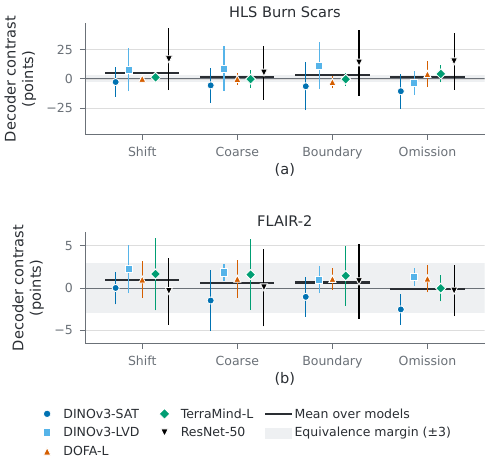}}{\phfig{5cm}{planned/figs/fig\_n15\_decoder.pdf}}
\caption{\textbf{Pretrained SAM decoder contrast per error type:} Paired drop under the pretrained SAM head minus the drop under the UNet decoder in points at \Ltwo, for every core model and type. A short black bar marks the mean over models. The layout is that of \Cref{fig:n7}. \ProvSlot{\ProvNFifteen}}
\label{fig:n15}
\end{figure}

\fi
\paragraph{Contrasts under the SegFormer Head (N7):}
Under the SegFormer head the \Ltwo type ranking by mean drop is \etype{\hlDecTypeRankburnOne}, \etype{\hlDecTypeRankburnTwo}, \etype{\hlDecTypeRankburnThree} and \etype{\hlDecTypeRankburnFour} on \burn.
On \flair it is \etype{\hlDecTypeRankflairOne}, \etype{\hlDecTypeRankflairTwo}, \etype{\hlDecTypeRankflairThree} and \etype{\hlDecTypeRankflairFour}\ifdefined\cvprbuild\else{} (\Cref{fig:n7})\fi{}.
\ifdefined\cvprbuild\else
The per-model contrasts on \burn are \hlDecInvSATburnEst{} [\hlDecInvSATburnCILo, \hlDecInvSATburnCIHi] for \sat, \hlDecInvLVDburnEst{} [\hlDecInvLVDburnCILo, \hlDecInvLVDburnCIHi] for \lvd and \hlDecInvDOFAburnEst{} [\hlDecInvDOFAburnCILo, \hlDecInvDOFAburnCIHi] for \dofa.
For \tm and \rn they are \hlDecInvTMburnEst{} [\hlDecInvTMburnCILo, \hlDecInvTMburnCIHi] and \hlDecInvRNburnEst{} [\hlDecInvRNburnCILo, \hlDecInvRNburnCIHi] points.
On \flair they are \hlDecInvSATflairEst{} [\hlDecInvSATflairCILo, \hlDecInvSATflairCIHi] for \sat, \hlDecInvLVDflairEst{} [\hlDecInvLVDflairCILo, \hlDecInvLVDflairCIHi] for \lvd and \hlDecInvDOFAflairEst{} [\hlDecInvDOFAflairCILo, \hlDecInvDOFAflairCIHi] for \dofa.
For \tm and \rn they are \hlDecInvTMflairEst{} [\hlDecInvTMflairCILo, \hlDecInvTMflairCIHi] and \hlDecInvRNflairEst{} [\hlDecInvRNflairCILo, \hlDecInvRNflairCIHi] points.
F1 under this head is \hlDecFOneburnEst{} [\hlDecFOneburnCILo, \hlDecFOneburnCIHi] on \burn and \hlDecFOneflairEst{} [\hlDecFOneflairCILo, \hlDecFOneflairCIHi] on \flair, with Holm $p$ \hlDecFOneburnPHolm{} and \hlDecFOneflairPHolm.
N6 under it is \hlDecNSixburnEst{} [\hlDecNSixburnCILo, \hlDecNSixburnCIHi] on \burn and \hlDecNSixflairEst{} [\hlDecNSixflairCILo, \hlDecNSixflairCIHi] on \flair, and reads \vd{\hlDecVerdictNSixburn} and \vd{\hlDecVerdictNSixflair}.
F2 under it is \hlDecFTwoburnEst{} [\hlDecFTwoburnCILo, \hlDecFTwoburnCIHi] on \burn and \hlDecFTwoflairEst{} [\hlDecFTwoflairCILo, \hlDecFTwoflairCIHi] on \flair, and reads \vd{\hlDecVerdictFTwoburn} and \vd{\hlDecVerdictFTwoflair}.
\fi

\paragraph{Contrasts under the Mask2Former Head (N9):}
The \Ltwo type ranking by mean drop under the Mask2Former head is \etype{\hlDecTypeRankMtwoFburnOne}, \etype{\hlDecTypeRankMtwoFburnTwo}, \etype{\hlDecTypeRankMtwoFburnThree} and \etype{\hlDecTypeRankMtwoFburnFour} on \burn.
On \flair it is \etype{\hlDecTypeRankMtwoFflairOne}, \etype{\hlDecTypeRankMtwoFflairTwo}, \etype{\hlDecTypeRankMtwoFflairThree} and \etype{\hlDecTypeRankMtwoFflairFour}\ifdefined\cvprbuild\else{} (\Cref{fig:n9})\fi{}.
\ifdefined\cvprbuild\else
The per-model contrasts on \burn are \hlDecInvMtwoFSATburnEst{} [\hlDecInvMtwoFSATburnCILo, \hlDecInvMtwoFSATburnCIHi] for \sat, \hlDecInvMtwoFLVDburnEst{} [\hlDecInvMtwoFLVDburnCILo, \hlDecInvMtwoFLVDburnCIHi] for \lvd and \hlDecInvMtwoFDOFAburnEst{} [\hlDecInvMtwoFDOFAburnCILo, \hlDecInvMtwoFDOFAburnCIHi] for \dofa.
For \tm and \rn they are \hlDecInvMtwoFTMburnEst{} [\hlDecInvMtwoFTMburnCILo, \hlDecInvMtwoFTMburnCIHi] and \hlDecInvMtwoFRNburnEst{} [\hlDecInvMtwoFRNburnCILo, \hlDecInvMtwoFRNburnCIHi] points.
On \flair they are \hlDecInvMtwoFSATflairEst{} [\hlDecInvMtwoFSATflairCILo, \hlDecInvMtwoFSATflairCIHi] for \sat, \hlDecInvMtwoFLVDflairEst{} [\hlDecInvMtwoFLVDflairCILo, \hlDecInvMtwoFLVDflairCIHi] for \lvd and \hlDecInvMtwoFDOFAflairEst{} [\hlDecInvMtwoFDOFAflairCILo, \hlDecInvMtwoFDOFAflairCIHi] for \dofa.
For \tm and \rn they are \hlDecInvMtwoFTMflairEst{} [\hlDecInvMtwoFTMflairCILo, \hlDecInvMtwoFTMflairCIHi] and \hlDecInvMtwoFRNflairEst{} [\hlDecInvMtwoFRNflairCILo, \hlDecInvMtwoFRNflairCIHi] points.
F1 under this head is \hlDecFOneMtwoFburnEst{} [\hlDecFOneMtwoFburnCILo, \hlDecFOneMtwoFburnCIHi] on \burn and \hlDecFOneMtwoFflairEst{} [\hlDecFOneMtwoFflairCILo, \hlDecFOneMtwoFflairCIHi] on \flair, with Holm $p$ \hlDecFOneMtwoFburnPHolm{} and \hlDecFOneMtwoFflairPHolm.
N6 under it is \hlDecNSixMtwoFburnEst{} [\hlDecNSixMtwoFburnCILo, \hlDecNSixMtwoFburnCIHi] on \burn and \hlDecNSixMtwoFflairEst{} [\hlDecNSixMtwoFflairCILo, \hlDecNSixMtwoFflairCIHi] on \flair, and reads \vd{\hlDecVerdictNSixMtwoFburn} and \vd{\hlDecVerdictNSixMtwoFflair}.
F2 under it is \hlDecFTwoMtwoFburnEst{} [\hlDecFTwoMtwoFburnCILo, \hlDecFTwoMtwoFburnCIHi] on \burn and \hlDecFTwoMtwoFflairEst{} [\hlDecFTwoMtwoFflairCILo, \hlDecFTwoMtwoFflairCIHi] on \flair, and reads \vd{\hlDecVerdictFTwoMtwoFburn} and \vd{\hlDecVerdictFTwoMtwoFflair}.
\fi

\paragraph{Contrasts under the Full Mask2Former Head (N11):}
With the full Mask2Former head the \Ltwo type ranking by mean drop is \etype{\hlDecTypeRankMtwoFFburnOne}, \etype{\hlDecTypeRankMtwoFFburnTwo}, \etype{\hlDecTypeRankMtwoFFburnThree} and \etype{\hlDecTypeRankMtwoFFburnFour} on \burn.
On \flair it is \etype{\hlDecTypeRankMtwoFFflairOne}, \etype{\hlDecTypeRankMtwoFFflairTwo}, \etype{\hlDecTypeRankMtwoFFflairThree} and \etype{\hlDecTypeRankMtwoFFflairFour}\ifdefined\cvprbuild\else{} (\Cref{fig:n11})\fi{}.
\ifdefined\cvprbuild\else
The per-model contrasts on \burn are \hlDecInvMtwoFFSATburnEst{} [\hlDecInvMtwoFFSATburnCILo, \hlDecInvMtwoFFSATburnCIHi] for \sat, \hlDecInvMtwoFFLVDburnEst{} [\hlDecInvMtwoFFLVDburnCILo, \hlDecInvMtwoFFLVDburnCIHi] for \lvd and \hlDecInvMtwoFFDOFAburnEst{} [\hlDecInvMtwoFFDOFAburnCILo, \hlDecInvMtwoFFDOFAburnCIHi] for \dofa.
For \tm and \rn they are \hlDecInvMtwoFFTMburnEst{} [\hlDecInvMtwoFFTMburnCILo, \hlDecInvMtwoFFTMburnCIHi] and \hlDecInvMtwoFFRNburnEst{} [\hlDecInvMtwoFFRNburnCILo, \hlDecInvMtwoFFRNburnCIHi] points.
On \flair they are \hlDecInvMtwoFFSATflairEst{} [\hlDecInvMtwoFFSATflairCILo, \hlDecInvMtwoFFSATflairCIHi] for \sat, \hlDecInvMtwoFFLVDflairEst{} [\hlDecInvMtwoFFLVDflairCILo, \hlDecInvMtwoFFLVDflairCIHi] for \lvd and \hlDecInvMtwoFFDOFAflairEst{} [\hlDecInvMtwoFFDOFAflairCILo, \hlDecInvMtwoFFDOFAflairCIHi] for \dofa.
For \tm and \rn they are \hlDecInvMtwoFFTMflairEst{} [\hlDecInvMtwoFFTMflairCILo, \hlDecInvMtwoFFTMflairCIHi] and \hlDecInvMtwoFFRNflairEst{} [\hlDecInvMtwoFFRNflairCILo, \hlDecInvMtwoFFRNflairCIHi] points.
F1 under this head is \hlDecFOneMtwoFFburnEst{} [\hlDecFOneMtwoFFburnCILo, \hlDecFOneMtwoFFburnCIHi] on \burn and \hlDecFOneMtwoFFflairEst{} [\hlDecFOneMtwoFFflairCILo, \hlDecFOneMtwoFFflairCIHi] on \flair, with Holm $p$ \hlDecFOneMtwoFFburnPHolm{} and \hlDecFOneMtwoFFflairPHolm.
N6 under it is \hlDecNSixMtwoFFburnEst{} [\hlDecNSixMtwoFFburnCILo, \hlDecNSixMtwoFFburnCIHi] on \burn and \hlDecNSixMtwoFFflairEst{} [\hlDecNSixMtwoFFflairCILo, \hlDecNSixMtwoFFflairCIHi] on \flair, and reads \vd{\hlDecVerdictNSixMtwoFFburn} and \vd{\hlDecVerdictNSixMtwoFFflair}.
F2 under it is \hlDecFTwoMtwoFFburnEst{} [\hlDecFTwoMtwoFFburnCILo, \hlDecFTwoMtwoFFburnCIHi] on \burn and \hlDecFTwoMtwoFFflairEst{} [\hlDecFTwoMtwoFFflairCILo, \hlDecFTwoMtwoFFflairCIHi] on \flair, and reads \vd{\hlDecVerdictFTwoMtwoFFburn} and \vd{\hlDecVerdictFTwoMtwoFFflair}.
\fi

\paragraph{Contrasts under the SAM Head (N10):}
Under the SAM head the \Ltwo type ranking by mean drop is \etype{\hlDecTypeRankSamburnOne}, \etype{\hlDecTypeRankSamburnTwo}, \etype{\hlDecTypeRankSamburnThree} and \etype{\hlDecTypeRankSamburnFour} on \burn.
On \flair it is \etype{\hlDecTypeRankSamflairOne}, \etype{\hlDecTypeRankSamflairTwo}, \etype{\hlDecTypeRankSamflairThree} and \etype{\hlDecTypeRankSamflairFour}\ifdefined\cvprbuild\else{} (\Cref{fig:n10})\fi{}.
\ifdefined\cvprbuild\else
The per-model contrasts on \burn are \hlDecInvSamSATburnEst{} [\hlDecInvSamSATburnCILo, \hlDecInvSamSATburnCIHi] for \sat, \hlDecInvSamLVDburnEst{} [\hlDecInvSamLVDburnCILo, \hlDecInvSamLVDburnCIHi] for \lvd and \hlDecInvSamDOFAburnEst{} [\hlDecInvSamDOFAburnCILo, \hlDecInvSamDOFAburnCIHi] for \dofa.
For \tm and \rn they are \hlDecInvSamTMburnEst{} [\hlDecInvSamTMburnCILo, \hlDecInvSamTMburnCIHi] and \hlDecInvSamRNburnEst{} [\hlDecInvSamRNburnCILo, \hlDecInvSamRNburnCIHi] points.
On \flair they are \hlDecInvSamSATflairEst{} [\hlDecInvSamSATflairCILo, \hlDecInvSamSATflairCIHi] for \sat, \hlDecInvSamLVDflairEst{} [\hlDecInvSamLVDflairCILo, \hlDecInvSamLVDflairCIHi] for \lvd and \hlDecInvSamDOFAflairEst{} [\hlDecInvSamDOFAflairCILo, \hlDecInvSamDOFAflairCIHi] for \dofa.
For \tm and \rn they are \hlDecInvSamTMflairEst{} [\hlDecInvSamTMflairCILo, \hlDecInvSamTMflairCIHi] and \hlDecInvSamRNflairEst{} [\hlDecInvSamRNflairCILo, \hlDecInvSamRNflairCIHi] points.
F1 under this head is \hlDecFOneSamburnEst{} [\hlDecFOneSamburnCILo, \hlDecFOneSamburnCIHi] on \burn and \hlDecFOneSamflairEst{} [\hlDecFOneSamflairCILo, \hlDecFOneSamflairCIHi] on \flair, with Holm $p$ \hlDecFOneSamburnPHolm{} and \hlDecFOneSamflairPHolm.
N6 under it is \hlDecNSixSamburnEst{} [\hlDecNSixSamburnCILo, \hlDecNSixSamburnCIHi] on \burn and \hlDecNSixSamflairEst{} [\hlDecNSixSamflairCILo, \hlDecNSixSamflairCIHi] on \flair, and reads \vd{\hlDecVerdictNSixSamburn} and \vd{\hlDecVerdictNSixSamflair}.
F2 under it is \hlDecFTwoSamburnEst{} [\hlDecFTwoSamburnCILo, \hlDecFTwoSamburnCIHi] on \burn and \hlDecFTwoSamflairEst{} [\hlDecFTwoSamflairCILo, \hlDecFTwoSamflairCIHi] on \flair, and reads \vd{\hlDecVerdictFTwoSamburn} and \vd{\hlDecVerdictFTwoSamflair}.
\fi

\paragraph{Contrasts under the Pretrained SAM Head (N15):}
The \Ltwo type ranking by mean drop under the pretrained SAM head is \etype{\hlDecTypeRankSamPreburnOne}, \etype{\hlDecTypeRankSamPreburnTwo}, \etype{\hlDecTypeRankSamPreburnThree} and \etype{\hlDecTypeRankSamPreburnFour} on \burn.
On \flair it is \etype{\hlDecTypeRankSamPreflairOne}, \etype{\hlDecTypeRankSamPreflairTwo}, \etype{\hlDecTypeRankSamPreflairThree} and \etype{\hlDecTypeRankSamPreflairFour}\ifdefined\cvprbuild\else{} (\Cref{fig:n15})\fi{}.
\ifdefined\cvprbuild\else
The per-model contrasts on \burn are \hlDecInvSamPreSATburnEst{} [\hlDecInvSamPreSATburnCILo, \hlDecInvSamPreSATburnCIHi] for \sat, \hlDecInvSamPreLVDburnEst{} [\hlDecInvSamPreLVDburnCILo, \hlDecInvSamPreLVDburnCIHi] for \lvd and \hlDecInvSamPreDOFAburnEst{} [\hlDecInvSamPreDOFAburnCILo, \hlDecInvSamPreDOFAburnCIHi] for \dofa.
For \tm and \rn they are \hlDecInvSamPreTMburnEst{} [\hlDecInvSamPreTMburnCILo, \hlDecInvSamPreTMburnCIHi] and \hlDecInvSamPreRNburnEst{} [\hlDecInvSamPreRNburnCILo, \hlDecInvSamPreRNburnCIHi] points.
On \flair they are \hlDecInvSamPreSATflairEst{} [\hlDecInvSamPreSATflairCILo, \hlDecInvSamPreSATflairCIHi] for \sat, \hlDecInvSamPreLVDflairEst{} [\hlDecInvSamPreLVDflairCILo, \hlDecInvSamPreLVDflairCIHi] for \lvd and \hlDecInvSamPreDOFAflairEst{} [\hlDecInvSamPreDOFAflairCILo, \hlDecInvSamPreDOFAflairCIHi] for \dofa.
For \tm and \rn they are \hlDecInvSamPreTMflairEst{} [\hlDecInvSamPreTMflairCILo, \hlDecInvSamPreTMflairCIHi] and \hlDecInvSamPreRNflairEst{} [\hlDecInvSamPreRNflairCILo, \hlDecInvSamPreRNflairCIHi] points.
F1 under this head is \hlDecFOneSamPreburnEst{} [\hlDecFOneSamPreburnCILo, \hlDecFOneSamPreburnCIHi] on \burn and \hlDecFOneSamPreflairEst{} [\hlDecFOneSamPreflairCILo, \hlDecFOneSamPreflairCIHi] on \flair, with Holm $p$ \hlDecFOneSamPreburnPHolm{} and \hlDecFOneSamPreflairPHolm.
N6 under it is \hlDecNSixSamPreburnEst{} [\hlDecNSixSamPreburnCILo, \hlDecNSixSamPreburnCIHi] on \burn and \hlDecNSixSamPreflairEst{} [\hlDecNSixSamPreflairCILo, \hlDecNSixSamPreflairCIHi] on \flair, and reads \vd{\hlDecVerdictNSixSamPreburn} and \vd{\hlDecVerdictNSixSamPreflair}.
F2 under it is \hlDecFTwoSamPreburnEst{} [\hlDecFTwoSamPreburnCILo, \hlDecFTwoSamPreburnCIHi] on \burn and \hlDecFTwoSamPreflairEst{} [\hlDecFTwoSamPreflairCILo, \hlDecFTwoSamPreflairCIHi] on \flair, and reads \vd{\hlDecVerdictFTwoSamPreburn} and \vd{\hlDecVerdictFTwoSamPreflair}.
\fi

\paragraph{Objective and Initialization:}
The matching objective changes the mean \Ltwo drop of the Mask2Former head by \hlDecFullVsLiteburnEst{} [\hlDecFullVsLiteburnCILo, \hlDecFullVsLiteburnCIHi] points on \burn and \hlDecFullVsLiteflairEst{} [\hlDecFullVsLiteflairCILo, \hlDecFullVsLiteflairCIHi] on \flair.
H8 reads \vd{\hlDecVerdictFullVsLiteburn} and \vd{\hlDecVerdictFullVsLiteflair}.
The released weights change the mean \Ltwo drop of the SAM head by \hlDecPretrainedVsScratchburnEst{} [\hlDecPretrainedVsScratchburnCILo, \hlDecPretrainedVsScratchburnCIHi] points on \burn and \hlDecPretrainedVsScratchflairEst{} [\hlDecPretrainedVsScratchflairCILo, \hlDecPretrainedVsScratchflairCIHi] on \flair.
That contrast reads \vd{\hlDecVerdictPretrainedVsScratchburn} and \vd{\hlDecVerdictPretrainedVsScratchflair}.

\ifdefined\cvprbuild\else
\begin{table}[t]
\centering\footnotesize\setlength{\tabcolsep}{3pt}
\ifdefined\preprintbuild\renewcommand{\arraystretch}{0.95}\fi%
\IfFileExists{planned/generated/table_n7_grid_burn.tex}{\begin{adjustbox}{max width=\columnwidth}%
\providecommand{\gcvdry}[1]{\textcolor{red}{#1}}
\providecommand{\ProvTableNSevenGridBurn}{Seeds per cell 5. Mean $\pm$ standard deviation over seeds, SegFormer head, \burn. Red cells are not run yet or still have a pending seed. Analysis of the finished runs.}
\begin{tabular}{@{}lllcccc@{}}
\toprule
Model & Type & Level & $n$ & mIoU & Drop & Retention (\%) \\
\midrule
\sat & Clean & -- & 5 & 80.0 $\pm$ 1.9 & -- & -- \\
 & Shift & L1 & 5 & 68.8 $\pm$ 13.6 & 11.2 $\pm$ 15.1 & 86.3 $\pm$ 18.4 \\
 & Shift & L2 & 5 & 64.9 $\pm$ 4.0 & 15.1 $\pm$ 3.9 & 81.2 $\pm$ 4.8 \\
 & Coarse & L1 & 5 & 75.3 $\pm$ 2.1 & 4.7 $\pm$ 2.5 & 94.2 $\pm$ 3.0 \\
 & Coarse & L2 & 5 & 54.9 $\pm$ 5.1 & 25.0 $\pm$ 4.3 & 68.6 $\pm$ 5.6 \\
 & Boundary & L1 & 5 & 70.1 $\pm$ 6.6 & 9.9 $\pm$ 8.2 & 87.8 $\pm$ 10.0 \\
 & Boundary & L2 & 5 & 63.7 $\pm$ 1.9 & 16.3 $\pm$ 2.2 & 79.7 $\pm$ 2.5 \\
 & Omission & L1 & 5 & 66.0 $\pm$ 4.5 & 14.0 $\pm$ 5.6 & 82.6 $\pm$ 6.7 \\
 & Omission & L2 & 5 & 58.9 $\pm$ 2.5 & 21.1 $\pm$ 2.7 & 73.7 $\pm$ 3.1 \\
\addlinespace[1.5pt]
\lvd & Clean & -- & 5 & 83.4 $\pm$ 1.0 & -- & -- \\
 & Shift & L1 & 5 & 81.8 $\pm$ 1.2 & 1.6 $\pm$ 1.1 & 98.1 $\pm$ 1.4 \\
 & Shift & L2 & 5 & 72.0 $\pm$ 1.3 & 11.3 $\pm$ 1.4 & 86.4 $\pm$ 1.5 \\
 & Coarse & L1 & 5 & 79.7 $\pm$ 0.8 & 3.6 $\pm$ 1.2 & 95.6 $\pm$ 1.4 \\
 & Coarse & L2 & 5 & 65.5 $\pm$ 2.1 & 17.9 $\pm$ 1.4 & 78.5 $\pm$ 1.8 \\
 & Boundary & L1 & 5 & 79.2 $\pm$ 1.3 & 4.2 $\pm$ 2.1 & 95.0 $\pm$ 2.5 \\
 & Boundary & L2 & 5 & 68.3 $\pm$ 2.0 & 15.1 $\pm$ 2.7 & 81.9 $\pm$ 3.1 \\
 & Omission & L1 & 5 & 72.4 $\pm$ 5.1 & 11.0 $\pm$ 4.2 & 86.8 $\pm$ 5.3 \\
 & Omission & L2 & 5 & 64.8 $\pm$ 4.5 & 18.6 $\pm$ 3.9 & 77.7 $\pm$ 4.8 \\
\addlinespace[1.5pt]
\dofa & Clean & -- & 5 & 86.8 $\pm$ 0.9 & -- & -- \\
 & Shift & L1 & 5 & 84.0 $\pm$ 1.2 & 2.9 $\pm$ 1.9 & 96.7 $\pm$ 2.2 \\
 & Shift & L2 & 5 & 72.5 $\pm$ 1.2 & 14.4 $\pm$ 1.0 & 83.5 $\pm$ 1.1 \\
 & Coarse & L1 & 5 & 81.9 $\pm$ 1.1 & 4.9 $\pm$ 1.7 & 94.4 $\pm$ 1.9 \\
 & Coarse & L2 & 5 & 64.1 $\pm$ 2.2 & 22.7 $\pm$ 2.5 & 73.8 $\pm$ 2.8 \\
 & Boundary & L1 & 5 & 83.0 $\pm$ 0.8 & 3.8 $\pm$ 0.7 & 95.6 $\pm$ 0.8 \\
 & Boundary & L2 & 5 & 71.7 $\pm$ 1.7 & 15.1 $\pm$ 2.2 & 82.6 $\pm$ 2.4 \\
 & Omission & L1 & 5 & 75.1 $\pm$ 3.0 & 11.7 $\pm$ 2.3 & 86.5 $\pm$ 2.7 \\
 & Omission & L2 & 5 & 55.7 $\pm$ 5.3 & 31.1 $\pm$ 4.7 & 64.2 $\pm$ 5.6 \\
\addlinespace[1.5pt]
\tm & Clean & -- & 5 & 87.4 $\pm$ 1.0 & -- & -- \\
 & Shift & L1 & 5 & 85.1 $\pm$ 1.1 & 2.3 $\pm$ 1.6 & 97.3 $\pm$ 1.7 \\
 & Shift & L2 & 5 & 75.5 $\pm$ 1.9 & 11.9 $\pm$ 2.6 & 86.4 $\pm$ 2.8 \\
 & Coarse & L1 & 5 & 83.7 $\pm$ 0.8 & 3.7 $\pm$ 1.0 & 95.8 $\pm$ 1.1 \\
 & Coarse & L2 & 5 & 68.2 $\pm$ 2.1 & 19.2 $\pm$ 2.6 & 78.0 $\pm$ 2.9 \\
 & Boundary & L1 & 5 & 84.6 $\pm$ 1.1 & 2.8 $\pm$ 2.0 & 96.8 $\pm$ 2.2 \\
 & Boundary & L2 & 5 & 74.7 $\pm$ 2.4 & 12.7 $\pm$ 3.0 & 85.4 $\pm$ 3.3 \\
 & Omission & L1 & 5 & 79.6 $\pm$ 1.6 & 7.8 $\pm$ 2.0 & 91.1 $\pm$ 2.2 \\
 & Omission & L2 & 5 & 68.6 $\pm$ 5.6 & 18.8 $\pm$ 5.6 & 78.5 $\pm$ 6.4 \\
\addlinespace[1.5pt]
\rn & Clean & -- & 5 & 88.2 $\pm$ 1.2 & -- & -- \\
 & Shift & L1 & 5 & 85.8 $\pm$ 1.0 & 2.4 $\pm$ 1.6 & 97.3 $\pm$ 1.8 \\
 & Shift & L2 & 5 & 76.7 $\pm$ 0.9 & 11.6 $\pm$ 1.9 & 86.9 $\pm$ 2.0 \\
 & Coarse & L1 & 5 & 84.6 $\pm$ 1.4 & 3.7 $\pm$ 1.9 & 95.9 $\pm$ 2.1 \\
 & Coarse & L2 & 5 & 70.9 $\pm$ 1.7 & 17.3 $\pm$ 1.5 & 80.4 $\pm$ 1.7 \\
 & Boundary & L1 & 5 & 83.2 $\pm$ 0.9 & 5.0 $\pm$ 1.0 & 94.3 $\pm$ 1.1 \\
 & Boundary & L2 & 5 & 74.7 $\pm$ 1.4 & 13.5 $\pm$ 1.7 & 84.7 $\pm$ 1.9 \\
 & Omission & L1 & 5 & 84.4 $\pm$ 1.7 & 3.8 $\pm$ 2.7 & 95.7 $\pm$ 3.0 \\
 & Omission & L2 & 5 & 74.3 $\pm$ 4.4 & 13.9 $\pm$ 3.8 & 84.2 $\pm$ 4.4 \\
\addlinespace[1.5pt]
\sat $\times$1/3 & Clean & -- & 5 & 83.6 $\pm$ 1.3 & -- & -- \\
 & Shift & L2 & 5 & 68.6 $\pm$ 3.0 & 15.0 $\pm$ 3.0 & 82.0 $\pm$ 3.5 \\
 & Coarse & L2 & 5 & 65.4 $\pm$ 1.5 & 18.1 $\pm$ 2.1 & 78.3 $\pm$ 2.3 \\
 & Boundary & L2 & 5 & 69.9 $\pm$ 2.1 & 13.6 $\pm$ 3.3 & 83.7 $\pm$ 3.7 \\
 & Omission & L2 & 5 & 61.6 $\pm$ 7.1 & 21.9 $\pm$ 7.4 & 73.8 $\pm$ 8.8 \\
\addlinespace[1.5pt]
\lvd $\times$1 & Clean & -- & 5 & 79.4 $\pm$ 1.4 & -- & -- \\
 & Shift & L2 & 5 & 64.9 $\pm$ 1.6 & 14.5 $\pm$ 1.7 & 81.8 $\pm$ 2.0 \\
 & Coarse & L2 & 5 & 60.8 $\pm$ 3.1 & 18.6 $\pm$ 2.4 & 76.5 $\pm$ 3.2 \\
 & Boundary & L2 & 5 & 64.1 $\pm$ 2.7 & 15.2 $\pm$ 3.2 & 80.8 $\pm$ 3.8 \\
 & Omission & L2 & 5 & 59.7 $\pm$ 2.6 & 19.6 $\pm$ 2.6 & 75.3 $\pm$ 3.2 \\
\bottomrule
\end{tabular}
\end{adjustbox}}{\tabletbd[6cm]{planned/generated/table\_n7\_grid\_burn.tex}}
\caption{\textbf{N7 grid on HLS Burn Scars:} Every N7 cell on \burn with the SegFormer head and its number of seeds $n$. Values are mean $\pm$ standard deviation over seeds of the test \miou, the paired drop in points and the retention. The first five blocks are the core models at their selected rates. The last two blocks are the matched-rate arm, \sat at $\times$1/3 and \lvd at $\times$1. \ProvSlot{\ProvNSeven}}
\label{tab:supp-n7-grid-burn}
\end{table}

\begin{table}[t]
\centering\footnotesize\setlength{\tabcolsep}{3pt}
\IfFileExists{planned/generated/table_n7_grid_flair.tex}{\begin{adjustbox}{max width=\columnwidth}%
\providecommand{\gcvdry}[1]{\textcolor{red}{#1}}
\providecommand{\ProvTableNSevenGridFlair}{Seeds per cell 5. Mean $\pm$ standard deviation over seeds, SegFormer head, \flair. Red cells are not run yet or still have a pending seed. Analysis of the finished runs.}
\begin{tabular}{@{}lllcccc@{}}
\toprule
Model & Type & Level & $n$ & mIoU & Drop & Retention (\%) \\
\midrule
\sat & Clean & -- & 5 & 50.5 $\pm$ 1.6 & -- & -- \\
 & Shift & L1 & 5 & 50.1 $\pm$ 1.1 & 0.4 $\pm$ 1.0 & 99.2 $\pm$ 2.0 \\
 & Shift & L2 & 5 & 47.1 $\pm$ 1.6 & 3.4 $\pm$ 1.3 & 93.3 $\pm$ 2.6 \\
 & Coarse & L1 & 5 & 49.7 $\pm$ 1.3 & 0.8 $\pm$ 1.0 & 98.4 $\pm$ 1.9 \\
 & Coarse & L2 & 5 & 44.9 $\pm$ 0.9 & 5.6 $\pm$ 0.9 & 88.9 $\pm$ 1.5 \\
 & Boundary & L1 & 5 & 49.3 $\pm$ 1.2 & 1.2 $\pm$ 0.8 & 97.6 $\pm$ 1.6 \\
 & Boundary & L2 & 5 & 44.8 $\pm$ 1.2 & 5.7 $\pm$ 0.7 & 88.6 $\pm$ 1.1 \\
 & Omission & L1 & 5 & 48.5 $\pm$ 1.0 & 2.1 $\pm$ 0.6 & 95.9 $\pm$ 1.0 \\
 & Omission & L2 & 5 & 44.9 $\pm$ 2.2 & 5.6 $\pm$ 2.0 & 88.8 $\pm$ 4.0 \\
\addlinespace[1.5pt]
\lvd & Clean & -- & 5 & 51.3 $\pm$ 0.9 & -- & -- \\
 & Shift & L1 & 5 & 50.4 $\pm$ 0.7 & 0.9 $\pm$ 1.1 & 98.3 $\pm$ 2.1 \\
 & Shift & L2 & 5 & 48.9 $\pm$ 0.6 & 2.4 $\pm$ 1.2 & 95.3 $\pm$ 2.2 \\
 & Coarse & L1 & 5 & 50.3 $\pm$ 0.6 & 0.9 $\pm$ 1.0 & 98.2 $\pm$ 1.9 \\
 & Coarse & L2 & 5 & 46.1 $\pm$ 0.4 & 5.2 $\pm$ 0.6 & 90.0 $\pm$ 1.1 \\
 & Boundary & L1 & 5 & 49.8 $\pm$ 0.4 & 1.5 $\pm$ 0.8 & 97.1 $\pm$ 1.5 \\
 & Boundary & L2 & 5 & 47.6 $\pm$ 0.8 & 3.7 $\pm$ 0.9 & 92.9 $\pm$ 1.7 \\
 & Omission & L1 & 5 & 50.0 $\pm$ 0.7 & 1.3 $\pm$ 0.6 & 97.4 $\pm$ 1.2 \\
 & Omission & L2 & 5 & 47.8 $\pm$ 0.5 & 3.5 $\pm$ 0.8 & 93.2 $\pm$ 1.4 \\
\addlinespace[1.5pt]
\dofa & Clean & -- & 5 & 47.6 $\pm$ 0.8 & -- & -- \\
 & Shift & L1 & 5 & 47.8 $\pm$ 0.8 & $-$0.2 $\pm$ 1.4 & 100.4 $\pm$ 2.8 \\
 & Shift & L2 & 5 & 44.3 $\pm$ 0.6 & 3.3 $\pm$ 1.2 & 93.1 $\pm$ 2.4 \\
 & Coarse & L1 & 5 & 47.2 $\pm$ 0.5 & 0.4 $\pm$ 1.1 & 99.1 $\pm$ 2.3 \\
 & Coarse & L2 & 5 & 41.3 $\pm$ 1.0 & 6.3 $\pm$ 1.0 & 86.8 $\pm$ 2.0 \\
 & Boundary & L1 & 5 & 46.4 $\pm$ 0.5 & 1.3 $\pm$ 0.8 & 97.4 $\pm$ 1.7 \\
 & Boundary & L2 & 5 & 42.4 $\pm$ 1.4 & 5.2 $\pm$ 1.9 & 89.0 $\pm$ 3.8 \\
 & Omission & L1 & 5 & 46.2 $\pm$ 0.9 & 1.4 $\pm$ 1.6 & 97.0 $\pm$ 3.2 \\
 & Omission & L2 & 5 & 44.2 $\pm$ 1.0 & 3.5 $\pm$ 1.2 & 92.7 $\pm$ 2.6 \\
\addlinespace[1.5pt]
\tm & Clean & -- & 5 & 41.9 $\pm$ 0.4 & -- & -- \\
 & Shift & L1 & 5 & 41.2 $\pm$ 0.5 & 0.7 $\pm$ 0.4 & 98.4 $\pm$ 1.0 \\
 & Shift & L2 & 5 & 37.0 $\pm$ 0.5 & 4.9 $\pm$ 0.8 & 88.3 $\pm$ 1.8 \\
 & Coarse & L1 & 5 & 40.1 $\pm$ 0.5 & 1.8 $\pm$ 0.8 & 95.6 $\pm$ 1.9 \\
 & Coarse & L2 & 5 & 36.6 $\pm$ 0.9 & 5.3 $\pm$ 1.2 & 87.3 $\pm$ 2.7 \\
 & Boundary & L1 & 5 & 39.6 $\pm$ 0.4 & 2.3 $\pm$ 0.4 & 94.5 $\pm$ 0.9 \\
 & Boundary & L2 & 5 & 34.3 $\pm$ 1.4 & 7.6 $\pm$ 1.4 & 81.8 $\pm$ 3.2 \\
 & Omission & L1 & 5 & 39.6 $\pm$ 0.8 & 2.4 $\pm$ 0.8 & 94.4 $\pm$ 1.9 \\
 & Omission & L2 & 5 & 36.1 $\pm$ 1.8 & 5.8 $\pm$ 1.8 & 86.1 $\pm$ 4.3 \\
\addlinespace[1.5pt]
\rn & Clean & -- & 5 & 44.6 $\pm$ 0.8 & -- & -- \\
 & Shift & L1 & 5 & 43.3 $\pm$ 1.9 & 1.3 $\pm$ 2.1 & 97.1 $\pm$ 4.8 \\
 & Shift & L2 & 5 & 40.9 $\pm$ 0.4 & 3.7 $\pm$ 0.7 & 91.8 $\pm$ 1.5 \\
 & Coarse & L1 & 5 & 43.3 $\pm$ 0.6 & 1.3 $\pm$ 0.6 & 97.0 $\pm$ 1.3 \\
 & Coarse & L2 & 5 & 39.6 $\pm$ 0.2 & 5.0 $\pm$ 0.8 & 88.8 $\pm$ 1.5 \\
 & Boundary & L1 & 5 & 42.9 $\pm$ 1.5 & 1.7 $\pm$ 1.8 & 96.2 $\pm$ 3.9 \\
 & Boundary & L2 & 5 & 41.3 $\pm$ 0.6 & 3.4 $\pm$ 0.4 & 92.5 $\pm$ 0.8 \\
 & Omission & L1 & 5 & 43.7 $\pm$ 0.7 & 0.9 $\pm$ 1.0 & 97.9 $\pm$ 2.1 \\
 & Omission & L2 & 5 & 40.6 $\pm$ 1.6 & 4.1 $\pm$ 2.3 & 91.0 $\pm$ 5.0 \\
\addlinespace[1.5pt]
\sat $\times$1/3 & Clean & -- & 5 & 48.0 $\pm$ 1.2 & -- & -- \\
 & Shift & L2 & 5 & 45.8 $\pm$ 0.5 & 2.2 $\pm$ 0.8 & 95.4 $\pm$ 1.5 \\
 & Coarse & L2 & 5 & 43.5 $\pm$ 1.0 & 4.5 $\pm$ 1.0 & 90.6 $\pm$ 1.9 \\
 & Boundary & L2 & 5 & 43.2 $\pm$ 1.5 & 4.9 $\pm$ 1.5 & 89.9 $\pm$ 3.1 \\
 & Omission & L2 & 5 & 43.0 $\pm$ 1.3 & 5.0 $\pm$ 0.7 & 89.6 $\pm$ 1.5 \\
\addlinespace[1.5pt]
\lvd $\times$1 & Clean & -- & 5 & 52.0 $\pm$ 0.6 & -- & -- \\
 & Shift & L2 & 5 & 49.1 $\pm$ 0.3 & 2.9 $\pm$ 0.5 & 94.5 $\pm$ 0.8 \\
 & Coarse & L2 & 5 & 46.1 $\pm$ 1.2 & 5.8 $\pm$ 1.6 & 88.8 $\pm$ 3.0 \\
 & Boundary & L2 & 5 & 47.8 $\pm$ 1.0 & 4.2 $\pm$ 1.0 & 92.0 $\pm$ 1.8 \\
 & Omission & L2 & 5 & 48.4 $\pm$ 1.0 & 3.6 $\pm$ 1.4 & 93.2 $\pm$ 2.7 \\
\bottomrule
\end{tabular}
\end{adjustbox}}{\tabletbd[6cm]{planned/generated/table\_n7\_grid\_flair.tex}}
\caption{\textbf{N7 grid on FLAIR-2:} Every N7 cell on \flair with the SegFormer head, in the layout of \Cref{tab:supp-n7-grid-burn}. \ProvSlot{\ProvNSeven}}
\label{tab:supp-n7-grid-flair}
\end{table}

\begin{table}[t]
\centering\footnotesize\setlength{\tabcolsep}{3pt}
\IfFileExists{planned/generated/table_n9_grid_burn.tex}{\begin{adjustbox}{max width=\columnwidth}%
\providecommand{\gcvdry}[1]{\textcolor{red}{#1}}
\providecommand{\ProvTableNNineGridBurn}{Seeds per cell 5. Mean $\pm$ standard deviation over seeds, Mask2Former head, \burn. Red cells are not run yet or still have a pending seed. Analysis of the finished runs.}
\begin{tabular}{@{}lllcccc@{}}
\toprule
Model & Type & Level & $n$ & mIoU & Drop & Retention (\%) \\
\midrule
\sat & Clean & -- & 5 & 60.0 $\pm$ 15.8 & -- & -- \\
 & Shift & L1 & 5 & 63.7 $\pm$ 11.7 & $-$3.7 $\pm$ 11.8 & 109.4 $\pm$ 24.4 \\
 & Shift & L2 & 5 & 59.4 $\pm$ 4.6 & 0.6 $\pm$ 18.9 & 105.6 $\pm$ 32.0 \\
 & Coarse & L1 & 5 & 58.1 $\pm$ 11.5 & 1.9 $\pm$ 13.5 & 100.4 $\pm$ 26.5 \\
 & Coarse & L2 & 5 & 53.2 $\pm$ 7.7 & 6.8 $\pm$ 17.8 & 93.8 $\pm$ 28.1 \\
 & Boundary & L1 & 5 & 58.8 $\pm$ 12.5 & 1.2 $\pm$ 22.6 & 105.0 $\pm$ 39.2 \\
 & Boundary & L2 & 5 & 51.9 $\pm$ 8.8 & 8.1 $\pm$ 18.6 & 91.8 $\pm$ 30.4 \\
 & Omission & L1 & 5 & 44.8 $\pm$ 0.0 & 15.2 $\pm$ 15.8 & 78.9 $\pm$ 20.6 \\
 & Omission & L2 & 5 & 44.8 $\pm$ 0.0 & 15.2 $\pm$ 15.8 & 78.9 $\pm$ 20.6 \\
\addlinespace[1.5pt]
\lvd & Clean & -- & 5 & 83.4 $\pm$ 0.6 & -- & -- \\
 & Shift & L1 & 5 & 80.2 $\pm$ 1.2 & 3.2 $\pm$ 1.0 & 96.2 $\pm$ 1.3 \\
 & Shift & L2 & 5 & 67.1 $\pm$ 3.6 & 16.3 $\pm$ 3.1 & 80.5 $\pm$ 3.9 \\
 & Coarse & L1 & 5 & 78.1 $\pm$ 1.1 & 5.2 $\pm$ 1.4 & 93.7 $\pm$ 1.7 \\
 & Coarse & L2 & 5 & 64.1 $\pm$ 1.9 & 19.3 $\pm$ 1.5 & 76.9 $\pm$ 1.9 \\
 & Boundary & L1 & 5 & 79.5 $\pm$ 1.7 & 3.9 $\pm$ 1.1 & 95.3 $\pm$ 1.3 \\
 & Boundary & L2 & 5 & 67.6 $\pm$ 1.8 & 15.8 $\pm$ 1.5 & 81.1 $\pm$ 1.9 \\
 & Omission & L1 & 5 & 72.2 $\pm$ 2.8 & 11.1 $\pm$ 3.0 & 86.7 $\pm$ 3.6 \\
 & Omission & L2 & 5 & 51.0 $\pm$ 9.0 & 32.4 $\pm$ 9.3 & 61.2 $\pm$ 11.1 \\
\addlinespace[1.5pt]
\dofa & Clean & -- & 5 & 86.4 $\pm$ 0.9 & -- & -- \\
 & Shift & L1 & 5 & 83.1 $\pm$ 0.4 & 3.3 $\pm$ 0.9 & 96.2 $\pm$ 1.0 \\
 & Shift & L2 & 5 & 73.3 $\pm$ 1.2 & 13.1 $\pm$ 1.7 & 84.9 $\pm$ 1.9 \\
 & Coarse & L1 & 5 & 82.0 $\pm$ 0.8 & 4.5 $\pm$ 1.5 & 94.8 $\pm$ 1.7 \\
 & Coarse & L2 & 5 & 65.9 $\pm$ 3.3 & 20.5 $\pm$ 3.3 & 76.3 $\pm$ 3.8 \\
 & Boundary & L1 & 5 & 81.9 $\pm$ 0.8 & 4.6 $\pm$ 1.1 & 94.7 $\pm$ 1.2 \\
 & Boundary & L2 & 5 & 69.8 $\pm$ 2.3 & 16.7 $\pm$ 2.3 & 80.7 $\pm$ 2.6 \\
 & Omission & L1 & 5 & 79.6 $\pm$ 2.0 & 6.9 $\pm$ 1.2 & 92.0 $\pm$ 1.4 \\
 & Omission & L2 & 5 & 59.2 $\pm$ 2.5 & 27.3 $\pm$ 3.2 & 68.5 $\pm$ 3.4 \\
\addlinespace[1.5pt]
\tm & Clean & -- & 5 & 88.2 $\pm$ 0.9 & -- & -- \\
 & Shift & L1 & 5 & 84.5 $\pm$ 1.0 & 3.7 $\pm$ 1.0 & 95.8 $\pm$ 1.2 \\
 & Shift & L2 & 5 & 76.8 $\pm$ 1.6 & 11.4 $\pm$ 2.2 & 87.1 $\pm$ 2.4 \\
 & Coarse & L1 & 5 & 84.6 $\pm$ 1.4 & 3.6 $\pm$ 1.5 & 95.9 $\pm$ 1.7 \\
 & Coarse & L2 & 5 & 67.0 $\pm$ 6.6 & 21.2 $\pm$ 6.0 & 76.0 $\pm$ 7.0 \\
 & Boundary & L1 & 5 & 84.6 $\pm$ 0.6 & 3.6 $\pm$ 1.2 & 96.0 $\pm$ 1.4 \\
 & Boundary & L2 & 5 & 76.4 $\pm$ 3.6 & 11.8 $\pm$ 4.1 & 86.6 $\pm$ 4.6 \\
 & Omission & L1 & 5 & 78.4 $\pm$ 8.3 & 9.8 $\pm$ 8.5 & 88.9 $\pm$ 9.5 \\
 & Omission & L2 & 5 & 65.2 $\pm$ 13.4 & 23.0 $\pm$ 12.7 & 73.8 $\pm$ 14.7 \\
\addlinespace[1.5pt]
\rn & Clean & -- & 5 & 63.7 $\pm$ 10.6 & -- & -- \\
 & Shift & L1 & 5 & 61.4 $\pm$ 9.3 & 2.2 $\pm$ 2.0 & 96.8 $\pm$ 2.9 \\
 & Shift & L2 & 5 & 59.6 $\pm$ 9.8 & 4.1 $\pm$ 6.6 & 94.1 $\pm$ 9.5 \\
 & Coarse & L1 & 5 & 62.6 $\pm$ 10.1 & 1.1 $\pm$ 1.2 & 98.4 $\pm$ 1.8 \\
 & Coarse & L2 & 5 & 63.6 $\pm$ 1.9 & 0.1 $\pm$ 9.5 & 102.5 $\pm$ 19.4 \\
 & Boundary & L1 & 5 & 62.5 $\pm$ 10.2 & 1.2 $\pm$ 2.1 & 98.3 $\pm$ 3.1 \\
 & Boundary & L2 & 5 & 55.8 $\pm$ 10.2 & 7.9 $\pm$ 9.3 & 88.5 $\pm$ 13.6 \\
 & Omission & L1 & 5 & 64.0 $\pm$ 13.6 & $-$0.3 $\pm$ 16.1 & 102.8 $\pm$ 26.5 \\
 & Omission & L2 & 5 & 49.4 $\pm$ 8.3 & 14.3 $\pm$ 18.9 & 82.2 $\pm$ 34.2 \\
\addlinespace[1.5pt]
\sat $\times$1/3 & Clean & -- & 5 & 81.3 $\pm$ 1.7 & -- & -- \\
 & Shift & L2 & 5 & 70.7 $\pm$ 1.6 & 10.6 $\pm$ 2.2 & 87.0 $\pm$ 2.5 \\
 & Coarse & L2 & 5 & 64.2 $\pm$ 1.5 & 17.2 $\pm$ 2.6 & 78.9 $\pm$ 2.8 \\
 & Boundary & L2 & 5 & 65.4 $\pm$ 1.8 & 15.9 $\pm$ 2.4 & 80.5 $\pm$ 2.6 \\
 & Omission & L2 & 5 & 50.5 $\pm$ 9.0 & 30.8 $\pm$ 8.5 & 62.0 $\pm$ 10.6 \\
\addlinespace[1.5pt]
\lvd $\times$1 & Clean & -- & 5 & 44.8 $\pm$ 0.0 & -- & -- \\
 & Shift & L2 & 5 & 44.8 $\pm$ 0.0 & 0.0 $\pm$ 0.0 & 100.0 $\pm$ 0.0 \\
 & Coarse & L2 & 5 & 44.8 $\pm$ 0.0 & 0.0 $\pm$ 0.0 & 100.0 $\pm$ 0.0 \\
 & Boundary & L2 & 5 & 44.8 $\pm$ 0.0 & 0.0 $\pm$ 0.0 & 100.0 $\pm$ 0.0 \\
 & Omission & L2 & 5 & 44.8 $\pm$ 0.0 & 0.0 $\pm$ 0.0 & 100.0 $\pm$ 0.0 \\
\bottomrule
\end{tabular}
\end{adjustbox}}{\tabletbd[6cm]{planned/generated/table\_n9\_grid\_burn.tex}}
\caption{\textbf{N9 grid on HLS Burn Scars:} Every N9 cell on \burn with the Mask2Former head, in the layout of \Cref{tab:supp-n7-grid-burn}. \ProvSlot{\ProvNNine}}
\label{tab:supp-n9-grid-burn}
\end{table}

\begin{table}[t]
\centering\footnotesize\setlength{\tabcolsep}{3pt}
\IfFileExists{planned/generated/table_n9_grid_flair.tex}{\begin{adjustbox}{max width=\columnwidth}%
\providecommand{\gcvdry}[1]{\textcolor{red}{#1}}
\providecommand{\ProvTableNNineGridFlair}{Seeds per cell 5. Mean $\pm$ standard deviation over seeds, Mask2Former head, \flair. Red cells are not run yet or still have a pending seed. Analysis of the finished runs.}
\begin{tabular}{@{}lllcccc@{}}
\toprule
Model & Type & Level & $n$ & mIoU & Drop & Retention (\%) \\
\midrule
\sat & Clean & -- & 5 & 50.2 $\pm$ 1.2 & -- & -- \\
 & Shift & L1 & 5 & 47.8 $\pm$ 1.0 & 2.3 $\pm$ 1.7 & 95.4 $\pm$ 3.2 \\
 & Shift & L2 & 5 & 45.8 $\pm$ 0.4 & 4.3 $\pm$ 1.4 & 91.4 $\pm$ 2.7 \\
 & Coarse & L1 & 5 & 49.1 $\pm$ 0.8 & 1.0 $\pm$ 0.7 & 98.0 $\pm$ 1.4 \\
 & Coarse & L2 & 5 & 43.5 $\pm$ 1.0 & 6.7 $\pm$ 0.6 & 86.7 $\pm$ 1.0 \\
 & Boundary & L1 & 5 & 48.0 $\pm$ 0.9 & 2.1 $\pm$ 0.6 & 95.8 $\pm$ 1.1 \\
 & Boundary & L2 & 5 & 43.1 $\pm$ 0.3 & 7.1 $\pm$ 1.0 & 85.9 $\pm$ 1.7 \\
 & Omission & L1 & 5 & 47.2 $\pm$ 1.0 & 3.0 $\pm$ 1.4 & 94.1 $\pm$ 2.6 \\
 & Omission & L2 & 5 & 42.8 $\pm$ 1.5 & 7.3 $\pm$ 1.9 & 85.5 $\pm$ 3.5 \\
\addlinespace[1.5pt]
\lvd & Clean & -- & 5 & 50.4 $\pm$ 0.5 & -- & -- \\
 & Shift & L1 & 5 & 49.7 $\pm$ 0.6 & 0.8 $\pm$ 0.5 & 98.5 $\pm$ 1.0 \\
 & Shift & L2 & 5 & 47.1 $\pm$ 0.7 & 3.4 $\pm$ 1.0 & 93.3 $\pm$ 1.9 \\
 & Coarse & L1 & 5 & 49.4 $\pm$ 0.5 & 1.1 $\pm$ 0.6 & 97.9 $\pm$ 1.2 \\
 & Coarse & L2 & 5 & 44.5 $\pm$ 0.7 & 6.0 $\pm$ 0.8 & 88.1 $\pm$ 1.5 \\
 & Boundary & L1 & 5 & 48.6 $\pm$ 0.9 & 1.9 $\pm$ 0.7 & 96.3 $\pm$ 1.4 \\
 & Boundary & L2 & 5 & 43.8 $\pm$ 1.3 & 6.6 $\pm$ 0.9 & 86.9 $\pm$ 1.8 \\
 & Omission & L1 & 5 & 48.7 $\pm$ 0.8 & 1.8 $\pm$ 0.8 & 96.5 $\pm$ 1.6 \\
 & Omission & L2 & 5 & 43.9 $\pm$ 1.2 & 6.6 $\pm$ 1.3 & 87.0 $\pm$ 2.5 \\
\addlinespace[1.5pt]
\dofa & Clean & -- & 5 & 48.2 $\pm$ 1.0 & -- & -- \\
 & Shift & L1 & 5 & 47.1 $\pm$ 0.5 & 1.1 $\pm$ 0.8 & 97.8 $\pm$ 1.6 \\
 & Shift & L2 & 5 & 43.7 $\pm$ 1.3 & 4.5 $\pm$ 1.8 & 90.7 $\pm$ 3.7 \\
 & Coarse & L1 & 5 & 47.1 $\pm$ 1.2 & 1.1 $\pm$ 1.1 & 97.7 $\pm$ 2.2 \\
 & Coarse & L2 & 5 & 41.5 $\pm$ 1.0 & 6.7 $\pm$ 1.4 & 86.0 $\pm$ 2.6 \\
 & Boundary & L1 & 5 & 46.2 $\pm$ 1.2 & 2.0 $\pm$ 1.3 & 95.8 $\pm$ 2.6 \\
 & Boundary & L2 & 5 & 42.2 $\pm$ 0.9 & 6.0 $\pm$ 1.6 & 87.6 $\pm$ 3.1 \\
 & Omission & L1 & 5 & 46.1 $\pm$ 1.2 & 2.1 $\pm$ 1.1 & 95.7 $\pm$ 2.4 \\
 & Omission & L2 & 5 & 42.5 $\pm$ 2.1 & 5.7 $\pm$ 2.4 & 88.3 $\pm$ 4.9 \\
\addlinespace[1.5pt]
\tm & Clean & -- & 5 & 41.4 $\pm$ 1.1 & -- & -- \\
 & Shift & L1 & 5 & 40.6 $\pm$ 1.6 & 0.8 $\pm$ 2.2 & 98.2 $\pm$ 5.1 \\
 & Shift & L2 & 5 & 36.8 $\pm$ 1.1 & 4.6 $\pm$ 1.8 & 88.9 $\pm$ 4.1 \\
 & Coarse & L1 & 5 & 40.6 $\pm$ 0.9 & 0.8 $\pm$ 0.5 & 98.1 $\pm$ 1.2 \\
 & Coarse & L2 & 5 & 35.6 $\pm$ 1.4 & 5.8 $\pm$ 1.5 & 86.0 $\pm$ 3.6 \\
 & Boundary & L1 & 5 & 38.9 $\pm$ 1.6 & 2.5 $\pm$ 1.3 & 93.9 $\pm$ 3.1 \\
 & Boundary & L2 & 5 & 35.7 $\pm$ 1.5 & 5.7 $\pm$ 1.3 & 86.3 $\pm$ 3.0 \\
 & Omission & L1 & 5 & 38.7 $\pm$ 1.4 & 2.7 $\pm$ 2.0 & 93.5 $\pm$ 4.7 \\
 & Omission & L2 & 5 & 35.0 $\pm$ 1.6 & 6.4 $\pm$ 1.4 & 84.5 $\pm$ 3.4 \\
\addlinespace[1.5pt]
\rn & Clean & -- & 5 & 44.0 $\pm$ 1.0 & -- & -- \\
 & Shift & L1 & 5 & 42.2 $\pm$ 1.1 & 1.8 $\pm$ 0.8 & 95.8 $\pm$ 1.8 \\
 & Shift & L2 & 5 & 40.2 $\pm$ 0.6 & 3.9 $\pm$ 1.5 & 91.2 $\pm$ 3.2 \\
 & Coarse & L1 & 5 & 42.6 $\pm$ 0.8 & 1.5 $\pm$ 1.0 & 96.7 $\pm$ 2.2 \\
 & Coarse & L2 & 5 & 38.2 $\pm$ 1.5 & 5.8 $\pm$ 2.5 & 86.9 $\pm$ 5.4 \\
 & Boundary & L1 & 5 & 41.8 $\pm$ 2.7 & 2.3 $\pm$ 2.1 & 94.8 $\pm$ 5.0 \\
 & Boundary & L2 & 5 & 39.1 $\pm$ 1.8 & 4.9 $\pm$ 2.0 & 88.9 $\pm$ 4.5 \\
 & Omission & L1 & 5 & 42.5 $\pm$ 0.7 & 1.6 $\pm$ 1.2 & 96.5 $\pm$ 2.7 \\
 & Omission & L2 & 5 & 40.0 $\pm$ 1.0 & 4.0 $\pm$ 1.2 & 90.9 $\pm$ 2.7 \\
\addlinespace[1.5pt]
\sat $\times$1/3 & Clean & -- & 5 & 46.2 $\pm$ 1.0 & -- & -- \\
 & Shift & L2 & 5 & 43.1 $\pm$ 0.9 & 3.1 $\pm$ 0.3 & 93.3 $\pm$ 0.6 \\
 & Coarse & L2 & 5 & 41.1 $\pm$ 0.8 & 5.1 $\pm$ 0.3 & 88.9 $\pm$ 0.5 \\
 & Boundary & L2 & 5 & 39.2 $\pm$ 1.1 & 7.1 $\pm$ 0.5 & 84.7 $\pm$ 1.2 \\
 & Omission & L2 & 5 & 40.0 $\pm$ 1.3 & 6.2 $\pm$ 1.5 & 86.5 $\pm$ 3.1 \\
\addlinespace[1.5pt]
\lvd $\times$1 & Clean & -- & 5 & 51.8 $\pm$ 0.6 & -- & -- \\
 & Shift & L2 & 5 & 47.6 $\pm$ 0.7 & 4.1 $\pm$ 0.6 & 92.0 $\pm$ 1.1 \\
 & Coarse & L2 & 5 & 45.2 $\pm$ 1.0 & 6.6 $\pm$ 1.2 & 87.3 $\pm$ 2.1 \\
 & Boundary & L2 & 5 & 46.0 $\pm$ 1.5 & 5.8 $\pm$ 1.3 & 88.9 $\pm$ 2.6 \\
 & Omission & L2 & 5 & 46.2 $\pm$ 0.7 & 5.6 $\pm$ 1.0 & 89.3 $\pm$ 1.8 \\
\bottomrule
\end{tabular}
\end{adjustbox}}{\tabletbd[6cm]{planned/generated/table\_n9\_grid\_flair.tex}}
\caption{\textbf{N9 grid on FLAIR-2:} Every N9 cell on \flair with the Mask2Former head, in the layout of \Cref{tab:supp-n7-grid-burn}. \ProvSlot{\ProvNNine}}
\label{tab:supp-n9-grid-flair}
\end{table}

\begin{table}[t]
\centering\footnotesize\setlength{\tabcolsep}{3pt}
\IfFileExists{planned/generated/table_n11_grid_burn.tex}{\begin{adjustbox}{max width=\columnwidth}%
\providecommand{\gcvdry}[1]{\textcolor{red}{#1}}
\providecommand{\ProvTableNElevenGridBurn}{Seeds per cell 5. Mean $\pm$ standard deviation over seeds, Full Mask2Former head, \burn. Red cells are not run yet or still have a pending seed. Analysis of the finished runs.}
\begin{tabular}{@{}lllcccc@{}}
\toprule
Model & Type & Level & $n$ & mIoU & Drop & Retention (\%) \\
\midrule
\sat & Clean & -- & 5 & 70.5 $\pm$ 4.1 & -- & -- \\
 & Shift & L1 & 5 & 67.3 $\pm$ 2.0 & 3.2 $\pm$ 3.2 & 95.6 $\pm$ 4.1 \\
 & Shift & L2 & 5 & 58.0 $\pm$ 2.7 & 12.5 $\pm$ 4.9 & 82.5 $\pm$ 6.0 \\
 & Coarse & L1 & 5 & 64.4 $\pm$ 5.0 & 6.1 $\pm$ 6.9 & 91.6 $\pm$ 9.4 \\
 & Coarse & L2 & 5 & 44.8 $\pm$ 0.0 & 25.8 $\pm$ 4.1 & 63.7 $\pm$ 3.6 \\
 & Boundary & L1 & 5 & 62.6 $\pm$ 3.8 & 8.0 $\pm$ 6.3 & 89.0 $\pm$ 8.3 \\
 & Boundary & L2 & 5 & 57.4 $\pm$ 4.0 & 13.2 $\pm$ 4.7 & 81.5 $\pm$ 6.1 \\
 & Omission & L1 & 5 & 67.2 $\pm$ 4.6 & 3.3 $\pm$ 8.0 & 95.8 $\pm$ 10.8 \\
 & Omission & L2 & 5 & 57.2 $\pm$ 1.8 & 13.3 $\pm$ 3.0 & 81.2 $\pm$ 3.4 \\
\addlinespace[1.5pt]
\lvd & Clean & -- & 5 & 79.0 $\pm$ 3.1 & -- & -- \\
 & Shift & L1 & 5 & 78.0 $\pm$ 1.4 & 1.1 $\pm$ 4.0 & 98.8 $\pm$ 5.1 \\
 & Shift & L2 & 5 & 66.2 $\pm$ 2.4 & 12.8 $\pm$ 2.5 & 83.8 $\pm$ 2.8 \\
 & Coarse & L1 & 5 & 76.8 $\pm$ 2.6 & 2.2 $\pm$ 3.2 & 97.3 $\pm$ 4.2 \\
 & Coarse & L2 & 5 & 44.8 $\pm$ 0.0 & 34.3 $\pm$ 3.1 & 56.7 $\pm$ 2.3 \\
 & Boundary & L1 & 5 & 77.8 $\pm$ 1.2 & 1.2 $\pm$ 3.0 & 98.6 $\pm$ 3.8 \\
 & Boundary & L2 & 5 & 53.0 $\pm$ 11.3 & 26.0 $\pm$ 12.6 & 67.3 $\pm$ 15.6 \\
 & Omission & L1 & 5 & 74.1 $\pm$ 4.7 & 4.9 $\pm$ 6.6 & 94.0 $\pm$ 8.1 \\
 & Omission & L2 & 5 & 64.9 $\pm$ 3.6 & 14.2 $\pm$ 4.5 & 82.1 $\pm$ 5.2 \\
\addlinespace[1.5pt]
\dofa & Clean & -- & 5 & 86.8 $\pm$ 1.0 & -- & -- \\
 & Shift & L1 & 5 & 84.3 $\pm$ 1.1 & 2.5 $\pm$ 1.7 & 97.1 $\pm$ 1.9 \\
 & Shift & L2 & 5 & 73.3 $\pm$ 1.8 & 13.5 $\pm$ 2.3 & 84.5 $\pm$ 2.5 \\
 & Coarse & L1 & 5 & 82.9 $\pm$ 1.2 & 3.9 $\pm$ 1.4 & 95.6 $\pm$ 1.6 \\
 & Coarse & L2 & 5 & 60.6 $\pm$ 5.1 & 26.2 $\pm$ 5.8 & 69.9 $\pm$ 6.5 \\
 & Boundary & L1 & 5 & 84.2 $\pm$ 1.5 & 2.5 $\pm$ 1.6 & 97.1 $\pm$ 1.9 \\
 & Boundary & L2 & 5 & 71.5 $\pm$ 1.5 & 15.3 $\pm$ 1.8 & 82.4 $\pm$ 2.0 \\
 & Omission & L1 & 5 & 84.1 $\pm$ 1.7 & 2.7 $\pm$ 2.1 & 96.9 $\pm$ 2.4 \\
 & Omission & L2 & 5 & 66.6 $\pm$ 7.1 & 20.2 $\pm$ 7.9 & 76.8 $\pm$ 8.9 \\
\addlinespace[1.5pt]
\tm & Clean & -- & 5 & 87.4 $\pm$ 1.5 & -- & -- \\
 & Shift & L1 & 5 & 85.2 $\pm$ 0.4 & 2.2 $\pm$ 1.7 & 97.5 $\pm$ 1.9 \\
 & Shift & L2 & 5 & 76.6 $\pm$ 0.6 & 10.8 $\pm$ 1.3 & 87.7 $\pm$ 1.3 \\
 & Coarse & L1 & 5 & 84.8 $\pm$ 0.7 & 2.5 $\pm$ 2.1 & 97.1 $\pm$ 2.3 \\
 & Coarse & L2 & 5 & 69.8 $\pm$ 2.4 & 17.6 $\pm$ 2.1 & 79.9 $\pm$ 2.4 \\
 & Boundary & L1 & 5 & 85.2 $\pm$ 0.4 & 2.2 $\pm$ 1.7 & 97.5 $\pm$ 1.9 \\
 & Boundary & L2 & 5 & 73.5 $\pm$ 1.9 & 13.9 $\pm$ 3.0 & 84.2 $\pm$ 3.2 \\
 & Omission & L1 & 5 & 83.0 $\pm$ 2.6 & 4.3 $\pm$ 2.6 & 95.0 $\pm$ 2.9 \\
 & Omission & L2 & 5 & 74.6 $\pm$ 2.4 & 12.8 $\pm$ 3.9 & 85.4 $\pm$ 4.2 \\
\addlinespace[1.5pt]
\rn & Clean & -- & 5 & 88.0 $\pm$ 0.4 & -- & -- \\
 & Shift & L1 & 5 & 85.3 $\pm$ 0.9 & 2.7 $\pm$ 0.9 & 96.9 $\pm$ 1.0 \\
 & Shift & L2 & 5 & 76.4 $\pm$ 1.5 & 11.5 $\pm$ 1.6 & 86.9 $\pm$ 1.8 \\
 & Coarse & L1 & 5 & 84.8 $\pm$ 1.0 & 3.2 $\pm$ 0.8 & 96.4 $\pm$ 0.9 \\
 & Coarse & L2 & 5 & 70.1 $\pm$ 3.3 & 17.9 $\pm$ 3.4 & 79.7 $\pm$ 3.8 \\
 & Boundary & L1 & 5 & 85.6 $\pm$ 0.4 & 2.3 $\pm$ 0.7 & 97.3 $\pm$ 0.7 \\
 & Boundary & L2 & 5 & 74.2 $\pm$ 1.2 & 13.7 $\pm$ 1.1 & 84.4 $\pm$ 1.2 \\
 & Omission & L1 & 5 & 87.0 $\pm$ 0.9 & 1.0 $\pm$ 0.9 & 98.9 $\pm$ 1.0 \\
 & Omission & L2 & 5 & 82.7 $\pm$ 2.3 & 5.2 $\pm$ 2.3 & 94.0 $\pm$ 2.6 \\
\addlinespace[1.5pt]
\sat $\times$1/3 & Clean & -- & 5 & 77.2 $\pm$ 2.5 & -- & -- \\
 & Shift & L2 & 5 & 64.3 $\pm$ 2.5 & 12.9 $\pm$ 3.4 & 83.4 $\pm$ 4.1 \\
 & Coarse & L2 & 5 & 51.9 $\pm$ 9.7 & 25.3 $\pm$ 9.5 & 67.2 $\pm$ 12.2 \\
 & Boundary & L2 & 5 & 62.9 $\pm$ 3.0 & 14.2 $\pm$ 3.6 & 81.6 $\pm$ 4.3 \\
 & Omission & L2 & 5 & 63.6 $\pm$ 3.0 & 13.5 $\pm$ 4.1 & 82.6 $\pm$ 5.0 \\
\addlinespace[1.5pt]
\lvd $\times$1 & Clean & -- & 5 & 68.7 $\pm$ 2.3 & -- & -- \\
 & Shift & L2 & 5 & 47.0 $\pm$ 5.0 & 21.7 $\pm$ 6.6 & 68.6 $\pm$ 9.0 \\
 & Coarse & L2 & 5 & 44.8 $\pm$ 0.0 & 24.0 $\pm$ 2.3 & 65.2 $\pm$ 2.2 \\
 & Boundary & L2 & 5 & 44.8 $\pm$ 0.0 & 24.0 $\pm$ 2.3 & 65.2 $\pm$ 2.2 \\
 & Omission & L2 & 5 & 57.9 $\pm$ 3.6 & 10.8 $\pm$ 3.4 & 84.3 $\pm$ 4.7 \\
\bottomrule
\end{tabular}
\end{adjustbox}}{\tabletbd[6cm]{planned/generated/table\_n11\_grid\_burn.tex}}
\caption{\textbf{N11 grid on HLS Burn Scars:} Every N11 cell on \burn with the full Mask2Former head, in the layout of \Cref{tab:supp-n7-grid-burn}. \ProvSlot{\ProvNEleven}}
\label{tab:supp-n11-grid-burn}
\end{table}

\begin{table}[t]
\centering\footnotesize\setlength{\tabcolsep}{3pt}
\IfFileExists{planned/generated/table_n11_grid_flair.tex}{\begin{adjustbox}{max width=\columnwidth}%
\providecommand{\gcvdry}[1]{\textcolor{red}{#1}}
\providecommand{\ProvTableNElevenGridFlair}{Seeds per cell 5. Mean $\pm$ standard deviation over seeds, Full Mask2Former head, \flair. Red cells are not run yet or still have a pending seed. Analysis of the finished runs.}
\begin{tabular}{@{}lllcccc@{}}
\toprule
Model & Type & Level & $n$ & mIoU & Drop & Retention (\%) \\
\midrule
\sat & Clean & -- & 5 & 46.5 $\pm$ 0.8 & -- & -- \\
 & Shift & L1 & 5 & 45.7 $\pm$ 1.5 & 0.7 $\pm$ 1.2 & 98.4 $\pm$ 2.5 \\
 & Shift & L2 & 5 & 42.5 $\pm$ 0.9 & 3.9 $\pm$ 1.1 & 91.6 $\pm$ 2.3 \\
 & Coarse & L1 & 5 & 45.9 $\pm$ 0.9 & 0.6 $\pm$ 0.4 & 98.8 $\pm$ 0.8 \\
 & Coarse & L2 & 5 & 39.1 $\pm$ 1.3 & 7.4 $\pm$ 1.4 & 84.1 $\pm$ 3.0 \\
 & Boundary & L1 & 5 & 44.5 $\pm$ 1.0 & 1.9 $\pm$ 1.3 & 95.9 $\pm$ 2.7 \\
 & Boundary & L2 & 5 & 41.9 $\pm$ 1.5 & 4.5 $\pm$ 1.4 & 90.3 $\pm$ 3.0 \\
 & Omission & L1 & 5 & 45.0 $\pm$ 1.7 & 1.4 $\pm$ 0.9 & 96.9 $\pm$ 2.1 \\
 & Omission & L2 & 5 & 42.9 $\pm$ 2.0 & 3.6 $\pm$ 1.4 & 92.2 $\pm$ 3.1 \\
\addlinespace[1.5pt]
\lvd & Clean & -- & 5 & 49.2 $\pm$ 0.6 & -- & -- \\
 & Shift & L1 & 5 & 48.5 $\pm$ 1.2 & 0.6 $\pm$ 0.7 & 98.7 $\pm$ 1.4 \\
 & Shift & L2 & 5 & 46.1 $\pm$ 0.4 & 3.0 $\pm$ 0.5 & 93.8 $\pm$ 1.1 \\
 & Coarse & L1 & 5 & 48.3 $\pm$ 0.8 & 0.8 $\pm$ 0.5 & 98.3 $\pm$ 1.0 \\
 & Coarse & L2 & 5 & 42.2 $\pm$ 0.9 & 7.0 $\pm$ 0.8 & 85.8 $\pm$ 1.6 \\
 & Boundary & L1 & 5 & 48.2 $\pm$ 0.8 & 1.0 $\pm$ 0.5 & 98.0 $\pm$ 1.0 \\
 & Boundary & L2 & 5 & 44.8 $\pm$ 1.6 & 4.3 $\pm$ 1.4 & 91.2 $\pm$ 2.9 \\
 & Omission & L1 & 5 & 48.0 $\pm$ 1.2 & 1.2 $\pm$ 0.7 & 97.5 $\pm$ 1.5 \\
 & Omission & L2 & 5 & 45.9 $\pm$ 1.1 & 3.3 $\pm$ 0.8 & 93.2 $\pm$ 1.7 \\
\addlinespace[1.5pt]
\dofa & Clean & -- & 5 & 48.0 $\pm$ 0.4 & -- & -- \\
 & Shift & L1 & 5 & 48.0 $\pm$ 0.5 & 0.0 $\pm$ 0.8 & 100.0 $\pm$ 1.6 \\
 & Shift & L2 & 5 & 43.7 $\pm$ 1.0 & 4.3 $\pm$ 1.0 & 91.0 $\pm$ 2.0 \\
 & Coarse & L1 & 5 & 47.3 $\pm$ 0.4 & 0.8 $\pm$ 0.5 & 98.4 $\pm$ 1.0 \\
 & Coarse & L2 & 5 & 41.2 $\pm$ 0.4 & 6.8 $\pm$ 0.7 & 85.8 $\pm$ 1.3 \\
 & Boundary & L1 & 5 & 47.0 $\pm$ 0.7 & 1.0 $\pm$ 0.5 & 97.9 $\pm$ 1.0 \\
 & Boundary & L2 & 5 & 44.7 $\pm$ 0.6 & 3.3 $\pm$ 0.8 & 93.1 $\pm$ 1.7 \\
 & Omission & L1 & 5 & 47.7 $\pm$ 1.1 & 0.3 $\pm$ 1.3 & 99.3 $\pm$ 2.7 \\
 & Omission & L2 & 5 & 44.8 $\pm$ 1.8 & 3.3 $\pm$ 1.7 & 93.2 $\pm$ 3.6 \\
\addlinespace[1.5pt]
\tm & Clean & -- & 5 & 42.8 $\pm$ 0.8 & -- & -- \\
 & Shift & L1 & 5 & 42.7 $\pm$ 0.8 & 0.1 $\pm$ 1.3 & 99.8 $\pm$ 3.0 \\
 & Shift & L2 & 5 & 37.7 $\pm$ 1.1 & 5.1 $\pm$ 1.0 & 88.0 $\pm$ 2.3 \\
 & Coarse & L1 & 5 & 42.5 $\pm$ 0.7 & 0.3 $\pm$ 1.0 & 99.3 $\pm$ 2.4 \\
 & Coarse & L2 & 5 & 34.9 $\pm$ 0.7 & 7.9 $\pm$ 1.2 & 81.5 $\pm$ 2.5 \\
 & Boundary & L1 & 5 & 41.6 $\pm$ 0.8 & 1.2 $\pm$ 0.8 & 97.1 $\pm$ 1.9 \\
 & Boundary & L2 & 5 & 38.3 $\pm$ 1.4 & 4.5 $\pm$ 1.3 & 89.5 $\pm$ 3.1 \\
 & Omission & L1 & 5 & 41.5 $\pm$ 1.2 & 1.3 $\pm$ 1.2 & 97.0 $\pm$ 2.7 \\
 & Omission & L2 & 5 & 38.3 $\pm$ 0.8 & 4.5 $\pm$ 1.1 & 89.4 $\pm$ 2.5 \\
\addlinespace[1.5pt]
\rn & Clean & -- & 5 & 45.8 $\pm$ 1.0 & -- & -- \\
 & Shift & L1 & 5 & 45.2 $\pm$ 0.4 & 0.6 $\pm$ 0.6 & 98.7 $\pm$ 1.3 \\
 & Shift & L2 & 5 & 40.3 $\pm$ 0.7 & 5.5 $\pm$ 1.5 & 88.1 $\pm$ 3.0 \\
 & Coarse & L1 & 5 & 45.2 $\pm$ 0.9 & 0.5 $\pm$ 1.7 & 98.9 $\pm$ 3.8 \\
 & Coarse & L2 & 5 & 37.6 $\pm$ 0.8 & 8.2 $\pm$ 1.6 & 82.2 $\pm$ 3.1 \\
 & Boundary & L1 & 5 & 43.6 $\pm$ 0.5 & 2.2 $\pm$ 1.4 & 95.3 $\pm$ 3.0 \\
 & Boundary & L2 & 5 & 41.3 $\pm$ 1.1 & 4.5 $\pm$ 1.6 & 90.3 $\pm$ 3.2 \\
 & Omission & L1 & 5 & 44.8 $\pm$ 1.5 & 0.9 $\pm$ 1.4 & 98.0 $\pm$ 3.0 \\
 & Omission & L2 & 5 & 41.3 $\pm$ 3.2 & 4.5 $\pm$ 3.0 & 90.2 $\pm$ 6.5 \\
\addlinespace[1.5pt]
\sat $\times$1/3 & Clean & -- & 5 & 44.0 $\pm$ 1.4 & -- & -- \\
 & Shift & L2 & 5 & 40.0 $\pm$ 1.2 & 4.0 $\pm$ 1.0 & 90.9 $\pm$ 2.0 \\
 & Coarse & L2 & 5 & 36.0 $\pm$ 2.2 & 8.0 $\pm$ 1.4 & 81.7 $\pm$ 3.4 \\
 & Boundary & L2 & 5 & 40.0 $\pm$ 1.7 & 4.0 $\pm$ 1.2 & 90.9 $\pm$ 2.7 \\
 & Omission & L2 & 5 & 39.7 $\pm$ 1.5 & 4.4 $\pm$ 1.1 & 90.1 $\pm$ 2.4 \\
\addlinespace[1.5pt]
\lvd $\times$1 & Clean & -- & 5 & 50.7 $\pm$ 0.4 & -- & -- \\
 & Shift & L2 & 5 & 46.8 $\pm$ 0.8 & 3.9 $\pm$ 1.1 & 92.3 $\pm$ 2.0 \\
 & Coarse & L2 & 5 & 43.5 $\pm$ 1.2 & 7.2 $\pm$ 1.2 & 85.8 $\pm$ 2.4 \\
 & Boundary & L2 & 5 & 48.0 $\pm$ 1.2 & 2.7 $\pm$ 1.2 & 94.7 $\pm$ 2.4 \\
 & Omission & L2 & 5 & 48.6 $\pm$ 0.9 & 2.1 $\pm$ 0.9 & 95.8 $\pm$ 1.7 \\
\bottomrule
\end{tabular}
\end{adjustbox}}{\tabletbd[6cm]{planned/generated/table\_n11\_grid\_flair.tex}}
\caption{\textbf{N11 grid on FLAIR-2:} Every N11 cell on \flair with the full Mask2Former head, in the layout of \Cref{tab:supp-n7-grid-burn}. \ProvSlot{\ProvNEleven}}
\label{tab:supp-n11-grid-flair}
\end{table}

\begin{table}[t]
\centering\footnotesize\setlength{\tabcolsep}{3pt}
\IfFileExists{planned/generated/table_n10_grid_burn.tex}{\begin{adjustbox}{max width=\columnwidth}%
\providecommand{\gcvdry}[1]{\textcolor{red}{#1}}
\providecommand{\ProvTableNTenGridBurn}{Seeds per cell 5. Mean $\pm$ standard deviation over seeds, SAM head, \burn. Red cells are not run yet or still have a pending seed. Analysis of the finished runs.}
\begin{tabular}{@{}lllcccc@{}}
\toprule
Model & Type & Level & $n$ & mIoU & Drop & Retention (\%) \\
\midrule
\sat & Clean & -- & 5 & 73.7 $\pm$ 1.8 & -- & -- \\
 & Shift & L1 & 5 & 70.7 $\pm$ 2.1 & 3.1 $\pm$ 3.4 & 95.9 $\pm$ 4.6 \\
 & Shift & L2 & 5 & 57.7 $\pm$ 8.1 & 16.0 $\pm$ 6.4 & 78.1 $\pm$ 9.3 \\
 & Coarse & L1 & 5 & 67.8 $\pm$ 2.1 & 5.9 $\pm$ 2.9 & 92.1 $\pm$ 3.8 \\
 & Coarse & L2 & 5 & 44.8 $\pm$ 0.0 & 29.0 $\pm$ 1.8 & 60.8 $\pm$ 1.5 \\
 & Boundary & L1 & 5 & 66.2 $\pm$ 5.9 & 7.6 $\pm$ 6.8 & 89.9 $\pm$ 8.9 \\
 & Boundary & L2 & 5 & 63.2 $\pm$ 2.0 & 10.6 $\pm$ 2.8 & 85.7 $\pm$ 3.6 \\
 & Omission & L1 & 5 & 64.1 $\pm$ 1.7 & 9.6 $\pm$ 2.3 & 87.0 $\pm$ 2.9 \\
 & Omission & L2 & 5 & 48.0 $\pm$ 5.9 & 25.7 $\pm$ 7.6 & 65.3 $\pm$ 9.8 \\
\addlinespace[1.5pt]
\lvd & Clean & -- & 5 & 82.5 $\pm$ 1.6 & -- & -- \\
 & Shift & L1 & 5 & 78.6 $\pm$ 1.6 & 3.9 $\pm$ 2.3 & 95.3 $\pm$ 2.7 \\
 & Shift & L2 & 5 & 68.7 $\pm$ 2.3 & 13.8 $\pm$ 3.7 & 83.3 $\pm$ 4.1 \\
 & Coarse & L1 & 5 & 77.9 $\pm$ 1.0 & 4.6 $\pm$ 0.9 & 94.4 $\pm$ 1.0 \\
 & Coarse & L2 & 5 & 54.8 $\pm$ 9.3 & 27.8 $\pm$ 9.3 & 66.4 $\pm$ 11.2 \\
 & Boundary & L1 & 5 & 78.7 $\pm$ 2.1 & 3.9 $\pm$ 2.9 & 95.3 $\pm$ 3.5 \\
 & Boundary & L2 & 5 & 68.6 $\pm$ 1.9 & 14.0 $\pm$ 1.5 & 83.1 $\pm$ 1.8 \\
 & Omission & L1 & 5 & 74.1 $\pm$ 3.2 & 8.4 $\pm$ 2.8 & 89.8 $\pm$ 3.5 \\
 & Omission & L2 & 5 & 61.4 $\pm$ 6.3 & 21.1 $\pm$ 5.6 & 74.3 $\pm$ 6.9 \\
\addlinespace[1.5pt]
\dofa & Clean & -- & 5 & 87.1 $\pm$ 1.6 & -- & -- \\
 & Shift & L1 & 5 & 84.0 $\pm$ 0.5 & 3.1 $\pm$ 2.0 & 96.5 $\pm$ 2.2 \\
 & Shift & L2 & 5 & 72.9 $\pm$ 1.2 & 14.2 $\pm$ 1.4 & 83.7 $\pm$ 1.5 \\
 & Coarse & L1 & 5 & 82.3 $\pm$ 0.8 & 4.8 $\pm$ 2.4 & 94.5 $\pm$ 2.7 \\
 & Coarse & L2 & 5 & 67.1 $\pm$ 2.6 & 20.0 $\pm$ 3.5 & 77.1 $\pm$ 3.8 \\
 & Boundary & L1 & 5 & 82.8 $\pm$ 1.0 & 4.4 $\pm$ 2.3 & 95.0 $\pm$ 2.6 \\
 & Boundary & L2 & 5 & 74.3 $\pm$ 2.0 & 12.8 $\pm$ 2.0 & 85.3 $\pm$ 2.2 \\
 & Omission & L1 & 5 & 81.8 $\pm$ 2.8 & 5.3 $\pm$ 3.1 & 93.9 $\pm$ 3.5 \\
 & Omission & L2 & 5 & 66.2 $\pm$ 2.2 & 20.9 $\pm$ 1.3 & 76.0 $\pm$ 1.6 \\
\addlinespace[1.5pt]
\tm & Clean & -- & 5 & 88.7 $\pm$ 0.8 & -- & -- \\
 & Shift & L1 & 5 & 84.3 $\pm$ 0.9 & 4.4 $\pm$ 0.4 & 95.1 $\pm$ 0.5 \\
 & Shift & L2 & 5 & 75.5 $\pm$ 1.8 & 13.1 $\pm$ 1.7 & 85.2 $\pm$ 1.9 \\
 & Coarse & L1 & 5 & 84.2 $\pm$ 0.6 & 4.5 $\pm$ 0.7 & 95.0 $\pm$ 0.8 \\
 & Coarse & L2 & 5 & 69.2 $\pm$ 1.5 & 19.4 $\pm$ 2.0 & 78.1 $\pm$ 2.1 \\
 & Boundary & L1 & 5 & 84.6 $\pm$ 1.1 & 4.1 $\pm$ 1.7 & 95.4 $\pm$ 1.8 \\
 & Boundary & L2 & 5 & 77.5 $\pm$ 1.2 & 11.1 $\pm$ 1.7 & 87.5 $\pm$ 1.8 \\
 & Omission & L1 & 5 & 81.7 $\pm$ 3.6 & 7.0 $\pm$ 3.3 & 92.1 $\pm$ 3.7 \\
 & Omission & L2 & 5 & 67.0 $\pm$ 10.5 & 21.7 $\pm$ 10.2 & 75.5 $\pm$ 11.6 \\
\addlinespace[1.5pt]
\rn & Clean & -- & 5 & 87.6 $\pm$ 0.5 & -- & -- \\
 & Shift & L1 & 5 & 86.6 $\pm$ 0.4 & 1.0 $\pm$ 0.9 & 98.9 $\pm$ 1.0 \\
 & Shift & L2 & 5 & 77.4 $\pm$ 1.1 & 10.3 $\pm$ 1.3 & 88.3 $\pm$ 1.5 \\
 & Coarse & L1 & 5 & 84.9 $\pm$ 1.2 & 2.7 $\pm$ 1.0 & 96.9 $\pm$ 1.1 \\
 & Coarse & L2 & 5 & 71.2 $\pm$ 2.2 & 16.5 $\pm$ 2.3 & 81.2 $\pm$ 2.6 \\
 & Boundary & L1 & 5 & 86.2 $\pm$ 0.8 & 1.4 $\pm$ 0.7 & 98.4 $\pm$ 0.8 \\
 & Boundary & L2 & 5 & 78.0 $\pm$ 1.0 & 9.7 $\pm$ 0.8 & 89.0 $\pm$ 0.9 \\
 & Omission & L1 & 5 & 85.8 $\pm$ 2.3 & 1.8 $\pm$ 2.5 & 97.9 $\pm$ 2.9 \\
 & Omission & L2 & 5 & 74.5 $\pm$ 3.8 & 13.1 $\pm$ 4.2 & 85.0 $\pm$ 4.7 \\
\addlinespace[1.5pt]
\sat $\times$1/3 & Clean & -- & 5 & 74.4 $\pm$ 3.2 & -- & -- \\
 & Shift & L2 & 5 & 64.5 $\pm$ 2.0 & 9.9 $\pm$ 4.1 & 86.9 $\pm$ 5.2 \\
 & Coarse & L2 & 5 & 58.1 $\pm$ 7.6 & 16.2 $\pm$ 8.6 & 78.3 $\pm$ 11.1 \\
 & Boundary & L2 & 5 & 67.6 $\pm$ 1.2 & 6.7 $\pm$ 2.7 & 91.1 $\pm$ 3.3 \\
 & Omission & L2 & 5 & 56.5 $\pm$ 8.1 & 17.9 $\pm$ 8.5 & 76.0 $\pm$ 11.1 \\
\addlinespace[1.5pt]
\lvd $\times$1 & Clean & -- & 5 & 74.6 $\pm$ 4.9 & -- & -- \\
 & Shift & L2 & 5 & 54.4 $\pm$ 9.5 & 20.1 $\pm$ 7.6 & 72.9 $\pm$ 10.5 \\
 & Coarse & L2 & 5 & 44.8 $\pm$ 0.0 & 29.8 $\pm$ 4.9 & 60.3 $\pm$ 4.2 \\
 & Boundary & L2 & 5 & 60.7 $\pm$ 2.2 & 13.9 $\pm$ 4.9 & 81.7 $\pm$ 5.9 \\
 & Omission & L2 & 5 & 45.0 $\pm$ 0.4 & 29.6 $\pm$ 4.8 & 60.5 $\pm$ 4.1 \\
\bottomrule
\end{tabular}
\end{adjustbox}}{\tabletbd[6cm]{planned/generated/table\_n10\_grid\_burn.tex}}
\caption{\textbf{N10 grid on HLS Burn Scars:} Every N10 cell on \burn with the SAM head, in the layout of \Cref{tab:supp-n7-grid-burn}. \ProvSlot{\ProvNTen}}
\label{tab:supp-n10-grid-burn}
\end{table}

\begin{table}[t]
\centering\footnotesize\setlength{\tabcolsep}{3pt}
\IfFileExists{planned/generated/table_n10_grid_flair.tex}{\begin{adjustbox}{max width=\columnwidth}%
\providecommand{\gcvdry}[1]{\textcolor{red}{#1}}
\providecommand{\ProvTableNTenGridFlair}{Seeds per cell 5. Mean $\pm$ standard deviation over seeds, SAM head, \flair. Red cells are not run yet or still have a pending seed. Analysis of the finished runs.}
\begin{tabular}{@{}lllcccc@{}}
\toprule
Model & Type & Level & $n$ & mIoU & Drop & Retention (\%) \\
\midrule
\sat & Clean & -- & 5 & 46.3 $\pm$ 1.5 & -- & -- \\
 & Shift & L1 & 5 & 44.9 $\pm$ 0.9 & 1.5 $\pm$ 2.0 & 96.9 $\pm$ 4.2 \\
 & Shift & L2 & 5 & 41.8 $\pm$ 1.5 & 4.5 $\pm$ 1.7 & 90.4 $\pm$ 3.6 \\
 & Coarse & L1 & 5 & 45.2 $\pm$ 0.9 & 1.1 $\pm$ 1.5 & 97.7 $\pm$ 3.2 \\
 & Coarse & L2 & 5 & 40.0 $\pm$ 1.5 & 6.3 $\pm$ 1.1 & 86.5 $\pm$ 2.2 \\
 & Boundary & L1 & 5 & 44.3 $\pm$ 1.0 & 2.0 $\pm$ 0.9 & 95.8 $\pm$ 2.0 \\
 & Boundary & L2 & 5 & 40.4 $\pm$ 1.0 & 6.0 $\pm$ 1.1 & 87.2 $\pm$ 2.1 \\
 & Omission & L1 & 5 & 44.2 $\pm$ 1.3 & 2.1 $\pm$ 1.1 & 95.4 $\pm$ 2.2 \\
 & Omission & L2 & 5 & 40.9 $\pm$ 1.3 & 5.5 $\pm$ 2.6 & 88.4 $\pm$ 5.3 \\
\addlinespace[1.5pt]
\lvd & Clean & -- & 5 & 46.2 $\pm$ 1.1 & -- & -- \\
 & Shift & L1 & 5 & 45.5 $\pm$ 1.1 & 0.6 $\pm$ 0.8 & 98.6 $\pm$ 1.7 \\
 & Shift & L2 & 5 & 43.1 $\pm$ 0.5 & 3.0 $\pm$ 0.7 & 93.5 $\pm$ 1.3 \\
 & Coarse & L1 & 5 & 45.3 $\pm$ 1.3 & 0.9 $\pm$ 0.4 & 98.0 $\pm$ 0.8 \\
 & Coarse & L2 & 5 & 41.2 $\pm$ 0.6 & 4.9 $\pm$ 1.0 & 89.3 $\pm$ 1.8 \\
 & Boundary & L1 & 5 & 44.4 $\pm$ 0.5 & 1.8 $\pm$ 1.4 & 96.2 $\pm$ 2.9 \\
 & Boundary & L2 & 5 & 41.2 $\pm$ 1.2 & 5.0 $\pm$ 1.5 & 89.2 $\pm$ 3.1 \\
 & Omission & L1 & 5 & 44.3 $\pm$ 1.4 & 1.9 $\pm$ 0.6 & 95.9 $\pm$ 1.3 \\
 & Omission & L2 & 5 & 41.5 $\pm$ 1.7 & 4.6 $\pm$ 1.2 & 90.0 $\pm$ 2.5 \\
\addlinespace[1.5pt]
\dofa & Clean & -- & 5 & 47.5 $\pm$ 0.4 & -- & -- \\
 & Shift & L1 & 5 & 46.9 $\pm$ 0.8 & 0.5 $\pm$ 0.6 & 98.9 $\pm$ 1.3 \\
 & Shift & L2 & 5 & 43.0 $\pm$ 0.9 & 4.4 $\pm$ 0.7 & 90.7 $\pm$ 1.5 \\
 & Coarse & L1 & 5 & 45.6 $\pm$ 0.8 & 1.8 $\pm$ 0.5 & 96.1 $\pm$ 1.0 \\
 & Coarse & L2 & 5 & 41.7 $\pm$ 0.9 & 5.7 $\pm$ 1.2 & 87.9 $\pm$ 2.4 \\
 & Boundary & L1 & 5 & 46.0 $\pm$ 0.6 & 1.5 $\pm$ 0.5 & 96.8 $\pm$ 1.0 \\
 & Boundary & L2 & 5 & 42.0 $\pm$ 0.6 & 5.5 $\pm$ 0.8 & 88.5 $\pm$ 1.7 \\
 & Omission & L1 & 5 & 45.4 $\pm$ 0.8 & 2.0 $\pm$ 0.8 & 95.7 $\pm$ 1.7 \\
 & Omission & L2 & 5 & 42.7 $\pm$ 1.5 & 4.8 $\pm$ 1.5 & 89.9 $\pm$ 3.2 \\
\addlinespace[1.5pt]
\tm & Clean & -- & 5 & 40.2 $\pm$ 0.5 & -- & -- \\
 & Shift & L1 & 5 & 39.5 $\pm$ 0.4 & 0.7 $\pm$ 0.7 & 98.2 $\pm$ 1.8 \\
 & Shift & L2 & 5 & 36.7 $\pm$ 1.3 & 3.4 $\pm$ 1.3 & 91.4 $\pm$ 3.1 \\
 & Coarse & L1 & 5 & 39.9 $\pm$ 0.9 & 0.3 $\pm$ 0.7 & 99.4 $\pm$ 1.7 \\
 & Coarse & L2 & 5 & 35.2 $\pm$ 0.6 & 5.0 $\pm$ 0.8 & 87.5 $\pm$ 1.9 \\
 & Boundary & L1 & 5 & 38.3 $\pm$ 0.6 & 1.9 $\pm$ 0.2 & 95.2 $\pm$ 0.5 \\
 & Boundary & L2 & 5 & 33.9 $\pm$ 1.0 & 6.3 $\pm$ 1.1 & 84.3 $\pm$ 2.7 \\
 & Omission & L1 & 5 & 38.8 $\pm$ 1.0 & 1.4 $\pm$ 0.7 & 96.6 $\pm$ 1.7 \\
 & Omission & L2 & 5 & 34.5 $\pm$ 2.0 & 5.6 $\pm$ 1.8 & 85.9 $\pm$ 4.5 \\
\addlinespace[1.5pt]
\rn & Clean & -- & 5 & 43.8 $\pm$ 1.2 & -- & -- \\
 & Shift & L1 & 5 & 42.9 $\pm$ 0.8 & 0.9 $\pm$ 1.7 & 98.1 $\pm$ 3.7 \\
 & Shift & L2 & 5 & 40.5 $\pm$ 1.5 & 3.3 $\pm$ 1.7 & 92.6 $\pm$ 4.0 \\
 & Coarse & L1 & 5 & 42.9 $\pm$ 1.1 & 0.8 $\pm$ 1.9 & 98.2 $\pm$ 4.3 \\
 & Coarse & L2 & 5 & 39.3 $\pm$ 1.0 & 4.5 $\pm$ 1.9 & 89.8 $\pm$ 4.2 \\
 & Boundary & L1 & 5 & 42.0 $\pm$ 1.5 & 1.8 $\pm$ 2.3 & 96.1 $\pm$ 5.0 \\
 & Boundary & L2 & 5 & 41.1 $\pm$ 0.6 & 2.7 $\pm$ 1.4 & 93.9 $\pm$ 3.0 \\
 & Omission & L1 & 5 & 43.2 $\pm$ 0.9 & 0.5 $\pm$ 1.9 & 98.9 $\pm$ 4.2 \\
 & Omission & L2 & 5 & 39.5 $\pm$ 1.5 & 4.2 $\pm$ 2.2 & 90.4 $\pm$ 4.9 \\
\addlinespace[1.5pt]
\sat $\times$1/3 & Clean & -- & 5 & 41.9 $\pm$ 0.6 & -- & -- \\
 & Shift & L2 & 5 & 38.5 $\pm$ 0.5 & 3.4 $\pm$ 0.4 & 91.8 $\pm$ 0.9 \\
 & Coarse & L2 & 5 & 36.9 $\pm$ 1.4 & 5.0 $\pm$ 1.0 & 88.1 $\pm$ 2.4 \\
 & Boundary & L2 & 5 & 36.9 $\pm$ 1.9 & 5.0 $\pm$ 2.4 & 88.1 $\pm$ 5.5 \\
 & Omission & L2 & 5 & 36.2 $\pm$ 1.4 & 5.7 $\pm$ 1.3 & 86.3 $\pm$ 3.0 \\
\addlinespace[1.5pt]
\lvd $\times$1 & Clean & -- & 5 & 48.8 $\pm$ 0.6 & -- & -- \\
 & Shift & L2 & 5 & 44.6 $\pm$ 0.3 & 4.2 $\pm$ 0.7 & 91.5 $\pm$ 1.2 \\
 & Coarse & L2 & 5 & 42.9 $\pm$ 1.0 & 5.8 $\pm$ 1.1 & 88.0 $\pm$ 2.2 \\
 & Boundary & L2 & 5 & 44.3 $\pm$ 1.0 & 4.5 $\pm$ 0.9 & 90.8 $\pm$ 1.7 \\
 & Omission & L2 & 5 & 44.5 $\pm$ 1.2 & 4.3 $\pm$ 1.7 & 91.3 $\pm$ 3.3 \\
\bottomrule
\end{tabular}
\end{adjustbox}}{\tabletbd[6cm]{planned/generated/table\_n10\_grid\_flair.tex}}
\caption{\textbf{N10 grid on FLAIR-2:} Every N10 cell on \flair with the SAM head, in the layout of \Cref{tab:supp-n7-grid-burn}. \ProvSlot{\ProvNTen}}
\label{tab:supp-n10-grid-flair}
\end{table}

\begin{table}[t]
\centering\footnotesize\setlength{\tabcolsep}{3pt}
\IfFileExists{planned/generated/table_n15_grid_burn.tex}{\begin{adjustbox}{max width=\columnwidth}%
\providecommand{\gcvdry}[1]{\textcolor{red}{#1}}
\providecommand{\ProvTableNFifteenGridBurn}{Seeds per cell 5. Mean $\pm$ standard deviation over seeds, Pretrained SAM head, \burn. Red cells are not run yet or still have a pending seed. Analysis of the finished runs.}
\begin{tabular}{@{}lllcccc@{}}
\toprule
Model & Type & Level & $n$ & mIoU & Drop & Retention (\%) \\
\midrule
\sat & Clean & -- & 5 & 64.1 $\pm$ 11.8 & -- & -- \\
 & Shift & L1 & 5 & 64.3 $\pm$ 11.1 & $-$0.1 $\pm$ 6.7 & 100.7 $\pm$ 9.8 \\
 & Shift & L2 & 5 & 52.8 $\pm$ 8.4 & 11.4 $\pm$ 8.8 & 83.5 $\pm$ 12.9 \\
 & Coarse & L1 & 5 & 64.5 $\pm$ 11.1 & $-$0.4 $\pm$ 15.6 & 104.1 $\pm$ 30.9 \\
 & Coarse & L2 & 5 & 47.9 $\pm$ 7.0 & 16.2 $\pm$ 13.0 & 76.9 $\pm$ 17.8 \\
 & Boundary & L1 & 5 & 64.4 $\pm$ 11.6 & $-$0.3 $\pm$ 5.7 & 100.8 $\pm$ 8.4 \\
 & Boundary & L2 & 5 & 55.5 $\pm$ 9.8 & 8.7 $\pm$ 16.3 & 90.1 $\pm$ 30.6 \\
 & Omission & L1 & 5 & 58.9 $\pm$ 9.8 & 5.2 $\pm$ 10.5 & 93.6 $\pm$ 18.4 \\
 & Omission & L2 & 5 & 54.3 $\pm$ 7.3 & 9.9 $\pm$ 15.6 & 88.2 $\pm$ 27.9 \\
\addlinespace[1.5pt]
\lvd & Clean & -- & 5 & 73.9 $\pm$ 16.5 & -- & -- \\
 & Shift & L1 & 5 & 65.6 $\pm$ 19.0 & 8.4 $\pm$ 17.5 & 90.1 $\pm$ 20.8 \\
 & Shift & L2 & 5 & 55.4 $\pm$ 14.6 & 18.5 $\pm$ 17.1 & 77.3 $\pm$ 21.0 \\
 & Coarse & L1 & 5 & 64.2 $\pm$ 17.8 & 9.7 $\pm$ 14.7 & 88.0 $\pm$ 18.3 \\
 & Coarse & L2 & 5 & 49.1 $\pm$ 9.7 & 24.8 $\pm$ 16.0 & 69.3 $\pm$ 19.9 \\
 & Boundary & L1 & 5 & 72.5 $\pm$ 15.5 & 1.5 $\pm$ 2.5 & 98.2 $\pm$ 3.0 \\
 & Boundary & L2 & 5 & 49.7 $\pm$ 11.0 & 24.2 $\pm$ 16.4 & 70.0 $\pm$ 20.4 \\
 & Omission & L1 & 5 & 67.7 $\pm$ 13.0 & 6.3 $\pm$ 5.8 & 92.4 $\pm$ 6.9 \\
 & Omission & L2 & 5 & 60.1 $\pm$ 9.7 & 13.8 $\pm$ 8.5 & 82.9 $\pm$ 10.6 \\
\addlinespace[1.5pt]
\dofa & Clean & -- & 5 & 85.4 $\pm$ 0.7 & -- & -- \\
 & Shift & L1 & 5 & 83.5 $\pm$ 1.3 & 1.9 $\pm$ 1.6 & 97.7 $\pm$ 1.9 \\
 & Shift & L2 & 5 & 74.1 $\pm$ 1.6 & 11.4 $\pm$ 1.5 & 86.7 $\pm$ 1.7 \\
 & Coarse & L1 & 5 & 81.5 $\pm$ 1.2 & 3.9 $\pm$ 0.9 & 95.4 $\pm$ 1.1 \\
 & Coarse & L2 & 5 & 64.8 $\pm$ 3.3 & 20.6 $\pm$ 3.6 & 75.9 $\pm$ 4.1 \\
 & Boundary & L1 & 5 & 81.8 $\pm$ 0.5 & 3.6 $\pm$ 0.8 & 95.8 $\pm$ 0.9 \\
 & Boundary & L2 & 5 & 73.7 $\pm$ 1.8 & 11.7 $\pm$ 2.0 & 86.3 $\pm$ 2.3 \\
 & Omission & L1 & 5 & 73.4 $\pm$ 16.3 & 12.0 $\pm$ 16.1 & 86.0 $\pm$ 18.9 \\
 & Omission & L2 & 5 & 55.9 $\pm$ 7.4 & 29.5 $\pm$ 7.0 & 65.4 $\pm$ 8.3 \\
\addlinespace[1.5pt]
\tm & Clean & -- & 5 & 87.8 $\pm$ 1.3 & -- & -- \\
 & Shift & L1 & 5 & 84.7 $\pm$ 1.6 & 3.1 $\pm$ 2.7 & 96.5 $\pm$ 3.0 \\
 & Shift & L2 & 5 & 76.7 $\pm$ 2.2 & 11.1 $\pm$ 2.7 & 87.4 $\pm$ 3.0 \\
 & Coarse & L1 & 5 & 83.5 $\pm$ 1.2 & 4.3 $\pm$ 1.7 & 95.1 $\pm$ 1.8 \\
 & Coarse & L2 & 5 & 69.0 $\pm$ 3.3 & 18.8 $\pm$ 4.2 & 78.6 $\pm$ 4.5 \\
 & Boundary & L1 & 5 & 83.6 $\pm$ 0.5 & 4.2 $\pm$ 1.7 & 95.2 $\pm$ 1.9 \\
 & Boundary & L2 & 5 & 77.5 $\pm$ 2.7 & 10.3 $\pm$ 3.7 & 88.3 $\pm$ 4.0 \\
 & Omission & L1 & 5 & 83.4 $\pm$ 1.5 & 4.4 $\pm$ 1.9 & 95.0 $\pm$ 2.2 \\
 & Omission & L2 & 5 & 71.0 $\pm$ 6.5 & 16.8 $\pm$ 6.6 & 80.9 $\pm$ 7.4 \\
\addlinespace[1.5pt]
\rn & Clean & -- & 5 & 79.4 $\pm$ 19.3 & -- & -- \\
 & Shift & L1 & 5 & 69.7 $\pm$ 22.7 & 9.7 $\pm$ 18.7 & 89.0 $\pm$ 21.3 \\
 & Shift & L2 & 5 & 51.0 $\pm$ 14.0 & 28.3 $\pm$ 21.1 & 67.9 $\pm$ 23.9 \\
 & Coarse & L1 & 5 & 76.5 $\pm$ 17.7 & 2.9 $\pm$ 2.1 & 96.7 $\pm$ 2.3 \\
 & Coarse & L2 & 5 & 55.5 $\pm$ 14.7 & 23.8 $\pm$ 18.9 & 72.9 $\pm$ 21.5 \\
 & Boundary & L1 & 5 & 77.4 $\pm$ 18.3 & 1.9 $\pm$ 1.2 & 97.8 $\pm$ 1.3 \\
 & Boundary & L2 & 5 & 51.2 $\pm$ 14.3 & 28.2 $\pm$ 21.2 & 68.0 $\pm$ 24.0 \\
 & Omission & L1 & 5 & 60.0 $\pm$ 20.8 & 19.4 $\pm$ 22.2 & 78.0 $\pm$ 25.1 \\
 & Omission & L2 & 5 & 44.8 $\pm$ 0.0 & 34.6 $\pm$ 19.3 & 60.7 $\pm$ 22.0 \\
\addlinespace[1.5pt]
\sat $\times$1/3 & Clean & -- & 5 & 77.9 $\pm$ 2.7 & -- & -- \\
 & Shift & L2 & 5 & 54.1 $\pm$ 12.8 & 23.7 $\pm$ 10.6 & 69.2 $\pm$ 14.3 \\
 & Coarse & L2 & 5 & 59.3 $\pm$ 8.2 & 18.6 $\pm$ 6.3 & 76.0 $\pm$ 8.9 \\
 & Boundary & L2 & 5 & 62.2 $\pm$ 9.8 & 15.7 $\pm$ 10.2 & 80.0 $\pm$ 13.1 \\
 & Omission & L2 & 5 & 55.9 $\pm$ 3.1 & 22.0 $\pm$ 5.6 & 72.0 $\pm$ 6.4 \\
\addlinespace[1.5pt]
\lvd $\times$1 & Clean & -- & 5 & 74.7 $\pm$ 2.3 & -- & -- \\
 & Shift & L2 & 5 & 44.8 $\pm$ 0.0 & 29.9 $\pm$ 2.3 & 60.0 $\pm$ 1.8 \\
 & Coarse & L2 & 5 & 46.6 $\pm$ 4.1 & 28.1 $\pm$ 3.8 & 62.4 $\pm$ 5.0 \\
 & Boundary & L2 & 5 & 50.5 $\pm$ 7.9 & 24.2 $\pm$ 9.2 & 67.8 $\pm$ 12.0 \\
 & Omission & L2 & 5 & 44.8 $\pm$ 0.0 & 29.9 $\pm$ 2.3 & 60.0 $\pm$ 1.8 \\
\bottomrule
\end{tabular}
\end{adjustbox}}{\tabletbd[6cm]{planned/generated/table\_n15\_grid\_burn.tex}}
\caption{\textbf{N15 grid on HLS Burn Scars:} Every N15 cell on \burn with the pretrained SAM head, in the layout of \Cref{tab:supp-n7-grid-burn}. \ProvSlot{\ProvNFifteen}}
\label{tab:supp-n15-grid-burn}
\end{table}

\begin{table}[t]
\centering\footnotesize\setlength{\tabcolsep}{3pt}
\IfFileExists{planned/generated/table_n15_grid_flair.tex}{\begin{adjustbox}{max width=\columnwidth}%
\providecommand{\gcvdry}[1]{\textcolor{red}{#1}}
\providecommand{\ProvTableNFifteenGridFlair}{Seeds per cell 5. Mean $\pm$ standard deviation over seeds, Pretrained SAM head, \flair. Red cells are not run yet or still have a pending seed. Analysis of the finished runs.}
\begin{tabular}{@{}lllcccc@{}}
\toprule
Model & Type & Level & $n$ & mIoU & Drop & Retention (\%) \\
\midrule
\sat & Clean & -- & 5 & 44.8 $\pm$ 1.3 & -- & -- \\
 & Shift & L1 & 5 & 44.4 $\pm$ 0.5 & 0.3 $\pm$ 1.1 & 99.4 $\pm$ 2.4 \\
 & Shift & L2 & 5 & 40.3 $\pm$ 2.2 & 4.4 $\pm$ 1.8 & 90.1 $\pm$ 4.0 \\
 & Coarse & L1 & 5 & 43.6 $\pm$ 2.0 & 1.2 $\pm$ 1.8 & 97.4 $\pm$ 4.1 \\
 & Coarse & L2 & 5 & 39.9 $\pm$ 1.3 & 4.8 $\pm$ 2.2 & 89.3 $\pm$ 4.7 \\
 & Boundary & L1 & 5 & 42.9 $\pm$ 1.4 & 1.9 $\pm$ 1.9 & 95.9 $\pm$ 4.1 \\
 & Boundary & L2 & 5 & 38.7 $\pm$ 1.1 & 6.0 $\pm$ 1.1 & 86.5 $\pm$ 2.3 \\
 & Omission & L1 & 5 & 42.5 $\pm$ 2.7 & 2.2 $\pm$ 1.8 & 95.0 $\pm$ 4.1 \\
 & Omission & L2 & 5 & 39.1 $\pm$ 1.9 & 5.7 $\pm$ 1.3 & 87.4 $\pm$ 2.9 \\
\addlinespace[1.5pt]
\lvd & Clean & -- & 5 & 45.1 $\pm$ 1.2 & -- & -- \\
 & Shift & L1 & 5 & 44.3 $\pm$ 1.1 & 0.8 $\pm$ 1.0 & 98.2 $\pm$ 2.2 \\
 & Shift & L2 & 5 & 41.1 $\pm$ 1.7 & 4.0 $\pm$ 1.7 & 91.2 $\pm$ 3.6 \\
 & Coarse & L1 & 5 & 44.2 $\pm$ 1.0 & 0.9 $\pm$ 0.3 & 97.9 $\pm$ 0.7 \\
 & Coarse & L2 & 5 & 38.5 $\pm$ 1.7 & 6.7 $\pm$ 1.1 & 85.2 $\pm$ 2.6 \\
 & Boundary & L1 & 5 & 43.7 $\pm$ 1.0 & 1.4 $\pm$ 0.5 & 96.8 $\pm$ 1.2 \\
 & Boundary & L2 & 5 & 39.8 $\pm$ 2.2 & 5.3 $\pm$ 1.3 & 88.3 $\pm$ 3.1 \\
 & Omission & L1 & 5 & 43.4 $\pm$ 1.0 & 1.7 $\pm$ 0.5 & 96.3 $\pm$ 1.1 \\
 & Omission & L2 & 5 & 40.2 $\pm$ 2.2 & 4.9 $\pm$ 1.3 & 89.1 $\pm$ 3.0 \\
\addlinespace[1.5pt]
\dofa & Clean & -- & 5 & 45.1 $\pm$ 0.9 & -- & -- \\
 & Shift & L1 & 5 & 44.4 $\pm$ 1.0 & 0.6 $\pm$ 1.8 & 98.6 $\pm$ 4.0 \\
 & Shift & L2 & 5 & 41.1 $\pm$ 0.7 & 3.9 $\pm$ 1.4 & 91.3 $\pm$ 3.0 \\
 & Coarse & L1 & 5 & 43.8 $\pm$ 1.1 & 1.2 $\pm$ 1.0 & 97.3 $\pm$ 2.3 \\
 & Coarse & L2 & 5 & 38.7 $\pm$ 0.8 & 6.4 $\pm$ 1.1 & 85.8 $\pm$ 2.1 \\
 & Boundary & L1 & 5 & 43.7 $\pm$ 0.9 & 1.4 $\pm$ 0.8 & 96.9 $\pm$ 1.7 \\
 & Boundary & L2 & 5 & 39.2 $\pm$ 1.0 & 5.9 $\pm$ 1.1 & 86.9 $\pm$ 2.3 \\
 & Omission & L1 & 5 & 42.9 $\pm$ 1.0 & 2.2 $\pm$ 1.2 & 95.1 $\pm$ 2.7 \\
 & Omission & L2 & 5 & 39.8 $\pm$ 1.3 & 5.3 $\pm$ 1.7 & 88.3 $\pm$ 3.6 \\
\addlinespace[1.5pt]
\tm & Clean & -- & 5 & 39.3 $\pm$ 1.2 & -- & -- \\
 & Shift & L1 & 5 & 39.0 $\pm$ 0.6 & 0.3 $\pm$ 0.8 & 99.3 $\pm$ 2.1 \\
 & Shift & L2 & 5 & 33.9 $\pm$ 2.7 & 5.4 $\pm$ 2.6 & 86.3 $\pm$ 6.5 \\
 & Coarse & L1 & 5 & 38.3 $\pm$ 1.5 & 1.0 $\pm$ 2.2 & 97.5 $\pm$ 5.5 \\
 & Coarse & L2 & 5 & 32.0 $\pm$ 1.4 & 7.2 $\pm$ 2.6 & 81.7 $\pm$ 6.2 \\
 & Boundary & L1 & 5 & 37.6 $\pm$ 0.6 & 1.7 $\pm$ 1.7 & 95.8 $\pm$ 4.4 \\
 & Boundary & L2 & 5 & 31.8 $\pm$ 1.0 & 7.5 $\pm$ 1.8 & 81.0 $\pm$ 4.2 \\
 & Omission & L1 & 5 & 37.1 $\pm$ 0.8 & 2.2 $\pm$ 1.1 & 94.6 $\pm$ 2.7 \\
 & Omission & L2 & 5 & 33.8 $\pm$ 2.0 & 5.4 $\pm$ 1.7 & 86.2 $\pm$ 4.3 \\
\addlinespace[1.5pt]
\rn & Clean & -- & 5 & 40.5 $\pm$ 1.3 & -- & -- \\
 & Shift & L1 & 5 & 38.0 $\pm$ 2.6 & 2.5 $\pm$ 2.0 & 93.7 $\pm$ 5.0 \\
 & Shift & L2 & 5 & 37.0 $\pm$ 3.2 & 3.5 $\pm$ 3.4 & 91.5 $\pm$ 8.2 \\
 & Coarse & L1 & 5 & 39.3 $\pm$ 2.7 & 1.2 $\pm$ 1.5 & 96.9 $\pm$ 3.9 \\
 & Coarse & L2 & 5 & 34.7 $\pm$ 2.9 & 5.8 $\pm$ 2.5 & 85.8 $\pm$ 6.1 \\
 & Boundary & L1 & 5 & 39.2 $\pm$ 3.7 & 1.3 $\pm$ 3.8 & 96.9 $\pm$ 9.2 \\
 & Boundary & L2 & 5 & 36.0 $\pm$ 2.2 & 4.5 $\pm$ 1.7 & 89.0 $\pm$ 4.1 \\
 & Omission & L1 & 5 & 38.9 $\pm$ 2.3 & 1.6 $\pm$ 2.0 & 96.1 $\pm$ 4.8 \\
 & Omission & L2 & 5 & 37.2 $\pm$ 1.0 & 3.3 $\pm$ 1.8 & 92.0 $\pm$ 4.3 \\
\addlinespace[1.5pt]
\sat $\times$1/3 & Clean & -- & 5 & 39.9 $\pm$ 1.8 & -- & -- \\
 & Shift & L2 & 5 & 37.1 $\pm$ 1.1 & 2.9 $\pm$ 1.2 & 92.9 $\pm$ 3.0 \\
 & Coarse & L2 & 5 & 35.3 $\pm$ 2.9 & 4.6 $\pm$ 1.3 & 88.3 $\pm$ 3.8 \\
 & Boundary & L2 & 5 & 34.5 $\pm$ 2.2 & 5.5 $\pm$ 0.6 & 86.3 $\pm$ 2.0 \\
 & Omission & L2 & 5 & 34.1 $\pm$ 3.0 & 5.8 $\pm$ 2.0 & 85.4 $\pm$ 5.1 \\
\addlinespace[1.5pt]
\lvd $\times$1 & Clean & -- & 5 & 47.3 $\pm$ 1.3 & -- & -- \\
 & Shift & L2 & 5 & 43.3 $\pm$ 0.4 & 4.0 $\pm$ 1.6 & 91.7 $\pm$ 3.1 \\
 & Coarse & L2 & 5 & 42.5 $\pm$ 0.9 & 4.8 $\pm$ 1.7 & 89.9 $\pm$ 3.3 \\
 & Boundary & L2 & 5 & 42.8 $\pm$ 0.6 & 4.6 $\pm$ 1.1 & 90.4 $\pm$ 2.0 \\
 & Omission & L2 & 5 & 42.4 $\pm$ 2.1 & 4.9 $\pm$ 1.6 & 89.7 $\pm$ 3.5 \\
\bottomrule
\end{tabular}
\end{adjustbox}}{\tabletbd[6cm]{planned/generated/table\_n15\_grid\_flair.tex}}
\caption{\textbf{N15 grid on FLAIR-2:} Every N15 cell on \flair with the pretrained SAM head, in the layout of \Cref{tab:supp-n7-grid-burn}. \ProvSlot{\ProvNFifteen}}
\label{tab:supp-n15-grid-flair}
\end{table}
\fi
\section{Matched Learning Rate for Every Model (N8)}
\label{sec:supp-n8}

\begin{table}[t]
\centering\footnotesize\setlength{\tabcolsep}{3pt}
\IfFileExists{planned/generated/table_n8.tex}{\begin{adjustbox}{max width=\columnwidth}%
\providecommand{\gcvdry}[1]{\textcolor{red}{#1}}
\providecommand{\ProvTableNEight}{Seeds per cell 5. Mean paired drop over the four \Ltwo types in points, every noisy run paired with the clean run of its own rate and seed, every arm on seeds 1 to 5 (the seeds N6 and N8 run). The rate effect and the contrasts carry the 95\,\% paired $t$ interval. Red cells are not run yet or rest on a block that is still running. Analysis of the finished runs.}
\begin{tabular}{@{}llccc@{}}
\toprule
Model & New arm & Drop $\times$1 & Drop $\times$1/3 & $\times$1/3 $-$ $\times$1 \\
\midrule
\multicolumn{5}{@{}l}{\textit{\burn}} \\
\sat & N6 ($\times$1/3) & 17.8 & 13.9 & $-$3.9 [$-$8.3, 0.6] \\
\lvd & N8 ($\times$1) & 19.5 & 14.1 & $-$5.4 [$-$8.2, $-$2.7] \\
\dofa & N8 ($\times$1/3) & 17.9 & 17.1 & $-$0.8 [$-$2.5, 0.9] \\
\tm & N8 ($\times$1/3) & 12.9 & 10.9 & $-$2.0 [$-$5.5, 1.5] \\
\rn & N8 ($\times$1/3) & 15.9 & 14.1 & $-$1.7 [$-$6.9, 3.4] \\
\addlinespace[1.5pt]
F2 at $\times$1/3 (H6) & N8 & \multicolumn{3}{c}{$-$0.2 [$-$3.4, 3.1]} \\
F2 at $\times$1 & A1 & \multicolumn{3}{c}{$-$0.5 [$-$5.4, 4.4]} \\
RQ1 at $\times$1 & N8 & \multicolumn{3}{c}{$-$1.7 [$-$5.2, 1.7]} \\
RQ1 at $\times$1/3 & N6 & \multicolumn{3}{c}{$-$0.2 [$-$3.7, 3.3]} \\
\midrule
\multicolumn{5}{@{}l}{\textit{\flair}} \\
\sat & N6 ($\times$1/3) & 6.5 & 5.1 & $-$1.4 [$-$2.6, $-$0.2] \\
\lvd & N8 ($\times$1) & 4.0 & 3.6 & $-$0.4 [$-$1.9, 1.2] \\
\dofa & N8 ($\times$1/3) & 4.3 & 3.9 & $-$0.4 [$-$1.3, 0.4] \\
\tm & N8 ($\times$1/3) & 5.2 & 4.2 & $-$1.0 [$-$3.3, 1.3] \\
\rn & N8 ($\times$1/3) & 4.2 & 3.8 & $-$0.4 [$-$2.0, 1.3] \\
\addlinespace[1.5pt]
F2 at $\times$1/3 (H6) & N8 & \multicolumn{3}{c}{0.2 [$-$1.3, 1.8]} \\
F2 at $\times$1 & A1 & \multicolumn{3}{c}{0.6 [$-$2.0, 3.1]} \\
RQ1 at $\times$1 & N8 & \multicolumn{3}{c}{2.5 [1.1, 4.0]} \\
RQ1 at $\times$1/3 & N6 & \multicolumn{3}{c}{1.5 [0.6, 2.5]} \\
\bottomrule
\end{tabular}
\end{adjustbox}}{\tabletbd[3cm]{planned/generated/table\_n8.tex}}
\caption{\textbf{Every core model at both learning-rate multipliers:} Mean paired drop over the four \Ltwo types in points on seeds 1 to 5 at the \gbtwo rate ($\times$1) and one third of it. The last column is the rate effect within a model, paired by seed, with its \CiLevel\,\% interval. The second column names the block that supplies the arm the main grid lacks. The top block shows \burn and the bottom block \flair. The last rows of each block give F2 and the RQ1 contrast at each common multiplier. \ProvSlot{\ProvNEight}}
\label{tab:n8}
\end{table}

\ifdefined\cvprbuild\else
\begin{figure}[t]
\centering
\IfFileExists{planned/figs/fig_n8_matched_lr.pdf}{\includegraphics[width=\linewidth]{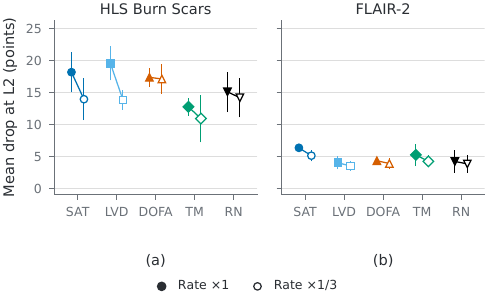}}{\phfig{4cm}{planned/figs/fig\_n8\_matched\_lr.pdf}}
\caption{\textbf{Drop at both learning-rate multipliers:} Mean paired drop over the four \Ltwo types in points per core model with \CiLevel\,\% intervals over seeds. Filled markers are the \gbtwo rate and open markers one third of it. The left panel shows \burn and the right panel \flair. Models are abbreviated as SAT (\sat), LVD (\lvd), DOFA (\dofa), TM (\tm) and RN (\rn). Each arm averages all seeds of its cells, so some arms differ from \Cref{tab:n8}. \ProvSlot{\ProvNEight}}
\label{fig:n8}
\end{figure}

\fi
Block N8 gives every core model both multipliers at the clean condition and \Ltwo \ifdefined\cvprbuild (\Cref{tab:n8})\else (\Cref{tab:n8}, \Cref{fig:n8})\fi{}.
At the common $\times$1/3 multiplier the F2 contrast is \hlMraFTwoThirdburnEst{} [\hlMraFTwoThirdburnCILo, \hlMraFTwoThirdburnCIHi] points on \burn and \hlMraFTwoThirdflairEst{} [\hlMraFTwoThirdflairCILo, \hlMraFTwoThirdflairCIHi] on \flair.
Its 90\,\% equivalence intervals are [\hlMraFTwoThirdburnTostCILo, \hlMraFTwoThirdburnTostCIHi] and [\hlMraFTwoThirdflairTostCILo, \hlMraFTwoThirdflairTostCIHi].
The two tests resolve \hlMraFTwoThirdburnMde{} and \hlMraFTwoThirdflairMde{} points.
On \burn the rate effect within a model is \hlMraRateDiffDOFAburnEst{} [\hlMraRateDiffDOFAburnCILo, \hlMraRateDiffDOFAburnCIHi] points for \dofa and \hlMraRateDiffTMburnEst{} [\hlMraRateDiffTMburnCILo, \hlMraRateDiffTMburnCIHi] for \tm.
For \rn it is \hlMraRateDiffRNburnEst{} [\hlMraRateDiffRNburnCILo, \hlMraRateDiffRNburnCIHi] points.
On \flair the three effects are \hlMraRateDiffDOFAflairEst{} [\hlMraRateDiffDOFAflairCILo, \hlMraRateDiffDOFAflairCIHi], \hlMraRateDiffTMflairEst{} [\hlMraRateDiffTMflairCILo, \hlMraRateDiffTMflairCIHi] and \hlMraRateDiffRNflairEst{} [\hlMraRateDiffRNflairCILo, \hlMraRateDiffRNflairCIHi] points.
At the common $\times$1 multiplier \sat minus \lvd is \hlMraFOneOneburnEst{} [\hlMraFOneOneburnCILo, \hlMraFOneOneburnCIHi] points on \burn and \hlMraFOneOneflairEst{} [\hlMraFOneOneflairCILo, \hlMraFOneOneflairCIHi] on \flair.
It reads \vd{\hlMraVerdictFOneburn} and \vd{\hlMraVerdictFOneflair}.
\ifdefined\cvprbuild\else

The \lvd clean seed-\CalSeed{} cell at $\times$1 on each dataset is the A2 probe run.
Its selected-type \Ltwo cells at seeds 1 and 2 are N2 cells, \etype{coarse} on \burn and \etype{omission} on \flair.
These \RunsNEightAliased{} cells are not trained again.
The clean \miou at the new multiplier is \hlMraCleanLVDOneburn{} for \lvd, \hlMraCleanDOFAThirdburn{} for \dofa, \hlMraCleanTMThirdburn{} for \tm and \hlMraCleanRNThirdburn{} for \rn on \burn.
On \flair it is \hlMraCleanLVDOneflair, \hlMraCleanDOFAThirdflair, \hlMraCleanTMThirdflair{} and \hlMraCleanRNThirdflair.
The multiplier with the smaller mean \Ltwo drop is \hlMraBestRateLVDburn{} for \lvd, \hlMraBestRateDOFAburn{} for \dofa, \hlMraBestRateTMburn{} for \tm and \hlMraBestRateRNburn{} for \rn on \burn.
On \flair it is \hlMraBestRateLVDflair, \hlMraBestRateDOFAflair, \hlMraBestRateTMflair{} and \hlMraBestRateRNflair.
\fi

\ifdefined\cvprbuild\else
\begin{table}[t]
\centering\footnotesize\setlength{\tabcolsep}{3pt}
\IfFileExists{planned/generated/table_n8_grid.tex}{\begin{adjustbox}{max width=\columnwidth}%
\providecommand{\gcvdry}[1]{\textcolor{red}{#1}}
\providecommand{\ProvTableNEightGrid}{Seeds per cell 5. Mean $\pm$ standard deviation over seeds, UNet head, at the rate each model did not have before N8, every noisy run paired with the clean run of its own rate and seed. Red cells are not run yet or still have a pending seed. Analysis of the finished runs.}
\begin{tabular}{@{}llllcccc@{}}
\toprule
Model & Rate & Dataset & Type & $n$ & mIoU & Drop & Retention (\%) \\
\midrule
\lvd & $\times$1 & \burn & Clean & 5 & 78.3 $\pm$ 2.4 & -- & -- \\
 &  &  & Shift \Ltwo & 5 & 63.5 $\pm$ 4.8 & 14.8 $\pm$ 3.6 & 81.1 $\pm$ 4.8 \\
 &  &  & Coarse \Ltwo & 5 & 52.0 $\pm$ 6.8 & 26.3 $\pm$ 6.3 & 66.4 $\pm$ 8.3 \\
 &  &  & Boundary \Ltwo & 5 & 63.5 $\pm$ 2.3 & 14.8 $\pm$ 3.3 & 81.1 $\pm$ 3.7 \\
 &  &  & Omission \Ltwo & 5 & 56.1 $\pm$ 7.4 & 22.2 $\pm$ 6.6 & 71.6 $\pm$ 8.5 \\
 &  & \flair & Clean & 5 & 51.2 $\pm$ 0.6 & -- & -- \\
 &  &  & Shift \Ltwo & 5 & 48.4 $\pm$ 0.2 & 2.7 $\pm$ 0.4 & 94.7 $\pm$ 0.8 \\
 &  &  & Coarse \Ltwo & 5 & 45.4 $\pm$ 0.4 & 5.8 $\pm$ 0.7 & 88.7 $\pm$ 1.3 \\
 &  &  & Boundary \Ltwo & 5 & 47.2 $\pm$ 1.4 & 3.9 $\pm$ 1.7 & 92.4 $\pm$ 3.3 \\
 &  &  & Omission \Ltwo & 5 & 47.7 $\pm$ 1.1 & 3.5 $\pm$ 1.3 & 93.2 $\pm$ 2.6 \\
\addlinespace[1.5pt]
\dofa & $\times$1/3 & \burn & Clean & 5 & 83.7 $\pm$ 1.2 & -- & -- \\
 &  &  & Shift \Ltwo & 5 & 73.2 $\pm$ 1.1 & 10.5 $\pm$ 1.3 & 87.5 $\pm$ 1.5 \\
 &  &  & Coarse \Ltwo & 5 & 63.4 $\pm$ 3.2 & 20.2 $\pm$ 3.1 & 75.8 $\pm$ 3.7 \\
 &  &  & Boundary \Ltwo & 5 & 70.1 $\pm$ 1.3 & 13.6 $\pm$ 1.0 & 83.8 $\pm$ 1.2 \\
 &  &  & Omission \Ltwo & 5 & 59.7 $\pm$ 6.7 & 24.0 $\pm$ 5.9 & 71.2 $\pm$ 7.3 \\
 &  & \flair & Clean & 5 & 46.7 $\pm$ 0.6 & -- & -- \\
 &  &  & Shift \Ltwo & 5 & 44.1 $\pm$ 0.7 & 2.6 $\pm$ 0.7 & 94.4 $\pm$ 1.5 \\
 &  &  & Coarse \Ltwo & 5 & 42.2 $\pm$ 1.0 & 4.6 $\pm$ 0.6 & 90.2 $\pm$ 1.4 \\
 &  &  & Boundary \Ltwo & 5 & 42.0 $\pm$ 1.2 & 4.7 $\pm$ 1.4 & 89.9 $\pm$ 2.9 \\
 &  &  & Omission \Ltwo & 5 & 43.2 $\pm$ 2.0 & 3.6 $\pm$ 2.4 & 92.4 $\pm$ 5.0 \\
\addlinespace[1.5pt]
\tm & $\times$1/3 & \burn & Clean & 5 & 85.1 $\pm$ 2.6 & -- & -- \\
 &  &  & Shift \Ltwo & 5 & 77.2 $\pm$ 1.2 & 7.9 $\pm$ 2.5 & 90.8 $\pm$ 2.8 \\
 &  &  & Coarse \Ltwo & 5 & 69.8 $\pm$ 2.3 & 15.3 $\pm$ 2.8 & 82.0 $\pm$ 2.9 \\
 &  &  & Boundary \Ltwo & 5 & 74.8 $\pm$ 2.6 & 10.4 $\pm$ 2.5 & 87.8 $\pm$ 2.8 \\
 &  &  & Omission \Ltwo & 5 & 75.1 $\pm$ 4.8 & 10.0 $\pm$ 6.8 & 88.4 $\pm$ 7.5 \\
 &  & \flair & Clean & 5 & 38.2 $\pm$ 1.3 & -- & -- \\
 &  &  & Shift \Ltwo & 5 & 35.1 $\pm$ 0.7 & 3.0 $\pm$ 0.9 & 92.1 $\pm$ 2.1 \\
 &  &  & Coarse \Ltwo & 5 & 33.7 $\pm$ 1.3 & 4.4 $\pm$ 0.6 & 88.4 $\pm$ 1.6 \\
 &  &  & Boundary \Ltwo & 5 & 32.9 $\pm$ 0.8 & 5.3 $\pm$ 1.3 & 86.3 $\pm$ 3.0 \\
 &  &  & Omission \Ltwo & 5 & 34.1 $\pm$ 0.8 & 4.1 $\pm$ 1.1 & 89.3 $\pm$ 2.6 \\
\addlinespace[1.5pt]
\rn & $\times$1/3 & \burn & Clean & 5 & 88.4 $\pm$ 0.9 & -- & -- \\
 &  &  & Shift \Ltwo & 5 & 75.9 $\pm$ 1.1 & 12.5 $\pm$ 1.6 & 85.9 $\pm$ 1.7 \\
 &  &  & Coarse \Ltwo & 5 & 70.2 $\pm$ 1.8 & 18.2 $\pm$ 2.5 & 79.4 $\pm$ 2.7 \\
 &  &  & Boundary \Ltwo & 5 & 76.0 $\pm$ 2.2 & 12.4 $\pm$ 2.6 & 86.0 $\pm$ 2.8 \\
 &  &  & Omission \Ltwo & 5 & 74.9 $\pm$ 6.0 & 13.5 $\pm$ 6.5 & 84.8 $\pm$ 7.2 \\
 &  & \flair & Clean & 5 & 44.2 $\pm$ 1.7 & -- & -- \\
 &  &  & Shift \Ltwo & 5 & 41.1 $\pm$ 1.1 & 3.1 $\pm$ 1.4 & 93.1 $\pm$ 3.0 \\
 &  &  & Coarse \Ltwo & 5 & 38.5 $\pm$ 1.3 & 5.7 $\pm$ 0.5 & 87.0 $\pm$ 0.7 \\
 &  &  & Boundary \Ltwo & 5 & 40.3 $\pm$ 1.9 & 3.9 $\pm$ 2.6 & 91.2 $\pm$ 5.5 \\
 &  &  & Omission \Ltwo & 5 & 41.7 $\pm$ 1.2 & 2.5 $\pm$ 0.7 & 94.3 $\pm$ 1.4 \\
\bottomrule
\end{tabular}
\end{adjustbox}}{\tabletbd[6cm]{planned/generated/table\_n8\_grid.tex}}
\caption{\textbf{N8 grid:} Every N8 cell with its number of seeds $n$. Values are mean $\pm$ standard deviation over seeds of the test \miou, the paired drop in points and the retention. The rate column is the multiplier of the \gbtwo learning rate that the model does not have in the main grid. \ProvSlot{\ProvNEight}}
\label{tab:supp-n8-grid}
\end{table}

\fi
\section{Pair Level, Seed Extension and Noise Robust Losses (N12, N13, N14)}
\label{sec:supp-lr}

\begin{table}[t]
\centering\footnotesize\setlength{\tabcolsep}{3pt}
\IfFileExists{planned/generated/table_n12.tex}{\begin{adjustbox}{max width=\columnwidth}%
\providecommand{\gcvdry}[1]{\textcolor{red}{#1}}
\providecommand{\ProvTableNTwelve}{Seeds per cell 5 (\Lpair{} cells). Mean paired drop over the four error types in points at \Lone, \Ltwo and the pair level \Lpair, every noisy run paired with the clean run of its own rate and seed. The \Lone and \Ltwo columns take the seeds of the \Lpair{} cells next to them (seeds 1 to 5 while \Lpair{} has not run). \Lpair{} $-$ \Ltwo pairs by model and seed, N6 (\sat $\times$1/3 $-$ \lvd $\times$1/3) and F2 by seed, with the 95\,\% $t$ interval. Red cells are not run yet or rest on a block that is still running. Analysis of the finished runs.}
\begin{tabular}{@{}lccc@{\hspace{8pt}}ccc@{}}
\toprule
& \multicolumn{3}{c}{\burn} & \multicolumn{3}{c}{\flair} \\
\cmidrule(lr){2-4}\cmidrule(l){5-7}
Model & \Lone & \Ltwo & \Lpair & \Lone & \Ltwo & \Lpair \\
\midrule
\sat & 3.8 & 17.8 & 22.2 & 2.1 & 6.5 & 10.1 \\
\lvd & 3.8 & 14.1 & 21.3 & 0.8 & 3.6 & 6.4 \\
\dofa & 4.1 & 17.9 & 24.0 & 1.0 & 4.3 & 7.9 \\
\tm & 3.7 & 12.9 & 21.6 & 1.6 & 5.2 & 7.7 \\
\rn & 3.2 & 15.9 & 22.4 & 0.9 & 4.2 & 7.3 \\
\midrule
Mean over models & 3.7 & 15.7 & 22.3 & 1.3 & 4.8 & 7.9 \\
\addlinespace[1.5pt]
\Lpair{} $-$ \Ltwo & \multicolumn{3}{c}{6.6 [5.3, 7.9]} & \multicolumn{3}{c}{3.1 [2.8, 3.4]} \\
N6 at \Lpair & \multicolumn{3}{c}{$-$0.9 [$-$5.0, 3.2]} & \multicolumn{3}{c}{2.2 [1.2, 3.2]} \\
F2 at \Lpair & \multicolumn{3}{c}{0.4 [$-$2.2, 3.0]} & \multicolumn{3}{c}{0.5 [$-$2.2, 3.3]} \\
\bottomrule
\end{tabular}
\end{adjustbox}}{\tabletbd[3cm]{planned/generated/table\_n12.tex}}
\caption{\textbf{Drop at the three levels:} Mean paired drop over the four types in points per core model at \Lone, \Ltwo and the pair level \Lpair{} on both core datasets. The \Lone and \Ltwo columns are main-grid cells on seeds 1 to 5, and the \Lpair column is block N12. The last rows give contrasts with their \CiLevel\,\% intervals in brackets. \ProvSlot{\ProvNTwelve}}
\label{tab:n12}
\end{table}

\ifdefined\cvprbuild\else
\begin{figure}[t]
\centering
\IfFileExists{planned/figs/fig_n12_lr.pdf}{\includegraphics[width=\linewidth]{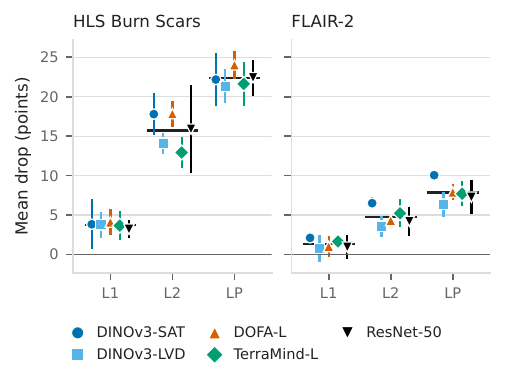}}{\phfig{4cm}{planned/figs/fig\_n12\_lr.pdf}}
\caption{\textbf{Drop at \Lone, \Ltwo and \Lpair{} per model:} Mean paired drop over the four types in points per core model with \CiLevel\,\% intervals over seeds. The three levels lie along the horizontal axis, with one marker per model. The left panel shows \burn and the right panel \flair. A black bar marks the mean over models. \ProvSlot{\ProvNTwelve}}
\label{fig:n12}
\end{figure}

\fi
\paragraph{Pair Level (N12):}
The level \Lpair{} targets a label \miou of \LrTargetMiou, the median over \LrTargetNPairs{} measured pairs.
At \Lpair{} the mean drop over models and types is \hlLrMeanDropburn{} points on \burn and \hlLrMeanDropflair{} on \flair \ifdefined\cvprbuild (\Cref{tab:n12})\else (\Cref{tab:n12}, \Cref{fig:n12})\fi{}.
The type ranking at \Lpair{} puts \etype{\hlLrTypeRankburnOne} first on \burn and \etype{\hlLrTypeRankflairOne} first on \flair. H9 reads \vd{\hlLrRankVerdictburn} and \vd{\hlLrRankVerdictflair}.
The drop at \Lpair{} minus the drop at \Ltwo reads \vd{\hlLrVerdictburn} and \vd{\hlLrVerdictflair}.
N6 at \Lpair{} reads \vd{\hlLrVerdictNSixburn} and \vd{\hlLrVerdictNSixflair}. F2 at \Lpair{} reads \vd{\hlLrVerdictFTwoburn} and \vd{\hlLrVerdictFTwoflair}.

\paragraph{Seed Extension of F2 (N13):}
At seeds 1 to 5 the F2 contrast on \burn resolves \hlFTwoExtFiveburnMde{} points and reads \vd{\hlFTwoExtFiveVerdictburn}.
With seeds 1 to 8 of every core model it is \hlFTwoExtburnEst{} [\hlFTwoExtburnCILo, \hlFTwoExtburnCIHi] points.
Its 90\,\% equivalence interval is [\hlFTwoExtburnTostCILo, \hlFTwoExtburnTostCIHi]. The test resolves \hlFTwoExtburnMde{} points and reads \vd{\hlFTwoExtVerdictburn}.

\ifdefined\cvprbuild\fi

\paragraph{Noise Robust Losses (N14):}
\ifdefined\cvprbuild\Cref{tab:n14}\else\mtab{tab:n14}\fi{} lists the clean \miou and the mean \Ltwo drop under each loss. %
The 90\,\% equivalence intervals of the generalized cross-entropy contrast are [\hlLossInvGceburnTostCILo, \hlLossInvGceburnTostCIHi] on \burn and [\hlLossInvGceflairTostCILo, \hlLossInvGceflairTostCIHi] on \flair.
For soft bootstrapping they are [\hlLossInvBootburnTostCILo, \hlLossInvBootburnTostCIHi] and [\hlLossInvBootflairTostCILo, \hlLossInvBootflairTostCIHi].
The four tests resolve \hlLossInvGceburnMde{} and \hlLossInvGceflairMde{} points for the first loss and \hlLossInvBootburnMde{} and \hlLossInvBootflairMde{} for the second.
\section{Extension Datasets and Real Label Pairs (D1 to D7)}
\label{sec:supp-ext}

\mtab{tab:ext-datasets} lists the seven extension datasets, and \Cref{tab:ext-results} and \Cref{fig:ext-real} their results. %
CloudSEN12, SpaceNet 2 and Kuro Siwo use the \gbtwo splits, and Chesapeake and EnviroAtlas draw fixed chips from the release tiles.
DFC2020 trains on its validation patches and tests on its test patches, and OpenEarthMap cuts the official split images into tiles.
Kuro Siwo has no optical bands, so the DINOv3 pair reads a false-color composite of the two SAR polarizations there.

\begin{figure*}[t]
\centering
\IfFileExists{planned/figs/fig_ext_real_noise.pdf}{\includegraphics[width=\textwidth]{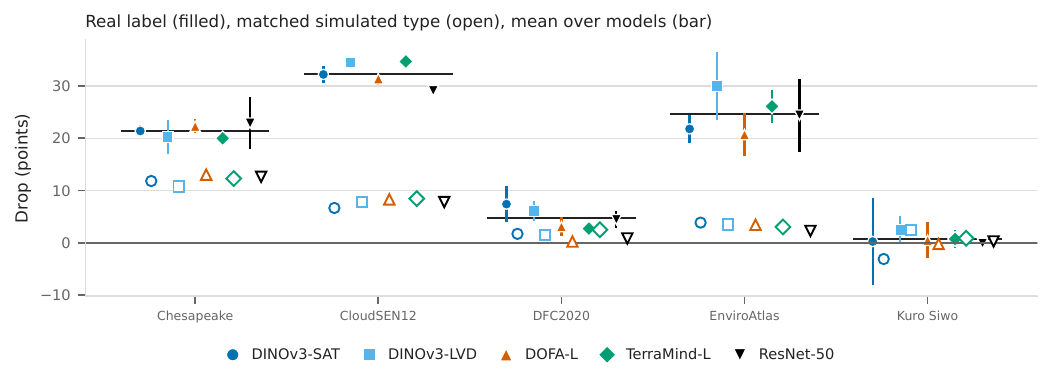}}{\phfig{4cm}{planned/figs/fig\_ext\_real\_noise.pdf from analysis/planned\_placeholders.py}}
\caption{\textbf{Real against matched simulated labels on the five pair datasets:} Paired drop in points under the real label (filled) and under its matched simulated type (open). Filled markers carry \CiLevel\,\% intervals over seeds, and a short black bar marks their mean over models.}
\label{fig:ext-real}
\end{figure*}

\begin{table*}[t]
\centering\footnotesize\setlength{\tabcolsep}{2.5pt}
\IfFileExists{planned/generated/table_ext_results.tex}{\begin{adjustbox}{max width=\textwidth}%
\providecommand{\gcvdry}[1]{\textcolor{red}{#1}}
\providecommand{\ProvTableExtResults}{Seeds per cell 5. Clean test mIoU, the mean paired drop over the four \Ltwo types and the paired drop under the dataset's real labels, in points, every noisy run paired with the clean run of its own setup (the real-label arm of DFC2020 and Kuro Siwo with the clean run on the tiles the real labels cover). Red cells are not run yet or rest on a sub-block that is still running. Analysis of the finished runs.}
\begin{tabular}{@{}llccccccccccccccc@{}}
\toprule
& & \multicolumn{3}{c}{\sat} & \multicolumn{3}{c}{\lvd} & \multicolumn{3}{c}{\dofa} & \multicolumn{3}{c}{\tm} & \multicolumn{3}{c}{\rn} \\
\cmidrule(lr){3-5} \cmidrule(lr){6-8} \cmidrule(lr){9-11} \cmidrule(lr){12-14} \cmidrule(lr){15-17}
Block & Dataset & Clean & Drop \Ltwo & Drop real & Clean & Drop \Ltwo & Drop real & Clean & Drop \Ltwo & Drop real & Clean & Drop \Ltwo & Drop real & Clean & Drop \Ltwo & Drop real \\
\midrule
D1 & Chesapeake & 81.3 & 10.2 & 21.4 & 79.3 & 10.2 & 20.3 & 83.0 & 12.0 & 22.3 & 82.0 & 9.9 & 20.0 & 82.5 & 8.7 & 22.9 \\
D2 & CloudSEN12 & 66.6 & 5.9 & 32.2 & 69.6 & 6.4 & 34.5 & 65.9 & 7.2 & 31.4 & 69.5 & 7.2 & 34.7 & 64.0 & 6.8 & 29.1 \\
D3 & DFC2020 & 40.5 & 2.9 & 7.4 & 43.2 & 2.5 & 6.1 & 36.5 & 1.4 & 3.1 & 38.9 & 3.4 & 2.7 & 38.8 & 1.2 & 4.5 \\
D4 & EnviroAtlas & 61.8 & 6.3 & 21.8 & 63.0 & 7.7 & 30.0 & 48.1 & 5.4 & 20.7 & 49.0 & 6.9 & 26.1 & 49.1 & 12.1 & 24.4 \\
D5 & SpaceNet 2 & 85.8 & 11.4 & -- & 86.0 & 10.4 & -- & 84.2 & 11.8 & -- & 83.4 & 11.7 & -- & 81.7 & 13.8 & -- \\
D6 & Kuro Siwo & 53.0 & 6.5 & 0.3 & 59.8 & 7.4 & 2.4 & 53.5 & 5.1 & 0.5 & 58.0 & 7.7 & 0.7 & 32.5 & 0.5 & $-$0.1 \\
D7 & OpenEarthMap & 63.5 & 7.6 & -- & 63.4 & 6.7 & -- & 61.0 & 6.8 & -- & 55.2 & 6.0 & -- & 55.6 & 5.4 & -- \\
\bottomrule
\end{tabular}
\end{adjustbox}}{\tabletbd[3cm]{planned/generated/table\_ext\_results.tex: run analysis/planned\_placeholders.py}}
\caption{\textbf{Main grid and real labels on the extension datasets:} Clean test \miou, mean paired drop over the four \Ltwo types and drop under the real label of the pair. Values are in points for every core model and dataset. A dash marks a dataset without a real pair. \ProvSlot{\ProvExtension}}
\label{tab:ext-results}
\end{table*}

\paragraph{Sub Blocks:}
Each extension dataset runs up to four sub-blocks with the \gbtwo recipe and the UNet decoder unless stated.
The core sub-block repeats the main grid on that dataset over five seeds.
The real sub-block trains the five models on the real label of the pair at every seed and exists on the five pair datasets only.
At the clean condition and \Ltwo the matched-rate sub-block repeats the DINOv3 pair at the swapped multipliers.
The decoder sub-block repeats the DINOv3 pair under the SegFormer, Mask2Former and SAM heads at the clean condition and \Ltwo.
The seven blocks hold \RunsExtensionTotal{} runs.
Chesapeake holds \RunsDOne, CloudSEN12 \RunsDTwo, DFC2020 \RunsDThree{} and EnviroAtlas \RunsDFour{} runs.
SpaceNet 2 holds \RunsDFive, Kuro Siwo \RunsDSix{} and OpenEarthMap \RunsDSeven{} runs.

\paragraph{Real Labels:}
The real label of a pair is the NLCD, s2cloudless, MODIS or CEMS layer.
It is aligned to the tile grid by nearest-neighbor reprojection or rasterized from polygons.
We then score it against the reference label on the training and validation tiles.
It is packed as one mask file with one realization, because a real label has no seed.
On DFC2020 and Kuro Siwo the real label covers a part of the tiles only.
The real arm and a clean baseline of its own therefore train and validate on the covered tiles there.
The real arm trains \RunsRealDOne{} runs on Chesapeake, \RunsRealDTwo{} on CloudSEN12 and \RunsRealDThree{} on DFC2020. It trains \RunsRealDFour{} on EnviroAtlas and \RunsRealDSix{} on Kuro Siwo.

\paragraph{Real Against Simulated Drop:}
The matched arm of every pair is the \etype{coarse} type at the level nearest to the measured label \miou, \hlRealMatchedLevelChes{} on Chesapeake.
On Chesapeake the real minus the matched simulated drop is \hlRealVsSynthChes{} [\hlRealVsSynthChesCILo, \hlRealVsSynthChesCIHi] points, with a flip rate of \RealFlipRateChes\,\% for the real label.
\Cref{fig:ext-real} shows the five pairs side by side.

\ifdefined\cvprbuild\else
\begin{figure*}[t]
\centering
\IfFileExists{planned/figs/fig_ext_synthetic.pdf}{\includegraphics[width=\textwidth]{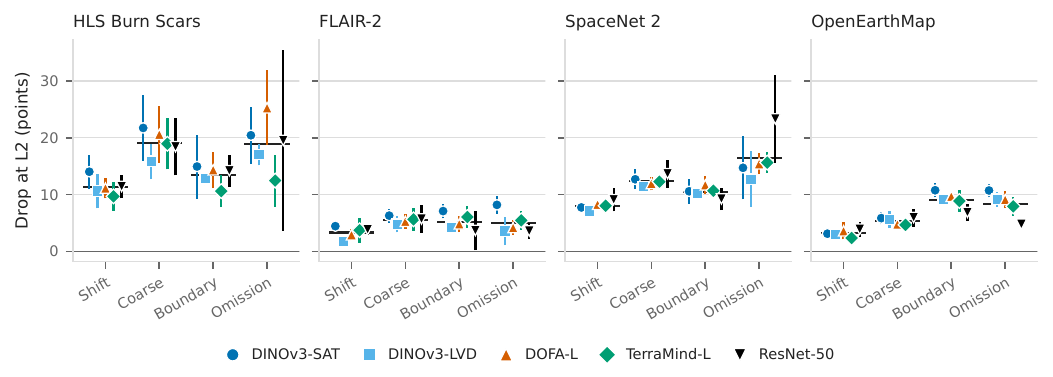}}{\phfig{4cm}{planned/figs/fig\_ext\_synthetic.pdf}}
\caption{\textbf{Error types on the core datasets and the two extension datasets without a pair:} Mean paired drop at \Ltwo per type and core model with \CiLevel\,\% intervals on \burn, \flair, SpaceNet 2 and OpenEarthMap. A short black bar marks the mean over models. \ProvSlot{\ProvExtension}}
\label{fig:ext-synth}
\end{figure*}

\fi
\paragraph{Verdicts per Dataset:}
The matched-rate RQ1 contrast reads \vd{\hlExtVerdictNSixChes} on Chesapeake, \vd{\hlExtVerdictNSixCsTwelve} on CloudSEN12, \vd{\hlExtVerdictNSixDfc} on DFC2020 and \vd{\hlExtVerdictNSixEnva} on EnviroAtlas.
It reads \vd{\hlExtVerdictNSixSnTwo} on SpaceNet 2, \vd{\hlExtVerdictNSixKuro} on Kuro Siwo and \vd{\hlExtVerdictNSixOem} on OpenEarthMap.
F2 reads \vd{\hlExtVerdictFTwoChes}, \vd{\hlExtVerdictFTwoCsTwelve}, \vd{\hlExtVerdictFTwoDfc} and \vd{\hlExtVerdictFTwoEnva} on the first four datasets.
On the last three it reads \vd{\hlExtVerdictFTwoSnTwo}, \vd{\hlExtVerdictFTwoKuro} and \vd{\hlExtVerdictFTwoOem}.
On Kuro Siwo \rn reaches a clean \miou far below the other models (\Cref{tab:ext-results}).
Several of its noisy cells sit at one constant value, so its small drop is a floor effect.
The decoder contrast of the DINOv3 pair reads \vd{\hlExtVerdictHFiveChes}, \vd{\hlExtVerdictHFiveCsTwelve}, \vd{\hlExtVerdictHFiveDfc} and \vd{\hlExtVerdictHFiveEnva} on the first four datasets.
On the last three it reads \vd{\hlExtVerdictHFiveSnTwo}, \vd{\hlExtVerdictHFiveKuro} and \vd{\hlExtVerdictHFiveOem}.
\ifdefined\cvprbuild\else
\Cref{fig:ext-synth} shows the \Ltwo drop per error type on the core datasets and the two extension datasets without a pair.
\fi

\begin{table*}[t]
\centering\footnotesize\setlength{\tabcolsep}{3pt}
\IfFileExists{planned/generated/table_hw_blocks.tex}{\begin{adjustbox}{max width=\textwidth}%
\begin{tabular}{@{}lrrrrrrrrrr@{}}
\toprule
Block & H100 NVL & RTX PRO 6000 & RTX 4090 & GB10 & RTX 5060 Ti & RTX 3080 Ti & RTX 3080 & Runs & Paired drops & Mixed pairs \\
\midrule
N7 & 392 & 23 & 1 & -- & -- & 39 & 95 & 550 & 480 & 40 \\
N8 & 59 & 60 & -- & -- & -- & 4 & 71 & 194 & 156 & 8 \\
N9 & 388 & 7 & -- & 12 & 43 & 64 & 36 & 550 & 480 & 43 \\
N10 & 356 & 2 & -- & 21 & 110 & 40 & 21 & 550 & 480 & 25 \\
N11 & 72 & 36 & -- & 111 & 259 & 72 & -- & 550 & 480 & 2 \\
N12 & 68 & 116 & -- & 32 & 16 & 48 & -- & 280 & 280 & 228 \\
N13 & 15 & -- & -- & -- & 20 & -- & 10 & 45 & 36 & 0 \\
N14 & 315 & 35 & -- & 5 & 15 & 60 & 70 & 500 & 400 & 0 \\
N15 & 188 & 29 & -- & 237 & 60 & 36 & -- & 550 & 480 & 0 \\
D1 & 390 & -- & -- & 10 & -- & 50 & -- & 450 & 385 & 0 \\
D2 & 400 & -- & -- & -- & -- & 50 & -- & 450 & 385 & 0 \\
D3 & 418 & -- & -- & -- & 2 & 44 & 11 & 475 & 385 & 0 \\
D4 & 390 & -- & -- & 10 & -- & 50 & -- & 450 & 385 & 0 \\
D5 & 355 & -- & -- & 25 & -- & 45 & -- & 425 & 360 & 0 \\
D6 & 305 & 15 & -- & 30 & 18 & 90 & 17 & 475 & 385 & 1 \\
D7 & 380 & -- & -- & -- & -- & 44 & 1 & 425 & 360 & 1 \\
\midrule
All & 4,491 & 323 & 1 & 493 & 543 & 736 & 332 & 6,919 & 5,917 & 348 \\
\bottomrule
\end{tabular}
\end{adjustbox}}{\tabletbd[3cm]{planned/generated/table\_hw\_blocks.tex: run scripts/make\_block\_hardware.py}}
\caption{\textbf{Finished runs per block and GPU class:} Runs of each block N7 to N15 and D1 to D7 by the GPU class that trained them. Paired drops counts the noisy runs with a clean baseline, and mixed pairs counts those whose clean run used another GPU class. A dash marks a class that ran none of the block.}
\label{tab:supp-hw-blocks}
\end{table*}

\paragraph{Hardware of the Later Blocks:}
Blocks N7 to N15 and D1 to D7 ran on a fleet of seven GPU classes that moved queued jobs between machines (\Cref{tab:supp-hw-blocks}).
The scheduler aims to keep the clean run and the noisy runs of one model, dataset and seed on one class.
\Cref{tab:supp-hw-blocks} counts the pairs that mix classes.
A pair that still mixes classes carries a hardware term, which \Cref{tab:supp-hwrep} measures for the N6 cells.

\section{Conditions of the Error Type Advice}
\label{sec:supp-advice}

The advice of \msec{sec:discussion} holds for region-type targets, the amorphous stuff classes \cite{kirillov2019panoptic} judged by area overlap. %
It excludes built-up classes and holds under six conditions.
The \etype{shift} type costs least in most non-built \flair classes \ifdefined\cvprbuild on point estimates\else on point estimates (\Cref{tab:supp-perclass-drop})\fi{}.
First, the \etype{shift} operator is a zero-mean random-direction offset per tile, and its one-sided variant still costs least at \Ltwo (\Cref{tab:supp-variants}).

Second, the ranking depends on the metric.
Under boundary \miou at the \BoundaryDPxBurn\,px band \etype{omission} is least harmful on both datasets on point estimates, separated from \etype{shift} on \flair only.
A product scored by boundary \miou at a wide band should weigh registration higher, on the \flair evidence (\Cref{sec:supp-boundary}).

Third, the ranking depends on the target.
On \flair buildings and impervious surfaces lose the most class IoU under \etype{coarse} and more under \etype{shift} \ifdefined\cvprbuild than under \etype{omission}\else than under \etype{omission} (\Cref{tab:supp-perclass-drop})\fi{}.
For built objects only the completeness part of the advice fails.
Fourth, \etype{omission} removes objects regardless of size, whereas rushed annotation misses small objects first.
A class-weighted loss (S2), a rare-class proxy, changes the \flair \etype{omission} drop by \hlSTwoDeltaSATflairEst{} [\hlSTwoDeltaSATflairCILo, \hlSTwoDeltaSATflairCIHi] points for \sat.
For \rn the change is \hlSTwoDeltaRNflairEst{} [\hlSTwoDeltaRNflairCILo, \hlSTwoDeltaRNflairCIHi] points. Neither change is detected.
Fifth, the \burn offsets and coarse blocks are stress tests, and \etype{shift} also ranks last at \Lone on both datasets.
Sixth, the advice assumes equal cost per type, whereas per-tile offsets are the type most amenable to automatic correction \cite{girard2019noisy,vargasmunoz2019correcting}.
Block S1 adds one real label error on a further dataset, Otsu water masks on \sfloods from a Sentinel-1 threshold.
They cost \hlSOneDropMeansenEst{} [\hlSOneDropMeansenCILo, \hlSOneDropMeansenCIHi] points at the \Ltwo target, below the \burn \Ltwo type means (\Cref{tab:supp-sen1floods}).
\fi
\section{Freezing and the Labeling Budget}
\label{sec:supp-budget}

\ifdefined\cvprbuild\else
\begin{figure}[t]
\centering
\IfFileExists{figs/fig_budget.pdf}{\includegraphics[width=\linewidth]{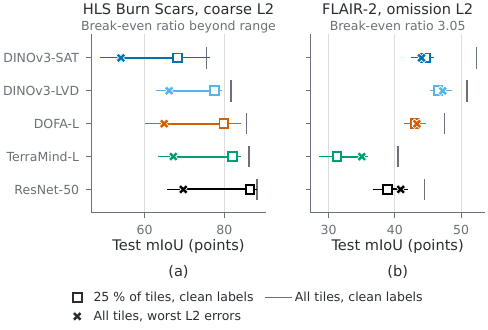}}{\phfig{5.5cm}{figs/fig\_budget.pdf from figs/make\_budget\_fig.py}}
\caption{\textbf{A clean subset against all tiles with errors:} Test \miou of each core model after training on \SubsetPct\,\% of the tiles with clean labels (open square). A cross marks the same model trained on all tiles with the selected \Ltwo error type. A line joins the two arms of a model. Bars are the \CiLevel\,\% $t$ interval over the seeds of the cell, \statFOneburnN{} for the DINOv3 noisy arm and \NSeeds{} otherwise. The gray tick is all tiles with clean labels. (a) \burn with \etype{coarse} and (b) \flair with \etype{omission}. The break-even cost ratio is the price ratio of clean to noisy labels at which the two arms tie. It lies beyond the measured range where the clean subset already wins.}
\label{fig:supp-budget}
\end{figure}

\begin{figure}[t]
\centering
\IfFileExists{figs/frozen_vs_full.pdf}{\includegraphics[width=\linewidth]{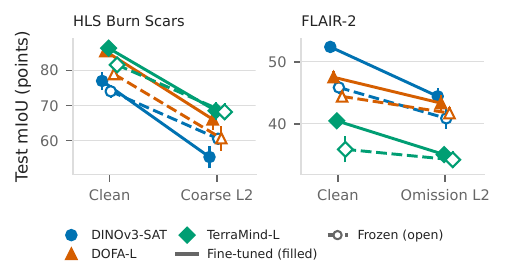}}{\phfig{3.0cm}{figs/frozen\_vs\_full.pdf}}
\caption{\textbf{Frozen against fine-tuned encoders at the selected \Ltwo type:} Clean and noisy test \miou of frozen (open markers) and fine-tuned (filled markers) \sat, \dofa and \tm. The left panel shows \burn with \etype{coarse} and the right panel \flair with \etype{omission}. Bars are the \CiLevel\,\% $t$ interval over the seeds of the cell.}
\label{fig:supp-frozen}
\end{figure}

\fi
\begin{table}[t]
\centering\footnotesize\setlength{\tabcolsep}{2.5pt}
\IfFileExists{generated/table_frozen.tex}{%
\providecommand{\gcvdry}[1]{\textcolor{red}{#1}}
\begin{tabular}{@{}lcccccccc@{}}
\toprule
 & \multicolumn{4}{c}{\burn} & \multicolumn{4}{c}{\flair} \\
\cmidrule(lr){2-5}\cmidrule(l){6-9}
 & \multicolumn{2}{c}{Clean} & \multicolumn{2}{c}{\etype{Coarse} \Ltwo} & \multicolumn{2}{c}{Clean} & \multicolumn{2}{c}{\etype{Omission} \Ltwo} \\
\cmidrule(lr){2-3}\cmidrule(lr){4-5}\cmidrule(lr){6-7}\cmidrule(l){8-9}
Model & FT & Frozen & FT & Frozen & FT & Frozen & FT & Frozen \\
\midrule
\sat & 77.0 & 74.0 & 55.3 & 60.6 & \textbf{52.4} & \textbf{45.9} & \textbf{44.4} & 40.9 \\
\dofa & 85.5 & 79.0 & 66.1 & 60.7 & 47.5 & 44.4 & 43.3 & \textbf{41.7} \\
\tm & \textbf{86.3} & \textbf{81.6} & \textbf{68.5} & \textbf{68.2} & 40.5 & 35.9 & 35.0 & 34.2 \\
\bottomrule
\end{tabular}
}{\tabletbd[2.2cm]{generated/table\_frozen.tex: run figs/make\_results\_tables.py}}
\caption{\textbf{Frozen against fine-tuned encoders:} Test \miou of the A3 models with the encoder fine-tuned (FT) and frozen, on clean labels and at the selected \Ltwo type of each dataset. Values are means over the seeds of the cell. On \burn the fine-tuned cells average \statFOneburnN{} seeds from two GPU classes (\Cref{sec:supp-hardware}), as does the fine-tuned \sat cell on \flair. F3 uses the \NSeeds{} common seeds on one class. Absolute values carry the reference-label error named in the main paper. Bold marks the best value of a column. \ProvSlot{\ResultsProvenance}}
\label{tab:supp-frozen}
\end{table}

\begin{table}[t]
\centering\footnotesize\setlength{\tabcolsep}{3pt}
\IfFileExists{generated/table_budget.tex}{%
\providecommand{\gcvdry}[1]{\textcolor{red}{#1}}
\begin{tabular}{@{}lcccccc@{}}
\toprule
 & \multicolumn{3}{c}{\burn, \etype{coarse} \Ltwo} & \multicolumn{3}{c}{\flair, \etype{omission} \Ltwo} \\
\cmidrule(lr){2-4}\cmidrule(l){5-7}
Model & Subset & Noisy & Clean & Subset & Noisy & Clean \\
\midrule
\sat & 68.3 & 55.3 & 77.0 & 44.6 & 44.4 & \textbf{52.4} \\
\lvd & 77.5 & 66.4 & 81.8 & \textbf{46.5} & \textbf{47.3} & 50.7 \\
\dofa & 79.9 & 66.1 & 85.5 & 43.1 & 43.3 & 47.5 \\
\tm & 82.0 & 68.5 & 86.3 & 31.3 & 35.0 & 40.5 \\
\rn & \textbf{86.4} & \textbf{70.5} & \textbf{88.3} & 38.9 & 40.9 & 44.5 \\
\bottomrule
\end{tabular}
}{\tabletbd[2.6cm]{generated/table\_budget.tex: run figs/make\_results\_tables.py}}
\caption{\textbf{A clean subset against all tiles with errors:} Test \miou of each core model after training on \SubsetPct\,\% of the tiles with clean labels (Subset). The other columns train on all tiles with the selected \Ltwo error type (Noisy) and on all tiles with clean labels (Clean). Values are means over the seeds of the cell. On \burn the Noisy and Clean cells average \statFOneburnN{} seeds from two GPU classes, as do those of the DINOv3 pair on \flair. F4 uses the \NSeeds{} common seeds on one class. Absolute values carry the reference-label error named in the main paper. Bold marks the best value of a column\ifdefined\cvprbuild\else , and the break-even cost ratios per model and seed are in \Cref{tab:supp-breakeven}\fi{}. \ProvSlot{\ResultsProvenance}}
\label{tab:supp-budget}
\end{table}

\ifdefined\cvprbuild\else
\begin{table}[t]
\centering\footnotesize\setlength{\tabcolsep}{4pt}
\IfFileExists{generated/table_break_even.tex}{%
\providecommand{\gcvdry}[1]{\textcolor{red}{#1}}
\providecommand{\ProvBreakEven}{Break-even clean to noisy cost ratio, one decimal. Per model from the cell means of that model, per seed from the model means within that seed, each by two-point log-linear interpolation of clean \miou between the \SubsetPct\,\% subset and the full set. $>$4.0 marks a subset arm that already beats the full noisy set. Analysis of the finished runs.}
\begin{tabular}{@{}lcc@{}}
\toprule
Unit & \burn & \flair \\
\midrule
\sat & $>$4.0 & $>$4.0 \\
\lvd & $>$4.0 & 3.1 \\
\dofa & $>$4.0 & 3.7 \\
\tm & $>$4.0 & 2.3 \\
\rn & $>$4.0 & 2.4 \\
\addlinespace[2pt]
Seed 1 & $>$4.0 & 2.8 \\
Seed 2 & $>$4.0 & 3.9 \\
Seed 3 & $>$4.0 & 2.5 \\
Seed 4 & $>$4.0 & 2.8 \\
Seed 5 & $>$4.0 & 3.5 \\
\bottomrule
\end{tabular}
}{\tabletbd[2.6cm]{generated/table\_break\_even.tex: run the analysis harness}}
\caption{\textbf{Break-even cost ratio per model and per seed:} The clean to noisy price ratio at which \SubsetPct\,\% clean tiles and all tiles with the selected \Ltwo errors tie. It comes from a two-point log-linear interpolation of clean \miou between the subset and the full set. Per model from that model's cell means, per seed from the model means within the seed. $>$4.0 marks a subset arm that already beats the full noisy set. \ProvSlot{\ResultsProvenance}}
\label{tab:supp-breakeven}
\end{table}

\fi
The frozen clean \miou is below the fine-tuned value for every model on both datasets \ifdefined\cvprbuild (\Cref{tab:supp-frozen})\else (\Cref{fig:supp-frozen}, \Cref{tab:supp-frozen})\fi{}.
In absolute noisy-label \miou the frozen encoder is ahead of or not separated from the fine-tuned one for \sat and \tm on \burn.
The values are \miouSATburnCoarseLTwoVarFrozenEncoderTrue{} against \miouSATburnCoarseLTwo{} and \miouTMburnCoarseLTwoVarFrozenEncoderTrue{} against \miouTMburnCoarseLTwo. These cell means average eight seeds for the fine-tuned arm, and \Cref{sec:supp-stats} gives the paired contrasts over the five common seeds.
The \sat fine-tuned arm runs at the \gbtwo rate. Against its matched-rate arm (\miouSATburnCoarseLTwoVarLrMultZeroThreeThreeThreeThree) the frozen \sat differs by \hlVeFrozenSATNSixburnEst{} [\hlVeFrozenSATNSixburnCILo, \hlVeFrozenSATNSixburnCIHi] points, paired over seeds 1 to 5, and on \flair by \hlVeFrozenSATNSixflairEst{} [\hlVeFrozenSATNSixflairCILo, \hlVeFrozenSATNSixflairCIHi].

\Cref{tab:supp-budget} gives the test \miou per model of the two F4 arms and of all tiles with clean labels.
The budget comparison carries four biases.
The milder \etype{omission} realizations on \flair help the noisy arm.
Clean validation labels help the clean-subset arm, and fewer gradient steps at equal epochs hurt it.
The selected type can also inflate the drop of the noisy arm.
The comparison \ifdefined\cvprbuild (\Cref{tab:supp-budget})\else (\Cref{fig:supp-budget} and \Cref{tab:supp-budget})\fi{} is therefore a point estimate of the value of clean labels rather than a bound.
A per-type reading uses the model means \ifdefined\cvprbuild of the main grid and \Cref{tab:supp-budget}\else of \Cref{tab:supp-grid-burn,tab:supp-grid-flair,tab:supp-budget}\fi{}. The clean subset trails all noisy tiles at \Lone for every type except \etype{omission} on \burn. It leads at \Ltwo for every type on \burn only.
At the selected type and \Lone, F4 is \hlVeFFourLOneburnEst{} [\hlVeFFourLOneburnCILo, \hlVeFFourLOneburnCIHi] points on \burn and \hlVeFFourLOneflairEst{} [\hlVeFFourLOneflairCILo, \hlVeFFourLOneflairCIHi] on \flair over the five core models and seeds 1 to 5.

\ifdefined\cvprbuild\else
\Cref{tab:supp-breakeven} gives the break-even ratio per model and per seed.
On \burn the subset arm beats the full noisy set for every model and seed, so the ratio exceeds \CostRatio{} throughout.
On \flair it spans \hlBreakEvenModelMinflair{} to \hlBreakEvenModelMaxflair{} over models and \hlBreakEvenSeedMinflair{} to \hlBreakEvenSeedMaxflair{} over seeds, and the mean-over-models value is \hlBreakEvenflair.
The ratio answers the procurement case in which both label sources are bought.
Where noisy labels already exist, the decision variable is the absolute cost of the clean sample against the F4 estimate.
A hybrid arm of all noisy tiles plus the clean sample remains open.

\fi
\section{Ranking at a Common Flip Rate}
\label{sec:supp-fliprate}

At equal label \miou the flip rates of the four types differ (\Cref{tab:supp-calibration}).
On \burn \etype{coarse} and \etype{omission} cost the most yet change the fewest pixels, \FlipBurnCoarseLTwo\,\% and \FlipBurnOmissionLTwo\,\% against \FlipBurnShiftLTwo\,\% for \etype{shift}.
On \flair \etype{shift} changes as many pixels as \etype{coarse} (\FlipFlairShiftLTwo\,\% against \FlipFlairCoarseLTwo\,\%) yet costs the least.
A ranking at a common flip rate can therefore only widen the gap between \etype{coarse} or \etype{omission} and \etype{shift}.
Each drop is interpolated linearly to the smallest \Ltwo flip rate, \hlFlipRankCommonPctburn\,\% on \burn and \hlFlipRankCommonPctflair\,\% on \flair.
Flip rates here are means over realizations 1 to 5, so they differ from the realization-\CalSeed{} values of \Cref{tab:supp-calibration}.
The ranking is then \hlFlipRankburn{} on \burn and \hlFlipRankflair{} on \flair.
On both datasets \etype{omission} thus moves ahead of \etype{coarse}, and \etype{shift} stays last.

\section{Statistical Details}
\label{sec:supp-stats}

The Welch $t$ test drops the seed pairing of each family. It agrees with the paired test for every family and dataset of the verdict table.
Its unadjusted $p$ values are \statFOneburnWelchP, \statFTwoburnWelchP, \statFThreeburnWelchP{} and \statFFourburnWelchP{} for F1 to F4 on \burn and \statFOneflairWelchP, \statFTwoflairWelchP, \statFThreeflairWelchP{} and \statFFourflairWelchP{} on \flair.
No verdict therefore depends on the pairing.

The linear mixed model of the \Ltwo drop has model, condition and their interaction as fixed effects and seed as a random effect \cite{seabold2010statsmodels}.
On \burn its condition effects against \etype{boundary} are \hlMixedCoefburnCoarse{} for \etype{coarse} (standard error \hlMixedSEburnCoarse), \hlMixedCoefburnOmission{} for \etype{omission} (\hlMixedSEburnOmission) and \hlMixedCoefburnShift{} for \etype{shift} (\hlMixedSEburnShift) points.
The fit includes the extension encoders, so its reference model is \cnxl.
Under the interaction these are effects at that reference model, so they do not equal the mean differences of \Cref{tab:supp-main}.
On \flair the fit is degenerate with non-finite standard errors. The paired contrasts are primary on both datasets.

Three quantities are specified here.
The clean deficit of the frozen encoders in F3 is reported descriptively.
For F4, the break-even cost ratio is the inverse of the clean fraction at which clean labels match the noisy \miou.
That fraction comes from log-linear interpolation between the subset and the full set.
The ranking of types at a common flip rate interpolates each drop linearly along the flip rate to the smallest \Ltwo flip rate.
This extrapolates below \Lone for a type whose \Lone flip rate exceeds that value.

The minimum detectable effects are computed at \MdePower\,\% power and two-sided $\alpha = \MdeAlpha$ with the seeds of each family, \statFTwoburnN{} for F2 on \burn.
They are \statFOneburnMde, \statFTwoburnMde, \statFThreeburnMde{} and \statFFourburnMde{} points for F1 to F4 on \burn and \statFOneflairMde, \statFTwoflairMde, \statFThreeflairMde{} and \statFFourflairMde{} points on \flair.
The N6 contrast has a minimum detectable effect of \hlVeNSixMdeburnEst{} points on \burn and \hlVeNSixMdeflairEst{} on \flair.
The N5 rule estimates the F1 minimum detectable effect at five seeds from the standard deviation of the contrast.
It reads no effect estimate, so it leaves the type I error of F1 close to nominal.
The plan set its look at the first \NFiveSeedsAt{} seeds. There the value is \hlVeNFiveMdeThreeburnEst{} on \burn and \hlVeNFiveMdeThreeflairEst{} on \flair, both below the \NFiveMde-point threshold.
The look was taken when seeds 1 to 5 were complete, where the value is \hlVeNFiveMdeFiveburnEst{} on \burn and \hlVeNFiveMdeFiveflairEst{} on \flair.
The \burn value exceeded the threshold, so seeds \NFiveSeeds{} of the DINOv3 pair were added on both datasets. This deviates from the planned look.
These seeds ran on a second GPU class, while both arms of each seed share one class (\Cref{tab:supp-hwmix}).
F1 now holds \statFOneburnN{} seeds on \burn and \statFOneflairN{} on \flair.
On seeds 1 to 5 alone it is \hlVeFOneFiveburnEst{} [\hlVeFOneFiveburnCILo, \hlVeFOneFiveburnCIHi] points on \burn and \hlVeFOneFiveflairEst{} [\hlVeFOneFiveflairCILo, \hlVeFOneFiveflairCIHi] on \flair.
Their unadjusted $p$ values are \hlVeFOneFiveburnP{} and \hlVeFOneFiveflairP.
An exact sign-flip test over the eight seeds gives $p = \hlVeFOneSignFlipburnP$ on \burn and \hlVeFOneSignFlipflairP{} on \flair.
At five seeds that test cannot go below $p = \hlVeFOneSignFlipburnFloorFive$.

Further contrasts are paired over all common seeds with \CiLevel\,\% intervals, outside any Holm family.
The 90\,\% interval of the N6 contrast is [\hlVeNSixTostburnCILo, \hlVeNSixTostburnCIHi] on \burn and [\hlVeNSixTostflairCILo, \hlVeNSixTostflairCIHi] on \flair. Both lie within the $\pm$\TostMargin-point margin of F1, an exploratory reading outside the plan.
The N6 cells carry the hardware term of \Cref{tab:supp-hwmix}.
In the type-averaged \Ltwo drop, \dofa minus \rn is \hlVeDOFARNburnEst{} [\hlVeDOFARNburnCILo, \hlVeDOFARNburnCIHi] on \burn and \hlVeDOFARNflairEst{} [\hlVeDOFARNflairCILo, \hlVeDOFARNflairCIHi] on \flair.
\tm minus \rn is \hlVeTMRNburnEst{} [\hlVeTMRNburnCILo, \hlVeTMRNburnCIHi] and \hlVeTMRNflairEst{} [\hlVeTMRNflairCILo, \hlVeTMRNflairCIHi].
No single \gfm therefore differs detectably from \rn, and F2 concerns their mean.
In noisy-label \miou at the selected \Ltwo type, frozen minus fine-tuned is \hlVeFrozenSATburnEst{} [\hlVeFrozenSATburnCILo, \hlVeFrozenSATburnCIHi] for \sat, \hlVeFrozenDOFAburnEst{} [\hlVeFrozenDOFAburnCILo, \hlVeFrozenDOFAburnCIHi] for \dofa and \hlVeFrozenTMburnEst{} [\hlVeFrozenTMburnCILo, \hlVeFrozenTMburnCIHi] for \tm on \burn.
On \flair it is \hlVeFrozenSATflairEst{} [\hlVeFrozenSATflairCILo, \hlVeFrozenSATflairCIHi], \hlVeFrozenDOFAflairEst{} [\hlVeFrozenDOFAflairCILo, \hlVeFrozenDOFAflairCIHi] and \hlVeFrozenTMflairEst{} [\hlVeFrozenTMflairCILo, \hlVeFrozenTMflairCIHi].
Over all common seeds, \lvd minus \sat in noisy-label \miou averaged over the four \Ltwo types is \hlVeNoisyLVDSATburnEst{} [\hlVeNoisyLVDSATburnCILo, \hlVeNoisyLVDSATburnCIHi] on \burn and \hlVeNoisyLVDSATflairEst{} [\hlVeNoisyLVDSATflairCILo, \hlVeNoisyLVDSATflairCIHi] on \flair.
Against \sat at the matched rate (N6, seeds 1 to 5) it is \hlVeNoisyLVDSATNSixburnEst{} [\hlVeNoisyLVDSATNSixburnCILo, \hlVeNoisyLVDSATNSixburnCIHi] and \hlVeNoisyLVDSATNSixflairEst{} [\hlVeNoisyLVDSATNSixflairCILo, \hlVeNoisyLVDSATNSixflairCIHi], with the hardware term of \Cref{tab:supp-hwmix}.
Over the eight seeds of the pair, the memorization rate of \sat minus \lvd at the selected type is \hlVeMemSATLVDburnEst{} [\hlVeMemSATLVDburnCILo, \hlVeMemSATLVDburnCIHi] points on \burn and \hlVeMemSATLVDflairEst{} [\hlVeMemSATLVDflairCILo, \hlVeMemSATLVDflairCIHi] on \flair.
The 90\,\% interval of F2 on \burn without the collapsed runs is [\hlVeFTwoCollTostburnCILo, \hlVeFTwoCollTostburnCIHi], within $\pm$\TostMargin{} points, the same exploratory reading.

A leave-one-seed-out check addresses the effect of selecting the worst type on seed \CalSeed.
Recomputed without seed \CalSeed{} (\hlWinnersCurseAThreeburnN{} seeds), the ablation contrasts on \burn are F3 \hlWinnersCurseAThreeburnEst{} [\hlWinnersCurseAThreeburnCILo, \hlWinnersCurseAThreeburnCIHi] and F4 \hlWinnersCurseAFourburnEst{} [\hlWinnersCurseAFourburnCILo, \hlWinnersCurseAFourburnCIHi] points.
On \flair they are F3 \hlWinnersCurseAThreeflairEst{} [\hlWinnersCurseAThreeflairCILo, \hlWinnersCurseAThreeflairCIHi] and F4 \hlWinnersCurseAFourflairEst{} [\hlWinnersCurseAFourflairCILo, \hlWinnersCurseAFourflairCIHi] points.
A5 gives \hlWinnersCurseAFiveburnEst{} [\hlWinnersCurseAFiveburnCILo, \hlWinnersCurseAFiveburnCIHi] on \burn and \hlWinnersCurseAFiveflairEst{} [\hlWinnersCurseAFiveflairCILo, \hlWinnersCurseAFiveflairCIHi] on \flair.
A6 gives \hlWinnersCurseASixburnEst{} [\hlWinnersCurseASixburnCILo, \hlWinnersCurseASixburnCIHi] on \burn.
The N6 arm against the main-grid \sat arm gives \hlWinnersCurseNSixburnEst{} [\hlWinnersCurseNSixburnCILo, \hlWinnersCurseNSixburnCIHi] on \burn and \hlWinnersCurseNSixflairEst{} [\hlWinnersCurseNSixflairCILo, \hlWinnersCurseNSixflairCIHi] on \flair.
Every sign is unchanged.
The \flair F4 and A5 intervals include zero without seed \CalSeed, and so do the A5, A6 and N6 intervals on \burn.
The check addresses the inflation of the selected type's effect. Choosing the type is a separate question.
On \burn \etype{coarse} and \etype{omission} are not separated, \hlTypeRankDiffburnOne{} [\hlTypeRankDiffCILoburnOne, \hlTypeRankDiffCIHiburnOne] points apart.

\begin{figure}[t]
\centering
\ifdefined\plannedbuild\includegraphics[width=\linewidth]{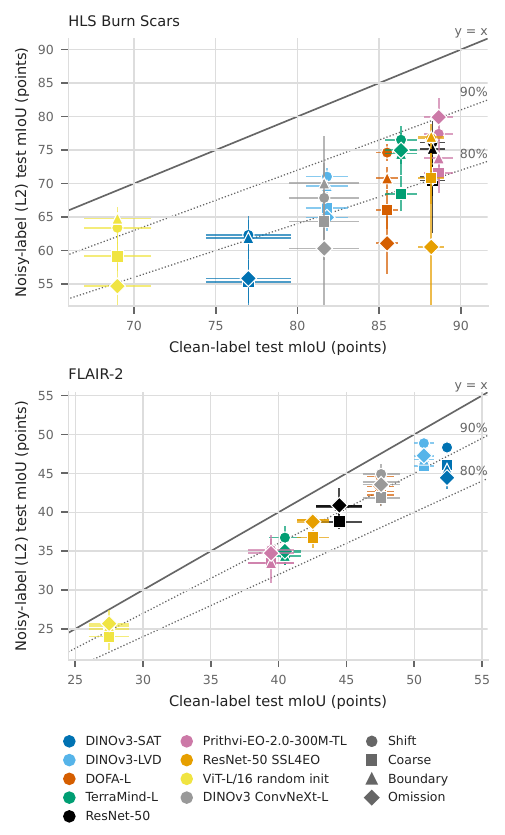}\else\includegraphics[width=\linewidth]{effective_robustness}\fi
\caption{\textbf{Clean against noisy \miou at \Ltwo:} One marker per \ifdefined\plannedbuild encoder\else core model\fi{} and error type, with \CiLevel\,\% intervals on both axes over the seeds of each cell. The cells have \NSeeds{} seeds, and \statFOneburnN{} for the core models on \burn and for the DINOv3 pair on \flair. The top panel shows \burn and the bottom panel \flair. The solid line is $y = x$ and the dotted lines mark 90\,\% and 80\,\% retention. A model above another model's interval at equal clean \miou retains more. The plot shows retention against clean \miou and fits no trend line across encoders.}
\label{fig:supp-er}
\end{figure}

\begin{table}[t]
\centering\footnotesize\setlength{\tabcolsep}{2pt}
\IfFileExists{generated/table_memorisation.tex}{%
\providecommand{\gcvdry}[1]{\textcolor{red}{#1}}
\begin{tabular}{@{}lcccc@{}}
\toprule
 & \multicolumn{3}{c}{\burn, \etype{coarse} \Ltwo} & \flair \\
\cmidrule(lr){2-4}\cmidrule(l){5-5}
Model & All & Burned & Background & All \\
\midrule
\sat & \textbf{72.6}$\mathord{\pm}$10.5 & \textbf{79.2} & 18.4 & 43.1$\mathord{\pm}$13.1 \\
\lvd & 81.0$\mathord{\pm}$5.2 & 88.0 & 20.4 & 25.2$\mathord{\pm}$7.6 \\
\dofa & 80.7$\mathord{\pm}$7.2 & 87.9 & 17.1 & 37.6$\mathord{\pm}$7.3 \\
\tm & 75.8$\mathord{\pm}$5.0 & 82.6 & \textbf{15.3} & 37.0$\mathord{\pm}$9.4 \\
\rn & 80.4$\mathord{\pm}$5.2 & 87.1 & 22.1 & \textbf{18.6}$\mathord{\pm}$5.6 \\
\sat (N6)$^{\ddagger}$ & 82.3$\mathord{\pm}$5.0 & 89.9 & 21.1 & 39.1$\mathord{\pm}$14.6 \\
\bottomrule
\end{tabular}
}{\tabletbd[2.4cm]{generated/table\_memorisation.tex: run figs/make\_results\_tables.py}}
\caption{\textbf{Memorization of corrupted labels:} Memorization rate \memrate{} of \ifdefined\plannedbuild\cref{eq:mem}\else Eq. (3) of the main paper\fi{} in percent at the selected checkpoint at the selected \Ltwo type. Values are means over \NSeeds{} seeds per cell, and \statFOneburnN{} for the core models on \burn and for the DINOv3 pair on \flair. The All columns add the \CiLevel\,\% $t$ half-width. On \burn the rate is given over all corrupted pixels (All) and separately on corrupted pixels whose clean class is burned or background. The \flair column is at \etype{\WorstFlair} \Ltwo. Bold marks the smallest rate of a column among the five core models. The last row is \sat at the \lvd learning rate (N6). A $\ddagger$ marks its cells with runs from two GPU classes (\ifdefined\plannedbuild\msec{sec:setup}\else Sec. 7 of the main paper\fi).} %
\label{tab:mem}
\end{table}

\ifdefined\cvprbuild\else
\begin{figure}[t]
\centering
\IfFileExists{figs/memorisation_dinov3.pdf}{\includegraphics[width=\linewidth]{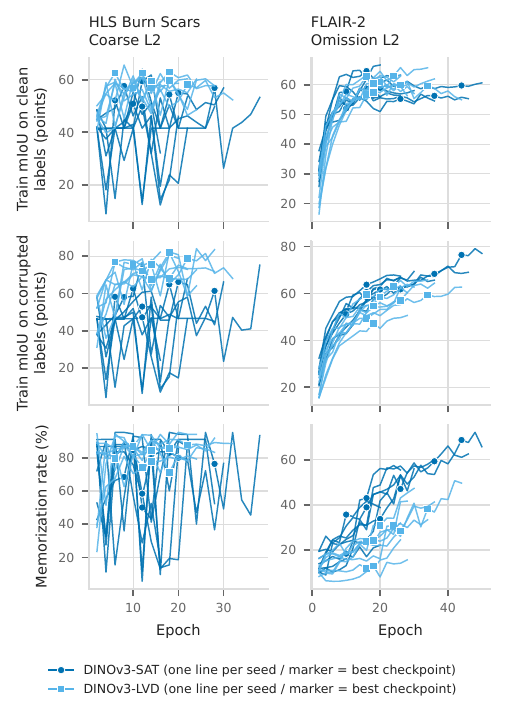}}{\phfig{4.0cm}{figs/memorisation\_dinov3.pdf from analysis/figures.py}}
\caption{\textbf{Memorization during training at the selected \Ltwo type:} Training \miou against the clean labels, training \miou against the corrupted labels and the memorization rate \memrate{} of \ifdefined\plannedbuild\cref{eq:mem}\else the main paper\fi. All three are measured on \MemCrops{} fixed training crops against the epoch. One thin line per seed with a marker at the selected checkpoint. The left column shows \burn and the right column \flair, with one row per quantity. Both encoders share each panel in their model colors, \sat at its \gbtwo rate and \lvd.}
\label{fig:supp-mem}
\end{figure}

\fi
On \burn no model crosses another model's interval at equal clean \miou (\Cref{fig:supp-er}), although some point estimates swap order.
On \flair the DINOv3 pair reverses at the selected rates, because \lvd ends \hlVeNoisyLVDSATflairEst{} [\hlVeNoisyLVDSATflairCILo, \hlVeNoisyLVDSATflairCIHi] points ahead at \Ltwo from a lower clean \miou.

On \flair the memorization rate rises throughout training\ifdefined\cvprbuild\else{} (\Cref{fig:supp-mem})\fi{}.
The matched-rate \sat arm reproduces \hlMemRateSATNSixflair\,\% of the corrupted pixels on \flair and \hlMemRateSATNSixburn\,\% on \burn (\Cref{tab:mem}).
The rates are upper bounds that include chance-level agreement, because no clean-model agreement with the corrupted masks was logged.
Both DINOv3 encoders reproduce most of the corrupted pixels on \burn from the first validation check, with \lvd above \sat, unlike on \flair.
On \burn the corrupted pixels are mostly burned pixels relabeled as background.
There the rate is \hlMemPerClassSATburnBurnScar\,\% for \sat and \hlMemPerClassLVDburnBurnScar\,\% for \lvd against \hlMemPerClassSATburnBackground\,\% and \hlMemPerClassLVDburnBackground\,\% on background pixels.
The rate on \burn therefore mixes the class prior with memorization.

\section{Extension Blocks}
\label{sec:supp-stretch}

\begin{table}[t]
\centering\footnotesize\setlength{\tabcolsep}{3pt}
\IfFileExists{generated/table_sen1floods.tex}{\begin{adjustbox}{max width=\columnwidth}%
\providecommand{\gcvdry}[1]{\textcolor{red}{#1}}
\providecommand{\ProvSenOneFloods}{Seeds per cell 5. Test mIoU on the hand-labeled test split after training on the hand labels and on the Otsu labels of the same tiles, mean $\pm$ 95\,\% $t$ half-width, and the paired drop in points. Bold marks the largest mIoU, the smallest drop and the largest retention of a column. Analysis of the finished runs.}
\begin{tabular}{@{}lcccc@{}}
\toprule
Model & Hand labels & Otsu labels & Drop & Retention (\%) \\
\midrule
\sat & 69.1 $\pm$ 3.0 & 61.9 $\pm$ 3.9 & 7.2 $\pm$ 6.5 & 89.7 \\
\lvd & 65.7 $\pm$ 7.1 & 60.4 $\pm$ 7.2 & \textbf{5.3} $\pm$ 12.4 & \textbf{92.8} \\
\dofa & 83.7 $\pm$ 1.3 & \textbf{74.9} $\pm$ 1.2 & 8.8 $\pm$ 1.9 & 89.4 \\
\tm & 87.5 $\pm$ 1.4 & 71.4 $\pm$ 1.7 & 16.1 $\pm$ 2.4 & 81.6 \\
\rn & \textbf{89.4} $\pm$ 0.4 & 73.7 $\pm$ 1.5 & 15.7 $\pm$ 1.3 & 82.4 \\
\bottomrule
\end{tabular}
\end{adjustbox}}{\tabletbd[2.2cm]{generated/table\_sen1floods.tex: run the analysis harness}}
\caption{\textbf{Hand against Otsu training labels on \sfloods (S1):} Test \miou on the hand-labeled test split after training on the hand labels and on the Otsu labels of the same chips. Cells give the mean $\pm$ \CiLevel\,\% $t$ half-width over \NSeeds{} seeds, with the paired drop in points and the retention. Bold marks the largest \miou, the smallest drop and the largest retention of a column. Against the hand masks of the training and validation chips, the Otsu masks score a label \miou of \hlOtsuLabelMiouFracAll{} and flip rate \hlOtsuFlipRateAll\,\%. That label \miou is a mask agreement, not a test score. \ProvSlot{\ResultsProvenance}}
\label{tab:supp-sen1floods}
\end{table}

\begin{table*}[t]
\centering\footnotesize\setlength{\tabcolsep}{4pt}
\IfFileExists{generated/table_stretch.tex}{%
\providecommand{\gcvdry}[1]{\textcolor{red}{#1}}
\providecommand{\ProvStretch}{Seeds per cell 5. Values are mean $\pm$ 95\,\% $t$ half-width, drops in points at \Ltwo. Subset mIoU is the clean test mIoU after training on 25\,\% of the tiles with clean labels. Bold marks the largest clean mIoU, the smallest drop and the largest subset mIoU per column within a dataset. Analysis of the finished runs.}
\begin{tabular}{@{}llccccccc@{}}
\toprule
Block & Model & Clean mIoU & Shift & Coarse & Boundary & Omission & Subset mIoU \\
\midrule
\multicolumn{8}{@{}l}{\textit{\burn}} \\
S3 & \prithvi & \textbf{88.6} $\pm$ 0.9 & 11.2 $\pm$ 1.1 & 17.1 $\pm$ 2.5 & 14.9 $\pm$ 4.3 & \textbf{8.7} $\pm$ 3.5 & 86.0 $\pm$ 2.1 \\
S4 & \rnssl & 88.2 $\pm$ 0.8 & 11.4 $\pm$ 1.2 & 17.4 $\pm$ 3.9 & 11.2 $\pm$ 1.9 & 27.6 $\pm$ 13.5 & \textbf{87.3} $\pm$ 1.2 \\
S5 & \vitr & 69.0 $\pm$ 2.0 & \textbf{5.6} $\pm$ 4.2 & \textbf{9.9} $\pm$ 2.9 & \textbf{4.2} $\pm$ 3.2 & 14.3 $\pm$ 10.1 & 61.7 $\pm$ 2.5 \\
S6 & \cnxl & 81.6 $\pm$ 2.1 & 13.8 $\pm$ 9.3 & 17.3 $\pm$ 3.7 & 11.6 $\pm$ 4.4 & 21.3 $\pm$ 11.5 & 74.6 $\pm$ 4.9 \\
\midrule
\multicolumn{8}{@{}l}{\textit{\flair}} \\
S3 & \prithvi & 39.4 $\pm$ 1.7 & 4.3 $\pm$ 2.3 & 6.0 $\pm$ 1.5 & 6.0 $\pm$ 2.9 & 4.7 $\pm$ 2.9 & 32.7 $\pm$ 1.2 \\
S4 & \rnssl & 42.5 $\pm$ 1.2 & 3.5 $\pm$ 1.2 & 5.8 $\pm$ 2.2 & 3.6 $\pm$ 1.3 & 3.7 $\pm$ 2.0 & 36.8 $\pm$ 2.0 \\
S5 & \vitr & 27.5 $\pm$ 1.5 & \textbf{2.5} $\pm$ 1.3 & \textbf{3.5} $\pm$ 2.8 & \textbf{2.1} $\pm$ 1.1 & \textbf{1.8} $\pm$ 2.2 & 21.1 $\pm$ 0.8 \\
S6 & \cnxl & \textbf{47.6} $\pm$ 1.4 & 2.6 $\pm$ 2.2 & 5.7 $\pm$ 1.0 & 3.6 $\pm$ 2.2 & 4.0 $\pm$ 2.4 & \textbf{43.5} $\pm$ 1.8 \\
\bottomrule
\end{tabular}
}{\tabletbd[3cm]{generated/table\_stretch.tex: run the analysis harness}}
\caption{\textbf{Extension encoders S3 to S6 on the main grid:} Clean test \miou and the paired drop of each error type at \Ltwo. Subset is the clean test \miou after training on \SubsetPct\,\% of the tiles with clean labels. Cells give the mean $\pm$ \CiLevel\,\% $t$ half-width over \NSeeds{} seeds. Bold marks the largest clean \miou, the smallest drop and the largest subset \miou per column within a dataset. A smaller drop from a lower clean \miou is not a better result. \vitr{} keeps the \MaxEpochs-epoch budget of every other model. \ProvSlot{\ResultsProvenance}}
\label{tab:supp-stretch}
\end{table*}

\Cref{tab:supp-sen1floods} gives block S1 on \sfloods \cite{bonafilia2020sen1floods11}.
The dataset pairs \TilePx\,px Sentinel-2 L1C chips at \ResSOneM\,m with hand-labeled water masks over \NSOneEvents{} flood events, \NSOneHandChips{} chips in all.
Each chip also ships a water mask obtained by Otsu thresholding \cite{otsu1979threshold} of the coincident Sentinel-1 scene, the S1OtsuLabelHand product of the release.
We use the hand-labeled training, validation and test splits of the release with \NSOneTrain, \NSOneVal{} and \NSOneTest{} chips.
The remaining hand-labeled chips form the release's held-out region split and are not used.
The hand arm trains and validates on the hand masks and the Otsu arm on the Otsu masks of the same chips.
Both are tested on the hand-labeled test split, and pixels without a label in either product are ignored.
Against the hand masks of the \hlOtsuNTilesAll{} training and validation chips the Otsu masks score a label \miou of \hlOtsuLabelMiouFracAll.
That label \miou places them at the \Ltwo target of the simulated types, with a flip rate of \hlOtsuFlipRateAll\,\%.
The water class carries the error, with an IoU of \hlOtsuIoUWaterAll{} against \hlOtsuIoUNotWaterAll{} for the other class.
Input bands follow \Cref{tab:supp-models}. Every model keeps its \burn learning rate and batch size because \gbtwo has no entry for this dataset.
Each cell holds \NSeeds{} seeds.
The drop averages \hlSOneDropMeansenEst{} [\hlSOneDropMeansenCILo, \hlSOneDropMeansenCIHi] points over \hlSOneDropMeansenN{} (model, seed) units.
It ranges from \hlSOneDropLVDsenEst{} [\hlSOneDropLVDsenCILo, \hlSOneDropLVDsenCIHi] for \lvd to \hlSOneDropTMsenEst{} [\hlSOneDropTMsenCILo, \hlSOneDropTMsenCIHi] for \tm.
The RGB-only DINOv3 pair starts far lower on the hand labels and loses less, so its retention is the highest.

\Cref{tab:supp-stretch} gives the main-grid cells of the four extension encoders.
Each contrast below is the \Ltwo drop averaged over the four types and paired by seed, with its \CiLevel\,\% $t$ interval.
\prithvi minus \dofa is \hlSThreeDropDOFAburnEst{} [\hlSThreeDropDOFAburnCILo, \hlSThreeDropDOFAburnCIHi] points on \burn and \hlSThreeDropDOFAflairEst{} [\hlSThreeDropDOFAflairCILo, \hlSThreeDropDOFAflairCIHi] on \flair.
Against \tm it is \hlVeSThreeTMburnEst{} [\hlVeSThreeTMburnCILo, \hlVeSThreeTMburnCIHi] and \hlVeSThreeTMflairEst{} [\hlVeSThreeTMflairCILo, \hlVeSThreeTMflairCIHi].
\rnssl minus \rn is \hlSFourDropRNburnEst{} [\hlSFourDropRNburnCILo, \hlSFourDropRNburnCIHi] on \burn and \hlSFourDropRNflairEst{} [\hlSFourDropRNflairCILo, \hlSFourDropRNflairCIHi] on \flair.
\vitr minus \sat is \hlSFiveDropSATburnEst{} [\hlSFiveDropSATburnCILo, \hlSFiveDropSATburnCIHi] on \burn and \hlSFiveDropSATflairEst{} [\hlSFiveDropSATflairCILo, \hlSFiveDropSATflairCIHi] on \flair.
Against \lvd it is \hlSFiveDropLVDburnEst{} [\hlSFiveDropLVDburnCILo, \hlSFiveDropLVDburnCIHi] and \hlSFiveDropLVDflairEst{} [\hlSFiveDropLVDflairCILo, \hlSFiveDropLVDflairCIHi].
\cnxl minus \sat is \hlSSixDropSATburnEst{} [\hlSSixDropSATburnCILo, \hlSSixDropSATburnCIHi] on \burn and \hlSSixDropSATflairEst{} [\hlSSixDropSATflairCILo, \hlSSixDropSATflairCIHi] on \flair.
Against \lvd it is \hlSSixDropLVDburnEst{} [\hlSSixDropLVDburnCILo, \hlSSixDropLVDburnCIHi] and \hlSSixDropLVDflairEst{} [\hlSSixDropLVDflairCILo, \hlSSixDropLVDflairCIHi].

\ifdefined\cvprbuild\else
\vitr shares the architecture and recipe of the DINOv3 pair.
Its learning rate is \VitrLrMultBurn{} the \sat rate on \burn and \VitrLrMultFlair{} on \flair.
The pick was made on clean seed \CalSeed{} among one, three and ten times that rate.
The picked probe run is its clean seed-\CalSeed{} cell.
It loses fewer points than \sat from a far lower clean \miou, as A6 found for \rn.
Under the shared \MaxEpochs-epoch budget it is a no-pretraining control and not a tuned from-scratch ViT.

The extension encoders also repeat the F4 budget contrast, \SubsetPct\,\% clean tiles minus all tiles at the selected \Ltwo type.
For \prithvi on \burn it is \hlSThreeBudgetburnEst{} [\hlSThreeBudgetburnCILo, \hlSThreeBudgetburnCIHi].
It is \hlSFourBudgetburnEst{} [\hlSFourBudgetburnCILo, \hlSFourBudgetburnCIHi] for \rnssl, \hlSFiveBudgetburnEst{} [\hlSFiveBudgetburnCILo, \hlSFiveBudgetburnCIHi] for \vitr and \hlSSixBudgetburnEst{} [\hlSSixBudgetburnCILo, \hlSSixBudgetburnCIHi] for \cnxl.
On \flair it is \hlSThreeBudgetflairEst{} [\hlSThreeBudgetflairCILo, \hlSThreeBudgetflairCIHi], \hlSFourBudgetflairEst{} [\hlSFourBudgetflairCILo, \hlSFourBudgetflairCIHi], \hlSFiveBudgetflairEst{} [\hlSFiveBudgetflairCILo, \hlSFiveBudgetflairCIHi] and \hlSSixBudgetflairEst{} [\hlSSixBudgetflairCILo, \hlSSixBudgetflairCIHi] in the same order.

\fi
Block S2 trains \sat and \rn on \flair with a class-weighted loss, clean and at \etype{omission} \Ltwo.
Each noisy run is paired with its own class-weighted clean run.
Against the main-grid drop of the same seed the change is \hlSTwoDeltaSATflairEst{} [\hlSTwoDeltaSATflairCILo, \hlSTwoDeltaSATflairCIHi] points for \sat and \hlSTwoDeltaRNflairEst{} [\hlSTwoDeltaRNflairCILo, \hlSTwoDeltaRNflairCIHi] for \rn.

\ifdefined\cvprbuild\else
The hardware term of each extension block is as follows.
In S1 the DINOv3 pair ran on the RTX PRO 6000 class and the other three models on GB10, with \hlHwBlockClassesSOne{} runs.
\dofa and \tm therefore ran off their planned class, but each model on one class.
S2 likewise ran on one class per model (\hlHwBlockClassesSTwo), so the S1 and S2 contrasts carry no hardware term.
S5 and S6 ran on one class (\hlHwBlockClassesSFive{} and \hlHwBlockClassesSSix) and S4 on two (\hlHwBlockClassesSFour).
Only S3 ran on three (\hlHwBlockClassesSThree), so the S3 and S4 cells carry a hardware term.
Every \prithvi cell spans the three classes (\Cref{tab:supp-manifest}).
Within its \Ltwo main-grid cells the per-class mean drops differ by \hlHwSThreeRangeMinLTwoburn{} to \hlHwSThreeRangeMaxLTwoburn{} points on \burn and \hlHwSThreeRangeMinLTwoflair{} to \hlHwSThreeRangeMaxLTwoflair{} on \flair.
The seed effect is included in these ranges.

\fi
\section{Experiment Blocks, Input Bands and Hardware}
\label{sec:supp-hardware}

\begin{table}[t]
\centering\footnotesize
\ifdefined\arxivbuild\renewcommand{\arraystretch}{0.9}\fi%
\setlength{\tabcolsep}{3pt}
\begin{tabular}{@{}lp{0.57\columnwidth}r@{}}
\toprule
Block & Question & Runs \\
\midrule
A1 & Main grid, all core models and conditions & \RunsAOne \\
A2 & Learning-rate choice for \lvd & \RunsATwo \\
A3 & Frozen encoder, \sat, \dofa, \tm & \RunsAThree \\
A4 & \SubsetPct\,\% of tiles with clean labels, all core models & \RunsAFour \\
A5 & Clean validation labels for stopping, \sat & \RunsAFive \\
A6 & \rn from random initialization, two worst types & \RunsASix \\
A7, N4 & RGB-only \dofa and \rn on \burn & \RunsASeven, \RunsNFour \\
N1, N2 & Learning-rate sensitivity, clean and noisy & \RunsNOne, \RunsNTwo \\
N3 & Fixed \MaxEpochs{} epochs without early stopping & \RunsNThree \\
N6 & \sat at the \lvd learning rate, clean and \Ltwo & \RunsNSix \\
N5 & Seeds \NFiveSeeds{} for the DINOv3 pair, conditional & \RunsNFive \\
S1 & Hand against Otsu labels on \sfloods & \RunsSOne \\
S2 & Class-weighted loss under \etype{omission} & \RunsSTwo \\
S3 to S6 & \prithvi, \rnssl, \vitr, \cnxl & \RunsSThree, \RunsSFour, \RunsSFive, \RunsSSix \\
\ifdefined\plannedbuild\PlannedBlocksRows\fi
\bottomrule
\end{tabular}
\caption{\textbf{Experiment blocks and their runs:} A blocks, N1 to N4 and N6 form the core design and S blocks are extensions. Runs count training runs over both datasets and all seeds. S3 to S6 each run the full main-grid conditions for one model. N6 reuses the seed \CalSeed{} clean runs of N1 and two selected-type seeds of N2, so it trains fewer runs than it has cells. A6 adds two learning-rate probe runs on clean seed \CalSeed{} to its grid cells. \ifdefined\plannedbuild\unskip\PlannedBlocksCaption{} \fi A2 counts the \lvd probe runs not reused as its clean seed-\CalSeed{} cell.}
\label{tab:supp-blocks}
\end{table}

\begin{figure}[t]
\centering
\IfFileExists{figs/model_roster.pdf}{\includegraphics[width=\linewidth]{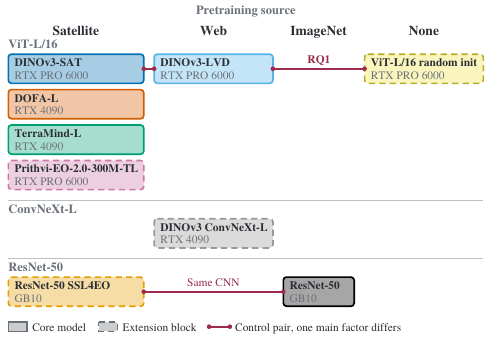}}{\phfig{6.6cm}{figs/model\_roster.pdf from figs/make\_model\_roster\_fig.py}}
\caption{\textbf{The model roster:} The nine encoders by architecture (rows) and pretraining source (columns). Each chip carries the fixed color of the model in every figure and the planned GPU class of its runs. Solid chips are the five core models and dashed chips the extension blocks S3 to S6. Wine connectors join the pairs read as controls, which differ mainly in one factor. These are the DINOv3 pair with its random-initialization anchor for RQ1 and the two \rn initializations.}
\label{fig:supp-roster}
\end{figure}

\begin{table}[t]
\centering\footnotesize\setlength{\tabcolsep}{3pt}
\begin{adjustbox}{max width=\columnwidth}
\begin{tabular}{@{}llll@{}}
\toprule
Model & \burn & \flair & \sfloods \\
\midrule
\sat, \lvd & RGB & RGB & RGB \\
\dofa & \NBurnBands{} HLS bands & RGB, NIR & \NBurnBands{} S2 bands \\
\tm & \NBurnBands{} HLS bands & RGB, NIR & \NSOneBands{} S2 bands \\
\rn & \NBurnBands{} HLS bands & RGB, NIR, elevation & \NBurnBands{} S2 bands \\
\prithvi & \NBurnBands{} HLS bands & RGB, NIR & Not run \\
\rnssl & \NBurnBands{} HLS bands & RGB, NIR, elevation & Not run \\
\vitr, \cnxl & RGB & RGB & Not run \\
\dofa and \rn RGB controls & RGB & Not run & Not run \\
\bottomrule
\end{tabular}
\end{adjustbox}
\caption{\textbf{Input bands per model:} Columns give the input bands on each dataset, which follow the \gbtwo configurations where they are public. NIR is near infrared. Elevation has no wavelength and is not given to the wavelength-conditioned \dofa or to \tm. On \sfloods \dofa and \rn keep the Sentinel-2 (S2) counterparts of their six \burn bands, because \gbtwo has no entry for this dataset. \tm receives the full S2 L1C set of its pretraining there. \prithvi reads the four \flair bands through the named HLS subset of its patch embedding, with the aerial NIR mapped to the NIR narrow weights. The \rn RGB control matches the RGB-only \rn that \gbtwo reports and runs on \burn only.}
\label{tab:supp-models}
\end{table}

\ifdefined\cvprbuild\else
\begin{table}[t]
\centering\footnotesize\setlength{\tabcolsep}{3pt}
\IfFileExists{generated/table_hardware.tex}{%
\begin{tabular}{@{}llr@{}}
\toprule
Model & GPU class & Runs \\
\midrule
DINOv3-SAT & RTX PRO 6000 Blackwell, 96\,GB & 200 \\
DINOv3-LVD & RTX PRO 6000 Blackwell, 96\,GB & 152 \\
DOFA-L & RTX 4090, 24\,GB & 140 \\
TerraMind-L & RTX 4090, 24\,GB & 130 \\
ResNet-50 & GB10, 128\,GB unified & 162 \\
Prithvi-EO-2.0-300M-TL & RTX PRO 6000 Blackwell, 96\,GB & 100 \\
ResNet-50 SSL4EO & GB10, 128\,GB unified & 100 \\
ViT-L/16 random init & RTX PRO 6000 Blackwell, 96\,GB & 104 \\
DINOv3 ConvNeXt-L & RTX 4090, 24\,GB & 100 \\
\bottomrule
\end{tabular}
}{\tabletbd{generated/table\_hardware.tex: run scripts/make\_config\_macros.py}}
\caption{\textbf{Planned GPU class per model:} Every run of a model was assigned the listed class in BF16 mixed precision. The cells that the mid-run reassignment moved across classes are listed in \Cref{sec:supp-hardware}, and the extension blocks that ran off their planned class in \Cref{sec:supp-stretch,sec:supp-hardware}. Runs are the planned runs of that model over the A, N1 to N5 and S blocks, without the \RunsNSix{} runs of N6 (\Cref{tab:supp-blocks}).}
\label{tab:supp-hardware}
\end{table}

\begin{table}[t]
\centering\footnotesize\setlength{\tabcolsep}{3pt}
\IfFileExists{generated/table_manifest_reflow.tex}{\begin{adjustbox}{max width=\columnwidth}%
\providecommand{\gcvdry}[1]{\textcolor{red}{#1}}
\providecommand{\ProvManifest}{Analysed runs per model 450 to 2332, all blocks, the five core models above the rule and the extension encoders S3 to S6 below it. Wall time and peak GPU memory are medians per run. A slash separates values that occur within one model. Analysis of the finished runs.}
\gdef\ManifestConstants{PyTorch 2.14.1, CUDA 13.0 and BF16 mixed precision}
\begin{tabular}{@{}lllccc@{}}
\toprule
Model & GPU & Driver & Runs & Wall (min) & Memory (GB) \\
\midrule
\sat & GB10 / H100 NVL / RTX 4090 / RTX 5060 Ti / RTX PRO 6000 & 580.178.04 / 595.71.05 / 595.84 / 595.91.07 / 595.99.02 / 621.01 & 2332 & 12.5 & 9.3 \\
\lvd & GB10 / H100 NVL / RTX 4090 / RTX 5060 Ti / RTX PRO 6000 & 580.178.04 / 595.71.05 / 595.84 / 595.91.07 / 595.99.02 / 621.01 & 2284 & 13.1 & 9.3 \\
\dofa & GB10 / H100 NVL / RTX 3080 / RTX 3080 Ti / RTX 4090 / RTX 5060 Ti / RTX PRO 6000 & 580.173.02 / 580.178.04 / 595.71.05 / 595.84 / 595.99.02 / 621.01 & 1195 & 14.1 & 7.6 \\
\tm & GB10 / H100 NVL / RTX 4090 / RTX 5060 Ti / RTX PRO 6000 & 580.173.02 / 580.178.04 / 595.71.05 / 595.84 / 595.99.02 / 621.01 & 1185 & 13.1 & 12.9 \\
\rn & GB10 / H100 NVL / RTX 3080 / RTX 3080 Ti / RTX 4090 / RTX 5060 Ti / RTX PRO 6000 & 580.173.02 / 580.178.04 / 595.71.05 / 595.84 / 595.99.02 / 621.01 & 1217 & 6.1 & 1.6 \\
\midrule
\prithvi & GB10 / H100 NVL / RTX 4090 / RTX 5060 Ti / RTX PRO 6000 & 580.173.02 / 580.178.04 / 595.71.05 / 595.84 / 595.91.07 / 621.01 & 450 & 19.1 & 7.0 \\
\rnssl & GB10 / H100 NVL / RTX 3080 Ti / RTX 4090 / RTX 5060 Ti / RTX PRO 6000 & 580.173.02 / 580.178.04 / 595.71.05 / 595.84 / 595.91.07 / 595.99.02 / 621.01 & 450 & 5.5 & 1.3 \\
\vitr & GB10 / H100 NVL / RTX 4090 / RTX 5060 Ti / RTX PRO 6000 & 580.173.02 / 580.178.04 / 595.71.05 / 595.84 / 595.91.07 / 621.01 & 454 & 13.9 & 9.3 \\
\cnxl & GB10 / H100 NVL / RTX 5060 Ti / RTX PRO 6000 & 580.173.02 / 580.178.04 / 595.71.05 / 595.91.07 / 621.01 & 450 & 12.8 & 5.2 \\
\bottomrule
\end{tabular}
\end{adjustbox}}{\tabletbd[2.4cm]{generated/table\_manifest\_reflow.tex: run scripts/reflow\_tables.py}}
\caption{\textbf{Run manifest per model:} GPU and driver of the analyzed runs over all blocks, their number, and the median wall time and peak GPU memory per run. The five core models are above the rule and the extension encoders S3 to S6 below it. Every run shares \ManifestConstants. A slash separates values that occur within one model. \ProvSlot{\ResultsProvenance}}
\label{tab:supp-manifest}
\end{table}

\fi
\begin{table*}[t]
\centering\footnotesize\setlength{\tabcolsep}{3pt}
\ifdefined\arxivbuild\renewcommand{\arraystretch}{0.9}\fi%
\IfFileExists{generated/table_hardware_mix.tex}{%
\providecommand{\gcvdry}[1]{\textcolor{red}{#1}}
\providecommand{\ProvHardwareMix}{60 cells. A and B are the mean test \miou (clean cells) or the mean paired drop (noisy cells) in points over the seeds run on each GPU class. The seeds differ between the classes, so A $-$ B mixes the seed effect with the hardware term. Analysis of the finished runs.}
\begin{tabular}{@{}lllllccc@{}}
\toprule
Block & Model & Cell & Class A (seeds) & Class B (seeds) & A & B & A $-$ B \\
\midrule
A1 & LVD & \burn \etype{boundary} L1 & RTX 4090 (3,4,5) & RTX PRO 6000 (1,2) & 3.1 & 4.6 & $-$1.5 \\
A1 & LVD & \burn \etype{shift} L1 & RTX 4090 (3,4,5) & RTX PRO 6000 (1,2) & 1.3 & 0.8 & 0.5 \\
A1 & LVD & \burn \etype{boundary} L2 & RTX 4090 (6,7,8) & RTX PRO 6000 (1,2,3,4,5) & 11.2 & 12.8 & $-$1.6 \\
A1 & LVD & \burn \etype{coarse} L2 & RTX 4090 (6,7,8) & RTX PRO 6000 (1,2,3,4,5) & 14.8 & 15.9 & $-$1.1 \\
A1 & LVD & \burn \etype{omission} L2 & RTX 4090 (6,7,8) & RTX PRO 6000 (1,2,3,4,5) & 16.5 & 17.1 & $-$0.6 \\
A1 & LVD & \burn \etype{shift} L2 & RTX 4090 (6,7,8) & RTX PRO 6000 (1,2,3,4,5) & 11.0 & 10.6 & 0.4 \\
A1 & LVD & \burn clean & RTX 4090 (6,7,8) & RTX PRO 6000 (1,2,3,4,5) & 81.5 & 82.0 & $-$0.6 \\
A1 & SAT & \burn \etype{boundary} L1 & RTX 4090 (3,4,5) & RTX PRO 6000 (1,2) & 0.8 & 6.9 & $-$6.1 \\
A1 & SAT & \burn \etype{shift} L1 & RTX 4090 (2,3,4,5) & RTX PRO 6000 (1) & $-$1.2 & 5.5 & $-$6.8 \\
A1 & SAT & \burn \etype{boundary} L2 & RTX 4090 (6,7,8) & RTX PRO 6000 (1,2,3,4,5) & 15.4 & 14.9 & 0.4 \\
A1 & SAT & \burn \etype{coarse} L2 & RTX 4090 (6,7,8) & RTX PRO 6000 (1,2,3,4,5) & 21.5 & 21.7 & $-$0.2 \\
A1 & SAT & \burn \etype{omission} L2 & RTX 4090 (6,7,8) & RTX PRO 6000 (1,2,3,4,5) & 22.3 & 20.5 & 1.9 \\
A1 & SAT & \burn \etype{shift} L2 & RTX 4090 (6,7,8) & RTX PRO 6000 (1,2,3,4,5) & 15.8 & 14.0 & 1.7 \\
A1 & SAT & \burn clean & RTX 4090 (6,7,8) & RTX PRO 6000 (1,2,3,4,5) & 78.9 & 75.8 & 3.1 \\
A1 & DOFA & \burn \etype{boundary} L1 & GB10 (4,5) & RTX 4090 (1,2,3) & 3.5 & 2.9 & 0.6 \\
A1 & DOFA & \burn \etype{shift} L1 & GB10 (4,5) & RTX 4090 (1,2,3) & 1.2 & 1.5 & $-$0.3 \\
A1 & DOFA & \burn \etype{boundary} L2 & RTX 4090 (1,2,3,4,5) & RTX 5060 Ti (6,7,8) & 14.4 & 15.2 & $-$0.9 \\
A1 & DOFA & \burn \etype{coarse} L2 & RTX 4090 (1,2,3,4,5) & RTX 5060 Ti (6,7,8) & 20.6 & 17.4 & 3.2 \\
A1 & DOFA & \burn \etype{omission} L2 & RTX 4090 (1,2,3,4,5) & RTX 5060 Ti (6,7,8) & 25.3 & 22.9 & 2.4 \\
A1 & DOFA & \burn \etype{shift} L2 & RTX 4090 (1,2,3,4,5) & RTX 5060 Ti (6,7,8) & 11.2 & 10.4 & 0.8 \\
A1 & DOFA & \burn clean & RTX 4090 (1,2,3,4,5) & RTX 5060 Ti (6,7,8) & 85.5 & 85.5 & 0.1 \\
A1 & RN & \burn \etype{boundary} L2 & GB10 (1,2,3,4,5) & RTX 3080 (7,8) & 14.2 & 13.5 & 0.7 \\
A1 & RN & \burn \etype{coarse} L2 & GB10 (1,2,3,4,5) & RTX 3080 (7,8) & 18.4 & 17.1 & 1.3 \\
A1 & RN & \burn \etype{omission} L2 & GB10 (1,2,3,4,5) & RTX 3080 (7,8) & 19.5 & 15.8 & 3.7 \\
A1 & RN & \burn \etype{shift} L2 & GB10 (1,2,3,4,5) & RTX 3080 (7,8) & 11.4 & 13.9 & $-$2.5 \\
A1 & RN & \burn clean & GB10 (1,2,3,4,5) & RTX 3080 (7,8) & 88.1 & 89.2 & $-$1.0 \\
A1 & TM & \burn \etype{boundary} L1 & GB10 (3,4,5) & RTX 4090 (1,2) & 2.6 & 1.7 & 0.8 \\
A1 & TM & \burn \etype{shift} L1 & GB10 (3,4,5) & RTX 4090 (1,2) & 3.1 & 3.8 & $-$0.7 \\
A1 & TM & \burn \etype{boundary} L2 & RTX 4090 (1,2,3,4,5) & H100 NVL (6,7,8) & 10.6 & 14.0 & $-$3.4 \\
A1 & TM & \burn \etype{coarse} L2 & RTX 4090 (1,2,3,4,5) & H100 NVL (6,7,8) & 19.0 & 16.0 & 2.9 \\
A1 & TM & \burn \etype{omission} L2 & RTX 4090 (1,2,3,4,5) & H100 NVL (6,7,8) & 12.5 & 9.6 & 2.9 \\
A1 & TM & \burn \etype{shift} L2 & RTX 4090 (1,2,3,4,5) & H100 NVL (6,7,8) & 9.7 & 10.0 & $-$0.3 \\
A1 & TM & \burn clean & RTX 4090 (1,2,3,4,5) & H100 NVL (6,7,8) & 86.2 & 86.6 & $-$0.5 \\
N6 & SAT & \burn \etype{boundary} L2 & RTX 4090 (1,2) & RTX PRO 6000 (3,4,5) & 12.3 & 9.9 & 2.3 \\
N6 & SAT & \burn \etype{omission} L2 & RTX 4090 (1,2) & RTX PRO 6000 (3,4,5) & 19.6 & 16.5 & 3.1 \\
N6 & SAT & \burn \etype{shift} L2 & RTX 4090 (1,2) & RTX PRO 6000 (3,4,5) & 13.4 & 10.9 & 2.5 \\
N6 & SAT & \burn clean & RTX 4090 (2) & RTX PRO 6000 (1,3,4,5) & 81.2 & 81.4 & $-$0.1 \\
A1 & LVD & \flair \etype{boundary} L1 & RTX 4090 (3,4,5) & RTX PRO 6000 (1,2) & 0.1 & 2.2 & $-$2.2 \\
A1 & LVD & \flair \etype{shift} L1 & RTX 4090 (3,4,5) & RTX PRO 6000 (1,2) & $-$0.6 & 1.6 & $-$2.3 \\
A1 & LVD & \flair \etype{boundary} L2 & RTX 4090 (6,7,8) & RTX PRO 6000 (1,2,3,4,5) & 3.4 & 4.3 & $-$0.9 \\
A1 & LVD & \flair \etype{coarse} L2 & RTX 4090 (6,7,8) & RTX PRO 6000 (1,2,3,4,5) & 4.8 & 4.8 & 0.0 \\
A1 & LVD & \flair \etype{omission} L2 & RTX 4090 (6,7,8) & RTX PRO 6000 (1,2,3,4,5) & 3.2 & 3.6 & $-$0.4 \\
A1 & LVD & \flair \etype{shift} L2 & RTX 4090 (6,7,8) & RTX PRO 6000 (1,2,3,4,5) & 2.1 & 1.7 & 0.4 \\
A1 & LVD & \flair clean & RTX 4090 (7,8) & RTX PRO 6000 (1,2,3,4,5,6) & 50.1 & 50.9 & $-$0.9 \\
A1 & SAT & \flair \etype{boundary} L1 & RTX 4090 (3,4,5) & RTX PRO 6000 (1,2) & 2.8 & 2.9 & $-$0.1 \\
A1 & SAT & \flair \etype{shift} L1 & RTX 4090 (3,4,5) & RTX PRO 6000 (1,2) & 2.0 & 0.4 & 1.6 \\
A1 & SAT & \flair \etype{boundary} L2 & RTX 4090 (6,7,8) & RTX PRO 6000 (1,2,3,4,5) & 6.5 & 7.1 & $-$0.6 \\
A1 & SAT & \flair \etype{coarse} L2 & RTX 4090 (6,7,8) & RTX PRO 6000 (1,2,3,4,5) & 6.4 & 6.3 & 0.1 \\
A1 & SAT & \flair \etype{omission} L2 & RTX 4090 (6,7,8) & RTX PRO 6000 (1,2,3,4,5) & 7.7 & 8.2 & $-$0.5 \\
A1 & SAT & \flair \etype{shift} L2 & RTX 4090 (6,7,8) & RTX PRO 6000 (1,2,3,4,5) & 3.5 & 4.4 & $-$0.9 \\
A1 & SAT & \flair clean & RTX 4090 (7,8) & RTX PRO 6000 (1,2,3,4,5,6) & 52.6 & 52.3 & 0.3 \\
A1 & DOFA & \flair \etype{boundary} L1 & GB10 (4,5) & RTX 4090 (1,2,3) & 2.5 & 1.3 & 1.2 \\
A1 & DOFA & \flair \etype{shift} L1 & GB10 (4,5) & RTX 4090 (1,2,3) & 0.1 & 0.5 & $-$0.4 \\
A1 & TM & \flair \etype{boundary} L1 & GB10 (4,5) & RTX 4090 (1,2,3) & 2.2 & 1.0 & 1.2 \\
A1 & TM & \flair \etype{shift} L1 & GB10 (4,5) & RTX 4090 (1,2,3) & 0.7 & 1.1 & $-$0.4 \\
N6 & SAT & \flair \etype{boundary} L2 & RTX 4090 (1,2,3) & RTX PRO 6000 (4,5) & 5.9 & 5.9 & 0.0 \\
N6 & SAT & \flair \etype{coarse} L2 & RTX 4090 (1,2,3) & RTX PRO 6000 (4,5) & 5.6 & 5.1 & 0.5 \\
N6 & SAT & \flair \etype{omission} L2 & RTX 4090 (3) & RTX PRO 6000 (1,2,4,5) & 5.1 & 6.4 & $-$1.3 \\
N6 & SAT & \flair \etype{shift} L2 & RTX 4090 (1,2,3) & RTX PRO 6000 (4,5) & 2.9 & 3.1 & $-$0.2 \\
N6 & SAT & \flair clean & RTX 4090 (2,3) & RTX PRO 6000 (1,4,5) & 50.8 & 50.3 & 0.6 \\
\bottomrule
\end{tabular}
}{\tabletbd[6cm]{generated/table\_hardware\_mix.tex: run the analysis harness}}
\caption{\textbf{Cells with runs on two GPU classes:} Every main-grid (A1) and matched-rate (N6) cell of a core model whose analyzed runs span two GPU classes, with the seeds run on each class. A and B are means over the seeds of that class. For clean cells they are the test \miou and for noisy cells the paired drop in points. The seeds differ between classes, so A $-$ B mixes the seed effect with the hardware term. Where a cell spans three classes, the two with the most seeds are printed. The \rn cells on \burn are such cells, because seed 6 ran on a third class. Models are abbreviated as SAT (\sat), LVD (\lvd), DOFA (\dofa), TM (\tm) and RN (\rn). \ProvSlot{\ResultsProvenance}}
\label{tab:supp-hwmix}
\end{table*}

\ifdefined\cvprbuild\else
\Cref{tab:supp-blocks} lists the experiment blocks, \Cref{fig:supp-roster} the nine encoders, \Cref{tab:supp-models} their input bands and \Cref{tab:supp-hardware,tab:supp-manifest} the planned GPU class and the recorded manifest.
\fi
\ifdefined\cvprbuild
\Cref{tab:supp-blocks} lists the experiment blocks, \Cref{fig:supp-roster} the nine encoders and \Cref{tab:supp-models} the input bands.
\fi

A mid-run reassignment moved \HwMixedModelsAOne{} core models across GPU classes for part of their runs.
The N5 seeds \hlSeedExtraSeeds{} of the DINOv3 pair also ran on a second class.
On \burn all three ran there, whereas on \flair seed 6 of the two clean cells ran on the planned class.
As a result \hlHwMixedAOneNburn{} main-grid cells on \burn and \hlHwMixedAOneNflair{} on \flair mix GPU classes.
They are the \HwMixedCondsAOne{} cells of \sat, \lvd, \dofa and \tm and the clean and \Ltwo cells of the DINOv3 pair. On \burn, seeds 6 to 8 of \dofa, \tm and \rn add their clean and \Ltwo cells.
The N6 cells with two classes are \hlHwNSixCellsburn{} on \burn. On \flair they are \hlHwNSixCellsflair.
All \hlHwSThreeCellsNburn{} \burn and \hlHwSThreeCellsNflair{} \flair cells of \prithvi hold runs from three GPU classes\ifdefined\cvprbuild\else{} (\Cref{sec:supp-stretch})\fi.
\Cref{tab:supp-hwmix} gives the within-cell means per class.
Over the noisy cells, the absolute difference between the class means of the drop averages \hlHwDropDiffMeanburn{} points on \burn and \hlHwDropDiffMeanflair{} on \flair. It is at most \hlHwDropDiffMaxburn{} and \hlHwDropDiffMaxflair.
The signed mean is \hlHwDropDiffSignedMeanburn{} on \burn and \hlHwDropDiffSignedMeanflair{} on \flair over cells whose class pairs and directions differ. It is therefore not an estimate of one class difference.
In \Lone cells with one or two seeds per class the seed effect dominates, and the largest differences sit there.
N5 adds seeds to the F1 arms only and not to N6, so the matched-rate contrast stays at five seeds with its hardware term.
\ifdefined\cvprbuild\else
Blocks S1, S3, S4 and S6 ran off the class planned in \Cref{tab:supp-hardware}.
\fi
\ifdefined\cvprbuild
The extension blocks S3, S4 and S6 ran off their planned GPU class.
\fi

The compute footprint of the main paper is the summed wall time of the \hlNRunsTotal{} finished runs. It is \hlGpuHours{} GPU-hours over seven GPU classes (\Cref{tab:supp-hw-blocks}).
S4 ran on \hlHwBlockClassesSFour{} rather than on GB10, with run counts in parentheses. S6 ran on \hlHwBlockClassesSSix{} rather than on \ifdefined\cvprbuild RTX 4090\else RTX 4090 (\Cref{tab:supp-manifest})\fi{}.
Learning rates are the clean-label values held fixed across conditions, which may disadvantage a model whose best rate shifts under noise \cite{xu2024specialized}.

Rows A, N1 to N6 and S of \Cref{tab:supp-blocks} plan \RunsTotal{} runs, \RunsCore{} in the core design and \RunsStretch{} in N5 and the extensions.
\ifdefined\cvprbuild\else
\Cref{tab:supp-hardware} assigns these runs, without N6, to GPU classes.
\fi
\ifdefined\cvprbuild
Each model was planned on one GPU class, and \Cref{tab:supp-hw-blocks,tab:supp-hwmix} show the classes the runs used.
\fi
Of the \hlNRunsTotal{} finished runs, \hlNRunsExcluded{} are excluded and \hlNRunsAnalysed{} enter the analysis. The excluded runs are the probe and timing runs and the hardware replicates of \Cref{tab:supp-hwrep} (blocks \hlNRunsExcludedBlocks).
Of the analyzed runs, \hlNRunsCoreAnalysed{} are the complete core design with N5 and \hlNRunsExtAnalysed{} the six extension blocks (\Cref{sec:supp-stretch}).
Blocks N7 to N15 add \hlNRunsPostReviewAnalysed{} analyzed runs and blocks D1 to D7 add \hlNRunsExtDatasetsAnalysed. Blocks N16, N18, N19 and N20 add \hlNRunsReviewAnalysed{} (\Cref{sec:supp-robust,sec:supp-layers}).

\ifdefined\plannedbuild\ifdefined\arxivbuild\IfFileExists{planned/sec/supp_arxiv_extra.tex}{%
\IfFileExists{planned/generated/macros_ax.tex}{%
\newcommand{\AxGsdChes}{1}  %
\newcommand{\AxClassesChes}{four}  %
\newcommand{\AxTilePxChes}{256}  %
\newcommand{\AxTrainChes}{1{,}000}  %
\newcommand{\AxValChes}{250}  %
\newcommand{\AxTestChes}{1{,}000}  %
\newcommand{\AxGsdCsTwelve}{10}  %
\newcommand{\AxClassesCsTwelve}{four}  %
\newcommand{\AxTilePxCsTwelve}{512}  %
\newcommand{\AxTrainCsTwelve}{1{,}000}  %
\newcommand{\AxValCsTwelve}{250}  %
\newcommand{\AxTestCsTwelve}{975}  %
\newcommand{\AxGsdDfc}{10}  %
\newcommand{\AxClassesDfc}{eight}  %
\newcommand{\AxTilePxDfc}{256}  %
\newcommand{\AxTrainDfc}{886}  %
\newcommand{\AxValDfc}{100}  %
\newcommand{\AxTestDfc}{5{,}128}  %
\newcommand{\AxGsdEnva}{1}  %
\newcommand{\AxClassesEnva}{four}  %
\newcommand{\AxTilePxEnva}{256}  %
\newcommand{\AxTrainEnva}{1{,}000}  %
\newcommand{\AxValEnva}{250}  %
\newcommand{\AxTestEnva}{1{,}000}  %
\newcommand{\AxGsdSnTwo}{0.3}  %
\newcommand{\AxClassesSnTwo}{two}  %
\newcommand{\AxTilePxSnTwo}{512}  %
\newcommand{\AxTrainSnTwo}{1{,}000}  %
\newcommand{\AxValSnTwo}{250}  %
\newcommand{\AxTestSnTwo}{2{,}961}  %
\newcommand{\AxGsdKuro}{10}  %
\newcommand{\AxClassesKuro}{three}  %
\newcommand{\AxTilePxKuro}{224}  %
\newcommand{\AxTrainKuro}{1{,}000}  %
\newcommand{\AxValKuro}{250}  %
\newcommand{\AxTestKuro}{2{,}000}  %
\newcommand{\AxGsdOem}{0.25--0.5}  %
\newcommand{\AxClassesOem}{eight}  %
\newcommand{\AxTilePxOem}{512}  %
\newcommand{\AxTrainOem}{1{,}000}  %
\newcommand{\AxValOem}{250}  %
\newcommand{\AxTestOem}{2{,}000}  %
\newcommand{\AxOptimizer}{AdamW}  %
\newcommand{\AxPrecision}{BF16 mixed}  %
\newcommand{\AxSchedule}{constant}  %
\newcommand{\AxCardTiles}{four}  %
\newcommand{\AxCardSeed}{1}  %
\newcommand{\AxMinPairs}{three}  %
\newcommand{\ProvTableAxModels}{Analysis of the finished runs}
\newcommand{\ProvTableAxHparams}{Analysis of the finished runs}
\newcommand{\ProvTableAxHparamsExt}{Analysis of the finished runs}
\newcommand{\ProvTableAxGridChes}{Seeds per cell 5. Analysis of the finished runs}
\newcommand{\ProvTableAxGridCstwelve}{Seeds per cell 5. Analysis of the finished runs}
\newcommand{\ProvTableAxGridDfc}{Seeds per cell 5. Analysis of the finished runs}
\newcommand{\ProvTableAxGridEnva}{Seeds per cell 5. Analysis of the finished runs}
\newcommand{\ProvTableAxGridSntwo}{Seeds per cell 5. Analysis of the finished runs}
\newcommand{\ProvTableAxGridKuro}{Seeds per cell 5. Analysis of the finished runs}
\newcommand{\ProvTableAxGridOem}{Seeds per cell 5. Analysis of the finished runs}
}{}
\IfFileExists{planned/generated/macros_ax_effects.tex}{%
\newcommand{\ProvTableAxEffectSizes}{Analysis of the finished runs}
}{}

\section{Dataset Cards}
\label{sec:ax-cards}

This section gives one card per dataset, ten in all.
Each card states the sensor, the classes, the splits used and the license, and shows \AxCardTiles{} training tiles.
The tiles are the first informative tiles of a shuffle of the training tile names with seed \AxCardSeed.
A tile counts as informative when its reference label holds at least two classes.
On a dataset with a real label pair the card also shows the label that the real arm trains on.
The cards make visible what a label \miou means on each dataset, which the calibration tables state only as a number.

\paragraph{HLS Burn Scars:}
In its \gbtwo version, \burn \cite{jakubik2023prithvi,simumba2025geobench2} holds harmonized Landsat and Sentinel-2 tiles at \ResBurnM\,m with \NBurnBands{} spectral bands.
Its label is a binary burn mask with a no-data value that training ignores.
We train on all \NBurnTrain{} training tiles, validate on \NBurnVal{} and test on \NBurnTest{} tiles of \TilePx\,px.
The dataset is released under CC BY 4.0 and has no second label, so it carries the simulated types only (\Cref{fig:ax-card-burn}).

\begin{figure}[t]
\centering
\IfFileExists{planned/figs/fig_card_burn_scars.pdf}{\includegraphics[width=\linewidth]{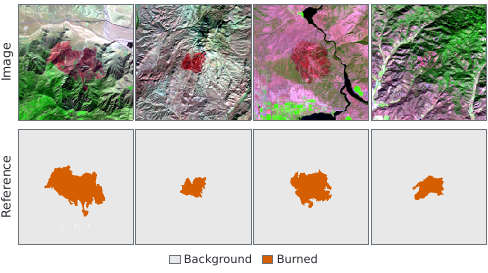}}{\phfig{4cm}{planned/figs/fig\_card\_burn\_scars.pdf from figs/make\_dataset\_cards.py}}
\caption{\textbf{Training set examples of HLS Burn Scars:} The top row shows a false-color composite of the second shortwave infrared, near infrared and red bands. The bottom row shows the reference label of the same tiles. Burn scars appear dark red in this composite.}
\label{fig:ax-card-burn}
\end{figure}

\paragraph{FLAIR-2:}
Aerial tiles of France at \ResFlairM\,m make up the \gbtwo version of \flair \cite{garioud2023flair2,simumba2025geobench2}, with RGB, near infrared and elevation channels.
Its label has \NFlairClasses{} land-cover classes.
The study trains on a fixed subset of \NFlairTrain{} tiles, validates on \NFlairVal{} and tests on the full \NFlairTest-tile split.
The dataset is released under the Etalab Open Licence 2.0 and has no second label (\Cref{fig:ax-card-flair}).

\begin{figure}[t]
\centering
\IfFileExists{planned/figs/fig_card_flair2.pdf}{\includegraphics[width=\linewidth]{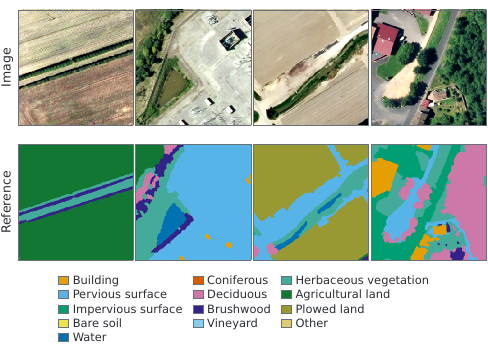}}{\phfig{4cm}{planned/figs/fig\_card\_flair2.pdf from figs/make\_dataset\_cards.py}}
\caption{\textbf{Training set examples of FLAIR-2:} The top row shows the RGB channels and the bottom row the reference label of the same tiles.}
\label{fig:ax-card-flair}
\end{figure}

\paragraph{Sen1Floods11:}
The hand-labeled part of \sfloods \cite{bonafilia2020sen1floods11} holds Sentinel-2 chips at \ResSOneM\,m with \NSOneBands{} bands and a binary water mask.
The release split gives \NSOneTrain{} training, \NSOneVal{} validation and \NSOneTest{} test chips of \TilePx\,px.
Each chip also carries an Otsu water mask from the coincident Sentinel-1 scene, the real label of this dataset.
The Otsu masks score a label \miou of \hlOtsuLabelMiouFracAll{} against the hand masks, at a flip rate of \hlOtsuFlipRateAll\,\%.
The mirror on Hugging Face states Apache 2.0 (\Cref{fig:ax-card-sen}).

\begin{figure}[t]
\centering
\IfFileExists{planned/figs/fig_card_sen1floods11.pdf}{\includegraphics[width=\linewidth]{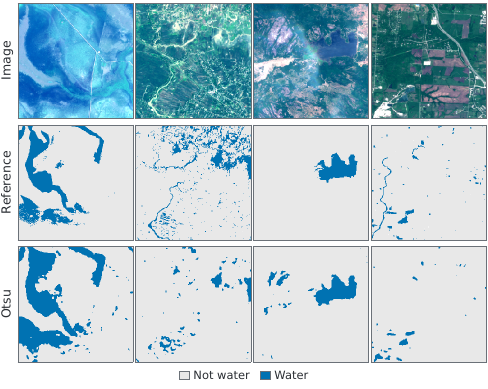}}{\phfig{4cm}{planned/figs/fig\_card\_sen1floods11.pdf from figs/make\_dataset\_cards.py}}
\caption{\textbf{Training set examples of Sen1Floods11:} The top row shows the RGB bands of the Sentinel-2 chip. The middle row shows the reference label and the bottom row the Otsu label of the same tiles. White marks pixels without a label.}
\label{fig:ax-card-sen}
\end{figure}

\paragraph{Chesapeake:}
Chesapeake \cite{robinson2019large} pairs NAIP aerial images at \AxGsdChes\,m with a land-cover label of \AxClassesChes{} classes.
The study draws \AxTilePxChes\,px chips once from the Delaware tiles of the release, \AxTrainChes{} for training, \AxValChes{} for validation and \AxTestChes{} for testing.
The real label is the 30\,m NLCD product that the release ships on the same grid.
It scores a label \miou of \RealLabelMiouChes{} against the reference label, at a flip rate of \RealFlipRateChes\,\%.
The dataset is released under CDLA-Permissive 1.0 (\Cref{fig:ax-card-ches}).

\begin{figure}[t]
\centering
\IfFileExists{planned/figs/fig_card_chesapeake.pdf}{\includegraphics[width=\linewidth]{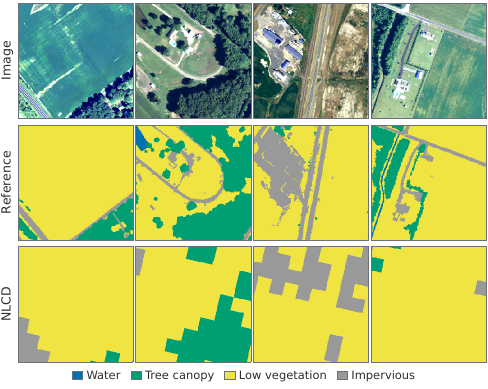}}{\phfig{4cm}{planned/figs/fig\_card\_chesapeake.pdf from figs/make\_dataset\_cards.py}}
\caption{\textbf{Training set examples of Chesapeake:} The top row shows the RGB channels of the NAIP image. The middle row shows the reference label and the bottom row the NLCD label of the same tiles.}
\label{fig:ax-card-ches}
\end{figure}

\paragraph{CloudSEN12:}
CloudSEN12 \cite{aybar2022cloudsen12,simumba2025geobench2} enters the study in its \gbtwo version, with Sentinel-2 patches at \AxGsdCsTwelve\,m and expert masks of \AxClassesCsTwelve{} classes.
The classes are clear sky, thick cloud, thin cloud and cloud shadow.
We train on \AxTrainCsTwelve{} patches, validate on \AxValCsTwelve{} and test on the \AxTestCsTwelve{} test patches.
The real label is the s2cloudless mask of the release, which has no thin cloud or shadow class.
It scores a label \miou of \RealLabelMiouCsTwelve{} at a flip rate of \RealFlipRateCsTwelve\,\%.
CloudSEN12 is released under CC0 1.0 (\Cref{fig:ax-card-cs}).

\begin{figure}[t]
\centering
\IfFileExists{planned/figs/fig_card_cloudsen12.pdf}{\includegraphics[width=\linewidth]{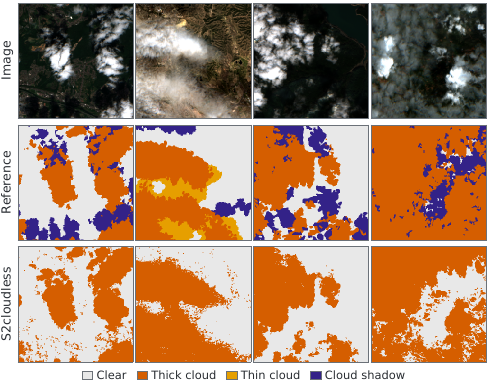}}{\phfig{4cm}{planned/figs/fig\_card\_cloudsen12.pdf from figs/make\_dataset\_cards.py}}
\caption{\textbf{Training set examples of CloudSEN12:} The top row shows the RGB bands of the Sentinel-2 patch. The middle row shows the reference label and the bottom row the s2cloudless label of the same tiles.}
\label{fig:ax-card-cs}
\end{figure}

\paragraph{DFC2020:}
DFC2020 \cite{schmitt2019sen12ms,schmitt2020weakly} holds Sentinel-2 and Sentinel-1 patches at \AxGsdDfc\,m with a semi-manual label of \AxClassesDfc{} land-cover classes.
The study splits the validation patches of the release into \AxTrainDfc{} training and \AxValDfc{} validation patches.
It tests on the \AxTestDfc{} test patches of the release.
The real label is the MODIS land-cover label of SEN12MS, resampled from 500\,m to the patch grid.
It scores a label \miou of \RealLabelMiouDfc{} at a flip rate of \RealFlipRateDfc\,\% on the patches it covers.
The dataset is released under CC BY 4.0 (\Cref{fig:ax-card-dfc}).

\begin{figure}[t]
\centering
\IfFileExists{planned/figs/fig_card_dfc2020.pdf}{\includegraphics[width=\linewidth]{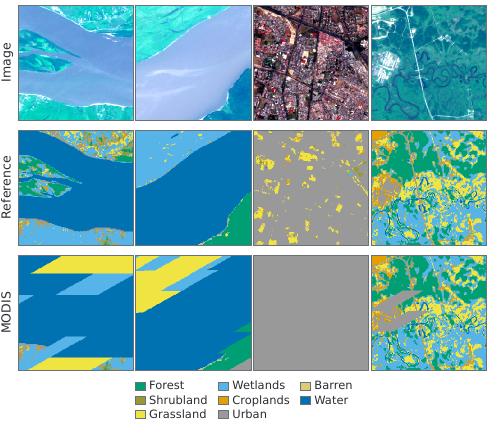}}{\phfig{4cm}{planned/figs/fig\_card\_dfc2020.pdf from figs/make\_dataset\_cards.py}}
\caption{\textbf{Training set examples of DFC2020:} The top row shows the RGB bands of the Sentinel-2 patch. The middle row shows the reference label and the bottom row the MODIS label of the same tiles. White marks pixels without a label.}
\label{fig:ax-card-dfc}
\end{figure}

\paragraph{EnviroAtlas:}
EnviroAtlas \cite{rolf2022resolving} pairs NAIP aerial images at \AxGsdEnva\,m with a land-cover label, collapsed here to the \AxClassesEnva{} Chesapeake classes.
The study draws \AxTrainEnva{} training and \AxValEnva{} validation chips of \AxTilePxEnva\,px from Pittsburgh.
Durham, Austin and Phoenix supply the \AxTestEnva{} test chips, so the test is across cities.
The real label is the 30\,m NLCD product, as on Chesapeake.
It scores a label \miou of \RealLabelMiouEnva{} at a flip rate of \RealFlipRateEnva\,\%.
EnviroAtlas is released under CC BY 4.0 (\Cref{fig:ax-card-enva}).

\begin{figure}[t]
\centering
\IfFileExists{planned/figs/fig_card_enviroatlas.pdf}{\includegraphics[width=\linewidth]{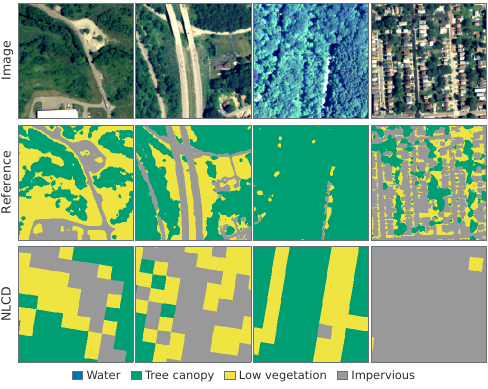}}{\phfig{4cm}{planned/figs/fig\_card\_enviroatlas.pdf from figs/make\_dataset\_cards.py}}
\caption{\textbf{Training set examples of EnviroAtlas:} The top row shows the RGB channels of the NAIP image. The middle row shows the reference label and the bottom row the NLCD label of the same tiles.}
\label{fig:ax-card-enva}
\end{figure}

\paragraph{SpaceNet 2:}
SpaceNet 2 \cite{vanetten2018spacenet,simumba2025geobench2} in its \gbtwo version holds pansharpened WorldView-3 chips at \AxGsdSnTwo\,m with building footprints.
Its label has \AxClassesSnTwo{} classes.
Each chip is cropped at its center to \AxTilePxSnTwo\,px.
The study trains on \AxTrainSnTwo{} chips, validates on \AxValSnTwo{} and tests on the \AxTestSnTwo{} test chips.
The dataset is released under CC BY-SA 4.0 and has no second label (\Cref{fig:ax-card-sn}).

\begin{figure}[t]
\centering
\IfFileExists{planned/figs/fig_card_spacenet2.pdf}{\includegraphics[width=\linewidth]{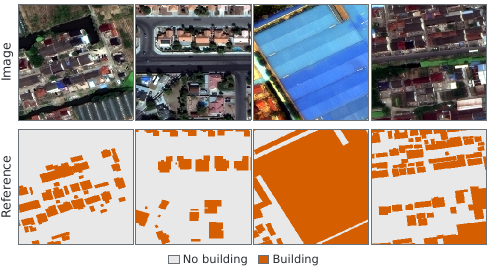}}{\phfig{4cm}{planned/figs/fig\_card\_spacenet2.pdf from figs/make\_dataset\_cards.py}}
\caption{\textbf{Training set examples of SpaceNet 2:} The top row shows the RGB bands of the WorldView-3 chip and the bottom row the reference label of the same tiles.}
\label{fig:ax-card-sn}
\end{figure}

\paragraph{Kuro Siwo:}
From Kuro Siwo \cite{bountos2024kurosiwo,simumba2025geobench2} we take the \gbtwo version, which holds Sentinel-1 patches at \AxGsdKuro\,m with two polarizations and an elevation layer.
The \AxClassesKuro{} classes of the expert label are no water, permanent water and flood.
We train on \AxTrainKuro{} patches of \AxTilePxKuro\,px, validate on \AxValKuro{} and test on the \AxTestKuro{} test patches.
The real label is the Copernicus EMS polygon set that the experts started from, rasterized on the patch grid.
It scores a label \miou of \RealLabelMiouKuro{} at a flip rate of \RealFlipRateKuro\,\% on the patches it covers.
Kuro Siwo is released under CC BY 4.0 (\Cref{fig:ax-card-kuro}).

\begin{figure}[t]
\centering
\IfFileExists{planned/figs/fig_card_kuro_siwo.pdf}{\includegraphics[width=\linewidth]{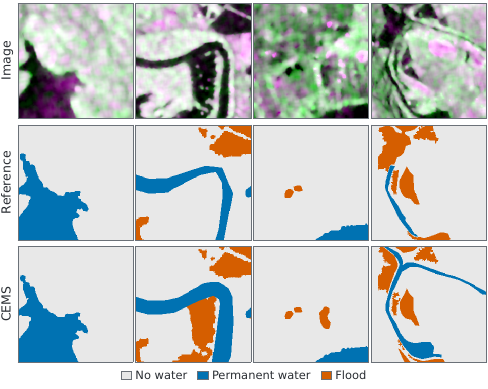}}{\phfig{4cm}{planned/figs/fig\_card\_kuro\_siwo.pdf from figs/make\_dataset\_cards.py}}
\caption{\textbf{Training set examples of Kuro Siwo:} The top row shows a false-color composite of the VV and VH backscatter in decibels. The middle row shows the reference label and the bottom row the Copernicus EMS (CEMS) label of the same tiles.}
\label{fig:ax-card-kuro}
\end{figure}

\paragraph{OpenEarthMap:}
OpenEarthMap \cite{xia2023openearthmap} holds aerial and satellite RGB images at \AxGsdOem\,m with a label of \AxClassesOem{} land-cover classes.
The study cuts the images of the official lists into tiles of \AxTilePxOem\,px and leaves out the images sourced from xBD.
It trains on \AxTrainOem{} tiles, validates on \AxValOem{} and tests on \AxTestOem{} tiles.
The license is CC BY 4.0 or CC BY-NC-SA 4.0 by region.
OpenEarthMap has no second label (\Cref{fig:ax-card-oem}).

\begin{figure}[t]
\centering
\IfFileExists{planned/figs/fig_card_openearthmap.pdf}{\includegraphics[width=\linewidth]{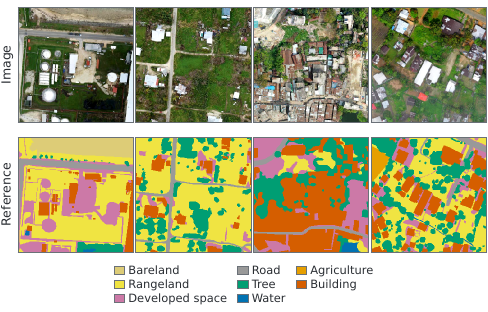}}{\phfig{4cm}{planned/figs/fig\_card\_openearthmap.pdf from figs/make\_dataset\_cards.py}}
\caption{\textbf{Training set examples of OpenEarthMap:} The top row shows the RGB image and the bottom row the reference label of the same tiles. White marks pixels without a label.}
\label{fig:ax-card-oem}
\end{figure}

\paragraph{Reading the Cards:}
The real labels differ in kind from one pair to the next.
NLCD and MODIS labels are coarse and blocky, as the \etype{coarse} type simulates.
Two classes are missing from the s2cloudless label.
The Otsu and Copernicus EMS labels move water outlines and miss small water bodies.
The pairs therefore span the simulated types without matching a single one, which is why \mtab{tab:ext-datasets} reports their label \miou. %

\section{Model Characteristics}
\label{sec:ax-models}

\begin{table*}[t]
\centering\footnotesize\setlength{\tabcolsep}{4pt}
\IfFileExists{planned/generated/table_ax_models.tex}{\begin{adjustbox}{max width=\textwidth}%
\providecommand{\gcvdry}[1]{\textcolor{red}{#1}}
\providecommand{\ProvTableAxModels}{Analysis of the finished runs}
\begin{tabular}{@{}lllllllc@{}}
\toprule
Model & Type & Param.\ (M) & Technique & Pretraining data & Bands, \burn & Bands, \flair & LR mult. \\
\midrule
\sat & ViT-L/16 & 323.4 & Self-distillation & SAT-493M satellite images & RGB & RGB & $\times$1 \\
\lvd & ViT-L/16 & 323.4 & Self-distillation & LVD-1689M web images & RGB & RGB & $\times$1/3 \\
\dofa & ViT-L/16 & 357.5 & Masked image modeling & Five Earth observation sensors & 6 bands & RGB, NIR & $\times$1 \\
\tm & ViT-L/16 & 323.9 & Masked token and pixel modeling & TerraMesh, nine modalities & 6 bands & RGB, NIR & $\times$1 \\
\rn & ResNet-50 & 43.8 & Supervised classification & ImageNet-1k & 6 bands & RGB, NIR, elevation & $\times$1 \\
\midrule
\prithvi & ViT-L/16 & 324.2 & Masked autoencoding & HLS time series & 6 bands & RGB, NIR & $\times$1 \\
\rnssl & ResNet-50 & 43.8 & Contrastive (MoCo) & SSL4EO-S12 Sentinel-2 & 6 bands & RGB, NIR, elevation & $\times$1 \\
\vitr & ViT-L/16 & 323.4 & None & None & RGB & RGB & $\times$1 / $\times$3 \\
\cnxl & ConvNeXt-L & 212.6 & Distillation from the ViT & LVD-1689M web images & RGB & RGB & $\times$1 \\
\bottomrule
\end{tabular}
\end{adjustbox}}{\tabletbd[3cm]{planned/generated/table\_ax\_models.tex: run scripts/make\_arxiv\_appendix.py}}
\caption{\textbf{Characteristics of the nine encoders:} Backbone type, parameter count and pretraining of each encoder, then its input bands on the two core datasets and its learning-rate multiplier. The parameter count covers the encoder with the UNet decoder on \burn, in millions. The multiplier scales the base learning rate of \Cref{tab:ax-hparams,tab:ax-hparams-ext} in the main grid, and two values give \burn and then \flair. NIR is near infrared. The first five rows are the core models and the last four the extension encoders. \ProvSlot{\ProvTableAxModels}}
\label{tab:ax-models}
\end{table*}

\Cref{tab:ax-models} lists the nine encoders with the factor that each one contributes to the design.
Six encoders share the ViT-L/16 backbone, so the contrasts among them vary pretraining at one capacity.
The two \rn rows hold capacity at a far smaller scale and differ in pretraining and in learning rate and batch size.
The DINOv3 encoders read RGB on every dataset, whereas the multispectral models read every band that their pretraining supports.
Input bands are therefore confounded with the encoder in each contrast between an RGB model and a multispectral model. %
\mtab{tab:all-models} gives the clean and noisy scores of the same nine rows. %

\section{Hyperparameters}
\label{sec:ax-hparams}

\begin{table*}[t]
\centering\footnotesize\setlength{\tabcolsep}{4pt}
\IfFileExists{planned/generated/table_ax_hparams.tex}{\begin{adjustbox}{max width=\textwidth}%
\providecommand{\gcvdry}[1]{\textcolor{red}{#1}}
\providecommand{\ProvTableAxHparams}{Analysis of the finished runs}
\begin{tabular}{@{}lccccccccc@{}}
\toprule
Dataset & Tile (px) & Crop (px) & Test grid & SAT & LVD & DOFA & TM & RN & LVD mult. \\
\midrule
\burn & 512 & 224 & 3 $\times$ 3 & 6.82 / 16 & 6.82 / 16 & 1.86 / 32 & 7.45 / 32 & 92.20 / 32 & $\times$1/3 \\
\flair & 512 & 224 & 3 $\times$ 3 & 2.02 / 32 & 2.02 / 32 & 0.69 / 8 & 2.16 / 32 & 19.60 / 16 & $\times$1/3 \\
\midrule
\sfloods & 512 & 224 & 3 $\times$ 3 & 6.82 / 16 & 6.82 / 16 & 1.86 / 32 & 7.45 / 32 & 92.20 / 32 & $\times$1 \\
\midrule
Chesapeake & 256 & 224 & 2 $\times$ 2 & 2.02 / 32 & 2.02 / 32 & 0.69 / 8 & 2.16 / 32 & 19.60 / 16 & $\times$1/3 \\
CloudSEN12 & 512 & 224 & 3 $\times$ 3 & 6.82 / 16 & 6.82 / 16 & 1.86 / 32 & 7.45 / 32 & 92.20 / 32 & $\times$1/3 \\
DFC2020 & 256 & 224 & 2 $\times$ 2 & 6.82 / 16 & 6.82 / 16 & 1.86 / 32 & 7.45 / 32 & 92.20 / 32 & $\times$1/3 \\
EnviroAtlas & 256 & 224 & 2 $\times$ 2 & 2.02 / 32 & 2.02 / 32 & 0.69 / 8 & 2.16 / 32 & 19.60 / 16 & $\times$1/3 \\
SpaceNet 2 & 512 & 224 & 3 $\times$ 3 & 2.02 / 32 & 2.02 / 32 & 0.69 / 8 & 2.16 / 32 & 19.60 / 16 & $\times$1/3 \\
Kuro Siwo & 224 & 224 & 1 $\times$ 1 & 6.82 / 16 & 6.82 / 16 & 1.86 / 32 & 7.45 / 32 & 92.20 / 32 & $\times$1/3 \\
OpenEarthMap & 512 & 224 & 3 $\times$ 3 & 2.02 / 32 & 2.02 / 32 & 0.69 / 8 & 2.16 / 32 & 19.60 / 16 & $\times$1/3 \\
\bottomrule
\end{tabular}
\end{adjustbox}}{\tabletbd[3cm]{planned/generated/table\_ax\_hparams.tex: run scripts/make\_arxiv\_appendix.py}}
\caption{\textbf{Training recipe per dataset:} Tile size, training crop size and the grid of crops per test tile, then the base learning rate and batch size of each core model. A model cell gives the learning rate in units of $10^{-5}$ and the batch size. The last column is the multiplier that \lvd runs at, and every other core model runs at its base rate. Models are abbreviated as SAT (\sat), LVD (\lvd), DOFA (\dofa), TM (\tm) and RN (\rn). \ProvSlot{\ProvTableAxHparams}}
\label{tab:ax-hparams}
\end{table*}

\begin{table}[t]
\centering\footnotesize\setlength{\tabcolsep}{4pt}
\IfFileExists{planned/generated/table_ax_hparams_ext.tex}{\begin{adjustbox}{max width=\columnwidth}%
\providecommand{\gcvdry}[1]{\textcolor{red}{#1}}
\providecommand{\ProvTableAxHparamsExt}{Analysis of the finished runs}
\begin{tabular}{@{}lcccc@{}}
\toprule
& \multicolumn{2}{c}{\burn} & \multicolumn{2}{c}{\flair} \\
\cmidrule(lr){2-3} \cmidrule(lr){4-5}
Model & LR / batch & Mult. & LR / batch & Mult. \\
\midrule
\prithvi & 0.81 / 8 & $\times$1 & 2.73 / 8 & $\times$1 \\
\rnssl & 9.77 / 8 & $\times$1 & 4.04 / 16 & $\times$1 \\
\vitr & 6.82 / 16 & $\times$1 & 2.02 / 32 & $\times$3 \\
\cnxl & 1.87 / 32 & $\times$1 & 0.85 / 8 & $\times$1 \\
\bottomrule
\end{tabular}
\end{adjustbox}}{\tabletbd[2cm]{planned/generated/table\_ax\_hparams\_ext.tex: run scripts/make\_arxiv\_appendix.py}}
\caption{\textbf{Training recipe of the extension encoders:} Base learning rate in units of $10^{-5}$, batch size and learning-rate multiplier on the two core datasets. \ProvSlot{\ProvTableAxHparamsExt}}
\label{tab:ax-hparams-ext}
\end{table}

\paragraph{Shared Settings:}
Every run uses the \AxOptimizer{} optimizer with weight decay \WeightDecay{} and a \AxSchedule{} learning rate.
Training runs for at most \MaxEpochs{} epochs in \AxPrecision{} precision on random crops of \CropPx\,px.
Validation runs every \ValEvery{} epochs on the labels of the training condition.
Training stops after \PatienceChecks{} validation checks without a gain, and the checkpoint with the best validation \miou is tested.

\paragraph{Rates per Dataset:}
\Cref{tab:ax-hparams} gives the base learning rate and batch size of each core model on each dataset.
The two core datasets use the published \gbtwo values.
A dataset without a published value reuses the values of the core dataset nearest in sensor, which the table shows as repeated rows.
The rates were therefore not tuned on any corrupted label, and none was tuned on an extension dataset.
\Cref{tab:ax-hparams-ext} gives the same values for the four extension encoders, each with its own learning-rate probe.

\section{Results by Dataset}
\label{sec:ax-grids}

\Cref{tab:ax-grid-ches,tab:ax-grid-cs,tab:ax-grid-dfc,tab:ax-grid-enva,tab:ax-grid-sn,tab:ax-grid-kuro,tab:ax-grid-oem} print every core cell of the seven extension datasets.
They follow the layout of \Cref{tab:supp-grid-burn,tab:supp-grid-flair} for the two core datasets.
Each table lists the clean cell and the four types at both levels for the five core models.
On a pair dataset it adds the run trained on the real label.
The real cell of DFC2020 and Kuro Siwo is paired with a clean run on the covered tiles, as \Cref{sec:supp-ext} describes.
\mtab{tab:ext-datasets} gives the runs per block. %

\begin{table}[t]
\centering\footnotesize\setlength{\tabcolsep}{3pt}
\IfFileExists{planned/generated/table_ax_grid_ches.tex}{\begin{adjustbox}{max width=\columnwidth}%
\providecommand{\gcvdry}[1]{\textcolor{red}{#1}}
\providecommand{\ProvTableAxGridChes}{Seeds per cell 5. Analysis of the finished runs}
\begin{tabular}{@{}lllcccc@{}}
\toprule
Model & Type & Level & $n$ & mIoU & Drop & Retention (\%) \\
\midrule
SAT & Clean & -- & 5 & 81.3 $\pm$ 0.7 & -- & -- \\
 & Shift & L1 & 5 & 80.7 $\pm$ 1.0 & 0.5 $\pm$ 1.3 & 99.3 $\pm$ 1.7 \\
 & Shift & L2 & 5 & 74.9 $\pm$ 0.9 & 6.4 $\pm$ 1.2 & 92.1 $\pm$ 1.4 \\
 & Coarse & L1 & 5 & 80.7 $\pm$ 0.5 & 0.6 $\pm$ 0.8 & 99.2 $\pm$ 1.0 \\
 & Coarse & L2 & 5 & 69.5 $\pm$ 0.8 & 11.8 $\pm$ 0.7 & 85.4 $\pm$ 0.8 \\
 & Boundary & L1 & 5 & 79.1 $\pm$ 0.6 & 2.2 $\pm$ 0.8 & 97.4 $\pm$ 1.0 \\
 & Boundary & L2 & 5 & 69.3 $\pm$ 0.5 & 12.0 $\pm$ 0.8 & 85.2 $\pm$ 0.8 \\
 & Omission & L1 & 5 & 78.4 $\pm$ 1.4 & 2.9 $\pm$ 2.0 & 96.4 $\pm$ 2.4 \\
 & Omission & L2 & 5 & 70.9 $\pm$ 3.6 & 10.4 $\pm$ 4.0 & 87.2 $\pm$ 4.9 \\
 & NLCD & Real & 5 & 59.9 $\pm$ 0.4 & 21.4 $\pm$ 0.8 & 73.7 $\pm$ 0.9 \\
\addlinespace[1.5pt]
LVD & Clean & -- & 5 & 79.3 $\pm$ 1.1 & -- & -- \\
 & Shift & L1 & 5 & 78.6 $\pm$ 0.8 & 0.7 $\pm$ 1.0 & 99.2 $\pm$ 1.2 \\
 & Shift & L2 & 5 & 73.5 $\pm$ 0.8 & 5.8 $\pm$ 1.1 & 92.7 $\pm$ 1.3 \\
 & Coarse & L1 & 5 & 77.7 $\pm$ 0.8 & 1.6 $\pm$ 1.5 & 98.0 $\pm$ 1.9 \\
 & Coarse & L2 & 5 & 68.5 $\pm$ 0.5 & 10.8 $\pm$ 1.0 & 86.4 $\pm$ 1.1 \\
 & Boundary & L1 & 5 & 75.8 $\pm$ 1.1 & 3.5 $\pm$ 1.9 & 95.6 $\pm$ 2.4 \\
 & Boundary & L2 & 5 & 68.0 $\pm$ 1.1 & 11.3 $\pm$ 1.6 & 85.8 $\pm$ 2.0 \\
 & Omission & L1 & 5 & 75.4 $\pm$ 3.1 & 3.9 $\pm$ 2.2 & 95.1 $\pm$ 2.8 \\
 & Omission & L2 & 5 & 66.3 $\pm$ 1.7 & 13.1 $\pm$ 1.3 & 83.5 $\pm$ 1.7 \\
 & NLCD & Real & 5 & 59.1 $\pm$ 1.9 & 20.3 $\pm$ 2.6 & 74.5 $\pm$ 3.0 \\
\addlinespace[1.5pt]
DOFA & Clean & -- & 5 & 83.0 $\pm$ 0.2 & -- & -- \\
 & Shift & L1 & 5 & 81.8 $\pm$ 0.4 & 1.2 $\pm$ 0.3 & 98.5 $\pm$ 0.4 \\
 & Shift & L2 & 5 & 75.0 $\pm$ 0.9 & 8.0 $\pm$ 1.0 & 90.3 $\pm$ 1.2 \\
 & Coarse & L1 & 5 & 80.9 $\pm$ 0.7 & 2.1 $\pm$ 0.7 & 97.5 $\pm$ 0.8 \\
 & Coarse & L2 & 5 & 70.0 $\pm$ 0.6 & 13.0 $\pm$ 0.6 & 84.3 $\pm$ 0.8 \\
 & Boundary & L1 & 5 & 79.8 $\pm$ 0.6 & 3.2 $\pm$ 0.5 & 96.2 $\pm$ 0.6 \\
 & Boundary & L2 & 5 & 70.1 $\pm$ 2.2 & 12.9 $\pm$ 2.2 & 84.5 $\pm$ 2.6 \\
 & Omission & L1 & 5 & 79.1 $\pm$ 1.0 & 4.0 $\pm$ 1.0 & 95.2 $\pm$ 1.2 \\
 & Omission & L2 & 5 & 69.0 $\pm$ 3.9 & 14.0 $\pm$ 3.9 & 83.2 $\pm$ 4.7 \\
 & NLCD & Real & 5 & 60.7 $\pm$ 1.2 & 22.3 $\pm$ 1.1 & 73.2 $\pm$ 1.3 \\
\addlinespace[1.5pt]
TM & Clean & -- & 5 & 82.0 $\pm$ 0.5 & -- & -- \\
 & Shift & L1 & 5 & 81.1 $\pm$ 0.5 & 1.0 $\pm$ 0.6 & 98.8 $\pm$ 0.8 \\
 & Shift & L2 & 5 & 74.6 $\pm$ 0.6 & 7.4 $\pm$ 0.9 & 91.0 $\pm$ 1.1 \\
 & Coarse & L1 & 5 & 80.3 $\pm$ 0.5 & 1.7 $\pm$ 0.6 & 97.9 $\pm$ 0.8 \\
 & Coarse & L2 & 5 & 69.8 $\pm$ 0.7 & 12.3 $\pm$ 0.6 & 85.0 $\pm$ 0.8 \\
 & Boundary & L1 & 5 & 78.0 $\pm$ 0.5 & 4.1 $\pm$ 0.6 & 95.0 $\pm$ 0.8 \\
 & Boundary & L2 & 5 & 71.6 $\pm$ 1.0 & 10.5 $\pm$ 1.2 & 87.2 $\pm$ 1.4 \\
 & Omission & L1 & 5 & 78.4 $\pm$ 1.7 & 3.7 $\pm$ 1.7 & 95.5 $\pm$ 2.1 \\
 & Omission & L2 & 5 & 72.5 $\pm$ 2.4 & 9.5 $\pm$ 2.2 & 88.4 $\pm$ 2.7 \\
 & NLCD & Real & 5 & 62.1 $\pm$ 0.8 & 20.0 $\pm$ 0.5 & 75.6 $\pm$ 0.7 \\
\addlinespace[1.5pt]
RN & Clean & -- & 5 & 82.5 $\pm$ 0.6 & -- & -- \\
 & Shift & L1 & 5 & 80.6 $\pm$ 1.0 & 1.8 $\pm$ 0.6 & 97.8 $\pm$ 0.7 \\
 & Shift & L2 & 5 & 73.8 $\pm$ 0.7 & 8.6 $\pm$ 0.7 & 89.5 $\pm$ 0.8 \\
 & Coarse & L1 & 5 & 80.8 $\pm$ 0.2 & 1.7 $\pm$ 0.7 & 97.9 $\pm$ 0.8 \\
 & Coarse & L2 & 5 & 69.8 $\pm$ 0.4 & 12.6 $\pm$ 0.9 & 84.7 $\pm$ 1.0 \\
 & Boundary & L1 & 5 & 80.6 $\pm$ 0.2 & 1.9 $\pm$ 0.7 & 97.7 $\pm$ 0.8 \\
 & Boundary & L2 & 5 & 73.6 $\pm$ 1.8 & 8.8 $\pm$ 1.8 & 89.3 $\pm$ 2.1 \\
 & Omission & L1 & 5 & 80.2 $\pm$ 1.2 & 2.2 $\pm$ 1.0 & 97.3 $\pm$ 1.2 \\
 & Omission & L2 & 5 & 77.9 $\pm$ 1.2 & 4.6 $\pm$ 1.0 & 94.4 $\pm$ 1.2 \\
 & NLCD & Real & 5 & 59.6 $\pm$ 3.4 & 22.9 $\pm$ 4.0 & 72.3 $\pm$ 4.6 \\
\bottomrule
\end{tabular}
\end{adjustbox}}{\tabletbd[6cm]{planned/generated/table\_ax\_grid\_ches.tex: run scripts/make\_arxiv\_appendix.py}}
\caption{\textbf{Every core cell on Chesapeake:} Test \miou, paired drop and retention for every core model, error type and level with the UNet decoder. Cells give the mean $\pm$ standard deviation over the $n$ seeds. The row named Real is the run trained on the real label of the pair, where one exists. A dash marks a quantity that does not exist for the clean row. The \miou and the drop are in points. Models are abbreviated as SAT (\sat), LVD (\lvd), DOFA (\dofa), TM (\tm) and RN (\rn). \ProvSlot{\ProvTableAxGridChes}}
\label{tab:ax-grid-ches}
\end{table}

\begin{table}[t]
\centering\footnotesize\setlength{\tabcolsep}{3pt}
\IfFileExists{planned/generated/table_ax_grid_cstwelve.tex}{\begin{adjustbox}{max width=\columnwidth}%
\providecommand{\gcvdry}[1]{\textcolor{red}{#1}}
\providecommand{\ProvTableAxGridCstwelve}{Seeds per cell 5. Analysis of the finished runs}
\begin{tabular}{@{}lllcccc@{}}
\toprule
Model & Type & Level & $n$ & mIoU & Drop & Retention (\%) \\
\midrule
SAT & Clean & -- & 5 & 66.6 $\pm$ 1.2 & -- & -- \\
 & Shift & L1 & 5 & 66.4 $\pm$ 0.8 & 0.1 $\pm$ 0.5 & 99.8 $\pm$ 0.7 \\
 & Shift & L2 & 5 & 61.6 $\pm$ 0.5 & 4.9 $\pm$ 1.4 & 92.6 $\pm$ 2.0 \\
 & Coarse & L1 & 5 & 65.8 $\pm$ 1.1 & 0.7 $\pm$ 1.8 & 99.0 $\pm$ 2.7 \\
 & Coarse & L2 & 5 & 59.9 $\pm$ 0.9 & 6.7 $\pm$ 1.4 & 90.0 $\pm$ 2.0 \\
 & Boundary & L1 & 5 & 65.2 $\pm$ 1.1 & 1.4 $\pm$ 1.8 & 97.9 $\pm$ 2.6 \\
 & Boundary & L2 & 5 & 61.3 $\pm$ 0.7 & 5.2 $\pm$ 0.8 & 92.2 $\pm$ 1.1 \\
 & Omission & L1 & 5 & 63.6 $\pm$ 1.2 & 3.0 $\pm$ 1.3 & 95.6 $\pm$ 1.9 \\
 & Omission & L2 & 5 & 59.9 $\pm$ 1.2 & 6.6 $\pm$ 2.1 & 90.1 $\pm$ 3.1 \\
 & s2cloudless & Real & 5 & 34.4 $\pm$ 0.4 & 32.2 $\pm$ 1.3 & 51.6 $\pm$ 1.2 \\
\addlinespace[1.5pt]
LVD & Clean & -- & 5 & 69.6 $\pm$ 0.7 & -- & -- \\
 & Shift & L1 & 5 & 68.5 $\pm$ 0.9 & 1.1 $\pm$ 1.1 & 98.5 $\pm$ 1.6 \\
 & Shift & L2 & 5 & 64.7 $\pm$ 0.8 & 4.8 $\pm$ 1.4 & 93.1 $\pm$ 2.0 \\
 & Coarse & L1 & 5 & 67.7 $\pm$ 0.8 & 1.9 $\pm$ 1.1 & 97.2 $\pm$ 1.5 \\
 & Coarse & L2 & 5 & 61.8 $\pm$ 0.5 & 7.8 $\pm$ 0.6 & 88.7 $\pm$ 0.8 \\
 & Boundary & L1 & 5 & 67.3 $\pm$ 0.6 & 2.2 $\pm$ 1.0 & 96.8 $\pm$ 1.4 \\
 & Boundary & L2 & 5 & 63.6 $\pm$ 0.5 & 6.0 $\pm$ 0.7 & 91.4 $\pm$ 0.9 \\
 & Omission & L1 & 5 & 66.9 $\pm$ 0.4 & 2.7 $\pm$ 1.0 & 96.1 $\pm$ 1.5 \\
 & Omission & L2 & 5 & 62.7 $\pm$ 0.3 & 6.9 $\pm$ 0.8 & 90.1 $\pm$ 1.1 \\
 & s2cloudless & Real & 5 & 35.1 $\pm$ 0.2 & 34.5 $\pm$ 0.8 & 50.4 $\pm$ 0.6 \\
\addlinespace[1.5pt]
DOFA & Clean & -- & 5 & 65.9 $\pm$ 0.5 & -- & -- \\
 & Shift & L1 & 5 & 65.5 $\pm$ 0.6 & 0.3 $\pm$ 0.5 & 99.5 $\pm$ 0.7 \\
 & Shift & L2 & 5 & 61.1 $\pm$ 0.3 & 4.8 $\pm$ 0.4 & 92.7 $\pm$ 0.5 \\
 & Coarse & L1 & 5 & 64.2 $\pm$ 1.1 & 1.7 $\pm$ 1.6 & 97.4 $\pm$ 2.4 \\
 & Coarse & L2 & 5 & 57.6 $\pm$ 0.8 & 8.3 $\pm$ 0.6 & 87.4 $\pm$ 0.9 \\
 & Boundary & L1 & 5 & 63.0 $\pm$ 0.9 & 2.8 $\pm$ 1.2 & 95.7 $\pm$ 1.7 \\
 & Boundary & L2 & 5 & 58.4 $\pm$ 1.0 & 7.5 $\pm$ 1.3 & 88.6 $\pm$ 2.0 \\
 & Omission & L1 & 5 & 62.2 $\pm$ 1.0 & 3.6 $\pm$ 0.7 & 94.5 $\pm$ 1.0 \\
 & Omission & L2 & 5 & 57.8 $\pm$ 1.4 & 8.1 $\pm$ 1.1 & 87.7 $\pm$ 1.7 \\
 & s2cloudless & Real & 5 & 34.5 $\pm$ 0.4 & 31.4 $\pm$ 0.9 & 52.3 $\pm$ 1.0 \\
\addlinespace[1.5pt]
TM & Clean & -- & 5 & 69.5 $\pm$ 0.5 & -- & -- \\
 & Shift & L1 & 5 & 68.5 $\pm$ 0.5 & 1.0 $\pm$ 1.0 & 98.5 $\pm$ 1.4 \\
 & Shift & L2 & 5 & 64.0 $\pm$ 0.6 & 5.6 $\pm$ 0.8 & 92.0 $\pm$ 1.2 \\
 & Coarse & L1 & 5 & 67.6 $\pm$ 0.9 & 1.9 $\pm$ 1.1 & 97.3 $\pm$ 1.6 \\
 & Coarse & L2 & 5 & 61.1 $\pm$ 0.3 & 8.5 $\pm$ 0.3 & 87.8 $\pm$ 0.4 \\
 & Boundary & L1 & 5 & 67.8 $\pm$ 0.4 & 1.7 $\pm$ 0.8 & 97.6 $\pm$ 1.1 \\
 & Boundary & L2 & 5 & 62.0 $\pm$ 0.7 & 7.5 $\pm$ 0.6 & 89.2 $\pm$ 0.9 \\
 & Omission & L1 & 5 & 66.0 $\pm$ 1.3 & 3.5 $\pm$ 1.8 & 95.0 $\pm$ 2.5 \\
 & Omission & L2 & 5 & 62.2 $\pm$ 1.0 & 7.3 $\pm$ 0.8 & 89.5 $\pm$ 1.1 \\
 & s2cloudless & Real & 5 & 34.8 $\pm$ 0.2 & 34.7 $\pm$ 0.5 & 50.1 $\pm$ 0.4 \\
\addlinespace[1.5pt]
RN & Clean & -- & 5 & 64.0 $\pm$ 0.7 & -- & -- \\
 & Shift & L1 & 5 & 62.7 $\pm$ 0.6 & 1.3 $\pm$ 0.7 & 97.9 $\pm$ 1.1 \\
 & Shift & L2 & 5 & 58.7 $\pm$ 1.6 & 5.3 $\pm$ 1.6 & 91.7 $\pm$ 2.5 \\
 & Coarse & L1 & 5 & 61.8 $\pm$ 0.4 & 2.2 $\pm$ 0.6 & 96.6 $\pm$ 0.9 \\
 & Coarse & L2 & 5 & 56.2 $\pm$ 1.2 & 7.8 $\pm$ 1.4 & 87.8 $\pm$ 2.1 \\
 & Boundary & L1 & 5 & 61.8 $\pm$ 1.0 & 2.3 $\pm$ 0.8 & 96.5 $\pm$ 1.2 \\
 & Boundary & L2 & 5 & 56.4 $\pm$ 3.4 & 7.6 $\pm$ 2.8 & 88.0 $\pm$ 4.5 \\
 & Omission & L1 & 5 & 60.3 $\pm$ 0.8 & 3.8 $\pm$ 0.8 & 94.1 $\pm$ 1.2 \\
 & Omission & L2 & 5 & 57.7 $\pm$ 0.7 & 6.3 $\pm$ 1.2 & 90.2 $\pm$ 1.8 \\
 & s2cloudless & Real & 5 & 34.9 $\pm$ 0.1 & 29.1 $\pm$ 0.7 & 54.6 $\pm$ 0.6 \\
\bottomrule
\end{tabular}
\end{adjustbox}}{\tabletbd[6cm]{planned/generated/table\_ax\_grid\_cstwelve.tex: run scripts/make\_arxiv\_appendix.py}}
\caption{\textbf{Every core cell on CloudSEN12:} Columns, rows and marks as in \Cref{tab:ax-grid-ches}. \ProvSlot{\ProvTableAxGridCstwelve}}
\label{tab:ax-grid-cs}
\end{table}

\begin{table}[t]
\centering\footnotesize\setlength{\tabcolsep}{3pt}
\IfFileExists{planned/generated/table_ax_grid_dfc.tex}{\begin{adjustbox}{max width=\columnwidth}%
\providecommand{\gcvdry}[1]{\textcolor{red}{#1}}
\providecommand{\ProvTableAxGridDfc}{Seeds per cell 5. Analysis of the finished runs}
\begin{tabular}{@{}lllcccc@{}}
\toprule
Model & Type & Level & $n$ & mIoU & Drop & Retention (\%) \\
\midrule
SAT & Clean & -- & 5 & 40.5 $\pm$ 1.0 & -- & -- \\
 & Shift & L1 & 5 & 39.3 $\pm$ 2.6 & 1.3 $\pm$ 2.3 & 96.9 $\pm$ 5.7 \\
 & Shift & L2 & 5 & 40.8 $\pm$ 0.6 & $-$0.2 $\pm$ 0.7 & 100.6 $\pm$ 1.8 \\
 & Coarse & L1 & 5 & 37.7 $\pm$ 5.1 & 2.8 $\pm$ 5.2 & 93.0 $\pm$ 12.8 \\
 & Coarse & L2 & 5 & 38.8 $\pm$ 1.0 & 1.7 $\pm$ 1.5 & 95.8 $\pm$ 3.6 \\
 & Boundary & L1 & 5 & 38.6 $\pm$ 1.8 & 1.9 $\pm$ 2.2 & 95.3 $\pm$ 5.2 \\
 & Boundary & L2 & 5 & 34.2 $\pm$ 2.0 & 6.3 $\pm$ 2.0 & 84.4 $\pm$ 4.7 \\
 & Omission & L1 & 5 & 37.5 $\pm$ 2.2 & 3.0 $\pm$ 1.8 & 92.6 $\pm$ 4.4 \\
 & Omission & L2 & 5 & 36.6 $\pm$ 1.3 & 4.0 $\pm$ 1.7 & 90.3 $\pm$ 4.0 \\
 & MODIS & Real & 5 & 32.5 $\pm$ 2.4 & 7.4 $\pm$ 2.8 & 81.5 $\pm$ 6.7 \\
\addlinespace[1.5pt]
LVD & Clean & -- & 5 & 43.2 $\pm$ 0.6 & -- & -- \\
 & Shift & L1 & 5 & 43.3 $\pm$ 0.8 & $-$0.1 $\pm$ 0.3 & 100.3 $\pm$ 0.7 \\
 & Shift & L2 & 5 & 43.0 $\pm$ 0.7 & 0.2 $\pm$ 0.8 & 99.5 $\pm$ 1.8 \\
 & Coarse & L1 & 5 & 42.9 $\pm$ 0.5 & 0.3 $\pm$ 0.6 & 99.3 $\pm$ 1.3 \\
 & Coarse & L2 & 5 & 41.6 $\pm$ 0.8 & 1.5 $\pm$ 0.8 & 96.4 $\pm$ 1.7 \\
 & Boundary & L1 & 5 & 42.4 $\pm$ 0.7 & 0.8 $\pm$ 0.8 & 98.2 $\pm$ 1.8 \\
 & Boundary & L2 & 5 & 38.0 $\pm$ 5.1 & 5.2 $\pm$ 5.3 & 88.0 $\pm$ 12.2 \\
 & Omission & L1 & 5 & 41.8 $\pm$ 0.8 & 1.4 $\pm$ 1.1 & 96.7 $\pm$ 2.5 \\
 & Omission & L2 & 5 & 40.0 $\pm$ 0.7 & 3.2 $\pm$ 1.0 & 92.7 $\pm$ 2.2 \\
 & MODIS & Real & 5 & 37.1 $\pm$ 1.5 & 6.1 $\pm$ 1.6 & 85.9 $\pm$ 3.6 \\
\addlinespace[1.5pt]
DOFA & Clean & -- & 5 & 36.5 $\pm$ 1.4 & -- & -- \\
 & Shift & L1 & 5 & 37.0 $\pm$ 0.2 & $-$0.5 $\pm$ 1.2 & 101.4 $\pm$ 3.4 \\
 & Shift & L2 & 5 & 37.2 $\pm$ 0.8 & $-$0.7 $\pm$ 1.5 & 102.0 $\pm$ 4.3 \\
 & Coarse & L1 & 5 & 37.0 $\pm$ 0.3 & $-$0.5 $\pm$ 1.5 & 101.5 $\pm$ 4.3 \\
 & Coarse & L2 & 5 & 36.3 $\pm$ 0.8 & 0.3 $\pm$ 1.1 & 99.4 $\pm$ 3.2 \\
 & Boundary & L1 & 5 & 35.3 $\pm$ 0.8 & 1.2 $\pm$ 1.6 & 96.9 $\pm$ 4.5 \\
 & Boundary & L2 & 5 & 33.7 $\pm$ 1.6 & 2.9 $\pm$ 2.2 & 92.3 $\pm$ 6.1 \\
 & Omission & L1 & 5 & 35.6 $\pm$ 1.0 & 0.9 $\pm$ 1.4 & 97.6 $\pm$ 4.0 \\
 & Omission & L2 & 5 & 33.3 $\pm$ 1.6 & 3.2 $\pm$ 2.3 & 91.4 $\pm$ 6.2 \\
 & MODIS & Real & 5 & 33.5 $\pm$ 1.2 & 3.1 $\pm$ 1.5 & 91.5 $\pm$ 4.0 \\
\addlinespace[1.5pt]
TM & Clean & -- & 5 & 38.9 $\pm$ 0.6 & -- & -- \\
 & Shift & L1 & 5 & 38.1 $\pm$ 0.4 & 0.8 $\pm$ 0.7 & 98.0 $\pm$ 1.7 \\
 & Shift & L2 & 5 & 37.6 $\pm$ 0.8 & 1.2 $\pm$ 1.1 & 96.9 $\pm$ 2.9 \\
 & Coarse & L1 & 5 & 38.2 $\pm$ 1.3 & 0.7 $\pm$ 1.6 & 98.2 $\pm$ 4.2 \\
 & Coarse & L2 & 5 & 36.4 $\pm$ 1.4 & 2.5 $\pm$ 1.9 & 93.6 $\pm$ 4.7 \\
 & Boundary & L1 & 5 & 37.7 $\pm$ 1.5 & 1.1 $\pm$ 2.0 & 97.1 $\pm$ 5.2 \\
 & Boundary & L2 & 5 & 34.7 $\pm$ 3.0 & 4.2 $\pm$ 3.5 & 89.3 $\pm$ 9.0 \\
 & Omission & L1 & 5 & 36.4 $\pm$ 2.4 & 2.5 $\pm$ 2.2 & 93.5 $\pm$ 5.6 \\
 & Omission & L2 & 5 & 33.0 $\pm$ 1.2 & 5.8 $\pm$ 1.1 & 85.0 $\pm$ 2.9 \\
 & MODIS & Real & 5 & 35.5 $\pm$ 0.9 & 2.7 $\pm$ 0.9 & 92.9 $\pm$ 2.4 \\
\addlinespace[1.5pt]
RN & Clean & -- & 5 & 38.8 $\pm$ 1.1 & -- & -- \\
 & Shift & L1 & 5 & 39.8 $\pm$ 0.4 & $-$1.1 $\pm$ 1.2 & 102.8 $\pm$ 3.2 \\
 & Shift & L2 & 5 & 38.9 $\pm$ 1.4 & $-$0.2 $\pm$ 1.3 & 100.4 $\pm$ 3.4 \\
 & Coarse & L1 & 5 & 39.6 $\pm$ 1.1 & $-$0.8 $\pm$ 1.1 & 102.1 $\pm$ 2.9 \\
 & Coarse & L2 & 5 & 37.9 $\pm$ 1.2 & 0.8 $\pm$ 2.0 & 97.9 $\pm$ 5.0 \\
 & Boundary & L1 & 5 & 39.3 $\pm$ 1.0 & $-$0.5 $\pm$ 1.9 & 101.4 $\pm$ 4.9 \\
 & Boundary & L2 & 5 & 36.2 $\pm$ 2.5 & 2.5 $\pm$ 2.4 & 93.5 $\pm$ 6.2 \\
 & Omission & L1 & 5 & 38.0 $\pm$ 2.3 & 0.8 $\pm$ 2.2 & 98.1 $\pm$ 5.5 \\
 & Omission & L2 & 5 & 37.2 $\pm$ 1.8 & 1.5 $\pm$ 2.2 & 96.1 $\pm$ 5.6 \\
 & MODIS & Real & 5 & 35.2 $\pm$ 1.6 & 4.5 $\pm$ 1.3 & 88.7 $\pm$ 3.3 \\
\bottomrule
\end{tabular}
\end{adjustbox}}{\tabletbd[6cm]{planned/generated/table\_ax\_grid\_dfc.tex: run scripts/make\_arxiv\_appendix.py}}
\caption{\textbf{Every core cell on DFC2020:} Columns, rows and marks as in \Cref{tab:ax-grid-ches}. The Real row is paired with a clean run on the tiles that the MODIS label covers. Its drop is therefore not the printed clean \miou minus its \miou. \ProvSlot{\ProvTableAxGridDfc}}
\label{tab:ax-grid-dfc}
\end{table}

\begin{table}[t]
\centering\footnotesize\setlength{\tabcolsep}{3pt}
\IfFileExists{planned/generated/table_ax_grid_enva.tex}{\begin{adjustbox}{max width=\columnwidth}%
\providecommand{\gcvdry}[1]{\textcolor{red}{#1}}
\providecommand{\ProvTableAxGridEnva}{Seeds per cell 5. Analysis of the finished runs}
\begin{tabular}{@{}lllcccc@{}}
\toprule
Model & Type & Level & $n$ & mIoU & Drop & Retention (\%) \\
\midrule
SAT & Clean & -- & 5 & 61.8 $\pm$ 0.4 & -- & -- \\
 & Shift & L1 & 5 & 61.1 $\pm$ 0.8 & 0.8 $\pm$ 1.2 & 98.8 $\pm$ 1.9 \\
 & Shift & L2 & 5 & 58.9 $\pm$ 1.3 & 3.0 $\pm$ 1.2 & 95.2 $\pm$ 2.0 \\
 & Coarse & L1 & 5 & 61.5 $\pm$ 0.3 & 0.3 $\pm$ 0.6 & 99.5 $\pm$ 1.0 \\
 & Coarse & L2 & 5 & 57.9 $\pm$ 1.0 & 3.9 $\pm$ 1.1 & 93.7 $\pm$ 1.7 \\
 & Boundary & L1 & 5 & 60.3 $\pm$ 0.6 & 1.5 $\pm$ 1.0 & 97.5 $\pm$ 1.6 \\
 & Boundary & L2 & 5 & 55.3 $\pm$ 2.0 & 6.5 $\pm$ 2.0 & 89.4 $\pm$ 3.2 \\
 & Omission & L1 & 5 & 57.7 $\pm$ 2.4 & 4.2 $\pm$ 2.1 & 93.3 $\pm$ 3.4 \\
 & Omission & L2 & 5 & 50.1 $\pm$ 6.8 & 11.8 $\pm$ 6.9 & 81.0 $\pm$ 11.1 \\
 & NLCD & Real & 5 & 40.0 $\pm$ 2.0 & 21.8 $\pm$ 2.2 & 64.8 $\pm$ 3.4 \\
\addlinespace[1.5pt]
LVD & Clean & -- & 5 & 63.0 $\pm$ 0.9 & -- & -- \\
 & Shift & L1 & 5 & 62.9 $\pm$ 0.6 & 0.1 $\pm$ 0.6 & 99.8 $\pm$ 1.0 \\
 & Shift & L2 & 5 & 60.6 $\pm$ 0.9 & 2.4 $\pm$ 1.2 & 96.2 $\pm$ 1.9 \\
 & Coarse & L1 & 5 & 62.7 $\pm$ 1.3 & 0.3 $\pm$ 1.1 & 99.6 $\pm$ 1.7 \\
 & Coarse & L2 & 5 & 59.5 $\pm$ 2.8 & 3.5 $\pm$ 2.7 & 94.4 $\pm$ 4.3 \\
 & Boundary & L1 & 5 & 59.8 $\pm$ 3.1 & 3.2 $\pm$ 3.3 & 94.9 $\pm$ 5.2 \\
 & Boundary & L2 & 5 & 52.1 $\pm$ 5.4 & 10.9 $\pm$ 5.5 & 82.8 $\pm$ 8.6 \\
 & Omission & L1 & 5 & 57.3 $\pm$ 7.4 & 5.7 $\pm$ 7.8 & 91.1 $\pm$ 12.4 \\
 & Omission & L2 & 5 & 48.8 $\pm$ 3.8 & 14.2 $\pm$ 3.9 & 77.5 $\pm$ 6.1 \\
 & NLCD & Real & 5 & 33.0 $\pm$ 4.9 & 30.0 $\pm$ 5.3 & 52.4 $\pm$ 8.2 \\
\addlinespace[1.5pt]
DOFA & Clean & -- & 5 & 48.1 $\pm$ 1.8 & -- & -- \\
 & Shift & L1 & 5 & 48.0 $\pm$ 3.1 & 0.1 $\pm$ 1.9 & 99.7 $\pm$ 4.0 \\
 & Shift & L2 & 5 & 47.5 $\pm$ 3.3 & 0.6 $\pm$ 2.6 & 98.8 $\pm$ 5.2 \\
 & Coarse & L1 & 5 & 49.1 $\pm$ 3.0 & $-$1.0 $\pm$ 2.7 & 102.2 $\pm$ 5.7 \\
 & Coarse & L2 & 5 & 44.6 $\pm$ 3.0 & 3.5 $\pm$ 2.0 & 92.8 $\pm$ 4.2 \\
 & Boundary & L1 & 5 & 48.8 $\pm$ 5.1 & $-$0.7 $\pm$ 3.6 & 101.2 $\pm$ 7.5 \\
 & Boundary & L2 & 5 & 43.4 $\pm$ 3.3 & 4.7 $\pm$ 2.8 & 90.2 $\pm$ 5.7 \\
 & Omission & L1 & 5 & 42.4 $\pm$ 3.6 & 5.7 $\pm$ 4.6 & 88.4 $\pm$ 9.3 \\
 & Omission & L2 & 5 & 35.1 $\pm$ 10.4 & 13.0 $\pm$ 11.6 & 73.5 $\pm$ 24.2 \\
 & NLCD & Real & 5 & 27.4 $\pm$ 2.3 & 20.7 $\pm$ 3.3 & 57.0 $\pm$ 5.9 \\
\addlinespace[1.5pt]
TM & Clean & -- & 5 & 49.0 $\pm$ 2.2 & -- & -- \\
 & Shift & L1 & 5 & 49.6 $\pm$ 2.8 & $-$0.6 $\pm$ 1.3 & 101.2 $\pm$ 2.6 \\
 & Shift & L2 & 5 & 46.7 $\pm$ 4.1 & 2.3 $\pm$ 2.9 & 95.2 $\pm$ 6.0 \\
 & Coarse & L1 & 5 & 49.9 $\pm$ 2.5 & $-$0.9 $\pm$ 3.0 & 101.9 $\pm$ 6.5 \\
 & Coarse & L2 & 5 & 45.9 $\pm$ 3.0 & 3.0 $\pm$ 4.7 & 94.1 $\pm$ 9.6 \\
 & Boundary & L1 & 5 & 44.0 $\pm$ 9.6 & 5.0 $\pm$ 8.1 & 89.4 $\pm$ 17.4 \\
 & Boundary & L2 & 5 & 41.5 $\pm$ 5.9 & 7.5 $\pm$ 5.6 & 84.7 $\pm$ 11.2 \\
 & Omission & L1 & 5 & 44.5 $\pm$ 8.8 & 4.5 $\pm$ 8.9 & 91.0 $\pm$ 18.5 \\
 & Omission & L2 & 5 & 34.4 $\pm$ 7.8 & 14.6 $\pm$ 6.0 & 69.9 $\pm$ 13.1 \\
 & NLCD & Real & 5 & 22.9 $\pm$ 1.0 & 26.1 $\pm$ 2.6 & 46.8 $\pm$ 3.1 \\
\addlinespace[1.5pt]
RN & Clean & -- & 5 & 49.1 $\pm$ 3.9 & -- & -- \\
 & Shift & L1 & 5 & 52.5 $\pm$ 2.1 & $-$3.4 $\pm$ 4.0 & 107.4 $\pm$ 8.9 \\
 & Shift & L2 & 5 & 41.4 $\pm$ 5.7 & 7.7 $\pm$ 6.9 & 84.7 $\pm$ 13.7 \\
 & Coarse & L1 & 5 & 52.3 $\pm$ 3.7 & $-$3.2 $\pm$ 3.3 & 106.8 $\pm$ 7.4 \\
 & Coarse & L2 & 5 & 46.8 $\pm$ 6.9 & 2.3 $\pm$ 6.8 & 95.5 $\pm$ 14.2 \\
 & Boundary & L1 & 5 & 46.1 $\pm$ 6.5 & 3.0 $\pm$ 7.0 & 94.2 $\pm$ 14.4 \\
 & Boundary & L2 & 5 & 31.1 $\pm$ 5.7 & 18.0 $\pm$ 2.3 & 62.9 $\pm$ 6.6 \\
 & Omission & L1 & 5 & 34.1 $\pm$ 5.3 & 15.0 $\pm$ 8.7 & 70.3 $\pm$ 14.6 \\
 & Omission & L2 & 5 & 28.8 $\pm$ 5.6 & 20.3 $\pm$ 5.8 & 58.8 $\pm$ 11.1 \\
 & NLCD & Real & 5 & 24.7 $\pm$ 2.4 & 24.4 $\pm$ 5.6 & 50.7 $\pm$ 8.1 \\
\bottomrule
\end{tabular}
\end{adjustbox}}{\tabletbd[6cm]{planned/generated/table\_ax\_grid\_enva.tex: run scripts/make\_arxiv\_appendix.py}}
\caption{\textbf{Every core cell on EnviroAtlas:} Columns, rows and marks as in \Cref{tab:ax-grid-ches}. \ProvSlot{\ProvTableAxGridEnva}}
\label{tab:ax-grid-enva}
\end{table}

\begin{table}[t]
\centering\footnotesize\setlength{\tabcolsep}{3pt}
\IfFileExists{planned/generated/table_ax_grid_sntwo.tex}{\begin{adjustbox}{max width=\columnwidth}%
\providecommand{\gcvdry}[1]{\textcolor{red}{#1}}
\providecommand{\ProvTableAxGridSntwo}{Seeds per cell 5. Analysis of the finished runs}
\begin{tabular}{@{}lllcccc@{}}
\toprule
Model & Type & Level & $n$ & mIoU & Drop & Retention (\%) \\
\midrule
SAT & Clean & -- & 5 & 85.8 $\pm$ 0.2 & -- & -- \\
 & Shift & L1 & 5 & 84.4 $\pm$ 0.4 & 1.4 $\pm$ 0.2 & 98.4 $\pm$ 0.2 \\
 & Shift & L2 & 5 & 78.1 $\pm$ 0.4 & 7.7 $\pm$ 0.5 & 91.0 $\pm$ 0.6 \\
 & Coarse & L1 & 5 & 83.1 $\pm$ 0.5 & 2.7 $\pm$ 0.5 & 96.9 $\pm$ 0.6 \\
 & Coarse & L2 & 5 & 73.1 $\pm$ 1.5 & 12.7 $\pm$ 1.4 & 85.2 $\pm$ 1.6 \\
 & Boundary & L1 & 5 & 82.8 $\pm$ 0.5 & 3.0 $\pm$ 0.5 & 96.5 $\pm$ 0.6 \\
 & Boundary & L2 & 5 & 75.2 $\pm$ 1.8 & 10.6 $\pm$ 1.8 & 87.7 $\pm$ 2.1 \\
 & Omission & L1 & 5 & 81.2 $\pm$ 0.7 & 4.7 $\pm$ 0.8 & 94.6 $\pm$ 0.9 \\
 & Omission & L2 & 5 & 71.1 $\pm$ 4.6 & 14.7 $\pm$ 4.5 & 82.8 $\pm$ 5.2 \\
\addlinespace[1.5pt]
LVD & Clean & -- & 5 & 86.0 $\pm$ 0.3 & -- & -- \\
 & Shift & L1 & 5 & 84.4 $\pm$ 0.5 & 1.6 $\pm$ 0.3 & 98.1 $\pm$ 0.3 \\
 & Shift & L2 & 5 & 78.9 $\pm$ 0.7 & 7.1 $\pm$ 0.6 & 91.7 $\pm$ 0.7 \\
 & Coarse & L1 & 5 & 84.1 $\pm$ 0.2 & 1.9 $\pm$ 0.2 & 97.8 $\pm$ 0.2 \\
 & Coarse & L2 & 5 & 74.6 $\pm$ 0.4 & 11.4 $\pm$ 0.6 & 86.7 $\pm$ 0.6 \\
 & Boundary & L1 & 5 & 83.2 $\pm$ 0.2 & 2.8 $\pm$ 0.3 & 96.7 $\pm$ 0.3 \\
 & Boundary & L2 & 5 & 75.8 $\pm$ 0.5 & 10.2 $\pm$ 0.5 & 88.2 $\pm$ 0.6 \\
 & Omission & L1 & 5 & 81.6 $\pm$ 0.6 & 4.5 $\pm$ 0.5 & 94.8 $\pm$ 0.6 \\
 & Omission & L2 & 5 & 73.3 $\pm$ 3.6 & 12.7 $\pm$ 3.9 & 85.2 $\pm$ 4.5 \\
\addlinespace[1.5pt]
DOFA & Clean & -- & 5 & 84.2 $\pm$ 0.5 & -- & -- \\
 & Shift & L1 & 5 & 82.5 $\pm$ 0.8 & 1.7 $\pm$ 1.0 & 98.0 $\pm$ 1.2 \\
 & Shift & L2 & 5 & 75.9 $\pm$ 0.5 & 8.3 $\pm$ 0.3 & 90.2 $\pm$ 0.3 \\
 & Coarse & L1 & 5 & 82.0 $\pm$ 0.4 & 2.2 $\pm$ 0.4 & 97.4 $\pm$ 0.5 \\
 & Coarse & L2 & 5 & 72.2 $\pm$ 0.5 & 12.0 $\pm$ 1.0 & 85.8 $\pm$ 1.1 \\
 & Boundary & L1 & 5 & 81.7 $\pm$ 0.2 & 2.5 $\pm$ 0.6 & 97.0 $\pm$ 0.7 \\
 & Boundary & L2 & 5 & 72.5 $\pm$ 1.1 & 11.7 $\pm$ 1.3 & 86.1 $\pm$ 1.5 \\
 & Omission & L1 & 5 & 78.3 $\pm$ 2.2 & 5.9 $\pm$ 1.7 & 93.0 $\pm$ 2.1 \\
 & Omission & L2 & 5 & 68.8 $\pm$ 1.3 & 15.4 $\pm$ 1.5 & 81.7 $\pm$ 1.7 \\
\addlinespace[1.5pt]
TM & Clean & -- & 5 & 83.4 $\pm$ 0.3 & -- & -- \\
 & Shift & L1 & 5 & 81.6 $\pm$ 0.5 & 1.8 $\pm$ 0.7 & 97.8 $\pm$ 0.8 \\
 & Shift & L2 & 5 & 75.4 $\pm$ 0.6 & 8.0 $\pm$ 0.7 & 90.4 $\pm$ 0.8 \\
 & Coarse & L1 & 5 & 81.3 $\pm$ 0.5 & 2.2 $\pm$ 0.6 & 97.4 $\pm$ 0.7 \\
 & Coarse & L2 & 5 & 71.2 $\pm$ 1.0 & 12.3 $\pm$ 0.9 & 85.3 $\pm$ 1.1 \\
 & Boundary & L1 & 5 & 80.2 $\pm$ 0.6 & 3.3 $\pm$ 0.7 & 96.1 $\pm$ 0.8 \\
 & Boundary & L2 & 5 & 72.8 $\pm$ 0.7 & 10.7 $\pm$ 0.6 & 87.2 $\pm$ 0.8 \\
 & Omission & L1 & 5 & 77.6 $\pm$ 1.3 & 5.9 $\pm$ 1.4 & 93.0 $\pm$ 1.7 \\
 & Omission & L2 & 5 & 67.8 $\pm$ 1.3 & 15.6 $\pm$ 1.5 & 81.3 $\pm$ 1.7 \\
\addlinespace[1.5pt]
RN & Clean & -- & 5 & 81.7 $\pm$ 1.5 & -- & -- \\
 & Shift & L1 & 5 & 80.3 $\pm$ 1.4 & 1.4 $\pm$ 1.8 & 98.3 $\pm$ 2.1 \\
 & Shift & L2 & 5 & 72.6 $\pm$ 0.3 & 9.1 $\pm$ 1.7 & 88.9 $\pm$ 1.8 \\
 & Coarse & L1 & 5 & 79.2 $\pm$ 1.0 & 2.6 $\pm$ 1.8 & 96.9 $\pm$ 2.2 \\
 & Coarse & L2 & 5 & 68.1 $\pm$ 0.5 & 13.7 $\pm$ 2.0 & 83.3 $\pm$ 2.2 \\
 & Boundary & L1 & 5 & 78.4 $\pm$ 2.3 & 3.4 $\pm$ 3.1 & 95.9 $\pm$ 3.7 \\
 & Boundary & L2 & 5 & 72.5 $\pm$ 0.6 & 9.2 $\pm$ 1.6 & 88.7 $\pm$ 1.7 \\
 & Omission & L1 & 5 & 78.6 $\pm$ 1.0 & 3.1 $\pm$ 1.6 & 96.2 $\pm$ 1.9 \\
 & Omission & L2 & 5 & 58.4 $\pm$ 6.2 & 23.3 $\pm$ 6.2 & 71.5 $\pm$ 7.4 \\
\bottomrule
\end{tabular}
\end{adjustbox}}{\tabletbd[6cm]{planned/generated/table\_ax\_grid\_sntwo.tex: run scripts/make\_arxiv\_appendix.py}}
\caption{\textbf{Every core cell on SpaceNet 2:} Columns, rows and marks as in \Cref{tab:ax-grid-ches}. \ProvSlot{\ProvTableAxGridSntwo}}
\label{tab:ax-grid-sn}
\end{table}

\begin{table}[t]
\centering\footnotesize\setlength{\tabcolsep}{3pt}
\IfFileExists{planned/generated/table_ax_grid_kuro.tex}{\begin{adjustbox}{max width=\columnwidth}%
\providecommand{\gcvdry}[1]{\textcolor{red}{#1}}
\providecommand{\ProvTableAxGridKuro}{Seeds per cell 5. Analysis of the finished runs}
\begin{tabular}{@{}lllcccc@{}}
\toprule
Model & Type & Level & $n$ & mIoU & Drop & Retention (\%) \\
\midrule
SAT & Clean & -- & 5 & 53.0 $\pm$ 4.3 & -- & -- \\
 & Shift & L1 & 5 & 50.8 $\pm$ 5.2 & 2.1 $\pm$ 2.4 & 95.9 $\pm$ 4.9 \\
 & Shift & L2 & 5 & 50.6 $\pm$ 1.7 & 2.3 $\pm$ 5.0 & 96.1 $\pm$ 9.0 \\
 & Coarse & L1 & 5 & 56.1 $\pm$ 1.9 & $-$3.1 $\pm$ 4.1 & 106.3 $\pm$ 7.9 \\
 & Coarse & L2 & 5 & 45.4 $\pm$ 0.9 & 7.6 $\pm$ 4.7 & 86.1 $\pm$ 7.5 \\
 & Boundary & L1 & 5 & 52.4 $\pm$ 4.1 & 0.6 $\pm$ 3.6 & 99.0 $\pm$ 6.9 \\
 & Boundary & L2 & 5 & 46.0 $\pm$ 2.3 & 6.9 $\pm$ 4.9 & 87.4 $\pm$ 8.1 \\
 & Omission & L1 & 5 & 48.1 $\pm$ 2.8 & 4.9 $\pm$ 3.8 & 91.0 $\pm$ 6.6 \\
 & Omission & L2 & 5 & 43.9 $\pm$ 2.8 & 9.1 $\pm$ 3.0 & 83.0 $\pm$ 4.6 \\
 & CEMS & Real & 5 & 48.8 $\pm$ 4.9 & 0.3 $\pm$ 6.7 & 100.1 $\pm$ 14.0 \\
\addlinespace[1.5pt]
LVD & Clean & -- & 5 & 59.8 $\pm$ 1.9 & -- & -- \\
 & Shift & L1 & 5 & 58.5 $\pm$ 1.2 & 1.3 $\pm$ 2.2 & 97.9 $\pm$ 3.5 \\
 & Shift & L2 & 5 & 56.2 $\pm$ 1.3 & 3.6 $\pm$ 2.5 & 94.1 $\pm$ 4.1 \\
 & Coarse & L1 & 5 & 57.4 $\pm$ 4.2 & 2.4 $\pm$ 4.7 & 96.0 $\pm$ 7.9 \\
 & Coarse & L2 & 5 & 50.7 $\pm$ 1.8 & 9.1 $\pm$ 3.2 & 84.9 $\pm$ 4.8 \\
 & Boundary & L1 & 5 & 57.5 $\pm$ 0.7 & 2.3 $\pm$ 2.0 & 96.2 $\pm$ 3.3 \\
 & Boundary & L2 & 5 & 51.7 $\pm$ 1.3 & 8.1 $\pm$ 0.8 & 86.5 $\pm$ 1.0 \\
 & Omission & L1 & 5 & 57.7 $\pm$ 1.8 & 2.0 $\pm$ 2.4 & 96.6 $\pm$ 4.1 \\
 & Omission & L2 & 5 & 50.8 $\pm$ 3.2 & 9.0 $\pm$ 2.1 & 84.9 $\pm$ 3.7 \\
 & CEMS & Real & 5 & 51.0 $\pm$ 1.5 & 2.4 $\pm$ 2.2 & 95.5 $\pm$ 4.0 \\
\addlinespace[1.5pt]
DOFA & Clean & -- & 5 & 53.5 $\pm$ 1.8 & -- & -- \\
 & Shift & L1 & 5 & 50.8 $\pm$ 3.3 & 2.7 $\pm$ 4.0 & 95.1 $\pm$ 7.2 \\
 & Shift & L2 & 5 & 50.7 $\pm$ 2.5 & 2.8 $\pm$ 1.2 & 94.7 $\pm$ 2.4 \\
 & Coarse & L1 & 5 & 53.7 $\pm$ 2.2 & $-$0.2 $\pm$ 1.0 & 100.3 $\pm$ 1.9 \\
 & Coarse & L2 & 5 & 48.4 $\pm$ 2.4 & 5.1 $\pm$ 3.8 & 90.6 $\pm$ 6.8 \\
 & Boundary & L1 & 5 & 53.6 $\pm$ 3.3 & $-$0.1 $\pm$ 2.4 & 100.1 $\pm$ 4.6 \\
 & Boundary & L2 & 5 & 50.1 $\pm$ 2.8 & 3.5 $\pm$ 2.9 & 93.6 $\pm$ 5.2 \\
 & Omission & L1 & 5 & 49.4 $\pm$ 5.8 & 4.1 $\pm$ 4.6 & 92.2 $\pm$ 8.8 \\
 & Omission & L2 & 5 & 44.3 $\pm$ 6.4 & 9.2 $\pm$ 5.0 & 82.6 $\pm$ 10.0 \\
 & CEMS & Real & 5 & 49.5 $\pm$ 1.6 & 0.5 $\pm$ 2.8 & 99.1 $\pm$ 5.5 \\
\addlinespace[1.5pt]
TM & Clean & -- & 5 & 58.0 $\pm$ 1.1 & -- & -- \\
 & Shift & L1 & 5 & 56.1 $\pm$ 1.9 & 1.9 $\pm$ 1.5 & 96.7 $\pm$ 2.6 \\
 & Shift & L2 & 5 & 54.5 $\pm$ 1.1 & 3.5 $\pm$ 2.1 & 93.9 $\pm$ 3.4 \\
 & Coarse & L1 & 5 & 57.1 $\pm$ 2.3 & 0.9 $\pm$ 2.6 & 98.5 $\pm$ 4.4 \\
 & Coarse & L2 & 5 & 49.2 $\pm$ 2.4 & 8.8 $\pm$ 2.6 & 84.8 $\pm$ 4.4 \\
 & Boundary & L1 & 5 & 54.4 $\pm$ 2.3 & 3.6 $\pm$ 2.8 & 93.8 $\pm$ 4.7 \\
 & Boundary & L2 & 5 & 50.4 $\pm$ 2.6 & 7.6 $\pm$ 3.3 & 87.0 $\pm$ 5.5 \\
 & Omission & L1 & 5 & 54.2 $\pm$ 1.8 & 3.8 $\pm$ 1.6 & 93.4 $\pm$ 2.8 \\
 & Omission & L2 & 5 & 47.1 $\pm$ 2.3 & 10.9 $\pm$ 3.2 & 81.3 $\pm$ 5.1 \\
 & CEMS & Real & 5 & 53.4 $\pm$ 1.1 & 0.7 $\pm$ 1.4 & 98.7 $\pm$ 2.5 \\
\addlinespace[1.5pt]
RN & Clean & -- & 5 & 32.5 $\pm$ 3.3 & -- & -- \\
 & Shift & L1 & 5 & 31.0 $\pm$ 0.0 & 1.5 $\pm$ 3.3 & 96.1 $\pm$ 8.7 \\
 & Shift & L2 & 5 & 31.0 $\pm$ 0.0 & 1.5 $\pm$ 3.3 & 96.1 $\pm$ 8.7 \\
 & Coarse & L1 & 5 & 32.2 $\pm$ 2.7 & 0.3 $\pm$ 4.8 & 100.0 $\pm$ 13.7 \\
 & Coarse & L2 & 5 & 34.3 $\pm$ 4.5 & $-$1.8 $\pm$ 6.6 & 106.7 $\pm$ 19.6 \\
 & Boundary & L1 & 5 & 32.4 $\pm$ 3.2 & 0.1 $\pm$ 0.2 & 99.8 $\pm$ 0.4 \\
 & Boundary & L2 & 5 & 31.6 $\pm$ 1.2 & 1.0 $\pm$ 3.8 & 97.8 $\pm$ 10.2 \\
 & Omission & L1 & 5 & 31.0 $\pm$ 0.0 & 1.5 $\pm$ 3.3 & 96.2 $\pm$ 8.7 \\
 & Omission & L2 & 5 & 31.0 $\pm$ 0.0 & 1.5 $\pm$ 3.3 & 96.1 $\pm$ 8.7 \\
 & CEMS & Real & 5 & 31.6 $\pm$ 1.4 & $-$0.1 $\pm$ 0.3 & 100.4 $\pm$ 0.8 \\
\bottomrule
\end{tabular}
\end{adjustbox}}{\tabletbd[6cm]{planned/generated/table\_ax\_grid\_kuro.tex: run scripts/make\_arxiv\_appendix.py}}
\caption{\textbf{Every core cell on Kuro Siwo:} Columns, rows and marks as in \Cref{tab:ax-grid-ches}. CEMS is the Copernicus EMS label, and its Real row is paired with a clean run on the tiles that it covers. Its drop is therefore not the printed clean \miou minus its \miou. \ProvSlot{\ProvTableAxGridKuro}}
\label{tab:ax-grid-kuro}
\end{table}

\begin{table}[t]
\centering\footnotesize\setlength{\tabcolsep}{3pt}
\IfFileExists{planned/generated/table_ax_grid_oem.tex}{\begin{adjustbox}{max width=\columnwidth}%
\providecommand{\gcvdry}[1]{\textcolor{red}{#1}}
\providecommand{\ProvTableAxGridOem}{Seeds per cell 5. Analysis of the finished runs}
\begin{tabular}{@{}lllcccc@{}}
\toprule
Model & Type & Level & $n$ & mIoU & Drop & Retention (\%) \\
\midrule
SAT & Clean & -- & 5 & 63.5 $\pm$ 0.6 & -- & -- \\
 & Shift & L1 & 5 & 63.3 $\pm$ 0.4 & 0.2 $\pm$ 0.6 & 99.7 $\pm$ 0.9 \\
 & Shift & L2 & 5 & 60.4 $\pm$ 0.9 & 3.1 $\pm$ 0.6 & 95.1 $\pm$ 0.9 \\
 & Coarse & L1 & 5 & 62.3 $\pm$ 0.7 & 1.2 $\pm$ 0.4 & 98.1 $\pm$ 0.7 \\
 & Coarse & L2 & 5 & 57.6 $\pm$ 0.6 & 5.9 $\pm$ 0.9 & 90.8 $\pm$ 1.3 \\
 & Boundary & L1 & 5 & 60.0 $\pm$ 0.6 & 3.4 $\pm$ 0.7 & 94.6 $\pm$ 1.1 \\
 & Boundary & L2 & 5 & 52.7 $\pm$ 0.7 & 10.8 $\pm$ 1.0 & 83.1 $\pm$ 1.5 \\
 & Omission & L1 & 5 & 59.6 $\pm$ 1.3 & 3.9 $\pm$ 1.3 & 93.8 $\pm$ 2.1 \\
 & Omission & L2 & 5 & 52.8 $\pm$ 0.9 & 10.7 $\pm$ 1.0 & 83.1 $\pm$ 1.5 \\
\addlinespace[1.5pt]
LVD & Clean & -- & 5 & 63.4 $\pm$ 0.4 & -- & -- \\
 & Shift & L1 & 5 & 62.6 $\pm$ 0.7 & 0.7 $\pm$ 0.9 & 98.9 $\pm$ 1.3 \\
 & Shift & L2 & 5 & 60.3 $\pm$ 1.1 & 3.0 $\pm$ 0.8 & 95.2 $\pm$ 1.3 \\
 & Coarse & L1 & 5 & 62.5 $\pm$ 0.8 & 0.9 $\pm$ 0.6 & 98.6 $\pm$ 0.9 \\
 & Coarse & L2 & 5 & 57.8 $\pm$ 1.1 & 5.6 $\pm$ 1.2 & 91.2 $\pm$ 1.9 \\
 & Boundary & L1 & 5 & 59.9 $\pm$ 1.0 & 3.4 $\pm$ 1.3 & 94.6 $\pm$ 2.0 \\
 & Boundary & L2 & 5 & 54.2 $\pm$ 1.2 & 9.2 $\pm$ 0.8 & 85.5 $\pm$ 1.3 \\
 & Omission & L1 & 5 & 60.5 $\pm$ 0.8 & 2.9 $\pm$ 0.9 & 95.4 $\pm$ 1.3 \\
 & Omission & L2 & 5 & 54.2 $\pm$ 0.9 & 9.2 $\pm$ 1.0 & 85.5 $\pm$ 1.6 \\
\addlinespace[1.5pt]
DOFA & Clean & -- & 5 & 61.0 $\pm$ 0.8 & -- & -- \\
 & Shift & L1 & 5 & 60.1 $\pm$ 0.8 & 0.9 $\pm$ 1.5 & 98.5 $\pm$ 2.4 \\
 & Shift & L2 & 5 & 57.4 $\pm$ 0.9 & 3.6 $\pm$ 1.2 & 94.1 $\pm$ 2.0 \\
 & Coarse & L1 & 5 & 59.0 $\pm$ 0.9 & 2.0 $\pm$ 1.4 & 96.7 $\pm$ 2.3 \\
 & Coarse & L2 & 5 & 56.2 $\pm$ 0.4 & 4.8 $\pm$ 0.5 & 92.1 $\pm$ 0.8 \\
 & Boundary & L1 & 5 & 58.3 $\pm$ 0.9 & 2.7 $\pm$ 1.0 & 95.6 $\pm$ 1.6 \\
 & Boundary & L2 & 5 & 51.3 $\pm$ 0.5 & 9.7 $\pm$ 0.5 & 84.1 $\pm$ 0.6 \\
 & Omission & L1 & 5 & 57.8 $\pm$ 1.3 & 3.1 $\pm$ 1.4 & 94.9 $\pm$ 2.3 \\
 & Omission & L2 & 5 & 51.9 $\pm$ 0.6 & 9.1 $\pm$ 1.2 & 85.1 $\pm$ 1.8 \\
\addlinespace[1.5pt]
TM & Clean & -- & 5 & 55.2 $\pm$ 0.8 & -- & -- \\
 & Shift & L1 & 5 & 54.3 $\pm$ 1.0 & 0.9 $\pm$ 0.7 & 98.4 $\pm$ 1.3 \\
 & Shift & L2 & 5 & 52.9 $\pm$ 0.4 & 2.4 $\pm$ 0.6 & 95.7 $\pm$ 1.1 \\
 & Coarse & L1 & 5 & 55.4 $\pm$ 1.0 & $-$0.2 $\pm$ 0.5 & 100.4 $\pm$ 0.9 \\
 & Coarse & L2 & 5 & 50.6 $\pm$ 0.4 & 4.7 $\pm$ 0.9 & 91.6 $\pm$ 1.6 \\
 & Boundary & L1 & 5 & 53.7 $\pm$ 0.6 & 1.5 $\pm$ 0.8 & 97.2 $\pm$ 1.5 \\
 & Boundary & L2 & 5 & 46.4 $\pm$ 1.5 & 8.9 $\pm$ 1.5 & 83.9 $\pm$ 2.7 \\
 & Omission & L1 & 5 & 52.3 $\pm$ 0.7 & 2.9 $\pm$ 0.5 & 94.7 $\pm$ 0.8 \\
 & Omission & L2 & 5 & 47.3 $\pm$ 0.6 & 7.9 $\pm$ 1.3 & 85.7 $\pm$ 2.2 \\
\addlinespace[1.5pt]
RN & Clean & -- & 5 & 55.6 $\pm$ 0.6 & -- & -- \\
 & Shift & L1 & 5 & 53.6 $\pm$ 2.2 & 2.0 $\pm$ 2.4 & 96.5 $\pm$ 4.4 \\
 & Shift & L2 & 5 & 51.7 $\pm$ 0.7 & 3.9 $\pm$ 1.1 & 93.0 $\pm$ 1.9 \\
 & Coarse & L1 & 5 & 53.7 $\pm$ 1.2 & 2.0 $\pm$ 0.9 & 96.5 $\pm$ 1.7 \\
 & Coarse & L2 & 5 & 49.7 $\pm$ 0.8 & 6.0 $\pm$ 1.3 & 89.3 $\pm$ 2.2 \\
 & Boundary & L1 & 5 & 53.3 $\pm$ 0.6 & 2.4 $\pm$ 1.0 & 95.8 $\pm$ 1.7 \\
 & Boundary & L2 & 5 & 48.8 $\pm$ 1.0 & 6.9 $\pm$ 1.2 & 87.7 $\pm$ 2.1 \\
 & Omission & L1 & 5 & 53.5 $\pm$ 0.8 & 2.1 $\pm$ 1.2 & 96.2 $\pm$ 2.2 \\
 & Omission & L2 & 5 & 50.9 $\pm$ 0.2 & 4.8 $\pm$ 0.6 & 91.4 $\pm$ 0.9 \\
\bottomrule
\end{tabular}
\end{adjustbox}}{\tabletbd[6cm]{planned/generated/table\_ax\_grid\_oem.tex: run scripts/make\_arxiv\_appendix.py}}
\caption{\textbf{Every core cell on OpenEarthMap:} Columns, rows and marks as in \Cref{tab:ax-grid-ches}. \ProvSlot{\ProvTableAxGridOem}}
\label{tab:ax-grid-oem}
\end{table}

\section{Effect Sizes of the Ablation Contrasts}
\label{sec:ax-effects}

The main text reports each contrast in points with an interval.
This section adds a standardized effect size, so contrasts with different seed spread can be compared on one scale.

\paragraph{Definition:}
Each contrast is a paired difference per model, dataset and seed between two arms that share the masks.
Cohen's $d$ is the mean of these differences divided by their standard deviation,
\begin{equation}
d = \frac{\bar{x}}{s_x},
\label{eq:ax-cohen}
\end{equation}
where $\bar{x}$ and $s_x$ are the mean and the sample standard deviation of the differences.
A positive value means that the first arm of the contrast has the larger drop, or the larger \miou for the subset contrast.
An effect size is printed from \AxMinPairs{} paired seeds on, because fewer seeds give an unstable deviation.
An extension dataset enters a figure once one model has that many pairs.

\newcommand{\gcvEffectsCaption}{\textbf{Effect size per contrast and dataset:} Mean paired change in points and Cohen's $d$ over the pooled model and seed pairs. Models counts the models with a pair and $n$ the pairs. The subset contrast compares test \miou and every other contrast the paired drop. A dash marks a contrast without a pair. \ProvSlot{\ProvTableAxEffectSizes}}
\ifdefined\preprintbuild
\begin{table}[t]
\centering\footnotesize\setlength{\tabcolsep}{3pt}
\IfFileExists{planned/generated/table_ax_effect_sizes_part1.tex}{%
\providecommand{\gcvdry}[1]{\textcolor{red}{#1}}
\providecommand{\ProvTableAxEffectSizes}{Analysis of the finished runs}
\begin{tabular}{@{}llcccc@{}}
\toprule
Contrast & Dataset & Models & $n$ & Change & Cohen's $d$ \\
\midrule
Frozen $-$ fine-tuned & \burn & 3 & 15 & $-$5.4 & $-$1.16 \\
 & \flair & 3 & 15 & $-$2.8 & $-$1.28 \\
\addlinespace[1.5pt]
SegFormer $-$ UNet & \burn & 5 & 25 & 1.4 & 0.42 \\
 & \flair & 5 & 25 & $-$0.1 & $-$0.07 \\
 & Chesapeake & 2 & 10 & 0.0 & $-$0.03 \\
 & CloudSEN12 & 2 & 10 & $-$0.1 & $-$0.07 \\
 & DFC2020 & 2 & 10 & $-$0.1 & $-$0.06 \\
 & EnviroAtlas & 2 & 10 & $-$3.0 & $-$1.05 \\
 & SpaceNet 2 & 2 & 10 & 0.4 & 0.26 \\
 & Kuro Siwo & 2 & 10 & 0.2 & 0.04 \\
 & OpenEarthMap & 2 & 10 & $-$0.1 & $-$0.12 \\
Mask2Former $-$ UNet & \burn & 5 & 25 & $-$1.4 & $-$0.14 \\
 & \flair & 5 & 25 & 0.8 & 0.56 \\
 & Chesapeake & 2 & 10 & 2.2 & 0.97 \\
 & CloudSEN12 & 2 & 10 & 0.3 & 0.27 \\
 & DFC2020 & 2 & 10 & 0.3 & 0.19 \\
 & EnviroAtlas & 2 & 10 & $-$2.3 & $-$0.58 \\
 & SpaceNet 2 & 2 & 10 & 5.5 & 1.50 \\
 & Kuro Siwo & 2 & 10 & 0.2 & 0.04 \\
 & OpenEarthMap & 2 & 10 & $-$0.1 & $-$0.09 \\
SAM $-$ UNet & \burn & 5 & 25 & 1.3 & 0.33 \\
 & \flair & 5 & 25 & 0.0 & 0.00 \\
 & Chesapeake & 2 & 10 & 0.1 & 0.07 \\
 & CloudSEN12 & 2 & 10 & 0.6 & 0.37 \\
 & DFC2020 & 2 & 10 & $-$0.5 & $-$0.30 \\
 & EnviroAtlas & 2 & 10 & $-$4.7 & $-$1.79 \\
 & SpaceNet 2 & 2 & 10 & 2.7 & 1.40 \\
 & Kuro Siwo & 2 & 10 & 2.2 & 0.72 \\
 & OpenEarthMap & 2 & 10 & $-$1.3 & $-$1.59 \\
Full Mask2Former $-$ UNet & \burn & 5 & 25 & 0.8 & 0.17 \\
 & \flair & 5 & 25 & 0.2 & 0.13 \\
Pretrained SAM $-$ UNet & \burn & 5 & 25 & 2.9 & 0.24 \\
 & \flair & 5 & 25 & 0.5 & 0.31 \\
\addlinespace[1.5pt]
GCE $-$ cross-entropy & \burn & 5 & 25 & $-$1.2 & $-$0.30 \\
 & \flair & 5 & 25 & 0.1 & 0.05 \\
Soft bootstrapping $-$ cross-entropy & \burn & 5 & 25 & $-$1.2 & $-$0.37 \\
 & \flair & 5 & 25 & 0.1 & 0.06 \\
\bottomrule
\end{tabular}
}{\input{planned/generated/table_ax_effect_sizes}}
\caption{\gcvEffectsCaption{} The table continues on the next page.}
\label{tab:ax-effects}
\end{table}
\IfFileExists{planned/generated/table_ax_effect_sizes_part2.tex}{\begin{table}[t]\ContinuedFloat
\centering\footnotesize\setlength{\tabcolsep}{3pt}
\providecommand{\gcvdry}[1]{\textcolor{red}{#1}}
\providecommand{\ProvTableAxEffectSizes}{Analysis of the finished runs}
\begin{tabular}{@{}llcccc@{}}
\toprule
Contrast & Dataset & Models & $n$ & Change & Cohen's $d$ \\
\midrule
Rate $\times$1/3 $-$ $\times$1 & \burn & 5 & 25 & $-$2.8 & $-$0.86 \\
 & \flair & 5 & 25 & $-$0.7 & $-$0.59 \\
 & Chesapeake & 2 & 10 & $-$0.6 & $-$0.36 \\
 & CloudSEN12 & 2 & 10 & 0.7 & 0.50 \\
 & DFC2020 & 2 & 10 & 0.3 & 0.11 \\
 & EnviroAtlas & 2 & 10 & 1.9 & 0.62 \\
 & SpaceNet 2 & 2 & 10 & $-$1.4 & $-$0.99 \\
 & Kuro Siwo & 2 & 10 & $-$0.4 & $-$0.11 \\
 & OpenEarthMap & 2 & 10 & $-$1.0 & $-$1.39 \\
\addlinespace[1.5pt]
Clean subset $-$ noisy tiles & \burn & 5 & 25 & 14.4 & 2.72 \\
 & \flair & 5 & 25 & $-$1.2 & $-$0.56 \\
\addlinespace[1.5pt]
Shift, \Ltwo $-$ \Lone & \burn & 5 & 25 & 9.7 & 3.00 \\
 & \flair & 5 & 25 & 2.6 & 1.89 \\
 & Chesapeake & 5 & 25 & 6.2 & 5.91 \\
 & CloudSEN12 & 5 & 25 & 4.3 & 4.01 \\
 & DFC2020 & 5 & 25 & 0.0 & 0.00 \\
 & EnviroAtlas & 5 & 25 & 3.8 & 0.82 \\
 & SpaceNet 2 & 5 & 25 & 6.5 & 5.72 \\
 & Kuro Siwo & 5 & 25 & 0.9 & 0.25 \\
 & OpenEarthMap & 5 & 25 & 2.3 & 1.43 \\
Coarse, \Ltwo $-$ \Lone & \burn & 5 & 25 & 15.7 & 3.83 \\
 & \flair & 5 & 25 & 4.7 & 3.49 \\
 & Chesapeake & 5 & 25 & 10.6 & 8.83 \\
 & CloudSEN12 & 5 & 25 & 6.1 & 4.59 \\
 & DFC2020 & 5 & 25 & 0.9 & 0.39 \\
 & EnviroAtlas & 5 & 25 & 4.1 & 1.22 \\
 & SpaceNet 2 & 5 & 25 & 10.1 & 10.15 \\
 & Kuro Siwo & 5 & 25 & 5.7 & 1.02 \\
 & OpenEarthMap & 5 & 25 & 4.2 & 3.01 \\
Boundary, \Ltwo $-$ \Lone & \burn & 5 & 25 & 10.4 & 4.31 \\
 & \flair & 5 & 25 & 3.4 & 2.01 \\
 & Chesapeake & 5 & 25 & 8.1 & 4.07 \\
 & CloudSEN12 & 5 & 25 & 4.7 & 2.79 \\
 & DFC2020 & 5 & 25 & 3.3 & 1.16 \\
 & EnviroAtlas & 5 & 25 & 7.1 & 1.02 \\
 & SpaceNet 2 & 5 & 25 & 7.5 & 3.93 \\
 & Kuro Siwo & 5 & 25 & 4.1 & 1.17 \\
 & OpenEarthMap & 5 & 25 & 6.4 & 3.78 \\
Omission, \Ltwo $-$ \Lone & \burn & 5 & 25 & 12.2 & 1.65 \\
 & \flair & 5 & 25 & 3.2 & 2.02 \\
 & Chesapeake & 5 & 25 & 7.0 & 1.85 \\
 & CloudSEN12 & 5 & 25 & 3.7 & 2.87 \\
 & DFC2020 & 5 & 25 & 1.8 & 0.92 \\
 & EnviroAtlas & 5 & 25 & 7.8 & 0.90 \\
 & SpaceNet 2 & 5 & 25 & 11.6 & 1.92 \\
 & Kuro Siwo & 5 & 25 & 4.7 & 1.18 \\
 & OpenEarthMap & 5 & 25 & 5.3 & 2.86 \\
\bottomrule
\end{tabular}

\caption{\textbf{Effect size per contrast and dataset, continued:} Columns as in the first part.}
\end{table}}{}
\else
\begin{table*}[t]
\centering\footnotesize\setlength{\tabcolsep}{3pt}
\IfFileExists{planned/generated/table_ax_effect_sizes_part1.tex}{\begin{adjustbox}{max width=\textwidth}%
\providecommand{\gcvdry}[1]{\textcolor{red}{#1}}
\providecommand{\ProvTableAxEffectSizes}{Analysis of the finished runs}
\begin{tabular}{@{}llcccc@{}}
\toprule
Contrast & Dataset & Models & $n$ & Change & Cohen's $d$ \\
\midrule
Frozen $-$ fine-tuned & \burn & 3 & 15 & $-$5.4 & $-$1.16 \\
 & \flair & 3 & 15 & $-$2.8 & $-$1.28 \\
\addlinespace[1.5pt]
SegFormer $-$ UNet & \burn & 5 & 25 & 1.4 & 0.42 \\
 & \flair & 5 & 25 & $-$0.1 & $-$0.07 \\
 & Chesapeake & 2 & 10 & 0.0 & $-$0.03 \\
 & CloudSEN12 & 2 & 10 & $-$0.1 & $-$0.07 \\
 & DFC2020 & 2 & 10 & $-$0.1 & $-$0.06 \\
 & EnviroAtlas & 2 & 10 & $-$3.0 & $-$1.05 \\
 & SpaceNet 2 & 2 & 10 & 0.4 & 0.26 \\
 & Kuro Siwo & 2 & 10 & 0.2 & 0.04 \\
 & OpenEarthMap & 2 & 10 & $-$0.1 & $-$0.12 \\
Mask2Former $-$ UNet & \burn & 5 & 25 & $-$1.4 & $-$0.14 \\
 & \flair & 5 & 25 & 0.8 & 0.56 \\
 & Chesapeake & 2 & 10 & 2.2 & 0.97 \\
 & CloudSEN12 & 2 & 10 & 0.3 & 0.27 \\
 & DFC2020 & 2 & 10 & 0.3 & 0.19 \\
 & EnviroAtlas & 2 & 10 & $-$2.3 & $-$0.58 \\
 & SpaceNet 2 & 2 & 10 & 5.5 & 1.50 \\
 & Kuro Siwo & 2 & 10 & 0.2 & 0.04 \\
 & OpenEarthMap & 2 & 10 & $-$0.1 & $-$0.09 \\
SAM $-$ UNet & \burn & 5 & 25 & 1.3 & 0.33 \\
 & \flair & 5 & 25 & 0.0 & 0.00 \\
 & Chesapeake & 2 & 10 & 0.1 & 0.07 \\
 & CloudSEN12 & 2 & 10 & 0.6 & 0.37 \\
 & DFC2020 & 2 & 10 & $-$0.5 & $-$0.30 \\
 & EnviroAtlas & 2 & 10 & $-$4.7 & $-$1.79 \\
 & SpaceNet 2 & 2 & 10 & 2.7 & 1.40 \\
 & Kuro Siwo & 2 & 10 & 2.2 & 0.72 \\
 & OpenEarthMap & 2 & 10 & $-$1.3 & $-$1.59 \\
Full Mask2Former $-$ UNet & \burn & 5 & 25 & 0.8 & 0.17 \\
 & \flair & 5 & 25 & 0.2 & 0.13 \\
Pretrained SAM $-$ UNet & \burn & 5 & 25 & 2.9 & 0.24 \\
 & \flair & 5 & 25 & 0.5 & 0.31 \\
\addlinespace[1.5pt]
GCE $-$ cross-entropy & \burn & 5 & 25 & $-$1.2 & $-$0.30 \\
 & \flair & 5 & 25 & 0.1 & 0.05 \\
Soft bootstrapping $-$ cross-entropy & \burn & 5 & 25 & $-$1.2 & $-$0.37 \\
 & \flair & 5 & 25 & 0.1 & 0.06 \\
\bottomrule
\end{tabular}
\hspace{1.5em}\end{adjustbox}}{\begin{adjustbox}{max width=\columnwidth}\input{planned/generated/table_ax_effect_sizes}\end{adjustbox}}
\caption{\gcvEffectsCaption}
\label{tab:ax-effects}
\end{table*}
\fi

\begin{figure}[t]
\centering
\IfFileExists{planned/figs/fig_ax_effect_frozen.pdf}{\includegraphics[width=\linewidth]{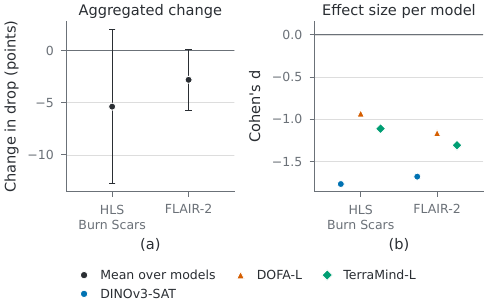}}{\phfig{4cm}{planned/figs/fig\_ax\_effect\_frozen.pdf from figs/make\_effect\_size\_figs.py}}
\caption{\textbf{Frozen encoder against full fine-tuning:} Panel (a) shows the mean change over models with its \CiLevel\,\% interval, and panel (b) Cohen's $d$ of the paired differences per model. The change is the paired drop at the selected \Ltwo type with a frozen encoder minus the drop under full fine-tuning. \ProvSlot{\ResultsProvenance}}
\label{fig:ax-effect-frozen}
\end{figure}

\begin{figure*}[t]
\centering
\IfFileExists{planned/figs/fig_ax_effect_decoders.pdf}{\includegraphics[width=\textwidth]{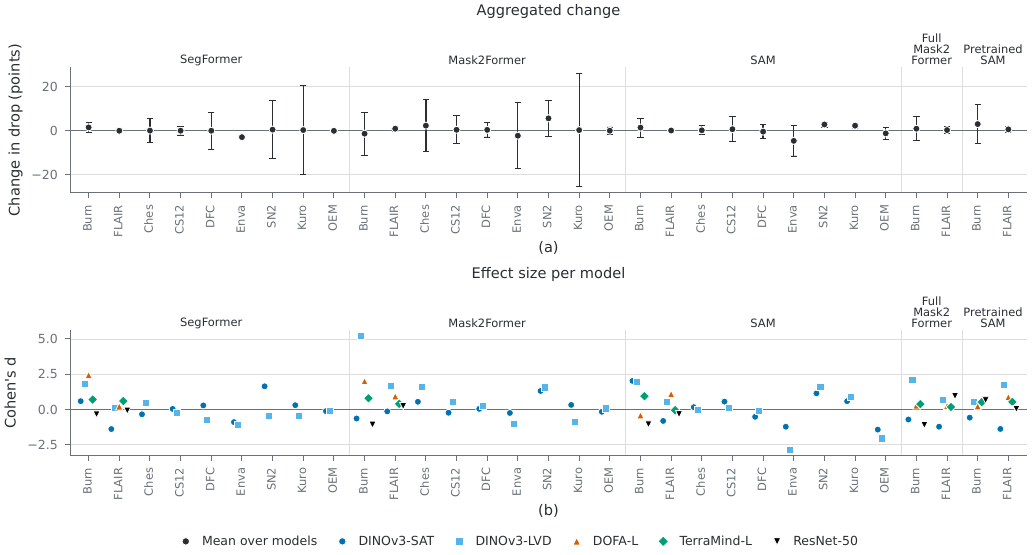}}{\phfig{4cm}{planned/figs/fig\_ax\_effect\_decoders.pdf from figs/make\_effect\_size\_figs.py}}
\caption{\textbf{Decoder heads against the UNet decoder:} Panel (a) shows the mean change over models with its \CiLevel\,\% interval, and panel (b) Cohen's $d$ of the paired differences per model. The change is the mean \Ltwo drop under the head minus the drop under the UNet decoder, per head and dataset. Burn is \burn and FLAIR is \flair. The other labels are Ches (Chesapeake), CS12 (CloudSEN12), DFC (DFC2020), Enva (EnviroAtlas), SN2 (SpaceNet 2), Kuro (Kuro Siwo) and OEM (OpenEarthMap). \ProvSlot{\ResultsProvenance}}
\label{fig:ax-effect-decoders}
\end{figure*}

\begin{figure}[t]
\centering
\IfFileExists{planned/figs/fig_ax_effect_losses.pdf}{\includegraphics[width=\linewidth]{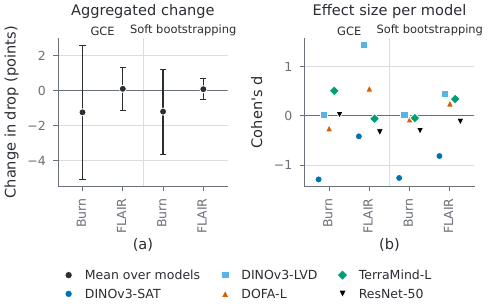}}{\phfig{4cm}{planned/figs/fig\_ax\_effect\_losses.pdf from figs/make\_effect\_size\_figs.py}}
\caption{\textbf{Noise-robust losses against the cross-entropy:} Panel (a) shows the mean change over models with its \CiLevel\,\% interval, and panel (b) Cohen's $d$ of the paired differences per model. The change is the mean \Ltwo drop under the loss minus the drop under the cross-entropy. Burn is \burn and FLAIR is \flair. GCE is generalized cross-entropy. \ProvSlot{\ProvNFourteen}}
\label{fig:ax-effect-losses}
\end{figure}

\begin{figure}[t]
\centering
\IfFileExists{planned/figs/fig_ax_effect_lr.pdf}{\includegraphics[width=\linewidth]{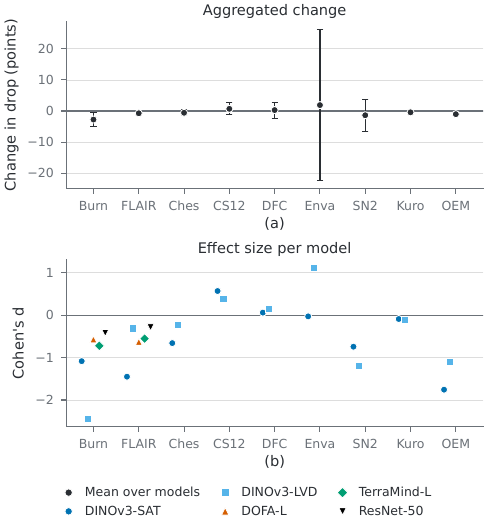}}{\phfig{4cm}{planned/figs/fig\_ax\_effect\_lr.pdf from figs/make\_effect\_size\_figs.py}}
\caption{\textbf{Learning-rate multiplier of one third against one:} Panel (a) shows the mean change over models with its \CiLevel\,\% interval, and panel (b) Cohen's $d$ of the paired differences per model. The change is the mean \Ltwo drop at the lower rate minus the drop at the higher rate, each paired within its rate. Burn is \burn and FLAIR is \flair. The other labels are Ches (Chesapeake), CS12 (CloudSEN12), DFC (DFC2020), Enva (EnviroAtlas), SN2 (SpaceNet 2), Kuro (Kuro Siwo) and OEM (OpenEarthMap). \ProvSlot{\ResultsProvenance}}
\label{fig:ax-effect-lr}
\end{figure}

\begin{figure}[t]
\centering
\IfFileExists{planned/figs/fig_ax_effect_subset.pdf}{\includegraphics[width=\linewidth]{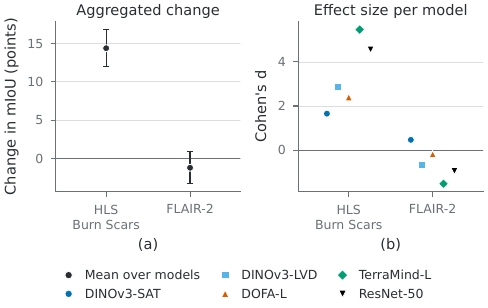}}{\phfig{4cm}{planned/figs/fig\_ax\_effect\_subset.pdf from figs/make\_effect\_size\_figs.py}}
\caption{\textbf{Clean subset against all noisy tiles:} Panel (a) shows the mean change over models with its \CiLevel\,\% interval, and panel (b) Cohen's $d$ of the paired differences per model. The change is the test \miou of the \SubsetPct\,\% clean subset arm minus that of all tiles at the selected \Ltwo type. \ProvSlot{\ResultsProvenance}}
\label{fig:ax-effect-subset}
\end{figure}

\begin{figure*}[t]
\centering
\IfFileExists{planned/figs/fig_ax_effect_levels.pdf}{\includegraphics[width=\textwidth]{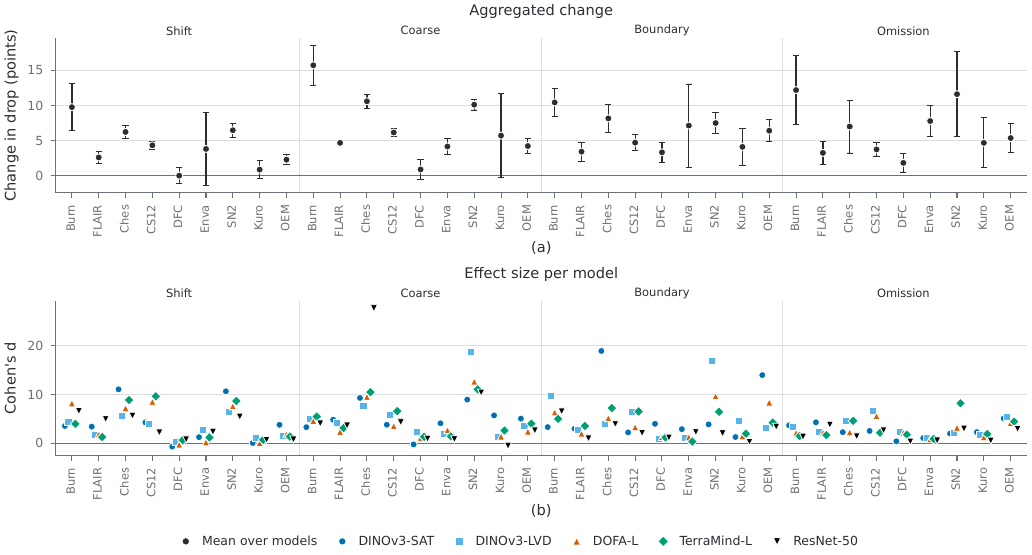}}{\phfig{4cm}{planned/figs/fig\_ax\_effect\_levels.pdf from figs/make\_effect\_size\_figs.py}}
\caption{\textbf{Level two against level one per error type:} Panel (a) shows the mean change over models with its \CiLevel\,\% interval, and panel (b) Cohen's $d$ of the paired differences per model. The change is the paired drop at \Ltwo minus the drop at \Lone. Burn is \burn and FLAIR is \flair. The other labels are Ches (Chesapeake), CS12 (CloudSEN12), DFC (DFC2020), Enva (EnviroAtlas), SN2 (SpaceNet 2), Kuro (Kuro Siwo) and OEM (OpenEarthMap). \ProvSlot{\ResultsProvenance}}
\label{fig:ax-effect-levels}
\end{figure*}

\paragraph{Frozen Encoder:}
Freezing the encoder lowers the paired drop on both core datasets (\Cref{fig:ax-effect-frozen}).
\Cref{tab:ax-effects} gives a negative effect size on both.
Every model in the contrast shares the sign.
The interval of the mean over the three models is wide and includes zero on both datasets.
The direction is thus consistent across models, while the size over models remains uncertain.

\paragraph{Decoder Heads:}
No decoder head shifts the paired drop in one direction across models and datasets (\Cref{fig:ax-effect-decoders}).
The per-model effect sizes scatter on both sides of zero, and the pooled values in \Cref{tab:ax-effects} range from small to large. %
\mtab{tab:decoders} gives the verdict of each head on the same runs. %

\paragraph{Noise Robust Losses:}
Neither noise-robust loss changes the paired drop consistently (\Cref{fig:ax-effect-losses}).
The intervals of the mean change include zero for both losses on both core datasets, and the pooled effect sizes are small.
The per-model values point in both directions, so a gain for one model does not transfer to the next.

\paragraph{Learning Rate:}
At the lower learning rate the paired drop is smaller on both core datasets (\Cref{fig:ax-effect-lr}).
\Cref{tab:ax-effects} gives negative effect sizes there.
The most negative per-model value belongs to \lvd on \burn.
A contrast between models at different rates therefore carries a rate term, which the matched-rate arms of \mtab{tab:verdicts} remove. %

\paragraph{Clean Subset:}
The clean subset beats all noisy tiles by a large margin on \burn and not on \flair (\Cref{fig:ax-effect-subset}).
Every model has a positive effect size on \burn (\Cref{tab:ax-effects}).
On \flair the values are near zero or negative.
The value of clean labels thus depends on the dataset, as the budget analysis in \Cref{sec:supp-budget} states in points.

\paragraph{Levels:}
The drop grows from \Lone to \Ltwo for every type and model on the core datasets (\Cref{fig:ax-effect-levels}).
These are among the largest effect sizes in \Cref{tab:ax-effects}, because the level effect is large against the seed spread.
The calibration target therefore orders the damage as intended.
The type comparison at one level relies on this order.

\clearpage
}{}\fi\fi
\ifdefined\plannedbuild
\ifdefined\cvprbuild
\else

\clearpage

\clearpage

\clearpage
\fi
\ifdefined\cvprbuild\def\axrealfigure{\begin{figure*}[t]}\else\def\axrealfigure{\begin{figure*}[p]}\fi
\section{Examples of the Real and Simulated Label Errors}
\label{sec:ax-real}

\ifdefined\cvprbuild\else
This section shows the real label of every pair dataset next to the reference label and the matched simulated label.
Each gallery holds training tiles of the real arm, chosen by a fixed rule from the tiles where the two labels disagree.
\fi
\ifdefined\cvprbuild
This section shows the real label of one pair dataset next to the reference label and the matched simulated label.
The gallery holds training tiles of the real arm, chosen by a fixed rule from the tiles where the two labels disagree.
\fi
The rule ranks the confusions between a reference class and a real label class by their pixel count.
Each of the most frequent confusions contributes the tile where it covers the largest share of pixels.
The label \miou under a difference map is the mean IoU over the classes present in that tile.
\Cref{fig:ax-real-confusion} then summarizes the same disagreement over all training tiles of the real arm.

\axrealfigure
\centering
\IfFileExists{planned/figs/fig_real_confusion.pdf}{\includegraphics[width=\textwidth]{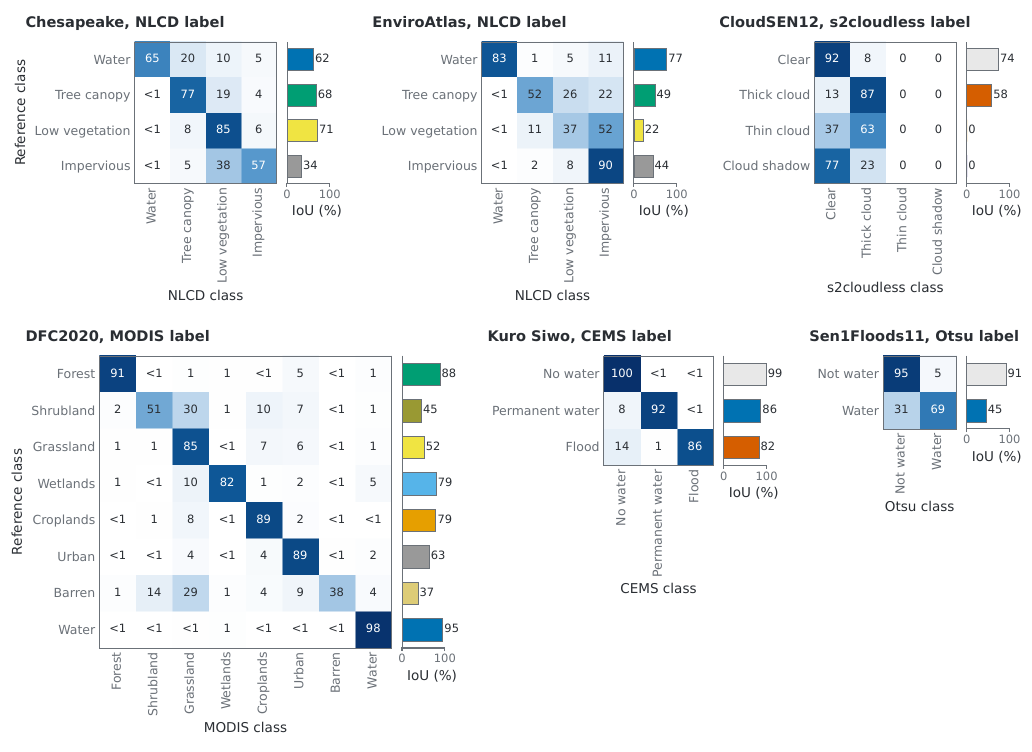}}{\phfig{8cm}{planned/figs/fig\_real\_confusion.pdf from figs/make\_real\_label\_gallery.py}}
\caption{\textbf{Confusion between the reference label and the real label:} Each matrix gives the share of the pixels of a reference class that the real label assigns to each class, in percent. Rows are reference classes and sum to 100. The bars beside a matrix give the IoU of the real label per class against the reference. Shares cover the training tiles of the real arm, and <1 marks a share below one percent.}
\label{fig:ax-real-confusion}
\end{figure*}

\paragraph{Confusion per Dataset:}
Real labels fail in different ways (\Cref{fig:ax-real-confusion}).
Both NLCD labels confuse the low vegetation and impervious classes most.
On CloudSEN12 the s2cloudless label has two classes only, so the thin cloud and shadow classes never appear in it.
The MODIS label moves shrubland and barren pixels to grassland.
By contrast, the Copernicus EMS and Otsu labels keep the background class and lose a part of the water classes.

\axrealfigure
\centering
\IfFileExists{planned/figs/fig_real_chesapeake.pdf}{\includegraphics[width=\textwidth]{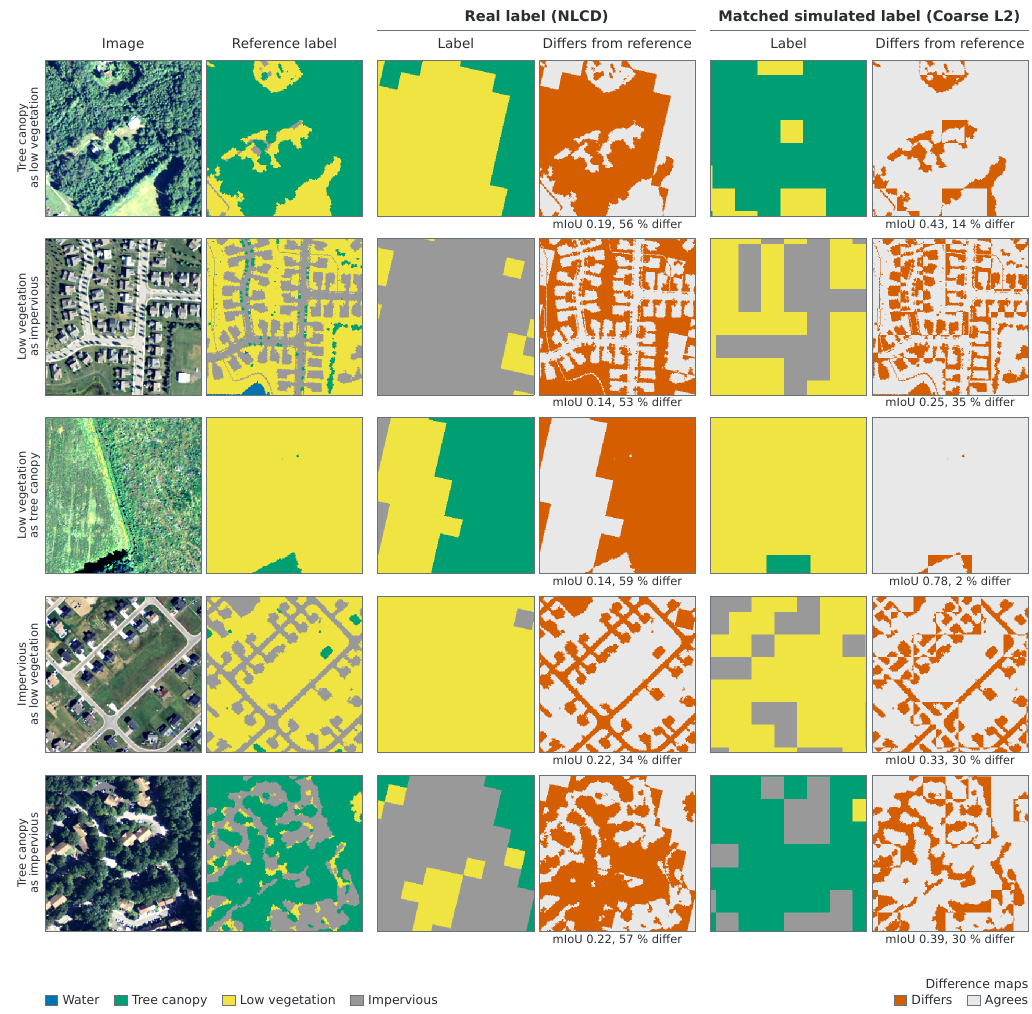}}{\phfig{8cm}{planned/figs/fig\_real\_chesapeake.pdf from figs/make\_real\_label\_gallery.py}}
\caption{\textbf{Real label errors on Chesapeake:} Each row is one training tile, named on the left by its largest confusion. The first two columns show the NAIP image and the reference label. The third column shows the NLCD label and the fourth the pixels where it differs from the reference. The last two columns show the matched simulated label and its difference map. The text under a difference map gives the label \miou of the tile and the share of differing pixels.}
\label{fig:ax-real-ches}
\end{figure*}

\paragraph{Chesapeake:}
The NLCD label is a grid of coarse blocks that ignores roads, houses and field edges (\Cref{fig:ax-real-ches}).
Whole blocks disagree with the reference.
The matched \etype{\hlRealMatchedTypeChes} label at \hlRealMatchedLevelChes{} keeps the dominant class of a block, so its differences stay near object outlines.
Over the training and validation tiles, the NLCD label scores a label \miou of \RealLabelMiouChes.

\ifdefined\cvprbuild\else
\begin{figure*}[p]
\centering
\IfFileExists{planned/figs/fig_real_enviroatlas.pdf}{\includegraphics[width=\textwidth]{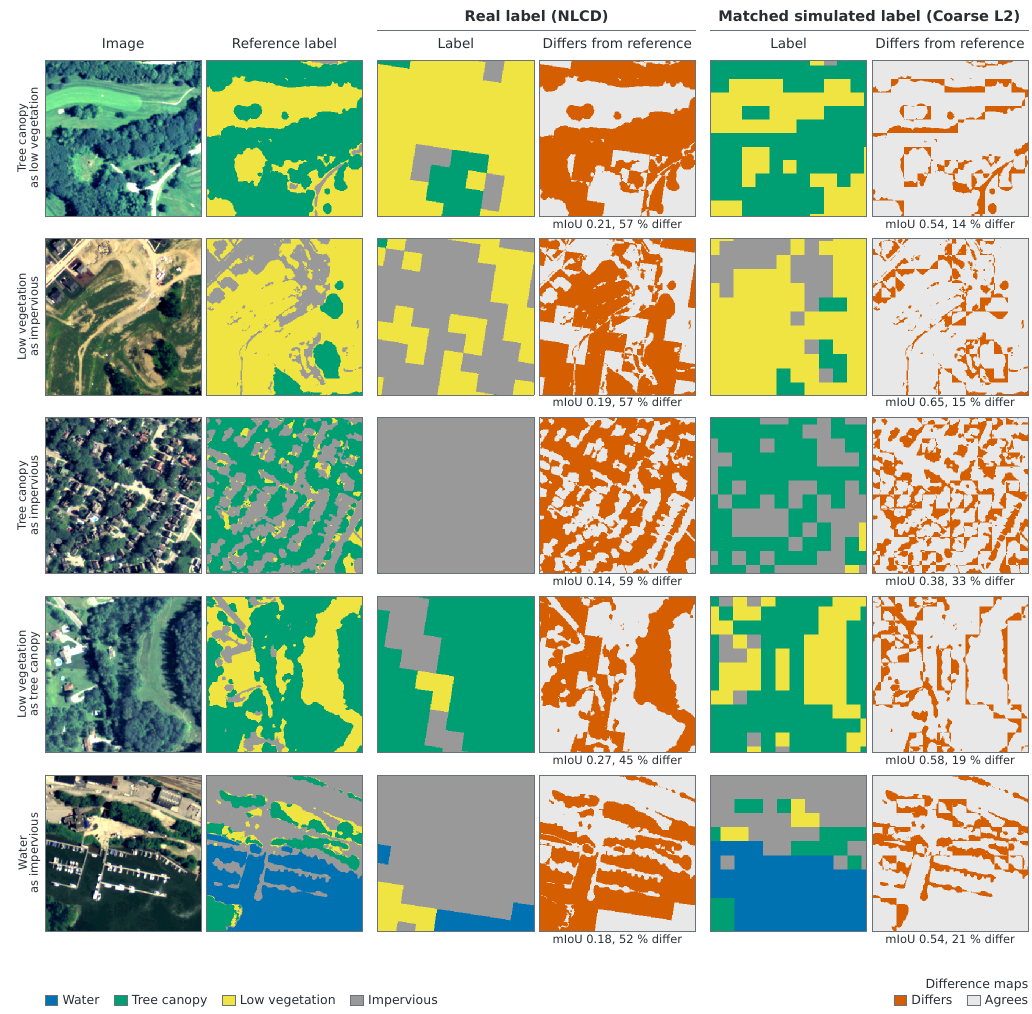}}{\phfig{8cm}{planned/figs/fig\_real\_enviroatlas.pdf from figs/make\_real\_label\_gallery.py}}
\caption{\textbf{Real label errors on EnviroAtlas:} Each row is one training tile, named on the left by its largest confusion. The first two columns show the NAIP image and the reference label. The third column shows the NLCD label and the fourth the pixels where it differs from the reference. The last two columns show the matched simulated label and its difference map. The text under a difference map gives the label \miou of the tile and the share of differing pixels.}
\label{fig:ax-real-enva}
\end{figure*}

\paragraph{EnviroAtlas:}
On EnviroAtlas the NLCD label shows the same block pattern as on Chesapeake (\Cref{fig:ax-real-enva}).
In the third row it assigns almost every pixel to the impervious class and removes the trees and lawns between houses.
The matched \etype{\hlRealMatchedTypeEnva} label at \hlRealMatchedLevelEnva{} keeps all classes of the tile.
The label \miou of the NLCD label over the training and validation tiles is \RealLabelMiouEnva.

\begin{figure*}[p]
\centering
\IfFileExists{planned/figs/fig_real_cloudsen12.pdf}{\includegraphics[width=\textwidth]{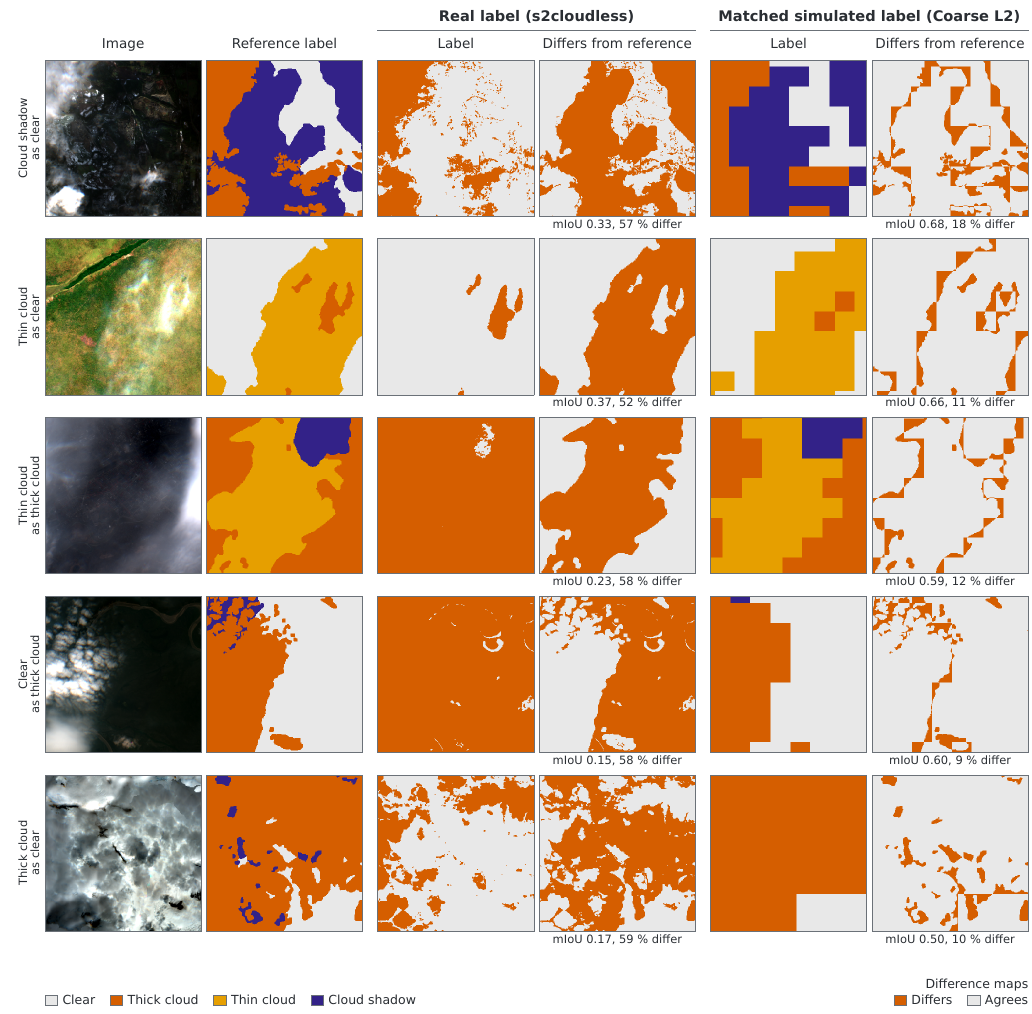}}{\phfig{8cm}{planned/figs/fig\_real\_cloudsen12.pdf from figs/make\_real\_label\_gallery.py}}
\caption{\textbf{Real label errors on CloudSEN12:} Each row is one training tile, named on the left by its largest confusion. The first two columns show the Sentinel-2 image and the reference label. The third column shows the s2cloudless label and the fourth the pixels where it differs from the reference. The last two columns show the matched simulated label and its difference map. The text under a difference map gives the label \miou of the tile and the share of differing pixels.}
\label{fig:ax-real-cs}
\end{figure*}

\paragraph{CloudSEN12:}
The s2cloudless label follows cloud outlines at full resolution, so its error is not a coarse one (\Cref{fig:ax-real-cs}).
It is a class error, because cloud shadow becomes clear and thin cloud becomes clear or thick cloud.
The matched \etype{\hlRealMatchedTypeCsTwelve} label at \hlRealMatchedLevelCsTwelve{} keeps all four classes and differs mainly along outlines.
The label \miou of the s2cloudless label over the training and validation tiles is \RealLabelMiouCsTwelve.

\begin{figure*}[p]
\centering
\IfFileExists{planned/figs/fig_real_dfc2020.pdf}{\includegraphics[width=\textwidth]{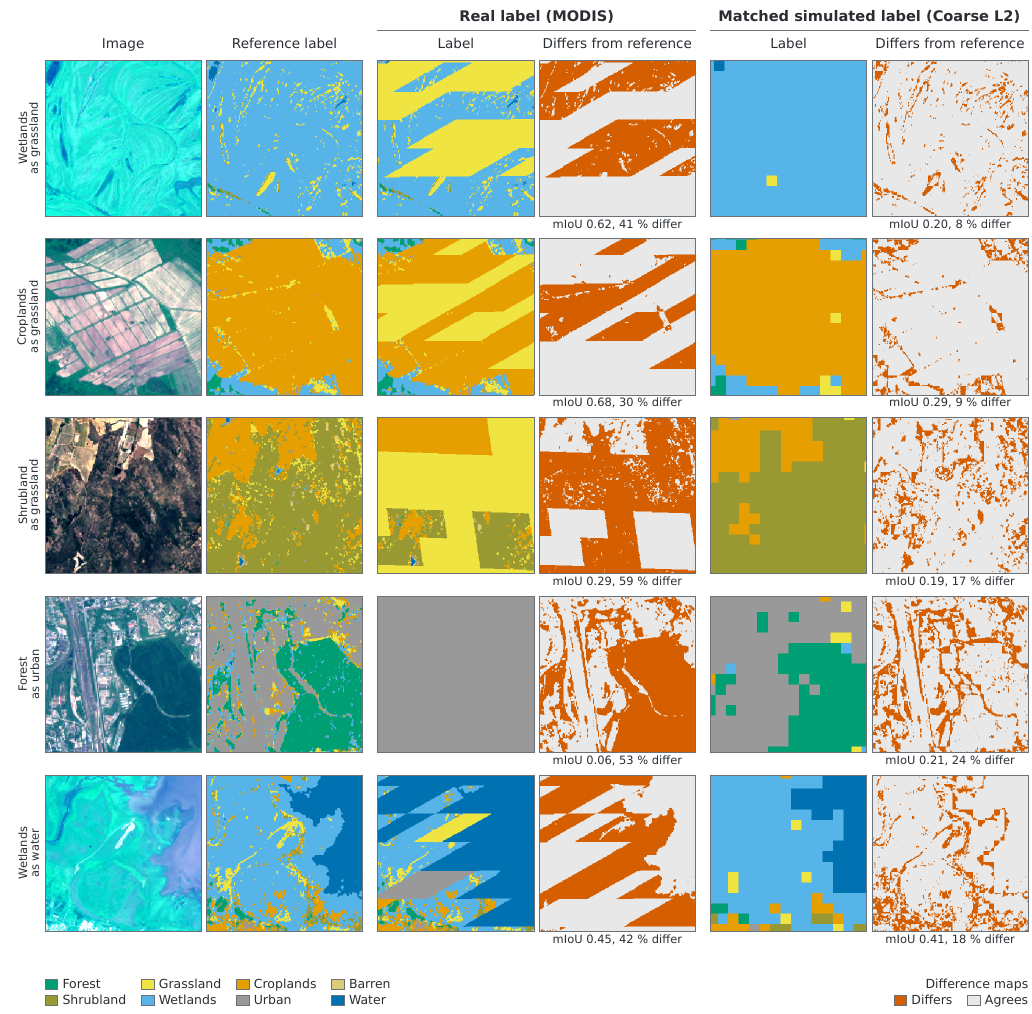}}{\phfig{8cm}{planned/figs/fig\_real\_dfc2020.pdf from figs/make\_real\_label\_gallery.py}}
\caption{\textbf{Real label errors on DFC2020:} Each row is one training tile, named on the left by its largest confusion. The first two columns show the Sentinel-2 image and the reference label. The third column shows the MODIS label and the fourth the pixels where it differs from the reference. The last two columns show the matched simulated label and its difference map. The text under a difference map gives the label \miou of the tile and the share of differing pixels.}
\label{fig:ax-real-dfc}
\end{figure*}

\paragraph{DFC2020:}
In the MODIS label each large slanted cell carries one class (\Cref{fig:ax-real-dfc}).
A cell can cover most of a tile, as in the fourth row where the forest is labeled urban.
The matched \etype{\hlRealMatchedTypeDfc} label at \hlRealMatchedLevelDfc{} uses far smaller blocks, so it removes fine structure and keeps the large regions.
The label \miou of the MODIS label over the training and validation tiles is \RealLabelMiouDfc.

\begin{figure*}[p]
\centering
\IfFileExists{planned/figs/fig_real_kuro_siwo.pdf}{\includegraphics[width=\textwidth]{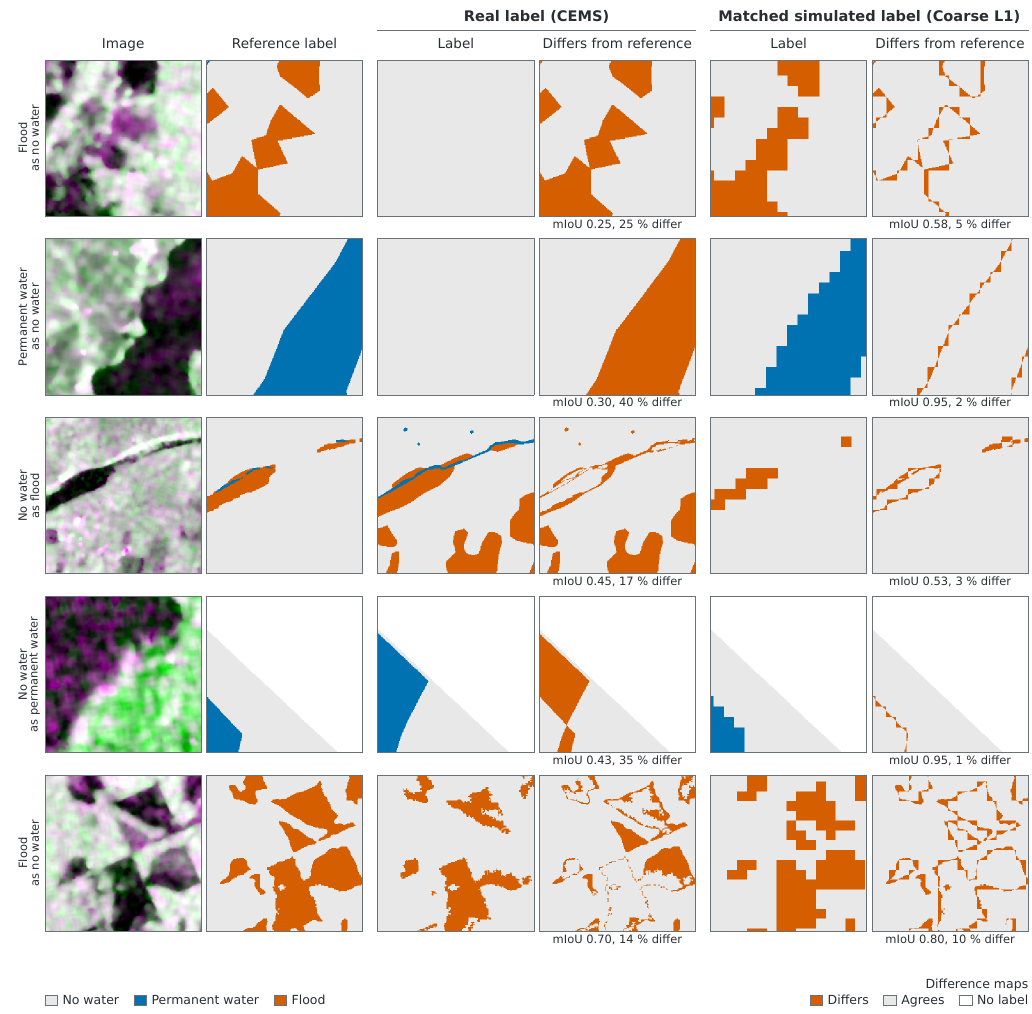}}{\phfig{8cm}{planned/figs/fig\_real\_kuro\_siwo.pdf from figs/make\_real\_label\_gallery.py}}
\caption{\textbf{Real label errors on Kuro Siwo:} Each row is one training tile, named on the left by its largest confusion. The first two columns show the SAR composite and the reference label. The third column shows the Copernicus EMS label and the fourth the pixels where it differs from the reference. The last two columns show the matched simulated label and its difference map. The text under a difference map gives the label \miou of the tile and the share of differing pixels. White marks pixels without a label.}
\label{fig:ax-real-kuro}
\end{figure*}

\paragraph{Kuro Siwo:}
The Copernicus EMS label differs by whole objects (\Cref{fig:ax-real-kuro}).
It misses complete flood and permanent water regions in some tiles and adds flood polygons in others.
The matched \etype{\hlRealMatchedTypeKuro} label at \hlRealMatchedLevelKuro{} keeps the regions and differs mainly along their outlines.
The label \miou over the tiles it covers is \RealLabelMiouKuro, so the chosen tiles are exceptions.

\begin{figure*}[p]
\centering
\IfFileExists{planned/figs/fig_real_sen1floods11.pdf}{\includegraphics[width=\textwidth]{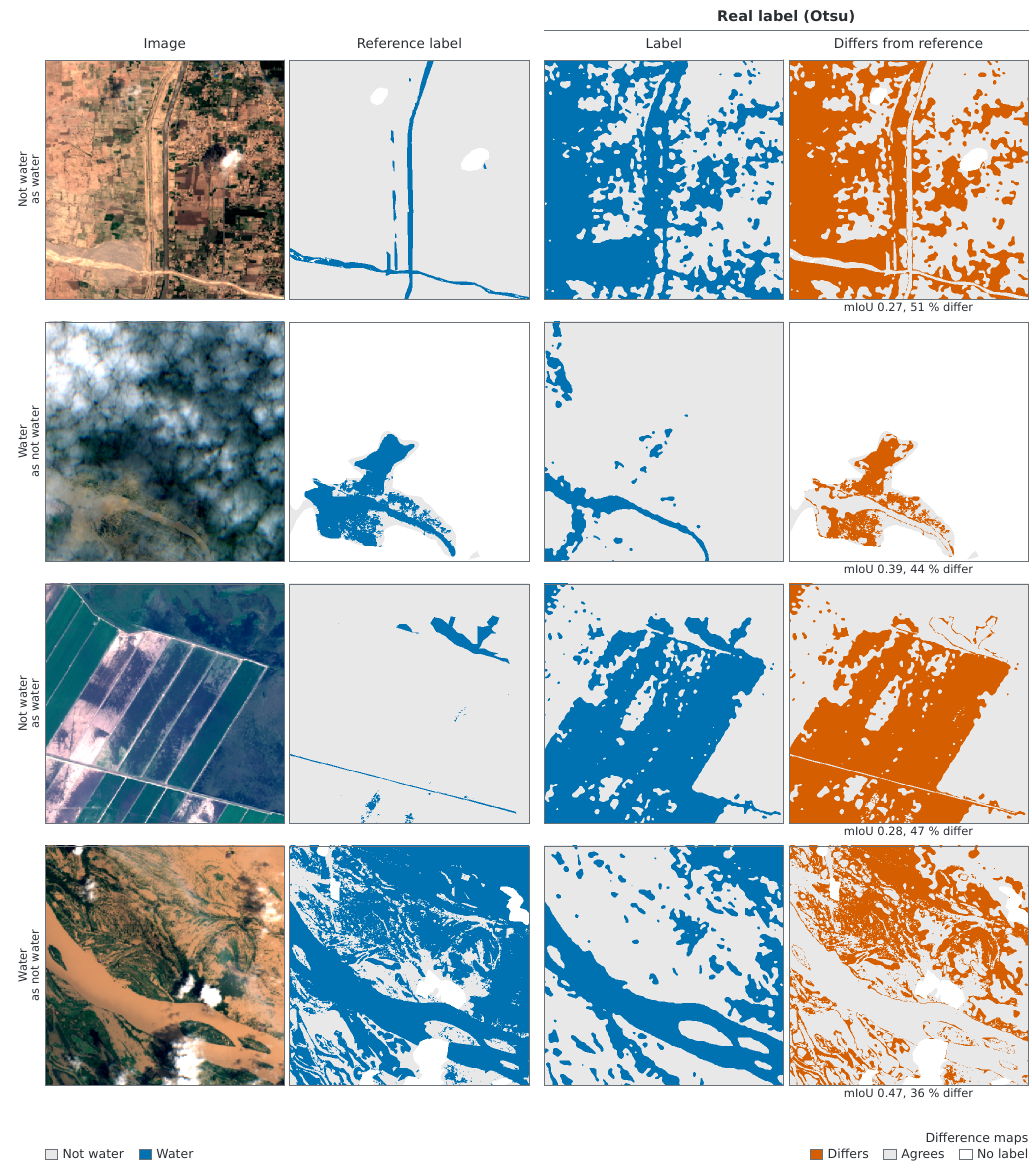}}{\phfig{8cm}{planned/figs/fig\_real\_sen1floods11.pdf from figs/make\_real\_label\_gallery.py}}
\caption{\textbf{Real label errors on \sfloods:} Each row is one training chip, named on the left by its largest confusion. The first two columns show the Sentinel-2 image and the hand label. The third column shows the Otsu label and the fourth the pixels where it differs from the hand label. The text under a difference map gives the label \miou of the chip and the share of differing pixels. White marks pixels without a label. This dataset has no simulated arm, so the figure has no matched label.}
\label{fig:ax-real-sen}
\end{figure*}

\paragraph{Sen1Floods11:}
The Otsu label errs in both directions (\Cref{fig:ax-real-sen}).
It marks large areas of land as water in the first and third rows.
It misses a part of the open water in the second and fourth rows.
The hand label leaves a part of some chips unlabeled, whereas the Otsu label covers the whole chip.
\fi

\ifdefined\cvprbuild\else
\clearpage
\fi
\ifdefined\cvprbuild\else%
\subsection{Simulated Label Errors}
\label{sec:ax-gallery}

\ifdefined\cvprbuild\else
This section shows what the simulated label errors look like on training tiles of every dataset that carries them.
\fi
\ifdefined\cvprbuild
This section shows what the simulated label errors look like on training tiles of \burn at \Ltwo.
\fi
\ifdefined\cvprbuild\else
Each dataset has one page at \Ltwo and one at \Lone.
Both pages show the same tiles.
\fi
\ifdefined\cvprbuild\Cref{fig:ax-gallery-burn_scars}\else Each page of \Crefrange{fig:ax-gallery-burn_scars}{fig:ax-gallery-openearthmap-l1}\fi{} holds one block per tile with the image, the reference label and the four corrupted labels.
The four error types are \etype{shift}, \etype{coarse}, \etype{boundary} and \etype{omission}, drawn in the class colors of \ifdefined\arxivbuild the dataset cards (\Cref{sec:ax-cards})\else the dataset\fi.
A small difference map under each corrupted label marks the labeled pixels that differ from the reference label.
The label \miou of the tile against its reference label and the share of changed pixels are printed beside the map.

The corrupted labels are the packed masks that the runs trained on, read from the first noise realization.
No mask was simulated again for these figures.
The tiles follow \ifdefined\arxivbuild the fixed random order of the dataset cards\else one fixed random order\fi{} over the training tiles.
A tile is shown when its reference label holds two classes or more and every error type changes it visibly at \Ltwo.
\ifdefined\cvprbuild\else
The tiles of EnviroAtlas, SpaceNet 2 and OpenEarthMap are the tiles of their dataset cards.
\fi
The numbers are per tile, so they need not match the calibrated label \miou of the whole training set.

\begin{figure*}[p]
\centering
\IfFileExists{planned/figs/fig_gallery_burn_scars.pdf}{\includegraphics[width=\textwidth]{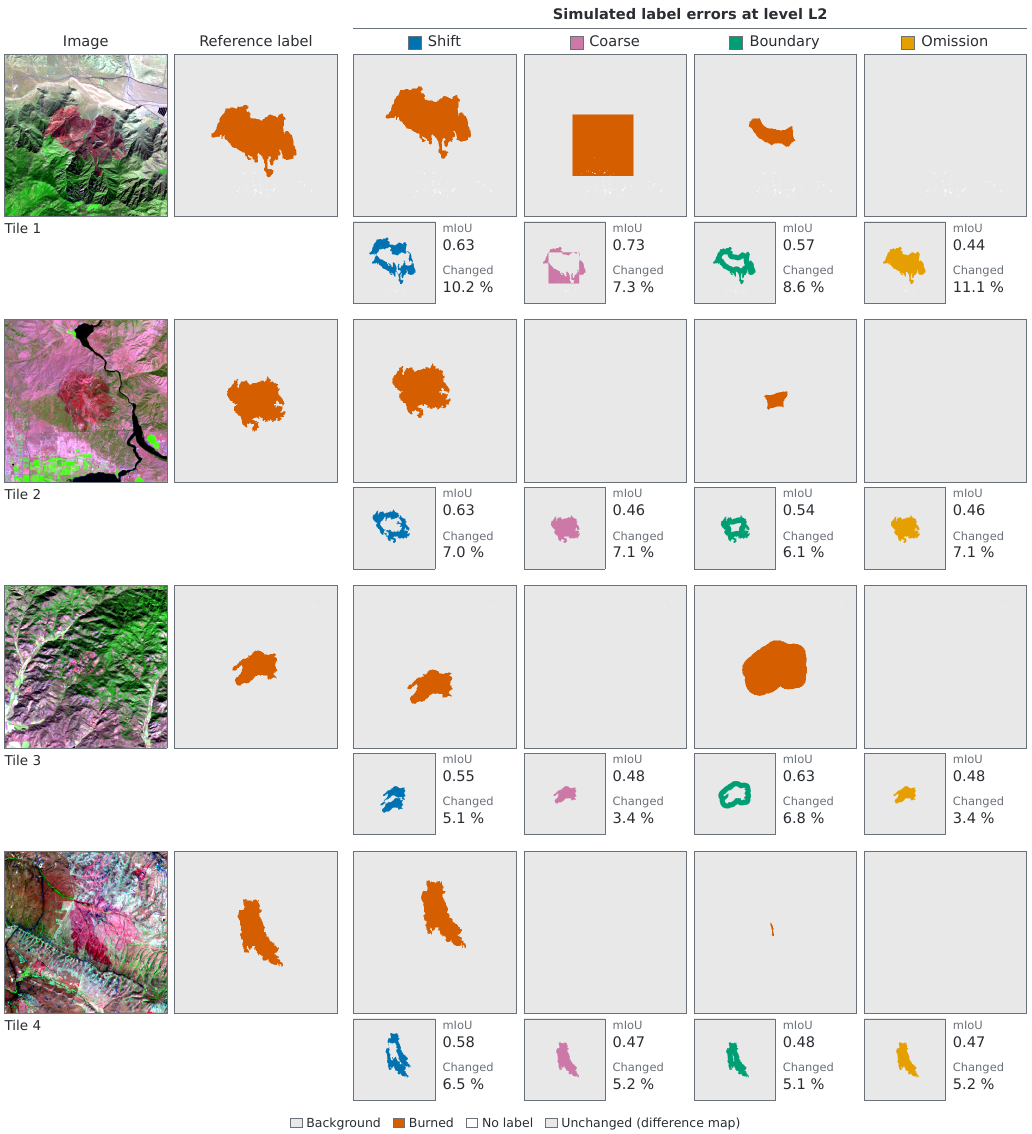}}{\phfig{8cm}{planned/figs/fig\_gallery\_burn\_scars.pdf from figs/make\_corruption\_gallery.py}}
\caption{\textbf{Simulated label errors on \burn at \Ltwo:} Each block is one training tile. The first column shows a false-color composite of the second shortwave infrared, near infrared and red bands and the second column the reference label. The next four columns show the labels under \etype{shift}, \etype{coarse}, \etype{boundary} and \etype{omission}. The small map under each corrupted label marks the changed pixels in the color of the error type. The label \miou of the tile and its share of changed pixels are printed beside the map. White marks pixels without a label.}
\label{fig:ax-gallery-burn_scars}
\end{figure*}

\ifdefined\cvprbuild\else
\begin{figure*}[p]
\centering
\IfFileExists{planned/figs/fig_gallery_burn_scars_L1.pdf}{\includegraphics[width=\textwidth]{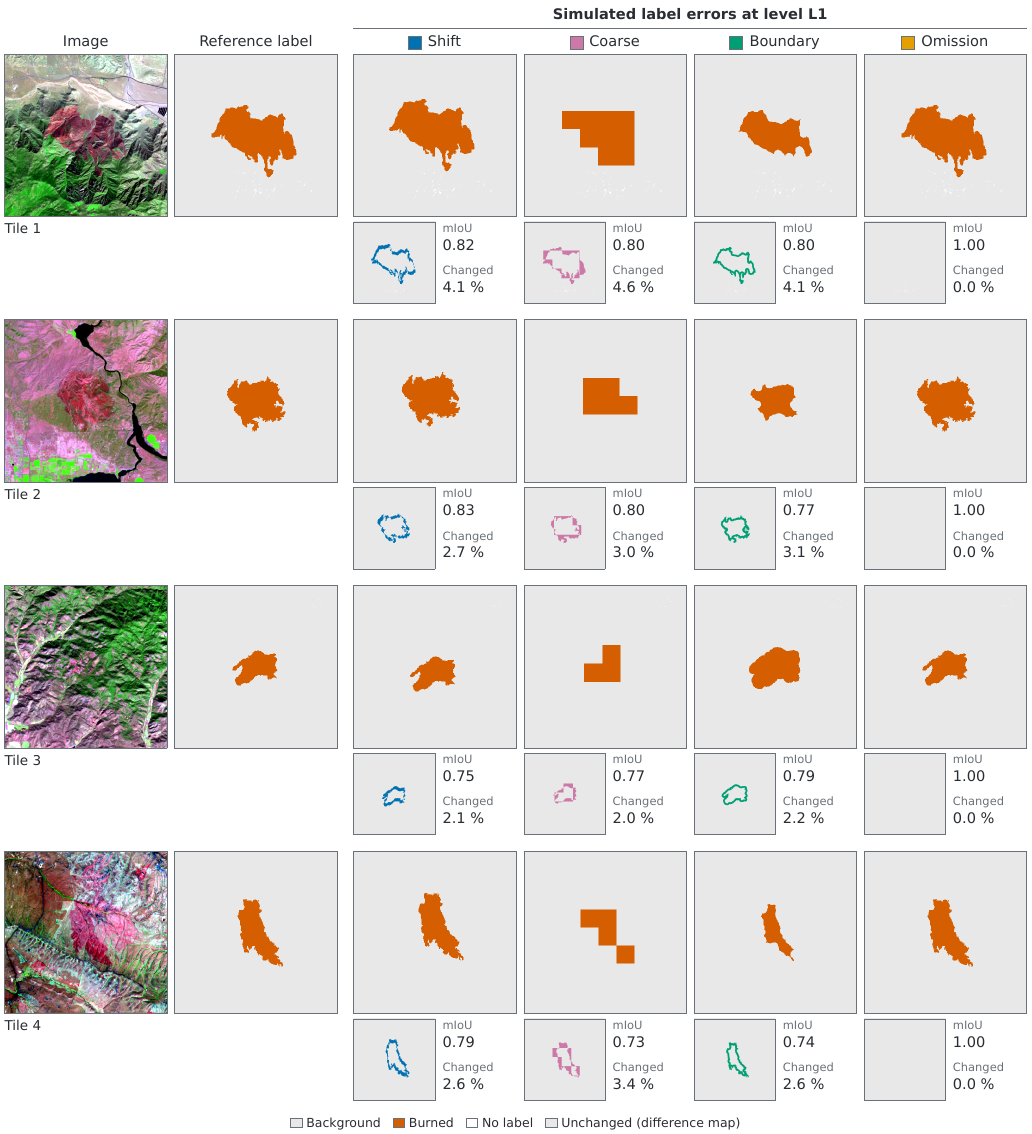}}{\phfig{8cm}{planned/figs/fig\_gallery\_burn\_scars\_L1.pdf from figs/make\_corruption\_gallery.py}}
\caption{\textbf{Simulated label errors on \burn at \Lone:} Each block is one training tile. The first column shows a false-color composite of the second shortwave infrared, near infrared and red bands and the second column the reference label. The next four columns show the labels under \etype{shift}, \etype{coarse}, \etype{boundary} and \etype{omission}. The small map under each corrupted label marks the changed pixels in the color of the error type. The label \miou of the tile and its share of changed pixels are printed beside the map. White marks pixels without a label.}
\label{fig:ax-gallery-burn_scars-l1}
\end{figure*}

\clearpage

\begin{figure*}[p]
\centering
\IfFileExists{planned/figs/fig_gallery_flair2.pdf}{\includegraphics[width=\textwidth]{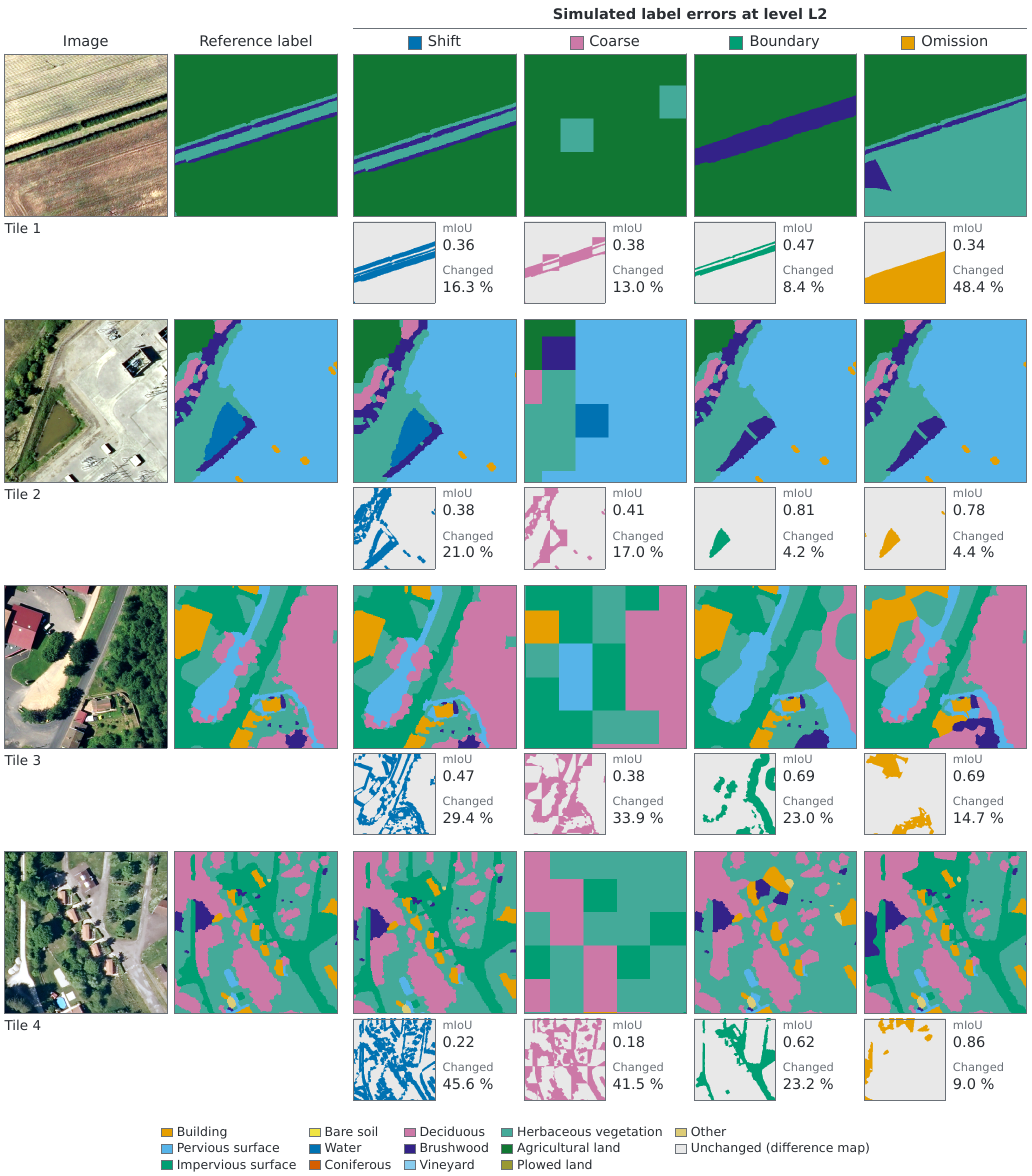}}{\phfig{8cm}{planned/figs/fig\_gallery\_flair2.pdf from figs/make\_corruption\_gallery.py}}
\caption{\textbf{Simulated label errors on \flair at \Ltwo:} Each block is one training tile. The first column shows the RGB channels of the aerial image and the second column the reference label. The next four columns show the labels under \etype{shift}, \etype{coarse}, \etype{boundary} and \etype{omission}. The small map under each corrupted label marks the changed pixels in the color of the error type. The label \miou of the tile and its share of changed pixels are printed beside the map.}
\label{fig:ax-gallery-flair2}
\end{figure*}

\begin{figure*}[p]
\centering
\IfFileExists{planned/figs/fig_gallery_flair2_L1.pdf}{\includegraphics[width=\textwidth]{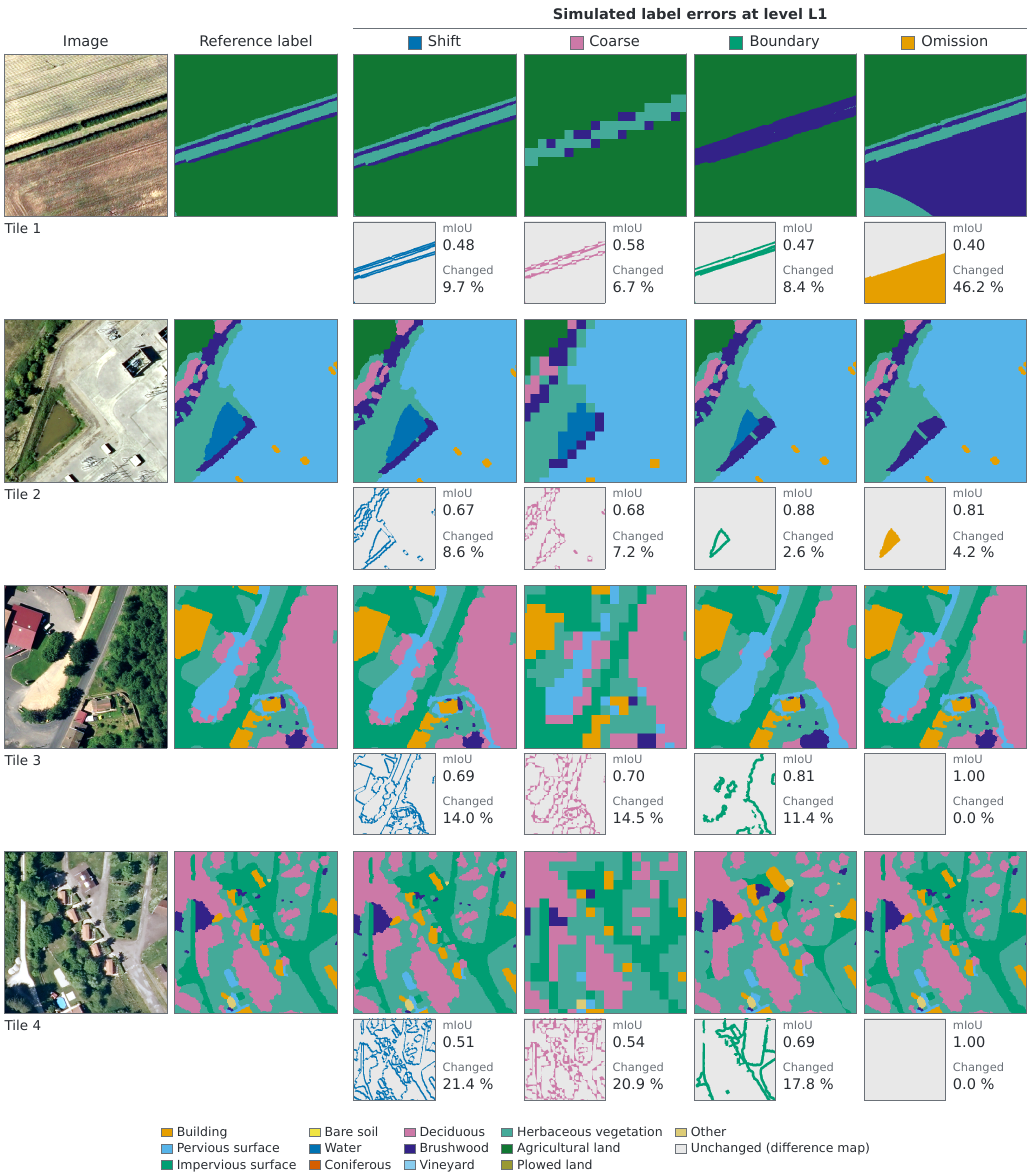}}{\phfig{8cm}{planned/figs/fig\_gallery\_flair2\_L1.pdf from figs/make\_corruption\_gallery.py}}
\caption{\textbf{Simulated label errors on \flair at \Lone:} Each block is one training tile. The first column shows the RGB channels of the aerial image and the second column the reference label. The next four columns show the labels under \etype{shift}, \etype{coarse}, \etype{boundary} and \etype{omission}. The small map under each corrupted label marks the changed pixels in the color of the error type. The label \miou of the tile and its share of changed pixels are printed beside the map.}
\label{fig:ax-gallery-flair2-l1}
\end{figure*}

\clearpage

\begin{figure*}[p]
\centering
\IfFileExists{planned/figs/fig_gallery_chesapeake.pdf}{\includegraphics[width=\textwidth]{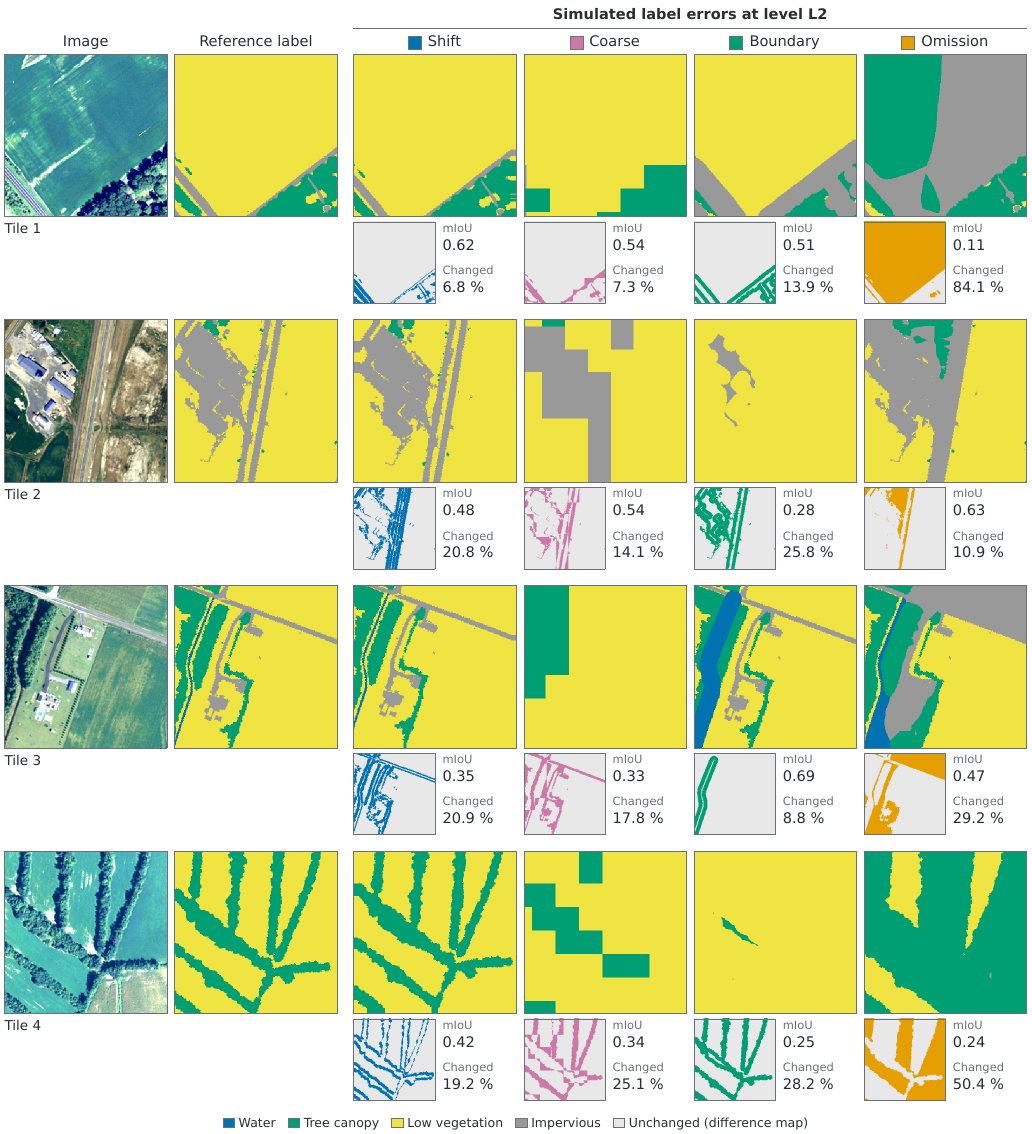}}{\phfig{8cm}{planned/figs/fig\_gallery\_chesapeake.pdf from figs/make\_corruption\_gallery.py}}
\caption{\textbf{Simulated label errors on Chesapeake at \Ltwo:} Each block is one training tile. The first column shows the RGB channels of the NAIP image and the second column the reference label. The next four columns show the labels under \etype{shift}, \etype{coarse}, \etype{boundary} and \etype{omission}. The small map under each corrupted label marks the changed pixels in the color of the error type. The label \miou of the tile and its share of changed pixels are printed beside the map.}
\label{fig:ax-gallery-chesapeake}
\end{figure*}

\begin{figure*}[p]
\centering
\IfFileExists{planned/figs/fig_gallery_chesapeake_L1.pdf}{\includegraphics[width=\textwidth]{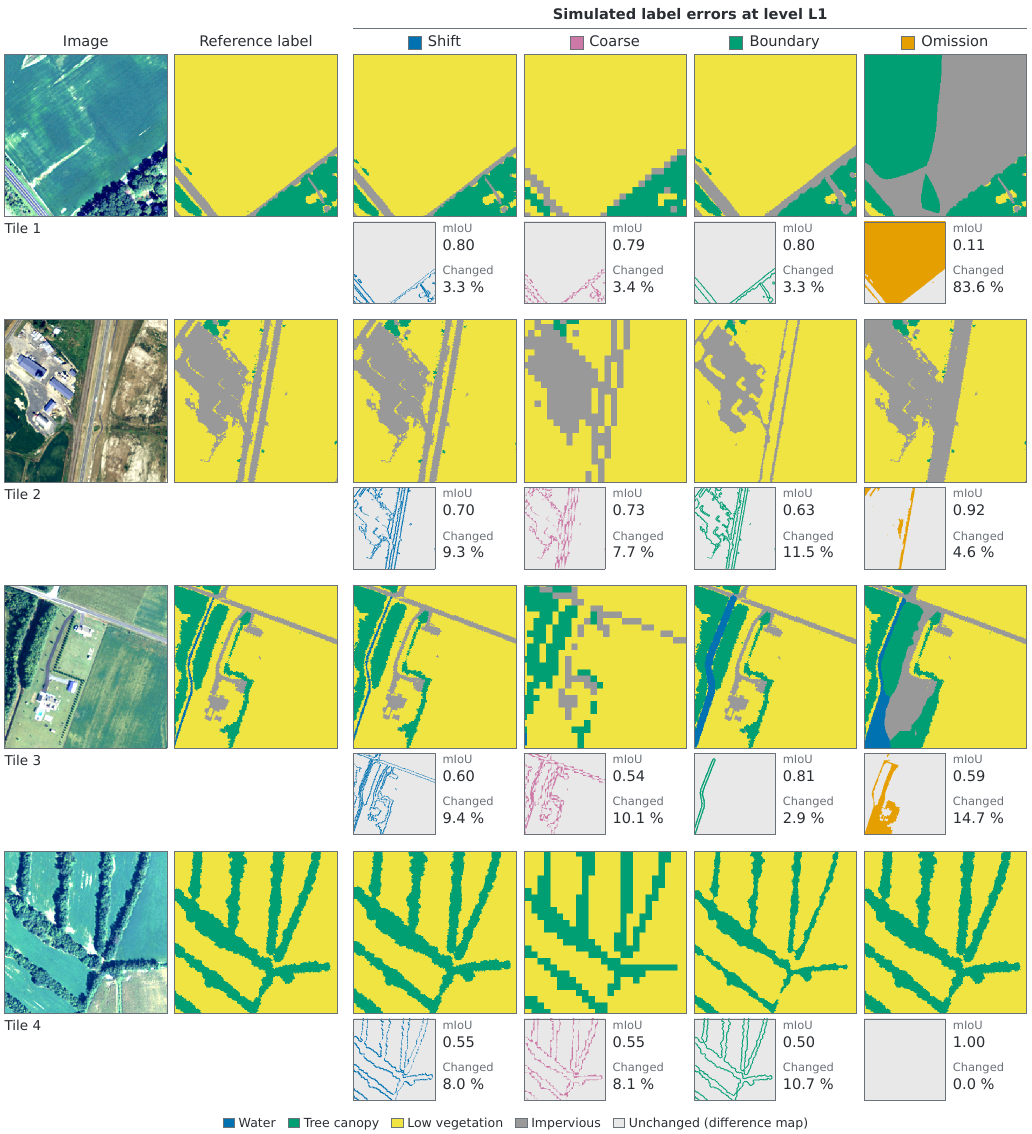}}{\phfig{8cm}{planned/figs/fig\_gallery\_chesapeake\_L1.pdf from figs/make\_corruption\_gallery.py}}
\caption{\textbf{Simulated label errors on Chesapeake at \Lone:} Each block is one training tile. The first column shows the RGB channels of the NAIP image and the second column the reference label. The next four columns show the labels under \etype{shift}, \etype{coarse}, \etype{boundary} and \etype{omission}. The small map under each corrupted label marks the changed pixels in the color of the error type. The label \miou of the tile and its share of changed pixels are printed beside the map.}
\label{fig:ax-gallery-chesapeake-l1}
\end{figure*}

\clearpage

\begin{figure*}[p]
\centering
\IfFileExists{planned/figs/fig_gallery_cloudsen12.pdf}{\includegraphics[width=\textwidth]{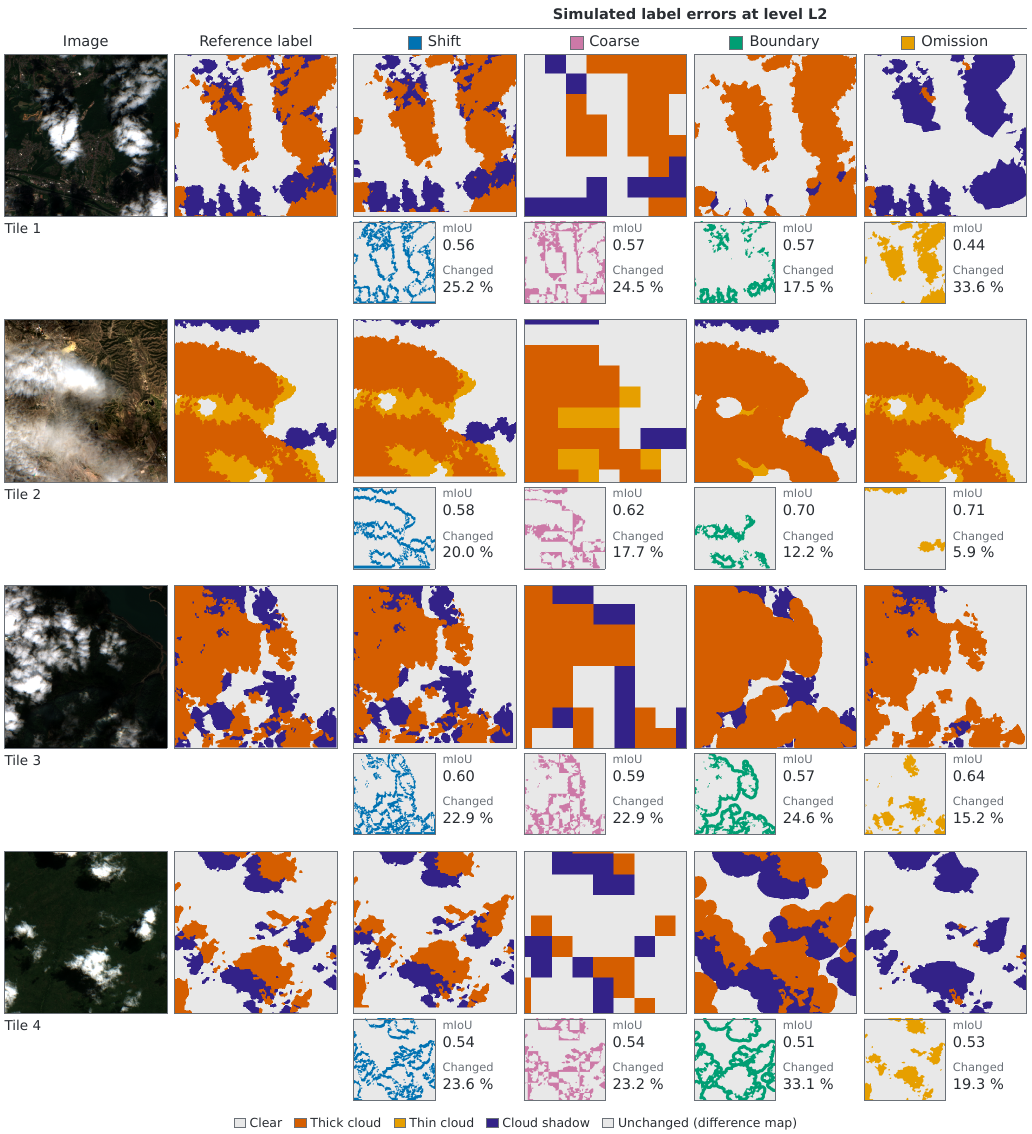}}{\phfig{8cm}{planned/figs/fig\_gallery\_cloudsen12.pdf from figs/make\_corruption\_gallery.py}}
\caption{\textbf{Simulated label errors on CloudSEN12 at \Ltwo:} Each block is one training tile. The first column shows the RGB bands of the Sentinel-2 patch and the second column the reference label. The next four columns show the labels under \etype{shift}, \etype{coarse}, \etype{boundary} and \etype{omission}. The small map under each corrupted label marks the changed pixels in the color of the error type. The label \miou of the tile and its share of changed pixels are printed beside the map.}
\label{fig:ax-gallery-cloudsen12}
\end{figure*}

\begin{figure*}[p]
\centering
\IfFileExists{planned/figs/fig_gallery_cloudsen12_L1.pdf}{\includegraphics[width=\textwidth]{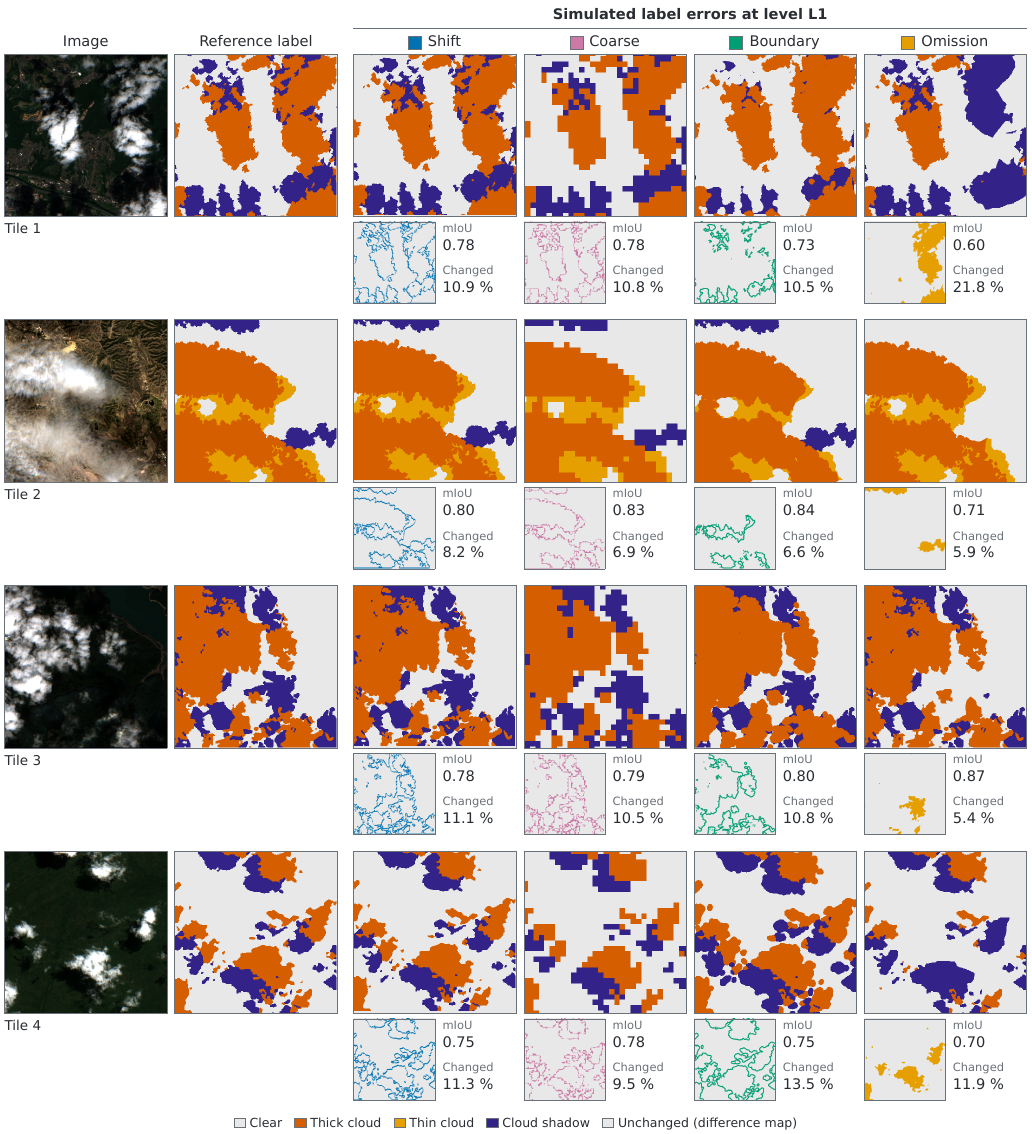}}{\phfig{8cm}{planned/figs/fig\_gallery\_cloudsen12\_L1.pdf from figs/make\_corruption\_gallery.py}}
\caption{\textbf{Simulated label errors on CloudSEN12 at \Lone:} Each block is one training tile. The first column shows the RGB bands of the Sentinel-2 patch and the second column the reference label. The next four columns show the labels under \etype{shift}, \etype{coarse}, \etype{boundary} and \etype{omission}. The small map under each corrupted label marks the changed pixels in the color of the error type. The label \miou of the tile and its share of changed pixels are printed beside the map.}
\label{fig:ax-gallery-cloudsen12-l1}
\end{figure*}

\clearpage

\begin{figure*}[p]
\centering
\IfFileExists{planned/figs/fig_gallery_dfc2020.pdf}{\includegraphics[width=\textwidth]{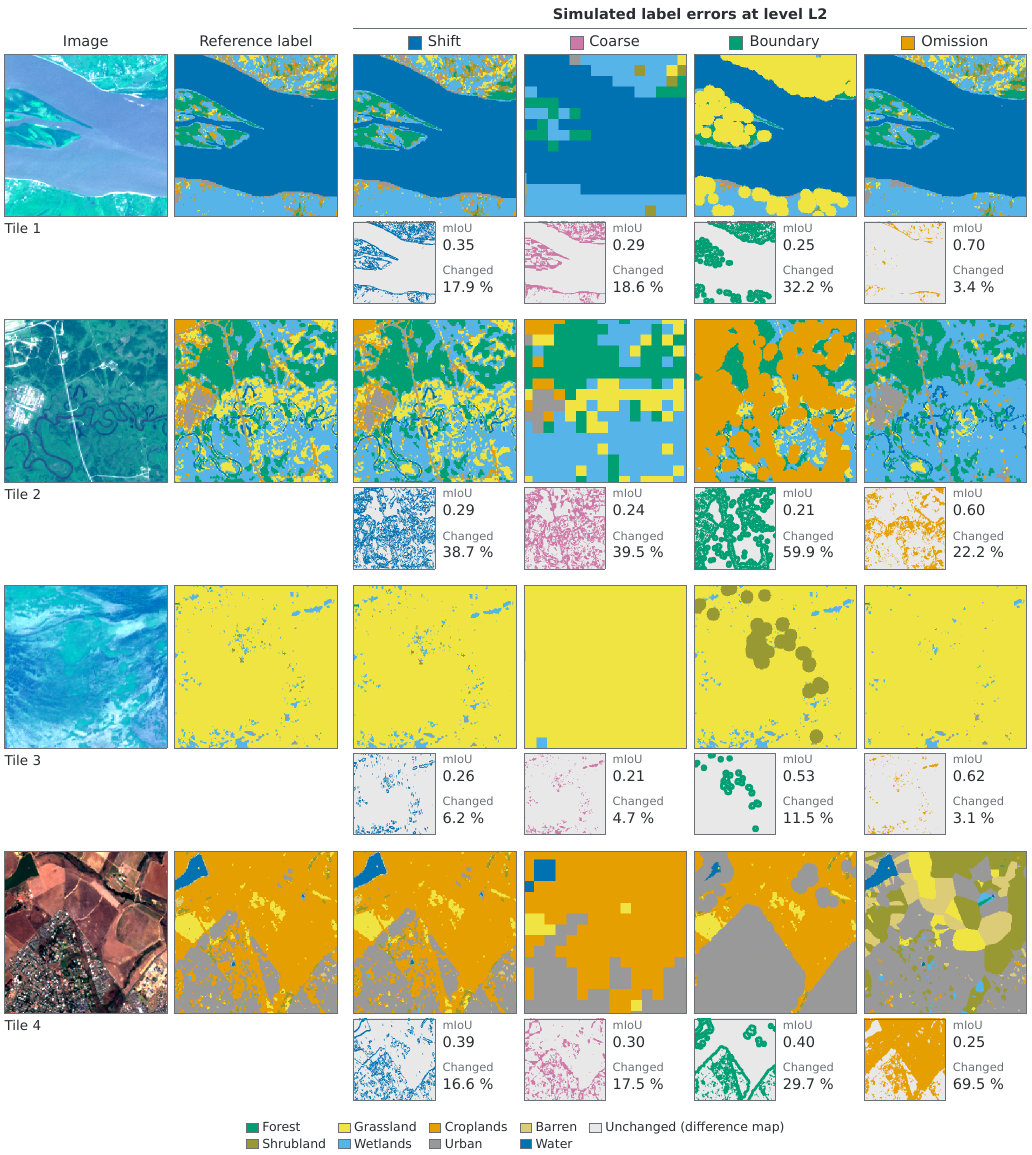}}{\phfig{8cm}{planned/figs/fig\_gallery\_dfc2020.pdf from figs/make\_corruption\_gallery.py}}
\caption{\textbf{Simulated label errors on DFC2020 at \Ltwo:} Each block is one training tile. The first column shows the RGB bands of the Sentinel-2 patch and the second column the reference label. The next four columns show the labels under \etype{shift}, \etype{coarse}, \etype{boundary} and \etype{omission}. The small map under each corrupted label marks the changed pixels in the color of the error type. The label \miou of the tile and its share of changed pixels are printed beside the map.}
\label{fig:ax-gallery-dfc2020}
\end{figure*}

\begin{figure*}[p]
\centering
\IfFileExists{planned/figs/fig_gallery_dfc2020_L1.pdf}{\includegraphics[width=\textwidth]{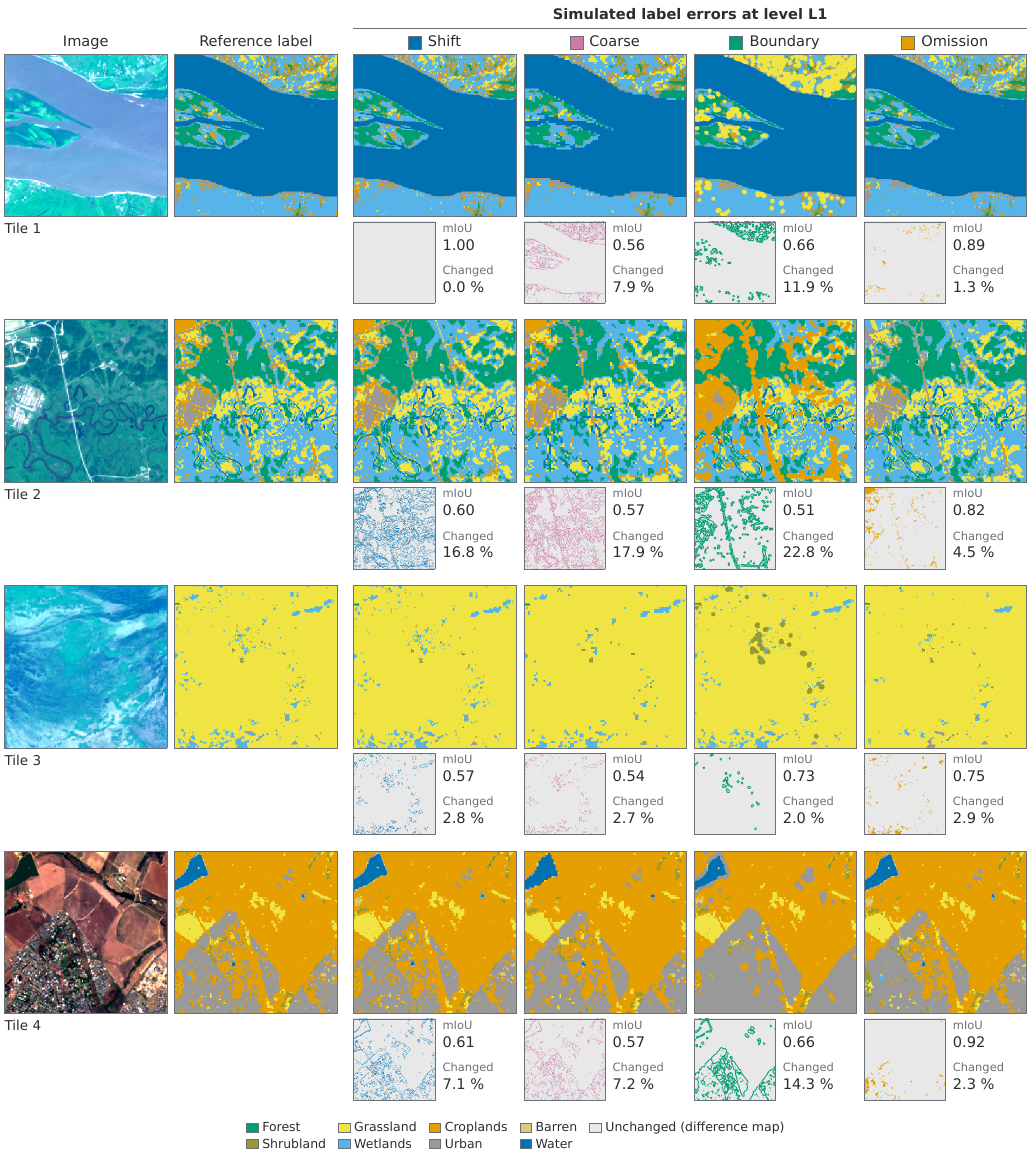}}{\phfig{8cm}{planned/figs/fig\_gallery\_dfc2020\_L1.pdf from figs/make\_corruption\_gallery.py}}
\caption{\textbf{Simulated label errors on DFC2020 at \Lone:} Each block is one training tile. The first column shows the RGB bands of the Sentinel-2 patch and the second column the reference label. The next four columns show the labels under \etype{shift}, \etype{coarse}, \etype{boundary} and \etype{omission}. The small map under each corrupted label marks the changed pixels in the color of the error type. The label \miou of the tile and its share of changed pixels are printed beside the map.}
\label{fig:ax-gallery-dfc2020-l1}
\end{figure*}

\clearpage

\begin{figure*}[p]
\centering
\IfFileExists{planned/figs/fig_gallery_enviroatlas.pdf}{\includegraphics[width=\textwidth]{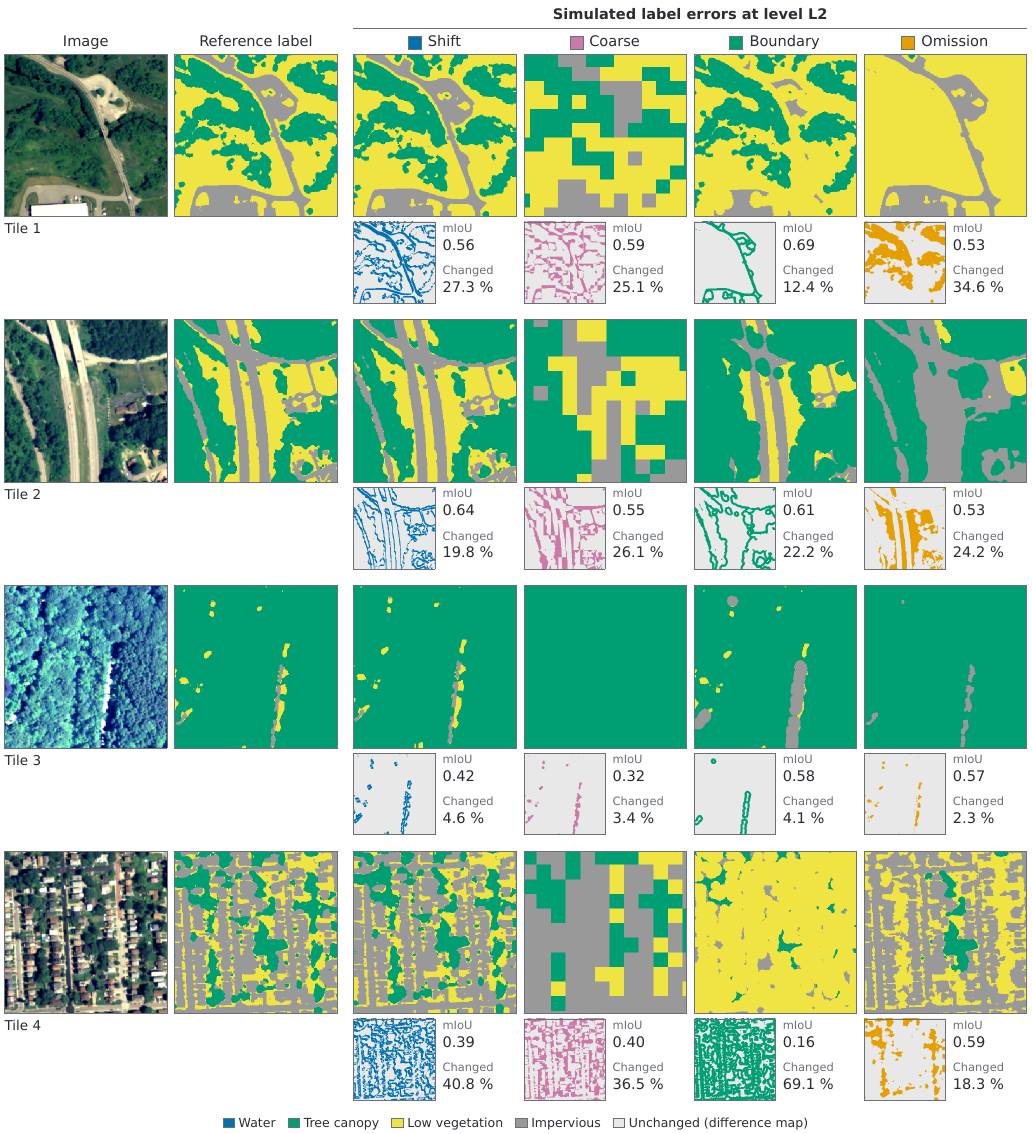}}{\phfig{8cm}{planned/figs/fig\_gallery\_enviroatlas.pdf from figs/make\_corruption\_gallery.py}}
\caption{\textbf{Simulated label errors on EnviroAtlas at \Ltwo:} Each block is one training tile. The first column shows the RGB channels of the NAIP image and the second column the reference label. The next four columns show the labels under \etype{shift}, \etype{coarse}, \etype{boundary} and \etype{omission}. The small map under each corrupted label marks the changed pixels in the color of the error type. The label \miou of the tile and its share of changed pixels are printed beside the map.}
\label{fig:ax-gallery-enviroatlas}
\end{figure*}

\begin{figure*}[p]
\centering
\IfFileExists{planned/figs/fig_gallery_enviroatlas_L1.pdf}{\includegraphics[width=\textwidth]{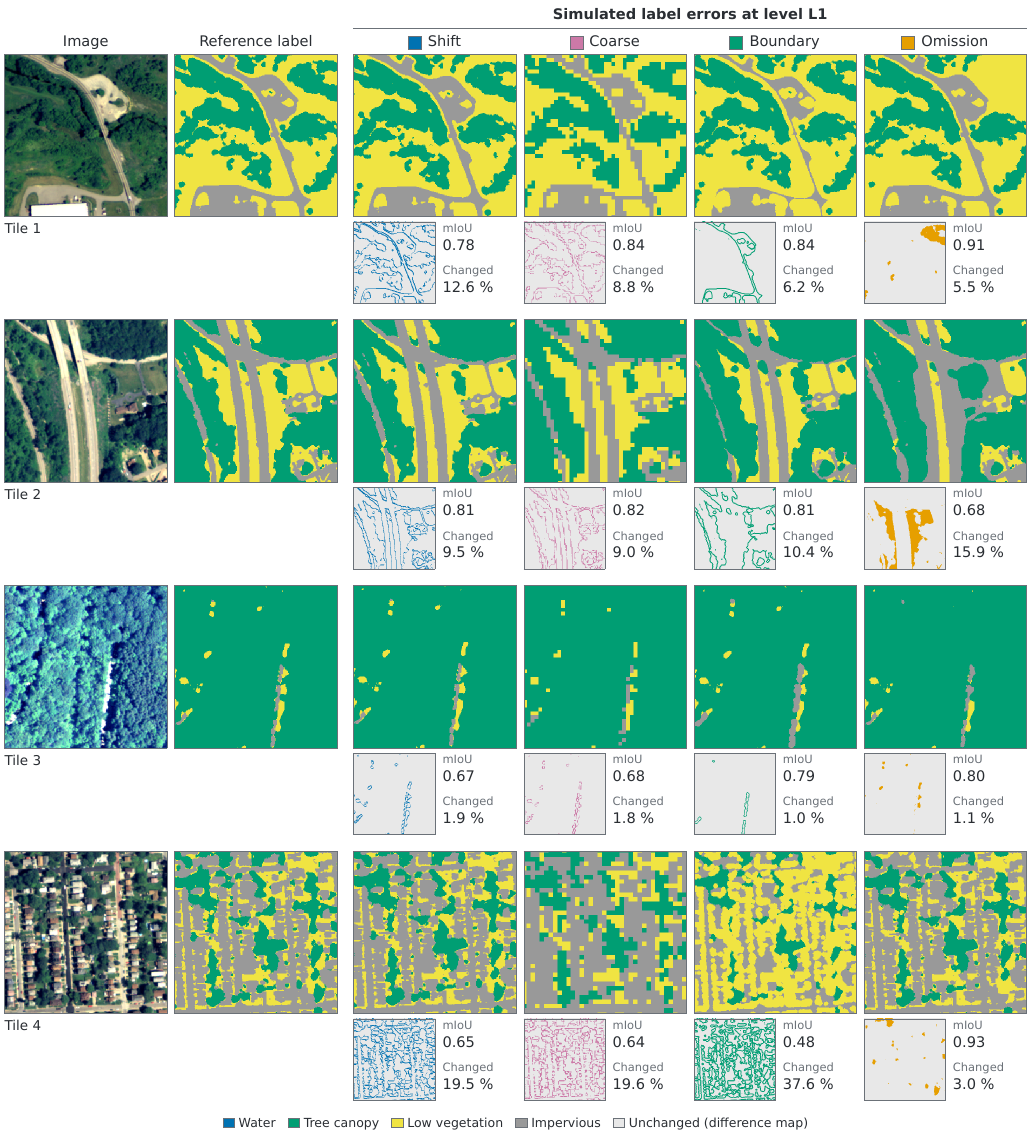}}{\phfig{8cm}{planned/figs/fig\_gallery\_enviroatlas\_L1.pdf from figs/make\_corruption\_gallery.py}}
\caption{\textbf{Simulated label errors on EnviroAtlas at \Lone:} Each block is one training tile. The first column shows the RGB channels of the NAIP image and the second column the reference label. The next four columns show the labels under \etype{shift}, \etype{coarse}, \etype{boundary} and \etype{omission}. The small map under each corrupted label marks the changed pixels in the color of the error type. The label \miou of the tile and its share of changed pixels are printed beside the map.}
\label{fig:ax-gallery-enviroatlas-l1}
\end{figure*}

\clearpage

\begin{figure*}[p]
\centering
\IfFileExists{planned/figs/fig_gallery_spacenet2.pdf}{\includegraphics[width=\textwidth]{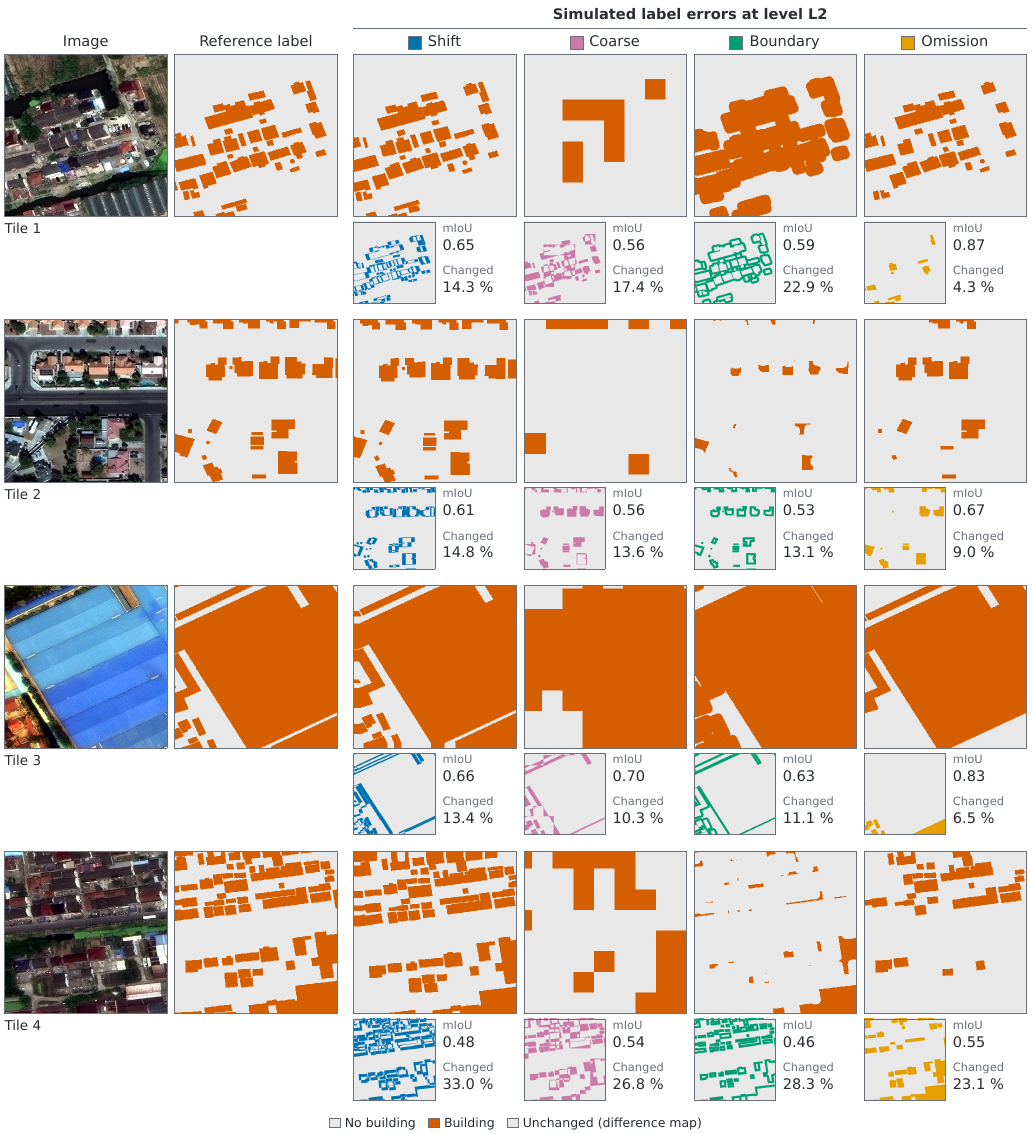}}{\phfig{8cm}{planned/figs/fig\_gallery\_spacenet2.pdf from figs/make\_corruption\_gallery.py}}
\caption{\textbf{Simulated label errors on SpaceNet 2 at \Ltwo:} Each block is one training tile. The first column shows the RGB bands of the WorldView-3 chip and the second column the reference label. The next four columns show the labels under \etype{shift}, \etype{coarse}, \etype{boundary} and \etype{omission}. The small map under each corrupted label marks the changed pixels in the color of the error type. The label \miou of the tile and its share of changed pixels are printed beside the map.}
\label{fig:ax-gallery-spacenet2}
\end{figure*}

\begin{figure*}[p]
\centering
\IfFileExists{planned/figs/fig_gallery_spacenet2_L1.pdf}{\includegraphics[width=\textwidth]{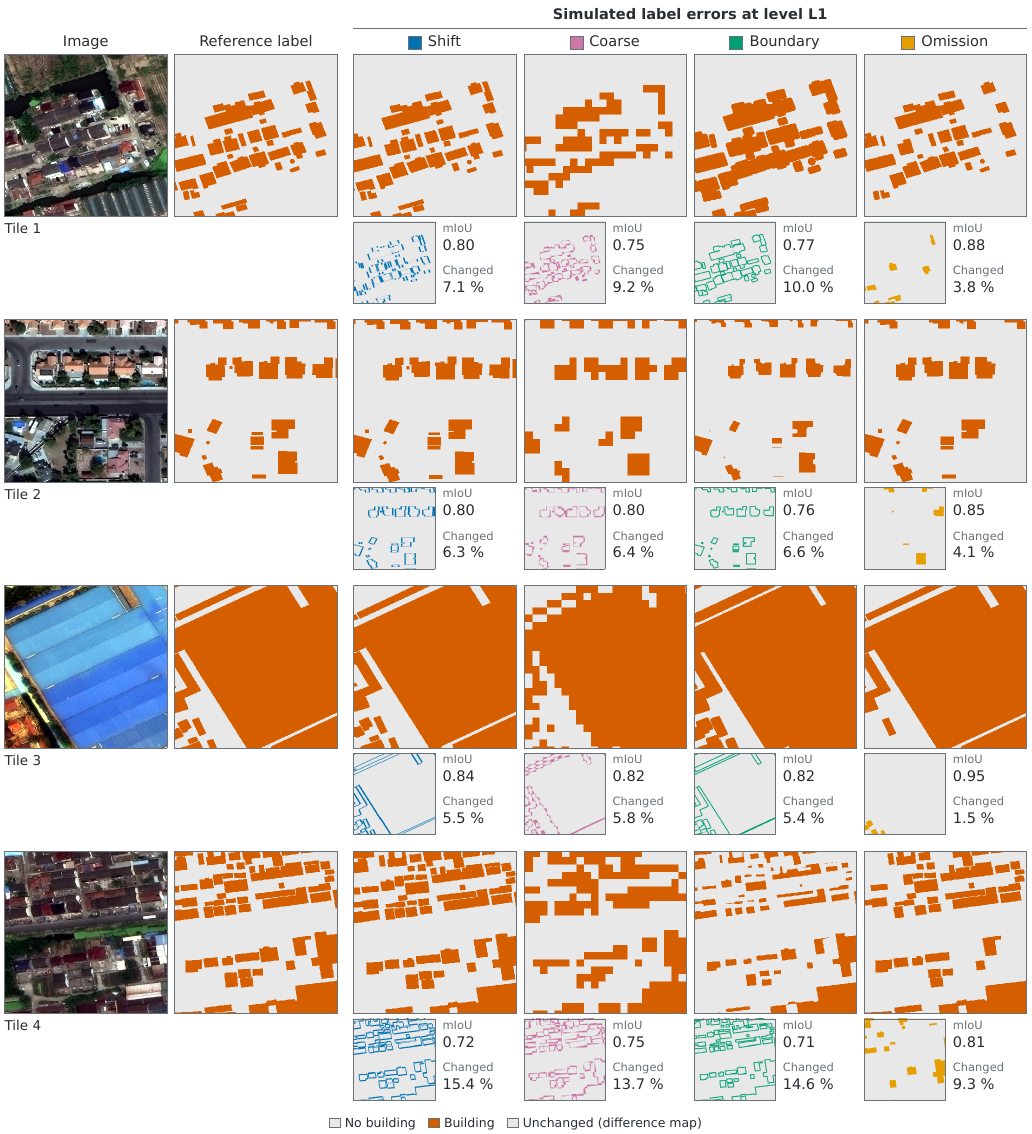}}{\phfig{8cm}{planned/figs/fig\_gallery\_spacenet2\_L1.pdf from figs/make\_corruption\_gallery.py}}
\caption{\textbf{Simulated label errors on SpaceNet 2 at \Lone:} Each block is one training tile. The first column shows the RGB bands of the WorldView-3 chip and the second column the reference label. The next four columns show the labels under \etype{shift}, \etype{coarse}, \etype{boundary} and \etype{omission}. The small map under each corrupted label marks the changed pixels in the color of the error type. The label \miou of the tile and its share of changed pixels are printed beside the map.}
\label{fig:ax-gallery-spacenet2-l1}
\end{figure*}

\clearpage

\begin{figure*}[p]
\centering
\IfFileExists{planned/figs/fig_gallery_kuro_siwo.pdf}{\includegraphics[width=\textwidth]{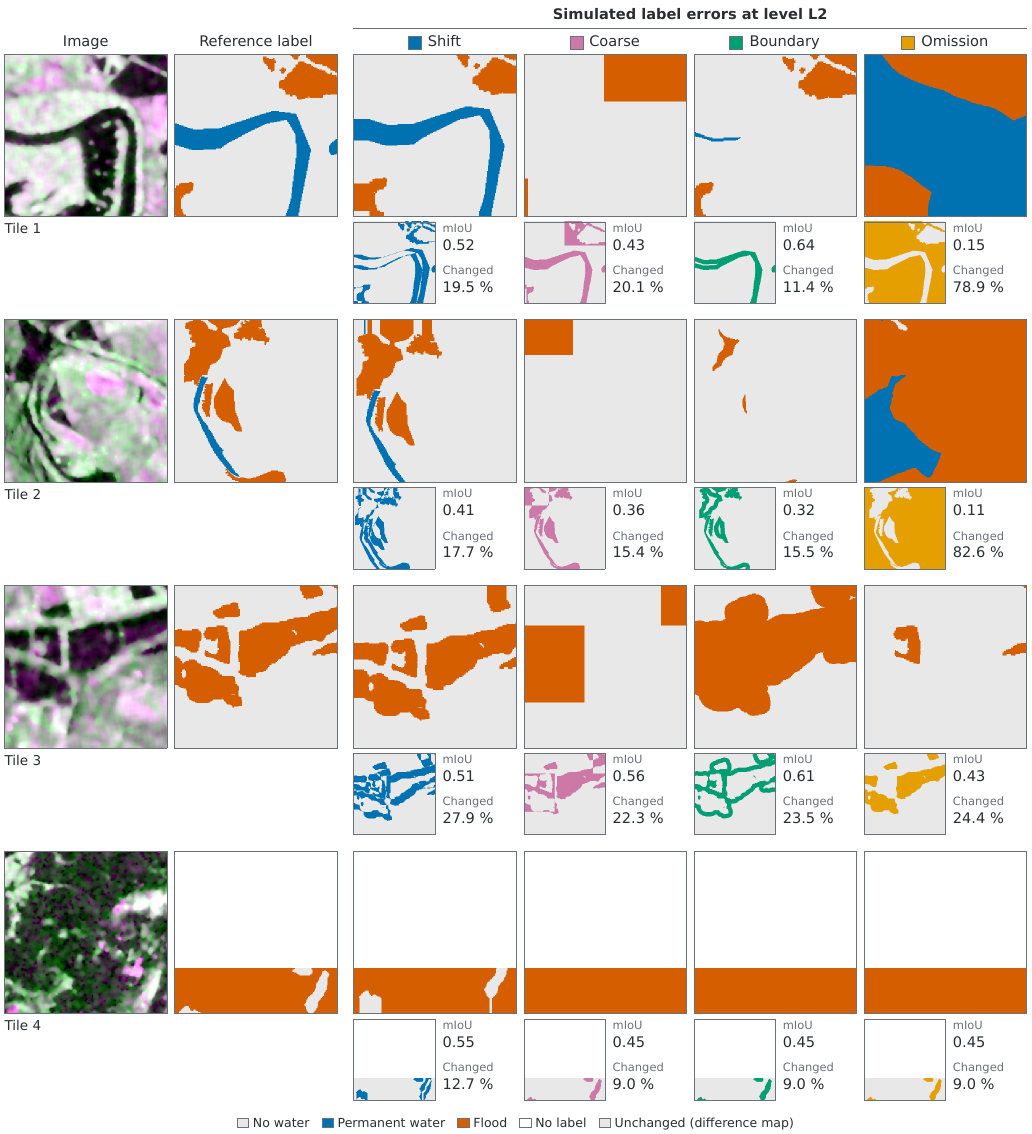}}{\phfig{8cm}{planned/figs/fig\_gallery\_kuro\_siwo.pdf from figs/make\_corruption\_gallery.py}}
\caption{\textbf{Simulated label errors on Kuro Siwo at \Ltwo:} Each block is one training tile. The first column shows a false-color composite of the VV and VH backscatter in decibels and the second column the reference label. The next four columns show the labels under \etype{shift}, \etype{coarse}, \etype{boundary} and \etype{omission}. The small map under each corrupted label marks the changed pixels in the color of the error type. The label \miou of the tile and its share of changed pixels are printed beside the map. White marks pixels without a label.}
\label{fig:ax-gallery-kuro_siwo}
\end{figure*}

\begin{figure*}[p]
\centering
\IfFileExists{planned/figs/fig_gallery_kuro_siwo_L1.pdf}{\includegraphics[width=\textwidth]{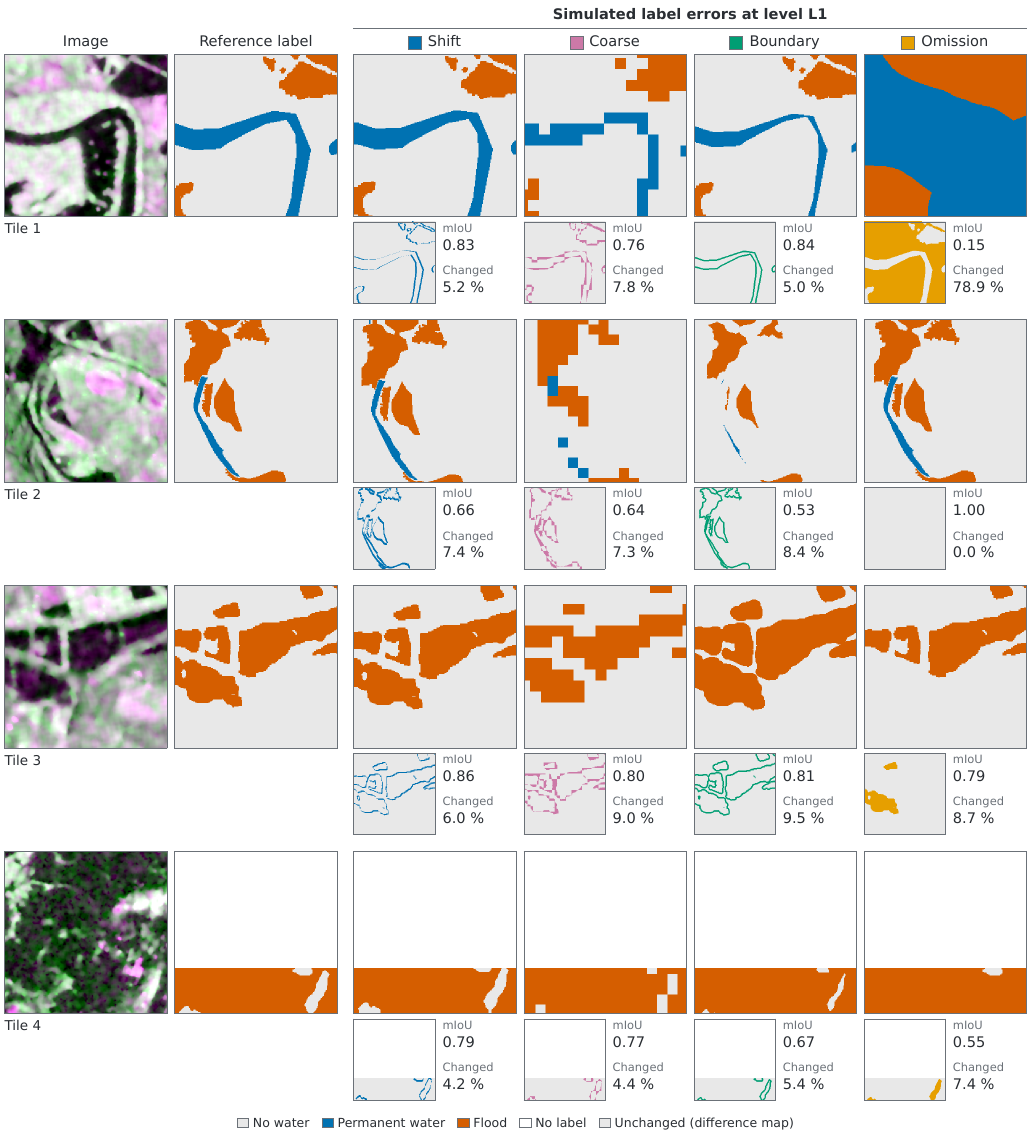}}{\phfig{8cm}{planned/figs/fig\_gallery\_kuro\_siwo\_L1.pdf from figs/make\_corruption\_gallery.py}}
\caption{\textbf{Simulated label errors on Kuro Siwo at \Lone:} Each block is one training tile. The first column shows a false-color composite of the VV and VH backscatter in decibels and the second column the reference label. The next four columns show the labels under \etype{shift}, \etype{coarse}, \etype{boundary} and \etype{omission}. The small map under each corrupted label marks the changed pixels in the color of the error type. The label \miou of the tile and its share of changed pixels are printed beside the map. White marks pixels without a label.}
\label{fig:ax-gallery-kuro_siwo-l1}
\end{figure*}

\clearpage

\begin{figure*}[p]
\centering
\IfFileExists{planned/figs/fig_gallery_openearthmap.pdf}{\includegraphics[width=\textwidth]{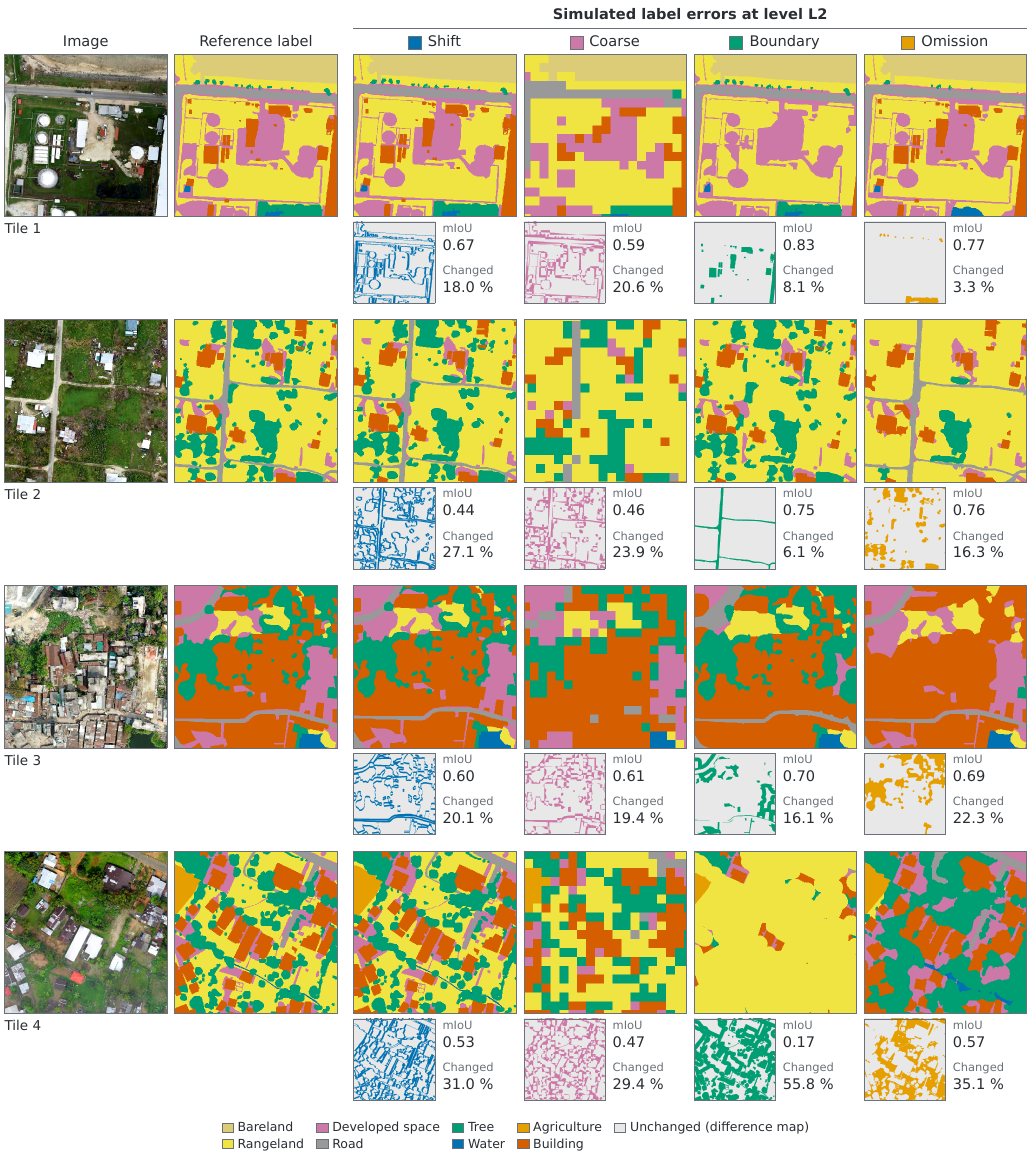}}{\phfig{8cm}{planned/figs/fig\_gallery\_openearthmap.pdf from figs/make\_corruption\_gallery.py}}
\caption{\textbf{Simulated label errors on OpenEarthMap at \Ltwo:} Each block is one training tile. The first column shows the RGB image and the second column the reference label. The next four columns show the labels under \etype{shift}, \etype{coarse}, \etype{boundary} and \etype{omission}. The small map under each corrupted label marks the changed pixels in the color of the error type. The label \miou of the tile and its share of changed pixels are printed beside the map.}
\label{fig:ax-gallery-openearthmap}
\end{figure*}

\begin{figure*}[p]
\centering
\IfFileExists{planned/figs/fig_gallery_openearthmap_L1.pdf}{\includegraphics[width=\textwidth]{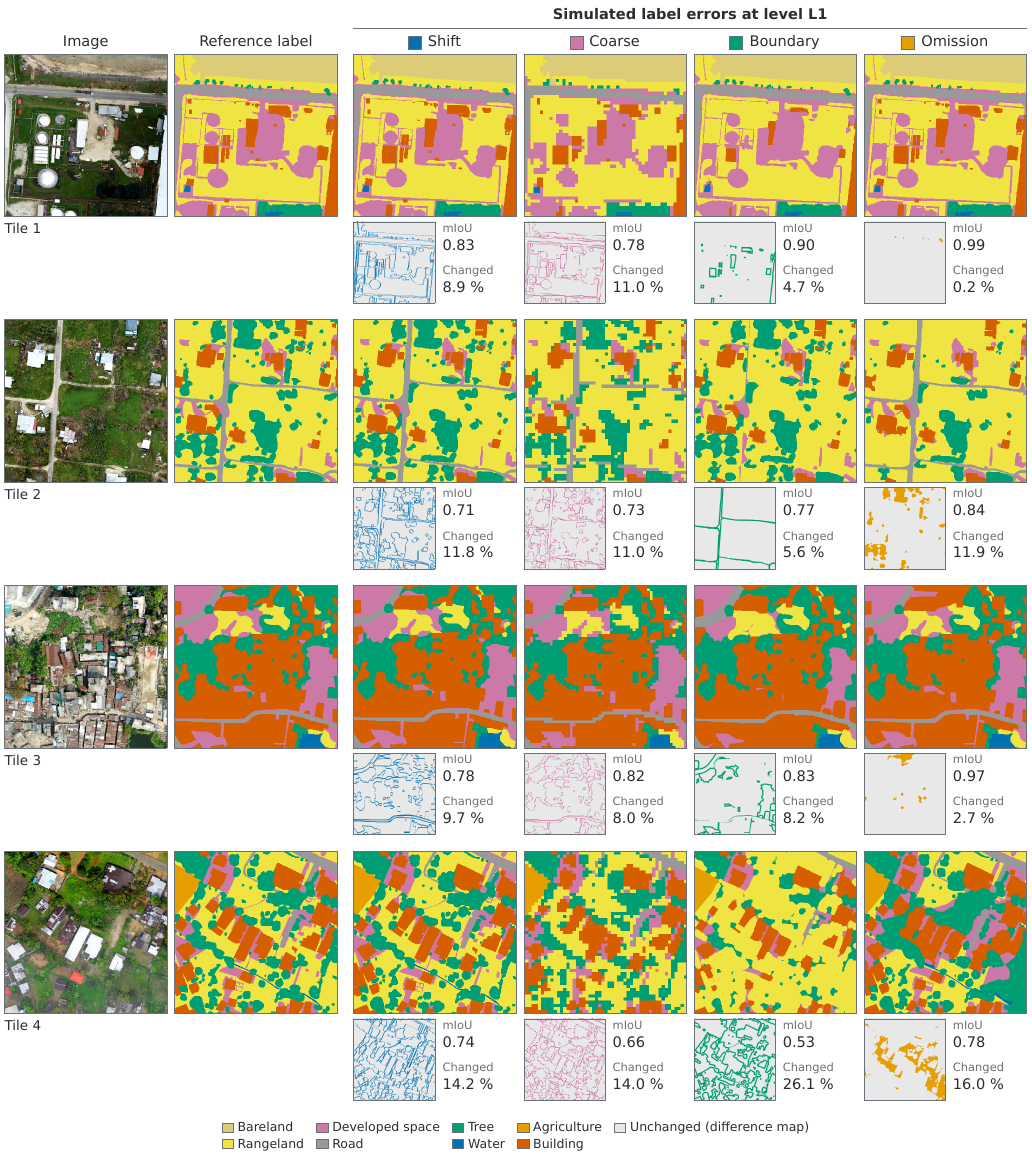}}{\phfig{8cm}{planned/figs/fig\_gallery\_openearthmap\_L1.pdf from figs/make\_corruption\_gallery.py}}
\caption{\textbf{Simulated label errors on OpenEarthMap at \Lone:} Each block is one training tile. The first column shows the RGB image and the second column the reference label. The next four columns show the labels under \etype{shift}, \etype{coarse}, \etype{boundary} and \etype{omission}. The small map under each corrupted label marks the changed pixels in the color of the error type. The label \miou of the tile and its share of changed pixels are printed beside the map.}
\label{fig:ax-gallery-openearthmap-l1}
\end{figure*}

\clearpage
\fi
\fi
\fi

\end{document}